\documentclass[letterpaper, 12pt]{cmuthesis}

\usepackage{libertine}
\usepackage{fullpage}
\usepackage{appendix}
\usepackage{chngcntr}
\usepackage{graphicx}
\usepackage{amsmath}
\usepackage{amssymb}
\usepackage{amsthm}
\usepackage{bm}
\usepackage{xspace}
\usepackage[font={small,bf,stretch=1.0}, format=plain]{caption}
\usepackage{subcaption}
\usepackage{pifont}
\usepackage{multirow}
\usepackage{algorithmicx}
\usepackage{algpseudocode}
\usepackage{algorithm}
\usepackage{hhline}
\usepackage{mathtools}
\usepackage[numbers,sort]{natbib}
\usepackage{macros}
\usepackage[pageanchor=true,plainpages=false, pdfpagelabels, bookmarks,bookmarksnumbered
]{hyperref}

\usepackage{setspace}
\usepackage{csquotes}
\usepackage{booktabs}
\usepackage{tabularx}
\usepackage{float}
\usepackage{tocloft}

\hypersetup{
  colorlinks   = true, %Colours links instead of ugly boxes
  linkcolor={red!50!black},
  citecolor={green!50!black},
  urlcolor={blue!80!black}
}

\AtBeginEnvironment{subappendices}{%
  \chapter*{Appendix}
  \addcontentsline{toc}{chapter}{Appendices}
}

\AtEndEnvironment{subappendices}{%
}

\begin {document} 
\frontmatter
\title{ %% {\it \huge Thesis Proposal}\\
  {\bf Machine Learning Systems\\ for Highly-Distributed and Rapidly-Growing Data}}
\author{Kevin Hsieh}
%\date{Thesis Proposal\\July 28, 2018}
\Year{2019}
\permission{All Rights Reserved}
\trnumber{}

%\committee{
%  Prof. Phillip B. Gibbons (Co-Chair) \\
%  Prof. Onur Mutlu (Co-Chair) \\
%  Prof. Gregory R. Ganger \\
%  Dr. Ganesh Ananthanarayanan \\
%  Dr. H. Brendan McMahan
%}

\maketitle

\begin{abstract}

The usability and practicality of any machine learning (ML) applications are largely influenced by two critical but hard-to-attain factors: low latency and low cost. Unfortunately, achieving low latency and low cost is very challenging when ML depends on real-world data that are highly distributed and rapidly growing (e.g., data collected by mobile phones and video cameras all over the world). Such real-world data pose many challenges in communication and computation. For example, when training data are distributed across data centers that span multiple continents, communication among data centers can easily overwhelm the limited wide-area network bandwidth, leading to prohibitively high latency and high cost.

In this dissertation, we demonstrate that the latency and cost of ML on highly-distributed and rapidly-growing data can be improved by one to two orders of magnitude by designing ML systems that exploit the characteristics of ML algorithms, ML model structures, and ML training/serving data. We support this thesis statement with three contributions. First, we design a system that provides both low-latency and low-cost ML serving (inferencing) over large-scale and continuously-growing datasets, such as videos. Second, we build a system that makes ML training over geo-distributed datasets as fast as training within a single data center. Third, we present a first detailed study and a system-level solution on a fundamental and largely overlooked problem: ML training over non-IID (i.e., not independent and identically distributed) data partitions (e.g., facial images collected by cameras varies according to the demographics of each camera's location).

\end{abstract}

\begin{acknowledgments}

I am grateful to everyone who enables me to pursue this challenging and exciting journey. First and foremost, I am indebted to my advisors, Phil Gibbons and Onur Mutlu, for their guidance, feedback, trust, and support throughout the years. I am incredibly fortunate to work with Phil, as his exemplary advice taught me various aspects of research. Phil guided me into a new research field when I had no clues, and he continuously provided valuable feedback in the most constructive and encouraging way. His passion for research gave me essential supports, especially when things are inevitably not working at times. His succinct writing and presentation were also great learning examples. Everything I learned from Phil will keep benefiting me in years to come.

I am equally fortunate to have the opportunity to work with Onur. After leading me into CMU, Onur put tremendous trust in me and support me throughout the years. I learned greatly from Onur, especially in striving for fundamental research and the highest clarity in writing and presentation. I always remembered how Onur shaped my view on research by showing me Dr. Hamming's ``You and your research''. The talk is indeed an excellent blueprint for a research journey. Onur also provided resources and opportunities for fruitful collaborations with the SAFARI research group and industrial collaborators. All of these helped me greatly along the way.

I am grateful to the members of my PhD committee: Greg Ganger, Ganesh Ananthanarayanan, and Brendan McMahan. Their valuable feedback and help brought this dissertation to complete. Over the years, Greg always gave me insightful feedback on countless occasions, and the research discussions with him gave me many different perspectives. Ganesh not only served in my PhD committee, but he was also my internship mentor in Microsoft Research. Ganesh guided me into a cutting-edge problem, which led to a central piece of the dissertation. I am also grateful to Brendan for all his time, efforts, and insightful comments that significantly improved many aspects of the dissertation. 

I want to give a big thank you to my internship mentors and industrial collaborators. Amar Phanishayee provided valuable feedback and vital support in the last few years of the journey. Eiman Ebrahimi taught me many research fundamentals when I interned in NVIDIA Research. Peter Bodik provided many important ideas and feedback when I was an intern in Microsoft Research. I also thank all the collaborators for their contributions and feedback: Victor Bahl, Niladrish Chatterjee, Steve Keckler, Gwangsun Kim, Mike O'Connor, Matthai Philipose, and Shivaram Venkataraman.

I thank my fellow PhD students in the SAFARI group and the PDL group, for their support and friendship. Special thanks to Nandita Vijaykumar for her invaluable feedback and help throughout the years, Kevin Chang and Hongyi Xin for their friendship and experience sharing, Aaron Harlap for his help and friendship (and NBA rumors), Cui Henggang for his help on multiple parameter server systems, Dimitris Konomis for his proof on the Gaia work, Donghyuk Lee for his kind support, Vivek Seshadri for his acute tips and feedback, as well as Samira Khan, Gennady Pekhimenko, and Yoongu Kim for their valuable suggestions in the early years. I also thank all the peers for their discussions, feedback, and support: Yixin Luo, Amirali Boroumand, Rachata Ausavarungnirun, Saugata Ghose, Jinliang Wei, Vignesh Balaji, Rajat Kateja, Jin Kyu Kim, Chris Fallin, Jianyu Wang, Lavanya Subramanian, Justin Meza, Yang Li, Jeremie Kim, Damla Senol, and Minesh Patel.

I benefited greatly from the PDL events, including the visit days and the retreats. These events connected me with many industrial collaborators who significantly influenced the course of my journey. Special thanks to all the people who made these events possible: Karen Lindenfelser, Joan Digney, Jason Boles, Chad Dougherty, Mitch Franzos, Garth Gibson, and Bill Courtright. I also thank the members and companies of the PDL Consortium, including Alibaba, Amazon, Datrium, Dell EMC, Facebook, Google, Hewlett Packard Labs, Hitachi, IBM Research, Intel Corporation, Micron, Microsoft Research, NetApp, Oracle Corporation, Salesforce, Samsung, Seagate Technology, and Two Sigma, for their interest, insights, feedback, and support.

This research was supported in part by our industrial partners: Google, Huawei, Intel, Microsoft, NVIDIA, Samsung, Seagate, and VMWare. This research was also partially supported by NSF (grants 1212962, 1320531, 1409723), Intel STC on Cloud Computing (ISTC-CC), Intel STC on Visual Cloud Systems (ISTC-VCS), and the Dept of Defense under contract FA8721-05-C-0003. 

Lastly, my most important thank you goes to my family. None of these would have been possible without their inspiring encouragement, deep understanding, endless love, and ultimate support.

\end{acknowledgments}

%\setlength{\cftbeforesecskip}{2pt}
%\begin{singlespacing}
\tableofcontents
\clearpage
\listoffigures
\clearpage
\listoftables
\clearpage

%\end{singlespacing}
%\thispagestyle{empty}

\mainmatter
%\pagebreak
%\pagenumbering{arabic} 

\let\cleardoublepage\clearpage

\chapter{Introduction}
\label{ch:introduction}

The explosive advancement of machine learning (ML) has been the engine
of many important applications such as image or video classification
(e.g.,\cite{DBLP:conf/nips/FromeCSBDRM13,
  DBLP:conf/cvpr/SzegedyLJSRAEVR15, DBLP:conf/cvpr/KarpathyTSLSF14}),
speech recognition (e.g.,~\cite{DBLP:journals/spm/X12a}),
recommendation systems (e.g.,~\cite{DBLP:conf/kdd/GemullaNHS11}), and
self-driving cars
(e.g.,~\cite{DBLP:journals/corr/BojarskiTDFFGJM16}). At its core, an
ML-driven application generally has two distinct phases: \emph{(i)
  training:} the process of searching for the best \emph{model} to
describe or explain training data (e.g., finding the neural network
model that can most accurately classify training images); and
\emph{(ii) serving:} using the pre-trained model to answer questions
for an input data (e.g., predicting the object class of an input
image). The success of both phases depends on two key factors:
\emph{low latency and low cost}.

Low latency is crucial for ML training and serving for three
reasons. First, latency of ML training is a key bottleneck for many ML
applications, as training an ML model can take days or even
months. Second, low-latency training enables fast iterations of
model/algorithm exploration, which is imperative for ML algorithm
developers to find high-accuracy models. Third, the latency of ML
serving largely determines the response time of an application,
which is critical for user-facing applications. Other than low
latency, low cost is equally important for ML applications. Both ML
training and serving can process large-scale datasets, which require
substantial computation and communication resources (e.g., thousands
of GPUs or heavy communication via wireless networks). The cost of
these resources can largely determine the practicality and feasibility
of an ML application.

Achieving low-latency and low-cost ML is particularly challenging when
ML depends on real-world, large-scale data. An example of this is the
large-scale deployment of many emerging ML applications, such as image
or video classification (e.g.,\cite{DBLP:conf/cvpr/KarpathyTSLSF14}),
speech recognition (e.g.,~\cite{DBLP:journals/spm/X12a}), and topic
modeling (e.g.,~\cite{BleiNJ03}), by large organizations like Google,
Microsoft, and Amazon. These data are generated at very rapid rates,
all over the world. As a result, the most timely and relevant ML data are
\emph{highly distributed} and \emph{rapidly growing}, which pose
three major challenges for low-latency and low-cost ML:

\begin{enumerate}

\item {\bf Computation challenge:} Large and rapidly-growing data
  requires corresponding computation power to process, and achieving
  low latency is challenging when the data quantity (e.g., a large
  amount of videos, genomics data, or user activity) overwhelms the
  computation power of an ML training or serving system. While
  pre-processing data may reduce user-facing latency, it can lead to
  excess monetary cost to the system, especially when some data are
  not relevant. For example, searching for people in a particular
  camera in a large enterprise may involve pre-processing videos in
  \emph{all} cameras, which requires drastically more machine time and
  thus more monetary cost in the cloud. Hence, how to tame latency and
  cost while dealing with rapidly-growing data is a fundamental
  challenge for many ML systems.

\item {\bf Communication challenge:} When ML data are highly
  distributed, massive communication overhead can drastically slow
  down an ML system and introduce substantial cost. For example, if
  training data are distributed in many data centers across multiple
  continents, communication among data centers can easily overwhelm
  the limited wide-area network (WAN) bandwidth, leading to
  prohibitively high latency. Furthermore, usage of WAN communication
  is very costly, and the cost of WAN communication can be much higher
  than the cost of machine time.

\item {\bf Statistical challenge:} Highly-distributed data are
  typically generated in different contexts, which can lead to
  significant differences in the distribution of the data across data
  partitions. For example, facial images collected by cameras will
  reflect the demographics of each camera's location, and images of
  kangaroos will be collected only from cameras in Australia or
  zoos. Such \emph{non-IID data} (i.e., not independent and
  identically distributed) pose a fundamental statistical challenge
  for ML training, because distributed ML systems assume each data
  partition is IID. Addressing the statistical challenge requires more
  frequent communication, which exacerbates the high-latency and
  high-cost problem associated with communication during training.
  
\end{enumerate}

For ML training, despite a significant amount of work that aims to
enable large-scale ML training applications (e.g.,~\cite{Mllib,
  LowGKBGH12, DBLP:conf/osdi/ChilimbiSAK14, DeanCMCDLMRSTYN12,
  HoCCLKGGGX13, CuiTWXDHHGGGX14, XingHDKWLZXKY15,
  DBLP:conf/osdi/LiAPSAJLSS14, DBLP:conf/cloud/WeiDQHCGGGX15,
  tensorflow2015-whitepaper, DBLP:conf/eurosys/CuiZGGX16}), the vast
majority of them assume \khi{that} the training data is centralized
within a data center and thus they do not address the challenge of
highly-distributed data. Few recent works
(e.g.,~\cite{DBLP:conf/aistats/McMahanMRHA17,
  DBLP:conf/ICLR/LinHMWD18, DBLP:conf/nips/SmithCST17}) that address
the communication challenge of highly-distributed data do not address
the statistical challenge directly. On the other hand, existing ML
serving systems (e.g.,~\cite{tensorflow_serving,
  DBLP:conf/wsdm/AgarwalLTXZ14, DBLP:conf/cidr/CrankshawBGLZFG15,
  DBLP:conf/nsdi/CrankshawWZFGS17}) mostly focus on serving smaller
querying data such as user preferences and images and they do not
address the computation challenge of large, rapidly-growing data
such as videos.

\section{Thesis Statement}

The goal of this dissertation is to enable low-latency and low-cost
ML training and serving on highly-distributed and rapidly-growing
data by proposing system-level solutions to tackle above
challenges. Our approach can be summarized as the following {\bf
  thesis statement}:

\begin{displayquote}
\emph{The latency and cost of ML training and serving on
  highly-distributed and rapidly-growing data can be improved by
  one to two orders of magnitude by designing ML systems that exploit
  the characteristics of ML algorithms, ML model structures, and ML
  training/serving data.}
\end{displayquote}

\section{Overview of Our Approach}

In line with the thesis statement, we take three directions to address
the aforementioned challenges: (i) we design and build a system to
provide both low-latency and low-cost ML serving over large-scale
rapidly-growing datasets (e.g. videos); (ii) we design and build a
low-latency and low-cost ML training system over geo-distributed
datasets; and (iii) we characterize the fundamental problem of ML
training over non-IID data partitions in detail, \kt{and we propose a
system-level solution to this problem.} We provide a brief overview of
each direction in the rest of this section.

\subsection{ML Serving over Large, Rapidly-Growing Datasets (e.g., Videos)}

Large volumes of videos are continuously recorded from cameras
deployed for traffic control and surveillance with the goal of
answering ``after the fact'' queries: {\em identify video frames with
  objects of certain classes (cars, bags)} from many days of recorded
video. While advancements in convolutional neural networks (\cnns)
have enabled \emph{serving} such queries with high accuracy, they are
too expensive and slow. Current systems for serving such queries on
large video datasets incur either high cost at video ingest time or
high latency at query time. We present {\focus}, a system providing
both low-cost and low-latency querying on large video
datasets. {\focus}' architecture flexibly and effectively divides the
query processing work between \emph{ingest time} and \emph{query
  time}. At ingest time (on live videos), {\focus} uses cheap
convolutional network classifiers (\cnns) to construct an
\emph{approximate index} of all possible object classes in each frame
(to handle queries for {\em any} class in the future). At query time,
     {\focus} leverages this approximate index to provide low latency,
     but compensates for the lower accuracy of the cheap \cnns through
     the judicious use of an expensive \cnn.  Experiments on
     commercial video streams show that {\focus} is $48\times$ (up to
     $92\times$) cheaper than using expensive \cnns for ingestion, and
     provides $125\times$ (up to $607\times$) lower query latency than
     a state-of-the-art video querying system
     (\sys{NoScope}~\cite{DBLP:journals/pvldb/KangEABZ17}).

\subsection{ML Training over Geo-Distributed Data}

ML is widely used to derive useful information from large-scale data
generated at increasingly rapid rates, \emph{all over the
  world}. Unfortunately, it is infeasible to move all this
globally-generated data to a centralized data center before running an
ML algorithm over it{\textemdash}moving large amounts of training data
over wide-area networks (WANs) can be extremely slow, and is also
subject to the constraints of privacy and data sovereignty laws. To
this end, we introduce a new, general geo-distributed ML training
system, {\gaia}, that decouples the communication \emph{within} a data
center from the communication \emph{between} data centers, enabling
different communication and consistency models for each.  We present a
new ML \synchronization model, \emph{\protocol (\protoabbrv)}, whose
key idea is to dynamically eliminate \emph{insignificant}
communication between data centers while still guaranteeing the
correctness of ML algorithms. Our experiments on our prototypes of
{\gaia} running across 11 Amazon EC2 global regions and on a cluster
that emulates EC2 WAN bandwidth show that, compared to two two
state-of-the-art distributed ML training systems, {\gaia} (1)
significantly improves performance, by 1.8--53.5$\times$, (2)
\khii{has performance within 0.94--1.40$\times$ of running the same ML
  algorithm on a local area network (LAN) in a single data center},
and (3) significantly reduces the monetary cost of running the same ML
algorithm on WANs, by 2.6--59.0$\times$.

%\subsection{Understanding The Non-IID Data Partition Problem for Decentralized ML}
\subsection{The Non-IID Data Partition Problem for Decentralized ML}

  Many large-scale machine learning (ML) applications need to train ML
  models over \emph{decentralized} datasets that are generated at
  different devices and locations. These decentralized datasets pose a
  fundamental challenge to ML because they are typically generated in
  very different contexts, which leads to significant differences in
  data distribution across devices/locations. In this work, we take a
  step toward better understanding this challenge, by presenting the
  first detailed experimental study of the impact of such
  \emph{non-IID data} on the decentralized training of deep neural
  networks (DNNs). Our study shows that: (i) the problem of non-IID
  data partitions is fundamental and pervasive, as it exists in all ML
  applications, DNN models, training datasets, and decentralized
  learning algorithms in our study; (ii) this problem is particularly
  difficult for DNN models with batch normalization layers; and (iii)
  the degree of deviation from IID (the skewness) is a key determinant
  of the difficulty level of the problem.  \kt{With these findings in
  mind, we present {\sscout}, a system-level approach that adapts the
  communication frequency of decentralized learning algorithms to the
  (skew-induced) accuracy loss between data partitions.  We also show
  that group normalization can recover much of the skew-induced
  accuracy loss of batch normalization.}
  
\section{Contributions}

This dissertation makes the following contributions.

\begin{enumerate}
  
\item We present a new architecture for low-cost and low-latency ML
  serving over large and rapidly-growing datasets (e.g., videos),
  based on a principled split of ingest and query functionalities. To
  this end, we propose techniques for efficient indexing with a cheap
  \cnn at ingest time, while ensuring high recall and precision by
  judiciously using expensive \cnns at query time. We demonstrate that
  our proposed system provides a new design point for ML serving
  systems that trade off between ingest cost and query latency: our
  system is significantly cheaper than an ingest-heavy design and
  significantly faster than query-optimized systems.

\item We propose a first general geo-distributed ML system that (1)
  differentiates the communication over a LAN from the communication
  over WANs to make efficient use of the scarce and heterogeneous WAN
  bandwidth, and (2) is general and flexible enough to deploy a wide
  range of ML algorithms while requiring \emph{no} change to the ML
  algorithms themselves. Our system is based on a new, efficient ML
  \synchronization model, \khi{\protocol (\protoabbrv)}, for
  communication between parameter servers across data centers over
  WANs. \protoabbrv guarantees \khi{that} each data center's view of
  the ML model parameters is approximately the same as the
  ``fully-consistent'' view and ensures that all significant updates
  are synchronized in time. We prove that \protoabbrv provides a
  theoretical guarantee on algorithm convergence for a widely used ML
  algorithm, stochastic gradient \khi{descent}. We build \khi{two}
  prototypes of our proposed system on CPU-based and GPU-based ML
  systems, and we demonstrate their effectiveness over 11 globally
  distributed regions with three popular ML algorithms. We show that
  our system provides significant performance improvements over
  \khi{two} \khii{state-of-the-art} distributed ML
  systems\khi{~\cite{CuiTWXDHHGGGX14, DBLP:conf/eurosys/CuiZGGX16}},
  and significantly reduces the communication overhead over WANs.

\item We build a deep understanding on the problem of non-IID data
  partitions for decentralized learning by conducting a first
  detailed empirical study. To our knowledge, our study is the
  first to show that the problem of non-IID data partitions is a
  fundamental and pervasive challenge for decentralized learning. We
  then make a new observation showing that the challenge of non-IID
  data partitions is particularly problematic for DNNs with batch
  normalization, even under the most conservative communication
  approach.  We discuss the root cause of this problem and we find
  that it can be addressed by using an alternative normalization
  technique. Third, we show that the difficulty level of this problem
  varies with the data skew. \kt{Finally, we design and evaluate \sscout,
  a system-level approach that adapts the communication frequency to
  reflect the skewness in the data, seeking to maximize communication
  savings while preserving model accuracy.}
  
\end{enumerate}

\section{Outline}

The rest of the dissertation is organized as
follows. Chapter~\ref{ch:background} describes necessary backgrounds
for ML training and serving systems. Chapter~\ref{ch:relatedwork}
discusses related work on low-latency and low-cost ML
systems. Chapter~\ref{ch:focus} presents
{\focus}~\cite{DBLP:conf/osdi/HsiehABVBPGM18}, our system that
provides both low-latency and low-cost ML serving (inferencing) over
large-scale and rapidly-growing datasets, such as
videos. Chapter~\ref{ch:gaia} presents
{\gaia}~\cite{DBLP:conf/nsdi/HsiehHVKGGM17}, our geo-distributed ML
training system that makes ML training over geo-distributed datasets
as fast as training within a single data
center. Chapter~\ref{ch:noniid} presents our study and solution on the
problem of non-IID data partitions for decentralized
learning~\cite{hsieh2019noniid}. Finally, Chapter~\ref{ch:conclusion} concludes the
dissertation and presents future research directions.

\chapter{Background}
\label{ch:background}

We first introduce the architectures of widely-used distributed ML
training systems, which serves as the background for our work on
low-latency and low-cost ML training over geo-distributed data and
arbitrarily skewed data partitions
(Section~\ref{subsec:ml_training_system}). We then provide a brief
overview of convolutional Neural Networks (\cnn), the state-of-the-art
approach to detecting and classifying objects in images, which serves
as the background for our work on latency-latency and low-cost an ML
serving system for video queries (Section~\ref{subsec:cnn}).

\section{Distributed ML Training Systems}
\label{subsec:ml_training_system}

While ML training algorithms have different types across different
domains, almost all have the same goal{\textemdash}searching for the
best \emph{model} (usually a set of \emph{parameters}) to describe or
explain the input \emph{data}~\cite{XingHDKWLZXKY15}. For example, the
goal of an image classification neural network is to find the
parameters (of the neural network) that can most accurately classify
the input images. Most ML training algorithms iteratively refine the
ML model until it \emph{converges} to fit the data. The correctness of
an ML training algorithm is thus determined by whether or not the
algorithm can \emph{accurately converge} to the best model for its
training data.

As the training data to an ML training algorithm is usually enormous,
processing all training data on a single machine can take an
unacceptably long time. Hence, the most common strategy to run a
large-scale ML training algorithm is to distribute the training data
among \emph{multiple} worker machines, and have each machine work on a
\emph{shard} of the training data in parallel with other machines. The
worker machines communicate with each other periodically to
\emph{synchronize} the updates from other machines. This strategy,
called \emph{data parallelism}~\cite{DBLP:conf/nips/ChuKLYBNO06}, is
widely used in many popular ML training systems (e.g.,~\cite{Mllib,
  Mahout, LowGKBGH12, DBLP:conf/osdi/ChilimbiSAK14, XingHDKWLZXKY15,
  DBLP:journals/corr/MengBYSVLFTAOXX15, tensorflow2015-whitepaper,
  DBLP:conf/osdi/LiAPSAJLSS14}).

There are many distributed ML training systems, such as ones using the
MapReduce~\cite{DBLP:conf/nips/ChuKLYBNO06} abstraction (e.g.,
MLlib~\cite{Mllib} and Mahout~\cite{Mahout}), ones using the graph
abstraction (e.g., GraphLab~\cite{LowGKBGH12} and
PowerGraph~\cite{DBLP:conf/osdi/GonzalezLGBG12}), and ones using the
parameter server abstraction (e.g., Petuum~\cite{XingHDKWLZXKY15} and
TensorFlow~\cite{tensorflow2015-whitepaper}). Among them, the
parameter server architecture provides a performance advantage over
other systems for many ML applications and has been widely adopted in
many distributed ML training systems.

\begin{figure}[h]
  \centering
  \includegraphics[width=0.50\textwidth]{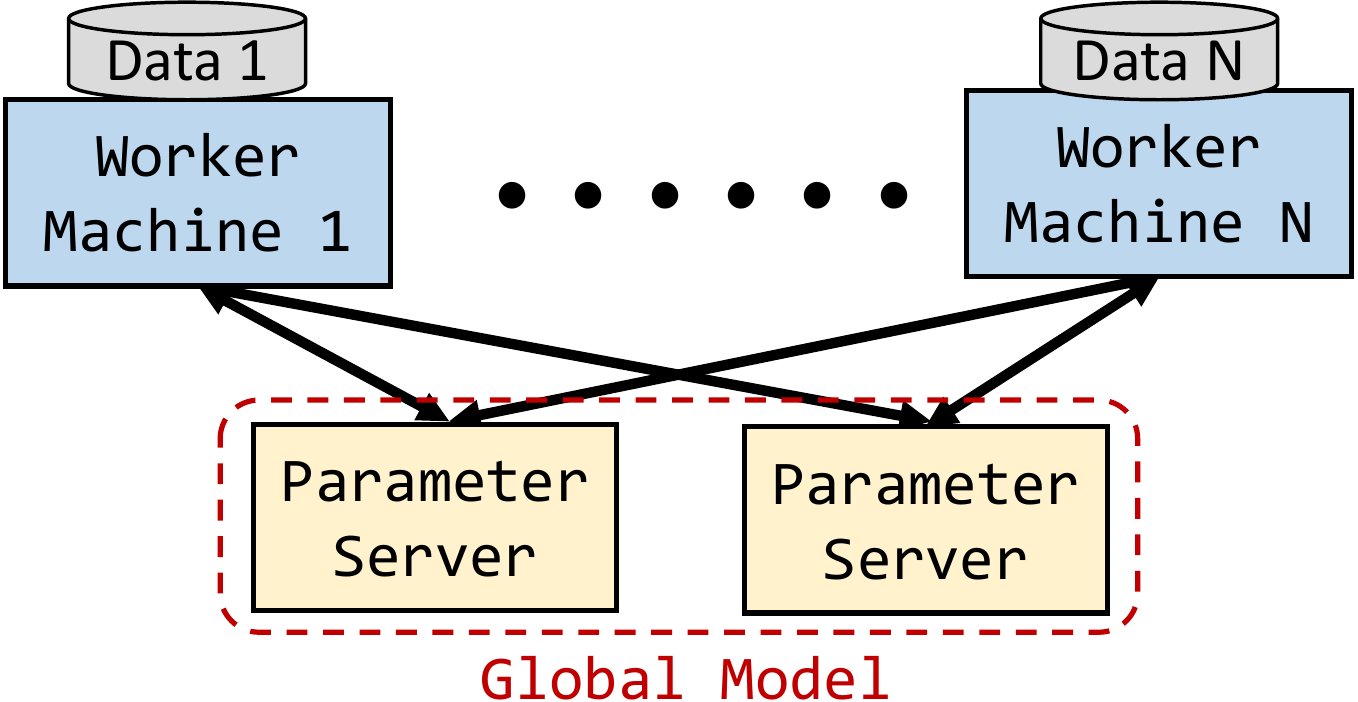}
  \caption{Overview of the parameter server architecture}
  \label{fig:parameter_server}
\end{figure}

Figure~\ref{fig:parameter_server} illustrates the high-level
overview of the parameter server (PS) architecture. In such an
architecture, each parameter server keeps a shard of the global model
parameters as a key-value store, and each worker machine communicates
with the parameter servers to \texttt{READ} and \texttt{UPDATE} the
corresponding parameters. The major benefit of this architecture is
that it allows ML programmers to view all model parameters as a
global shared memory, and leave the parameter servers to handle the
synchronization.

Synchronization among workers in a distributed ML training system is a
critical operation. Each worker needs to see other workers' updates to
the global model to compute more accurate updates using fresh
information. However, synchronization is a high-cost operation that
can significantly slow down the workers and reduce the benefits of
parallelism. The trade-off between \emph{fresher updates} and
\emph{communication overhead} leads to three major \synchronization
models: (1) \textbf{Bulk Synchronous Parallel
  (BSP)}~\cite{DBLP:journals/cacm/Valiant90}, which synchronizes all
updates after each worker goes through its shard of data; all workers
need to see the most up-to-date model before proceeding to the next
iteration, (2) \textbf{Stale Synchronous Parallel
  (SSP)}~\cite{HoCCLKGGGX13}, which allows the fastest worker to be
ahead of the slowest worker by up to a bounded number of iterations,
so the fast workers may proceed with a \emph{bounded stale} (i.e.,
old) model, and (3) \textbf{Total Asynchronous Parallel
  (TAP)}~\cite{DBLP:conf/nips/RechtRWN11}, which removes the
synchronization between workers completely; all workers keep running
based on the results of best-effort communication (i.e., each
sends/receives as many updates as possible). Both BSP and SSP
guarantee algorithm convergence~\cite{HoCCLKGGGX13, dai2015analysis},
while there is no such guarantee for TAP. Most state-of-the-art
parameter servers implement both BSP and SSP
(e.g.,~\cite{HoCCLKGGGX13, CuiTWXDHHGGGX14, XingHDKWLZXKY15,
  CuiCHKLKWDGGGX14, DBLP:conf/osdi/LiAPSAJLSS14,
  tensorflow2015-whitepaper, DBLP:conf/eurosys/CuiZGGX16}).

\section{Convolutional Neural Networks}
\label{subsec:cnn}

Convolution Neural Networks (\cnns) are the state-of-the-art method for many computer vision tasks
such as object detection and classification
(e.g.,~\cite{DBLP:conf/nips/KrizhevskySH12,
  DBLP:conf/cvpr/SzegedyLJSRAEVR15, DBLP:conf/cvpr/HeZRS16, DBLP:journals/corr/RedmonF16, DBLP:conf/cvpr/LinDGHHB17}).

\begin{figure}[h]
  \centering
  \includegraphics[width=0.70\textwidth]{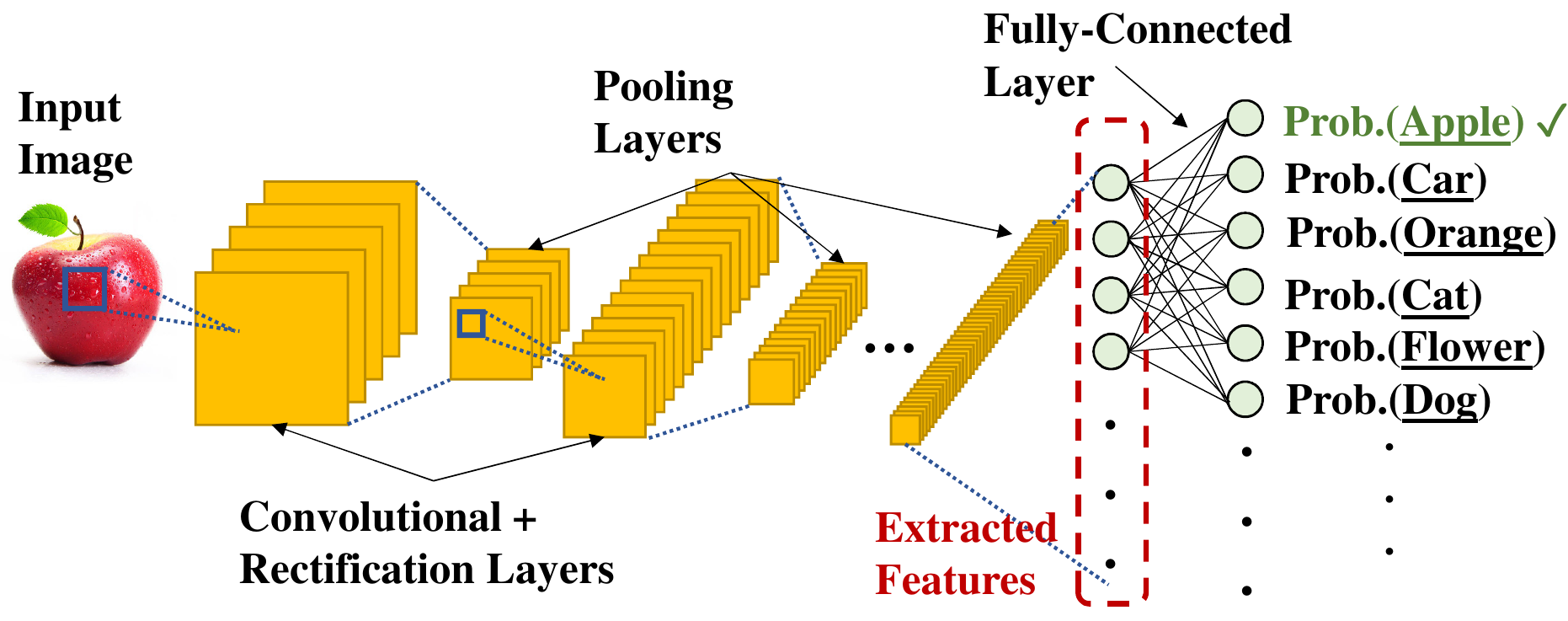}
  \caption{Architecture of an image classification \cnn.}
  \label{fig:cnn}
\end{figure}

Figure~\ref{fig:cnn} illustrates the architecture of a representative 
image classification \cnn. 
Broadly, \cnns consist of different types of layers including convolutional layers, pooling layers
and fully-connected layers. The output from the final layer of a classification \cnn are the probabilities of all object classes (e.g., dog, flower, car), and the class with the highest probability is the predicted class for the object in the input image.

%As the convolutional, rectification, and pooling layers are designed to detect visual features,
The output of the penultimate (i.e., previous-to-last) layer can be
considered as ``representative features'' of the input
image~\cite{DBLP:conf/nips/KrizhevskySH12}. The features are a
real-valued vector, with lengths between $512$ and $4096$ in
state-of-the-art classifier \cnns
(e.g.,~\cite{DBLP:conf/nips/KrizhevskySH12, Simonyan15,
  DBLP:conf/cvpr/SzegedyLJSRAEVR15, DBLP:conf/cvpr/HeZRS16}).
It has been shown that images with
similar feature vectors (i.e., small Euclidean distances) are visually
similar~\cite{DBLP:conf/nips/KrizhevskySH12,
  DBLP:conf/eccv/BabenkoSCL14, DBLP:conf/cvpr/RazavianASC14, DBLP:conf/iccv/BabenkoL15}.

Because {\em inference} using state-of-the-art \cnns is computationally expensive (and slow), 
there are two main techniques that have been developed to reduce the cost of inference.
First, \emph{compression} is a set of techniques that can dramatically reduce the cost of inference at the expense of accuracy.
Such techniques include removing some expensive convolutional layers~\cite{Simonyan15}, matrix pruning~\cite{pruning1,pruning2}, and others~\cite{lowrank,fitnets}. 
For example, ResNet18, which is a ResNet152 variant with only 18 layers is $8\times$ cheaper.
Likewise, Tiny YOLO~\cite{DBLP:journals/corr/RedmonF16}, a shallower variant of the YOLO object detector, is 5$\times$ cheaper than YOLOv2. 
Second, a more recent technique is \cnn \emph{specialization}~\cite{mcdnn}, where the \cnns are trained on a subset of a dataset specific to a particular context, also making them much cheaper. 

\chapter{Related Work}
\label{ch:relatedwork}

We discuss related work that are related to low-latency and low-cost
ML systems on large-scale, highly-distributed data.

\section{Distributed ML Training Systems with Centralized Data}

There are many distributed ML systems that aim to enable large-scale
ML applications (e.g.,~\cite{Mllib, Mahout, LowGKBGH12,
  DBLP:conf/osdi/ChilimbiSAK14, DeanCMCDLMRSTYN12, HoCCLKGGGX13,
  CuiTWXDHHGGGX14, XingHDKWLZXKY15, CuiCHKLKWDGGGX14,
  DBLP:conf/osdi/LiAPSAJLSS14, DBLP:conf/nips/LiASY14,
  DBLP:conf/cloud/WeiDQHCGGGX15, tensorflow2015-whitepaper,
  AhmedAGNS12, DBLP:conf/eurosys/KimHLZDGX16,
  DBLP:conf/eurosys/CuiZGGX16, DBLP:journals/corr/GoyalDGNWKTJH17,
  DBLP:journals/corr/ChenLLLWWXXZZ15,
  DBLP:conf/nips/RechtRWN11}). These systems successfully demonstrate
their effectiveness \khi{on a} large number of machines by employing
various synchronization models and system optimizations. However, all
of them assume \khi{that} the network communication happens
\emph{within} a data center and do not tackle the challenges \khii{of}
highly-distributed data.

\section{Distributed ML Training with Decentralized Data}

Few recent works (e.g.,~\cite{DBLP:conf/aistats/McMahanMRHA17,
  DBLP:conf/ICLR/LinHMWD18, DBLP:conf/ccs/ShokriS15,
  DBLP:conf/icml/TangLYZL18}) aim to enable low-latency ML training on
highly-distributed, decentralized data. For example, federated
learning~\cite{DBLP:conf/aistats/McMahanMRHA17} coordinates mobile
devices to train an ML model using wireless networks while keeping
training data in local devices. Their major focus is to reduce
communication overhead among training nodes, and they either
\emph{(i)} assume the data partitions are IID or \emph{(ii)} conduct
only a limited study on non-IID data partitions. Some recent
work~\cite{DBLP:journals/corr/abs-1806-00582,
  DBLP:conf/nips/SmithCST17} investigates the problem of non-IID data
partitions. For example, instead of training a global model to fit
non-IID data partitions, federated multi-task
learning~\cite{DBLP:conf/nips/SmithCST17} proposes training local
models for each data partition while leveraging other data partitions
to improve the model accuracy.  However, this approach does not solve
the problem for global models, which are essential when a local model
is unavailable (e.g., a brand new partition without training data) or
ineffective (e.g., a partition with too few training examples for a
class, such as kangaroos in Europe).  Zhao et al.'s
study~\cite{DBLP:journals/corr/abs-1806-00582} discusses the problem
of \fedavg~\cite{DBLP:conf/aistats/McMahanMRHA17} over non-IID data
partitions, but this study does not discuss the implication for other
decentralized learning algorithms, ML applications, DNN models, and
datasets.

\section{Communication-Efficient ML Training Algorithms} 

A large body of prior work proposes ML training algorithms to reduce
the dependency on intensive parameter updates to enable more efficient
parallel computation (e.g.,~\cite{DBLP:conf/nips/JaggiSTTKHJ14,
  DBLP:journals/jmlr/ZhangDW13, DBLP:conf/nips/ZinkevichWSL10,
  DBLP:conf/icml/TakacBRS13, DBLP:conf/uai/NeiswangerWX14,
  DBLP:conf/icml/ShamirS014, DBLP:conf/icml/ZhangL15a}). These work
can be potentially useful in addressing the communication challenge of
highly-distributed data. However, these ML algorithms are not general
and their applicability depends on applications. Besides, they do not
address the statistical challenge of non-IID data partitions as they
assume the data partitions is IID. In contrast, our goal is to propose
generic system-level solutions \khii{that do not require any changes
  to ML algorithms}, and we aim to propose solutions that work on
non-IID data partitions.

\section{Low-Latency ML Serving Systems} 

Some prior work proposes ML serving systems to achieve low-latency
responses (e.g.,~\cite{tensorflow_serving,
  DBLP:conf/wsdm/AgarwalLTXZ14, DBLP:conf/cidr/CrankshawBGLZFG15,
  DBLP:conf/nsdi/CrankshawWZFGS17,
  DBLP:journals/pvldb/KangEABZ17}). Among them, some
works~\cite{DBLP:conf/wsdm/AgarwalLTXZ14,
  DBLP:conf/cidr/CrankshawBGLZFG15} focus on linear ML models that are
fast but often are less accurate than the computationally intensive
models such as deep neural networks (DNNs). The ones that provide
low-latency serving with DNNs~\cite{tensorflow_serving,
  DBLP:conf/nsdi/CrankshawWZFGS17} mostly focus on serving smaller
querying data such as user preferences and images and they do not
address the computation challenge of massive data such as
videos. Recent work~\cite{DBLP:journals/pvldb/KangEABZ17} that
addresses the latency of serving large-scale data like videos provides
significantly improvement in latency, but the latencies are still slow
(e.g., {\em 5 hours} to query a month-long video on a GPU,
Chapter~\ref{ch:focus}). Hence, a lot more efforts are needed to
address the computation challenge of serving massive data in ML
applications.

\chapter{ML Serving over Large, Rapidly-Growing Datasets: A Case Study of Video Queries}
\label{ch:focus}

Cameras are ubiquitous, with millions of them deployed by public and private entities at traffic intersections, enterprise offices, and retail stores. Videos from these cameras are continuously recorded \cite{genetec, avigilon}, with the main purpose of answering ``after-the-fact'' queries \kff{such as}: {\em identify video frames with objects of certain classes (like cars or bags)} from many days of recorded video. Because the results from these video analytics queries \kff{may be needed quickly in many use cases}, achieving low latency is crucial. % across many cameras.

%To be useful, the system needs to answer

%Storing the videos is important for regulatory as well as competitive reasons.
%However, the main reason for storing the videos is to enable answering ``after-the-fact'' questions.
%Examples include forensic and investigative analytics in the wake of a terrorist attack, traffic violations, and movement patterns of customers in grocery stores.

%Computer vision techniques have improved vastly.

%Advances in convolutional neural networks (\cnns) backed by copious training data and hardware accelerators (e.g., GPUs~\cite{k80}) have led to high accuracy in the computer vision tasks like object detection and classification. For instance, the ResNet$152$ object classifier \cnn~\cite{DBLP:conf/cvpr/HeZRS16} won the ImageNet challenge that evaluates classification correctness on $1,000$ classes using a public image dataset with labeled ground truths \cite{ILSVRC15}. % with a \ga{XX} accuracy.
%For each image, these classifiers return a ranked list of $1,000$ classes in decreasing order of confidence.

Advances in convolutional neural networks (\cnns) backed by copious training data and hardware accelerators (e.g., GPUs~\cite{P100}) have led to highly accurate results in tasks like object detection and classification \kff{of} images. \kv{For instance, the ResNet$152$ classifier \cnn~\cite{DBLP:conf/cvpr/HeZRS16}, winner of the ImageNet challenge 2015~\cite{ILSVRC15}, surpasses human-level performance in classifying $1,000$ object classes on a public image dataset that has labeled ground truths~\cite{DBLP:conf/iccv/HeZRS15}.}% \ga{Aren't the labeled ground truths by humans too?}

Despite the accuracy of image classifier \cnns (like ResNet152) and object detectors (like YOLOv2~\cite{DBLP:journals/corr/RedmonF16}), using them for video analytics queries is both expensive and slow. For example, %even {\em after} down-sampling the video streams from 30 frames per second (fps) to 1 fps,
even {\em after} using various motion detection techniques to filter out frames with no moving objects, %using the YOLOv2 \cnn %ResNet$152$ classifier 
%on the remaining frames at {\em query-time} 
using an object detector such as YOLOv2~\cite{DBLP:journals/corr/RedmonF16} to identify frames with a given class (e.g., ambulance) on a month-long traffic video requires $\approx{}190$ hours on a high-end GPU (NVIDIA P100~\cite{P100}) and costs over $\$380$ in the Azure cloud \kff{(\sys{Standard\_NC6s\_v2} instances)}. %$280$ GPU hours and costs $\$250$ in the Azure cloud 
%\sv{do we have YOLO numbers here}
To achieve a query latency of say one minute on $190$ %$280$ 
GPU hours of work would require tens of thousands of GPUs detecting objects in the video frames in parallel, which is \kff{two to three} orders of magnitude more than what is typically provisioned (few tens or hundreds \kff{of GPUs}) by traffic jurisdictions or retail stores. %This leads to high query latencies.
% While these queries can be parallelized across frames, achieving a latency of a few seconds for the above query would require a million GPUs which is many orders of magnitude more than what is typically provisioned by traffic jurisdictions or retail stores.
%%Note that the above cost and latency values are {\em after} using motion detection techniques to exclude frames with no moving objects.
Recent work like \sys{NoScope}~\cite{DBLP:journals/pvldb/KangEABZ17} has significantly improved the filtering of frames by using techniques \kff{like} lightweight binary classifiers for the queried class (e.g., ambulance) before running heavy \cnns.
However, the latencies are still \kff{long, e.g., it takes {\em 5 hours}} to query a month-long video on a GPU, in our evaluations.
Moreover, \kff{videos from many cameras often need to be queried, which increases the latency and the GPU requirements even more.}
%\ko{it still needs to go through \emph{all} the video frames at query time. As a result, the query latencies are still slow (average 30 minutes to query on a month-long video with a 10-GPU cluster). }
%much higher than what is suitable for interactive use 
%(average 5 minutes on an 12-hour video in our experiments). %\sv{Can we put some concrete number here like takes 1hr with 10 GPU cluster etc.} %Further, the binary classifiers in approaches like \sys{NoScope} need to be retrained for each class that is queried.
% While computer vision techniques have improved significantly in the past years, applying them to large volumes of video is very expensive. One way to execute user queries is to perform all video analytics at \emph{query-time}, i.e., only after the user has issued the query. However, this approach would have very high resource demand at query-time and would provide very slow responses. For example, while processing one month of video using ResNet$152$ is embarrassingly parallel, achieving one second latency would require approximately one million GPUs which is several orders of magnitude more than these organizations typically provision for.

%We believe that enabling {\em low-latency and low-cost querying over large video datasets} will make video analytics more useful and open up many new opportunities.
The objective of our work is to enable {\em low-latency and low-cost querying over large, continuously-growing video datasets}. 

%%%A natural approach to enabling low latency querying is doing all classifications with ResNet152 at \emph{ingest-time}, i.e., on the \emph{live} videos, and storing the results in an index of object classes to video frames. Any queries for specific classes (e.g., cars) will thus involve only a simple index lookup at {\em query-time}. There are, however, at least two problems with this approach. First, the cost to index all the video (using ResNet152) at ingest-time, e.g., \$250/month/stream in the above example, is prohibitively high. Second, most of this ingest-time cost is wasteful because typically only a small fraction of recorded videos get queried \cite{video-trends}. Following a theft, the police would query a few days of video from a handful of surveillance cameras, but not all the videos.
A natural approach to enable low latency queries is doing most of the work at \emph{ingest-time}, i.e., on the \emph{live} video \kff{that is} being captured. If object detection, using say YOLO, were performed on frames at ingest-time, queries for specific classes (e.g., ambulance) \kff{would} involve only a simple \emph{index} lookup \kff{to find video frames with the queried object class}. There are, however, two main shortcomings with this approach. First, most of the ingest-time work may be wasteful because typically only a small fraction of recorded frames ever get queried \cite{video-trends}, e.g., only after an incident that needs \kff{investigation}.
%For example, following an incident, the investigators would only query a few days of video from a handful of surveillance cameras.
Second, filtering techniques that use binary classifiers (as in \sys{NoScope}~\cite{DBLP:journals/pvldb/KangEABZ17}) are ineffective at ingest-time because {\em any} of a number of object classes could be queried later and running even lightweight binary classifiers for many classes \kff{can be} prohibitively expensive. %\ga{True?}

{\noindent{\bf Objectives \& Techniques.} We present {\focus}, a system to support low-latency, low-cost queries on large video datasets.
To address the above challenges and shortcomings, {\focus} has the following goals:
$(a)$ provide \kff{low-cost} indexing of {\em multiple} object classes in the video at ingest-time,
$(b)$ achieve high accuracy and low latency for queries, and
$(c)$ enable trade-offs between the cost at ingest-time and the latency at query-time.
{\focus} takes as inputs from the user a \emph{ground-truth CNN} (or ``GT-CNN'', e.g., YOLO) and the desired accuracy of results that {\focus} needs to achieve relative to the GT-CNN.
With these inputs, {\focus} uses three key techniques to achieve the above goals: $(1)$~an {\em approximate} indexing scheme at ingest-time using cheap \cnns, $(2)$~redundancy elimination by {\em clustering} similar objects, and $(3)$~a tunable mechanism for judiciously {\em trading off} ingest cost and query latency.

%%{\focus} uses four key techniques -- cheap \cnns for ingest, using top-{\sf K} results from these ingest-time \cnns as an approximate index, clustering similar objects, and judicious selection of system and model parameters.
%balances the cost of video analytics between ingest- and query-time (to reduce potential wasted work when video is queried rarely).
%We describe our core techniques below.

%$(1)$ low cost of indexing video at ingest-time,
%$(2)$ provide results to queries with high accuracy and low latency, and
%$(3)$ balance the potential wasted work at ingest-time against the latency at query-time.

%We seek to resolve this dichotomy between query time and ingest time processing, i.e., reduce the cost of video analysis at ingest time, while still enabling low-latency queries on large datasets.
%balance the  {\em potential wasted work of video pre-processing (if the videos are not queried)} against the  {\em processing latency at query time}.

%We seek to resolve this dichotomy between query time and ingest time processing, i.e., balance the  {\em potential wasted work of video pre-processing (if the videos are not queried)} against the  {\em processing latency at query time}.

%\pb{START alternate version}

{\bf (1)} {\em Approximate indexing using a cheap ingest \cnn.}
To make video ingestion cheap, {\focus} uses \emph{compressed} and \emph{specialized} versions of the GT-\cnn that have fewer convolutional layers \cite{Simonyan15}, use smaller image sizes, and are trained to \kff{recognize} the classes specific to each video stream. 
%First, to make video ingestion cheap, {\focus} uses \emph{compressed} and \emph{specialized} versions of \cnns, to create an ingest-time index of object classes to frames. CNN compression (e.g.,~\cite{Simonyan15}) creates new CNNs with fewer convolutional layers and smaller input images. Specialization~\cite{mcdnn, DBLP:journals/corr/ShenHPK17} trains those CNNs on a smaller set of object classes specific to each video stream so that those cheaper CNNs can classify these video-specific objects more accurately. Together, these techniques result in highly efficient CNNs for video indexing.
%reduce the cost of video ingestion by using cheaper compressed and specialized versions of \cnns to create an ingest-time index of object classes to video frames. CNN {\em compression} uses fewer convolutional layers and smaller image resolutions to reduce the cost of executing the CNN. For example, the ResNet18 classifier has only $18$ layers in contrast to the ResNet$152$ model's $152$ layers.
%CNN {\em specialization} relies on the observation that a specific video camera contains only a small number of object classes and their appearance is much more constrained.
%Targeting the \cnns to this constrained set that is specific to each camera allows us to train efficient versions of even smaller models like ResNet18. %Together, these techniques achieve \ga{XX} reduction in GPU cycles.
The cheap ingest \cnns, however, are less accurate than the expensive GT-\cnn, both in terms of {\em recall} and {\em precision}. We define recall as the fraction of frames in the video that contain objects of the queried class that were {\em actually} returned in the query's results. Precision, on the other hand, is the fraction of frames in the query's results that contain objects of the queried class.
%$%, reducing both {\em recall} and {\em precision} of the results.
%$%To increase recall -- not missing objects of the queried class -- we index each object with the ``top-{\sf K}'' classification results of the cheap CNN, instead of just the top-most result.
%$%This way, even if the cheap \cnn's top-most result doesn't match the expensive \cnn, we will likely capture the correct object class in its top-{\sf K} results.

Using a cheap \cnn to filter frames \kff{upfront} risks incorrectly eliminating frames.
To overcome this potential loss in recall, {\focus} relies on an empirical observation: while the top (i.e., most confident) classification results of the cheap \cnns and expensive GT-\cnn often
do not match, the top result of the expensive \cnn often falls within the {\em top-{\sf K}} most confident results of the cheap \cnn. Therefore, at ingest-time, {\focus} indexes each frame with the ``top-{\sf K}'' results of the cheap CNN, instead of just the top result.
%To increase precision -- not returning objects of classes different from the queried class -- at query-time, we classify matching objects out of the top-{\sf K} ingest index with the expensive \cnn.
To increase precision, at query-time, after filtering frames using the top-{\sf K} index,
we apply the GT-\cnn and return only frames \kff{that actually contains the queried object class}.

%\sv{Fix frame vs. object here}
{\bf (2)} {\em Redundancy elimination via clustering.} To reduce the query-time latency of using the expensive GT-\cnn, {\focus} relies on the significant similarity between objects in videos.
%, both near and far-apart in time.
For example, a car \kff{moving across a camera} will look very similar in consecutive frames.
%Also, different cars of the same model and color will look similar even across days.
{\focus} leverages this similarity by clustering the objects at ingest-time. 
%(using image feature vectors)
\kff{We} classify {\em only} the cluster centroids with the GT-\cnn at query-time, and assign the same class to all objects in the cluster. This considerably reduces query latency. Clustering, in fact, identifies redundant objects even across non-contiguous and temporally-distant frames. 

%clusters similar objects from the live video, classifies the cluster centroid objects with a cheap \cnn, and indexes each centroid with the top {\sf K} classification results from the ingest \cnn (based on classification confidence). At query-time, when objects for a class X is queried, {\focus} looks up the ingest index for centroids that match class X and classifies the centroids with an expensive \cnn. For centroids whose classification from the expensive \cnn equals class X, it returns all the objects from the corresponding clusters.

%\ga{TODO} Finally, using the cheap \cnn at ingest-time and expensive \cnn at query-time allows us to achieve different trade offs between ingest- and query-time costs. The cheaper the ingest \cnns, the lower their accuracy. To maintain high accuracy of queries, we need to counteract the ingest-time inaccuracy by running more object cluster centroids through the expensive \cnn at query-time. Among the different architectures of \cnns such as ResNet \cite{DBLP:conf/cvpr/HeZRS16}, AlexNet \cite{DBLP:conf/nips/KrizhevskySH12}, and VGG \cite{DBLP:journals/corr/SimonyanZ14a}, we carefully select the ingest \cnn along with the clustering and specialization parameters to balance between the costs at ingest and query -- sharply improve one of the two costs for a small worsening of the other.%, and not waste too much ingest cost when queries are rare.

{\bf (3)} {\em \kff{Trading} off ingest cost vs. query latency.} {\focus} \kff{intelligently} chooses its parameters (including {\sf K} and the cheap ingest-time \cnn) to meet user-specified targets on precision and recall. 
% chooses its parameters to meet user-specified accuracy.
%It considers the different ways to compress the GT-CNN along with the specialization and clustering parameters, and only considers combinations that meet the target precision and recall.
Among the parameter choices that meet the accuracy targets, it allows the user to trade off between ingest cost and query latency. For example, using a cheaper ingest \cnn reduces the ingest cost but increases the query latency as {\focus} needs to use a larger {\sf K} for the top-{\sf K} index to achieve the accuracy targets. {\focus} automatically identifies the ``sweet spot'' in parameters, which sharply improves one of ingest cost or query latency for a small worsening of the other. It also allows for policies to balance the two, depending on the fraction of videos the application expects to get queried.

In summary, {\focus}' ingest-time and query-time operations are as follows. At ingest-time, \kff{{\focus}} classifies the detected objects using a cheap \cnn, clusters similar objects, and indexes each cluster centroid using the top-{\sf K} most confident classification results,
 where {\sf K} is auto-selected based on the user-specified precision, recall, and cost/latency trade-off point.
 At query-time, {\focus} looks up the ingest index for \kff{cluster} centroids that match the class X requested by the user and classifies them using the GT-CNN. \kff{Finally, {\focus} returns all objects from the clusters that are classified as class X to the user.}

\smallskip\noindent{\bf Evaluation Highlights.} We build {\focus} and evaluate it on fourteen 12-hour videos from three domains -- traffic cameras, surveillance cameras, and news. We compare against two baselines: \sys{``Ingest-heavy''}, which uses the heavy GT-CNN for \kff{ingest}, and \sys{``NoScope''}, \kff{a recent} state-of-the-art video querying system~\cite{DBLP:journals/pvldb/KangEABZ17}. % on all the video frames at query time. %
%For the frames with no moving objects, we skip any further processing and assign them the result from their previous frame. 
%, which is one of the core techniques in recent prior work~\cite{DBLP:journals/pvldb/KangEABZ17}.
%%%%%%%%%%
%Figure~\ref{fig:result_trade_off} shows a representative result for a surveillance video on a university street, with YOLOv2~\cite{DBLP:journals/corr/RedmonF16} as \kff{the} GT-\cnn. %the \video{auburn\_c} traffic video.
%By zooming in, we see that
%{\focus} (the {\focus-Balance} point) is simultaneously 86$\times$ cheaper
%than the \sys{Ingest-all} baseline in its GPU consumption {\em and}
%56$\times$ faster than the \sys{Query-all} baseline in query
%latency, all the while achieving at least 95\% precision and recall.
%(Also shown are two alternatives provided by {\focus} that offer slightly different trade-offs.)
\kff{We use YOLOv2~\cite{DBLP:journals/corr/RedmonF16} as the GT-\cnn.}
\kff{On average, across all the videos}, {\focus} is $48\times$ (up to 92$\times$) cheaper
than \sys{Ingest-heavy} and $125\times$ (up to 607$\times$) faster than
\sys{NoScope}, all the while achieving $\geq$ 99\% precision and recall. 
In other words, the latency to query a month-long video drops from $5$ hours to only $2.4$ minutes, at an ingest cost of $\$8$/month/stream. \kff{Figure~\ref{fig:result_trade_off} also shows representative results with different trade-off alternatives for a surveillance video.}
%%In other words, the ingest cost/month/stream comes down from \$380 to \$8, and the latency to query a month-long video drops from $5$ hours to $2$ minutes.} %\ga{Aren't our videos 12 hours each?}

 \begin{figure}[h]
   \vspace{0.05in}
  \centering
  \includegraphics[width=0.88\textwidth]{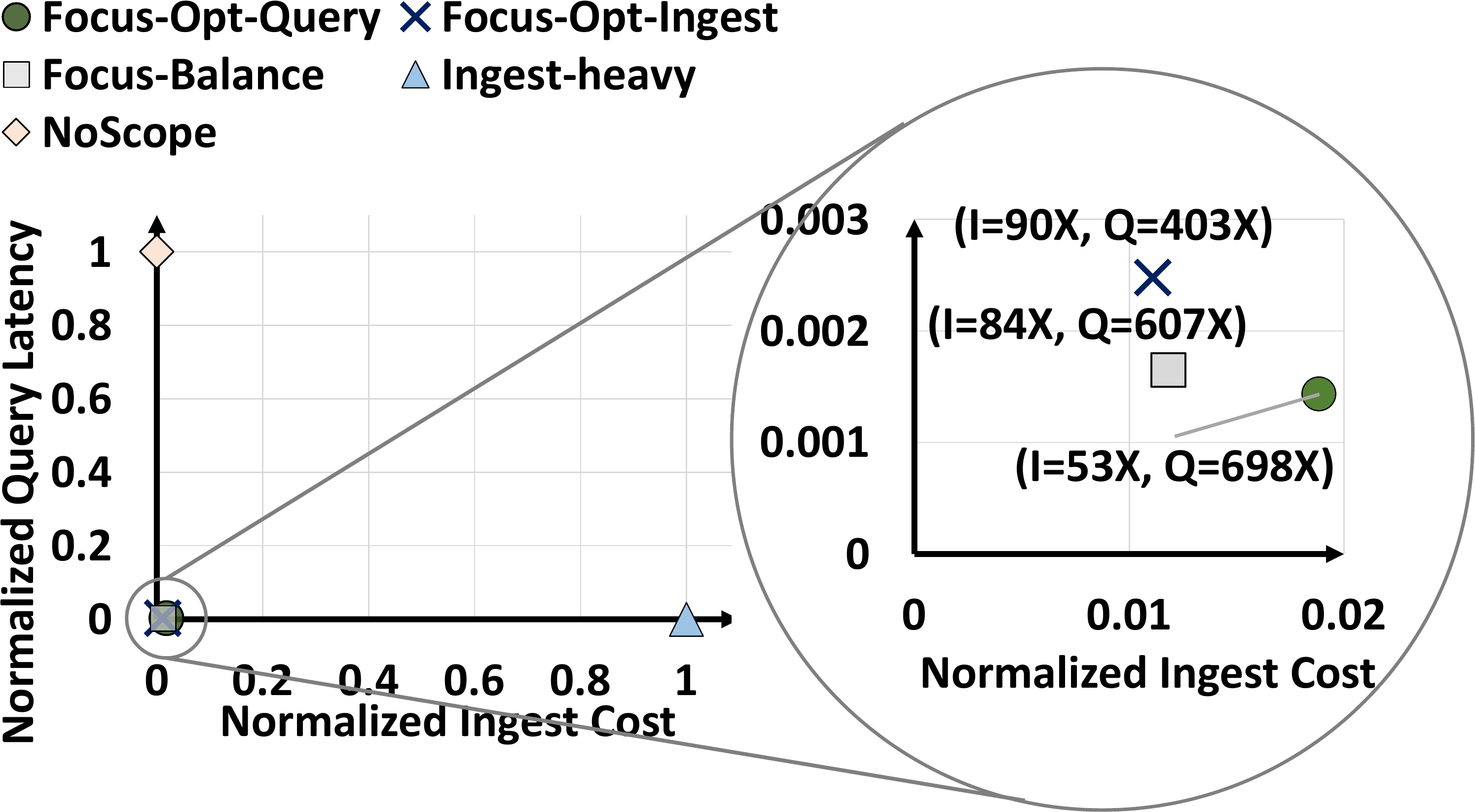}
  \caption[Effectiveness of {\focus} at reducing both ingest cost and
    query latency, for an example surveillance video]{Effectiveness of
    {\focus} at reducing both ingest cost and
    query latency, for an example surveillance video.  We compare
    against two baselines: \sys{``Ingest-heavy''} that uses the
    YOLOv2~\cite{DBLP:journals/corr/RedmonF16} object detector \cnn
    for ingestion, and \sys{``NoScope''}, the state-of-the-art video
    querying system~\cite{DBLP:journals/pvldb/KangEABZ17}. On the
    left, we see that {\focus} (the {\focus-Balance} point) is
    simultaneously 84$\times$ cheaper than \sys{Ingest-heavy} in its
    cost \kff{(the $I$ value)} {\em and} 607$\times$ faster than
    \sys{NoScope} in query latency \kff{(the $Q$ value)}, all the
    while achieving at least 99\% precision and recall \kff{(not
    plotted)}. Zooming in, also shown are two alternative \kff{{\focus}
      designs} offering different trade-offs, {\focus-Opt-Query} and
    {\focus-Opt-Ingest}, each with at least 99\% precision and recall.
  }
  \label{fig:result_trade_off}
\end{figure}

%\kv{In addition, we also compare against a recent prior work, \sys{NoScope}~\cite{DBLP:journals/pvldb/KangEABZ17}, across various recall/precision targets (95-99\%). {\focus} is both faster (1.8 to 34.9$\times$) and cheaper (1.4 to 33.5$\times$) than \sys{NoScope} for queries even on the most frequent object classes. However, {\em arguably}, a system for querying videos is more useful for {\em rarer} classes: querying for ``ambulance'' in a traffic video is more interesting than querying for ``car''. When querying such less frequent object classes, {\focus} is $12-200\times$ faster than \sys{NoScope}.} %See \xref{sec:evaluation} for the full details.
%%%%%%%%%%

%%%% Text from before placing figure in the Intro %%%%%%%%%%
%On average, {\focus} is $58\times$ (up to 98$\times$) cheaper
%than the \sys{Ingest-all} baseline in its GPU consumption and
%$37\times$ (up to 57$\times$) faster than the \sys{Query-all} baseline
%in query latency, all the while achieving a precision and recall both of
%$\geq 95\%$.
%%%%%%%%%%%%%%

%\ga{TODO} We make the following contributions.
%\begin{enumerate}
%\item Formulate the problem of interactive querying of video data via balancing ingest time and query time load.
%\item Techniques for cheap ingestion of videos using compressed and video-specific specialization of \cnns.
%\item Leverage the similarity of objects in a video to cluster them using \cnn features for low latency querying.
%\item End-to-end system for low latency querying of large video datasets.
%\end{enumerate}

\section{Characterizing Real-world Videos}
\label{subsec:char}

We aim to support queries of the form: {\em find all frames in the video that contain objects of class X}. We identify some key characteristics of real-world videos towards supporting these queries: $(i)$ large portions of videos can be excluded (\xref{subsubsec:filtering}), \kff{$(ii)$ only a limited set of object classes occur in each video (\xref{subsubsec:limited}), and
$(iii)$ objects of the same class have similar feature vectors (\xref{subsubsec:features}).}
The design of {\focus} is based on these characteristics.

\kff{We analyze six 12-hour videos from three domains: traffic cameras, surveillance cameras, and news channels (\xref{sec:methdology} provides the details.)}
%We analyzed 12 hours of video from each of six video streams from
%each and these streams span
%traffic cameras, %(\texttt{Auburn (Alabama), Jackson Hole (Wyoming)})
%surveillance cameras, %(\texttt{Lausanne (Switzerland), Sittard (Netherlands)}), 
%and news channels. %(\texttt{CNN, MSNBC}). 
In this chapter, we use results from YOLOv2 \cite{DBLP:journals/corr/RedmonF16}, trained to classify 80 object classes based on the COCO \cite{DBLP:conf/eccv/LinMBHPRDZ14} dataset, as the ground truth.% We detect the objects in each frame of these videos (using background subtraction~\cite{DBLP:conf/avbs/KaewTraKulPongB15}), and classify each object with the ResNet152 \cnn \cite{DBLP:conf/cvpr/HeZRS16} among the supported $1,000$ object classes. In this paper, we use results from the costly ResNet152 \cnn as ground truth.

\subsection{Excluding large portions of videos}
\label{subsubsec:filtering}

We find considerable potential to avoid processing large portions of videos at query-time. Not all the frames in a video are relevant to a query because each query looks only for a {\em specific class} of objects. %Figure~\ref{fig:frame_fraction_per_query} shows the distribution of the fraction of frames that contain each class of objects.
In our video sets, an object class occurs in only $0.16$\% of the frames on average, and even the most frequent object classes occur in no more than $26\%-78\%$ of the frames. This is because while there are usually some dominant classes (e.g., cars in a traffic camera, people in a news channel), most other classes are rare. 
Overall, the above data suggests considerable potential to speed up query latencies by {\em indexing} frames \kff{using} the object classes. 
Also, in our experience, a system for querying videos is more useful for less frequent classes: querying for ``ambulance'' in a traffic video is more interesting than querying for something commonplace like ``car''.
%\sv{I removed the line about rare classes because in eval we exclude them.}

%To understand the correlation between eventful frames and time, we plot the percentage of objects each hour has in Figure~\ref{fig:obj_over_time}. As we see, most objects show up in certain time in some traffic and surveillance videos (e.g., the first half of \texttt{Auburn} and \texttt{Jackson Hole}, and the second half of \texttt{Lausanne}). This is because in these videos, there are very few objects at night times. However, such characteristic does not hold for surveillance videos on busy streets (e.g., \texttt{Church Street}) or news channels (e.g., \texttt{CNN} and \texttt{MSNBC}), which suggests we \emph{cannot} apply some simple rules on all videos.

%\begin{figure}[t!]
%  \centering
%  \includegraphics[width=0.48\textwidth]{focus/Figures/frame_fraction_per_query.pdf}
%  \caption{The fraction of frames matches a query. \ga{Include 25th and 75th percentiles. If there is no difference among the videos, we may not need a graph?}}
%  \label{fig:frame_fraction_per_query}
%\end{figure}
\subsection{Limited set of object classes in each video}
\label{subsubsec:limited}

%We next focus on the classes of objects that occur in each of the videos and the disparity in frequency among them.

Most video streams have a limited set of objects because each video has its own context (e.g., traffic cameras can have automobiles, pedestrians or bikes, but not airplanes). %It is rare that a video stream contains objects of {\em all} the classes recognized by state-of-the-art \cnns.

\begin{figure}[h]
  \centering
  \includegraphics[width=0.7\textwidth]{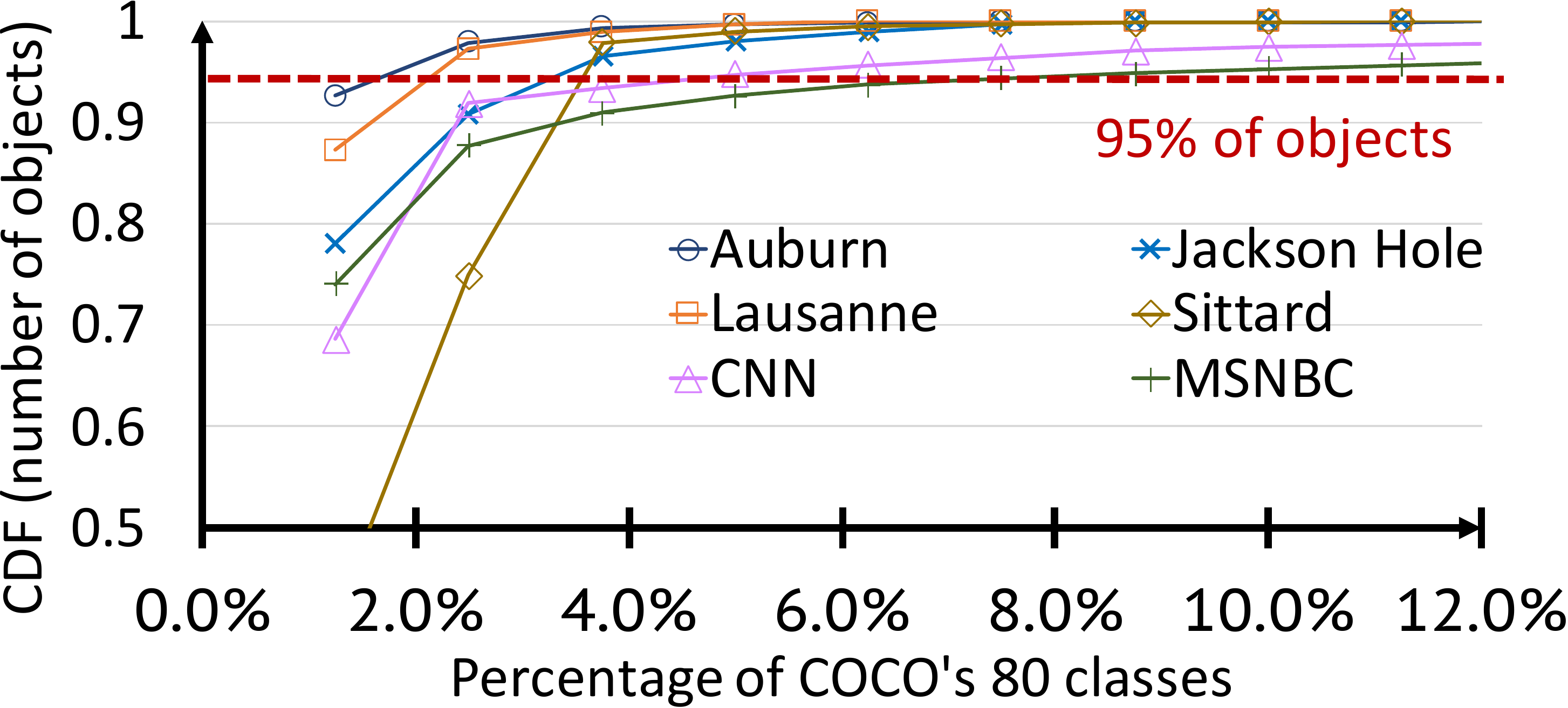}
  \caption[CDF of frequency of object classes]{CDF of frequency of object classes. The x-axis is the fraction of classes out of the $80$ classes recognized by the COCO~\cite{DBLP:conf/eccv/LinMBHPRDZ14} dataset (truncated to $12\%$). \ignore{\ga{UPDATE - YOLO.}}}
  \label{fig:class_num_cdf}
\end{figure}

Figure~\ref{fig:class_num_cdf} shows the cumulative distribution function (CDF) of the frequency of object classes in our videos (as classified by YOLOv2). We make two observations.
% Two points: One is 
First, $2\%-10\%$ of the most frequent object classes cover $\geq 95\%$ of the objects in all
video streams. In fact, for some videos like \texttt{Auburn} and \texttt{Jackson Hole} we find that only $11\%-19\%$ object classes
occur in the entire video. Thus, for each video stream, if we can \textit{automatically} determine its most frequent object 
classes, we can train efficient \cnns \emph{specialized} for these classes. % (\xref{subsec:cnn}).
Second, \kff{a closer analysis reveals that there is little overlap between the object classes} among different videos.
On average, the Jaccard index~\cite{jaccard} (i.e., intersection over union) 
between the videos based on their object classes is only $0.46$. This implies that we need to specialize \cnns
for each video stream separately to achieve the most benefits.

%\ignore{\ga{Update the numbers with YOLO.}} First, objects of only $11\%-19\%$ (not graphed) of the $80$ object
%classes occur in the less busy videos (\texttt{Auburn},
%\texttt{Jackson Hole}, \texttt{Lausanne}, and \texttt{Sittard}). Even
%in the busier videos (\texttt{CNN}, and \texttt{MSNBC}), objects of
%only $75\%$ of the classes appear. Also, there is little overlap
%between the classes of objects among the different videos. 
%On average,
%the Jaccard indexes~\cite{jaccard} (i.e., intersection over union)
%between the videos based on their object classes is only
%$0.46$. Second, even among the object classes that do occur, a small
%fraction of classes disproportionately
%dominate. 
%Figure~\ref{fig:class_num_cdf} shows that $2\%-10\%$ of the
%most frequent object classes cover $\geq 95\%$ of the objects in each
%video stream.  Thus, for each video stream, we can
%\textit{automatically} determine its most frequent object 
%classes and train efficient \cnns \emph{specialized} for these classes.% (\xref{subsec:cnn}).

%Thus, we can {\em specialize} the \cnns to only the classes that occur in {\em each} video, and further focus on only the {\em dominantly occurring} classes, while still allowing for the other classes to be queried.

%   It means \emph{using the state-of-the-art \cnns to detect all the classes may not be necessary if we can treat the dominant classes differently from the rare classes}.
% In another word, \emph{there are many object classes never show up in each video stream, but these classes vary among video streams}.

\subsection{Feature vectors for finding duplicate objects}
\label{subsubsec:features}

Objects moving in the video often stay in the frame for several seconds; for example, a pedestrian might take 15 seconds to cross a street.
Instead of classifying {\em each instance} of the same object across the frames, we would like to \emph{inexpensively} find duplicate objects and only classify one of them using a \cnn (and apply the same label to all duplicates). Thus, given $n$ duplicate objects, \kff{we would like} only one \cnn classification operation instead of $n$.

%We consider two ways to find duplicate objects: 1) using pixel values and 2) feature vectors extracted using a cheap \cnn.
%While pixel values are an obvious choice, they are very sensitive to even small changes in the object.
Comparing pixel values across frames is an obvious technique to identify duplicate objects, however, this technique turns out to be highly sensitive to even small changes in the camera's view of an object. Instead,
feature vectors extracted from the \cnns (\xref{subsec:cnn}) are more robust because they are specifically trained to extract visual features for classification.
%In our videos, we observe that the average distance between the feature vectors of the same class from ResNet18 is \ga{XX} lower than between feature vectors of different object classes, which shows the potential for finding object duplicates.
We verify the robustness of feature vectors using the following analysis. 
In each video, for each object $i$, we find its \emph{nearest} neighbor $j$ using feature vectors from a cheap \cnn (ResNet18) and compute the fraction of object pairs that belong to the same class.
This fraction is over 99\% in each of our videos, which shows the promise of using feature vectors from cheap \cnns to identify duplicate objects {\em even} across frames that are {\em not} temporally contiguous. %\ga{Quantify this?}

\section{Overview of {\focusSec}}
\label{sec:architecture}

%\subsubsection*{Problem Formulation}

%% Query Formulation
The goal of {\focus} is to {\em index live video streams} by the object classes occurring in them and enable answering ``after-the-fact'' queries later on the stored videos of the form: \emph{find all frames that contain objects of class X}. Optionally, the query can be restricted to a subset of cameras and a time range. Such a query formulation is the basis for many widespread applications and could be used either on its own (such as for detecting all cars or bicycles in the video) or used as a basis for further processing (e.g., finding all collisions between cars and bicycles).% In the scenarios of our focus, the specific user queries -- object classes that are queried for -- are not known at ingest-time. \ga{???}
%The above query formulation can be augmented with additional filters like time window of interest (e.g., last 24 hours) that can be implemented using well-studied temporal indexes.

\noindent{\bf System Configuration.} {\focus} is designed to work with a wide variety of current and future \cnn{}s.  The user (system administrator) provides a \emph{ground-truth CNN} (GT-\cnn), which
serves as the accuracy baseline for {\focus}, but is far too costly to run on every video frame.  Through a sequence of techniques, {\focus} provides results of nearly-comparable accuracy but at greatly reduced cost. In this chapter, we use YOLOv2 \cite{DBLP:journals/corr/RedmonF16} % ResNet152 image classifier 
as the default GT-\cnn.

%% Goals - accuracy
%\noindent{\bf Accuracy target:} {\focus} is designed to produce results for queries with an accuracy comparable to a sophisticated and accurate \cnn (e.g., the ResNet152 image classifier) on all the video frames. % that won the ImageNet competition.
%\noindent{\bf Accuracy target:}
\kv{Because different applications require different accuracies, {\focus} permits the user to specify the accuracy target, while providing reasonable defaults.}
The accuracy target is specified in terms of \emph{precision}, i.e., fraction of frames output by the query that actually contain an object of class X according to GT-\cnn, and \emph{recall}, i.e., fraction of frames that contain objects of class X according to GT-\cnn that were actually returned by the query.

\begin{figure*}[h!]
\centering
\includegraphics[width=0.9\textwidth]{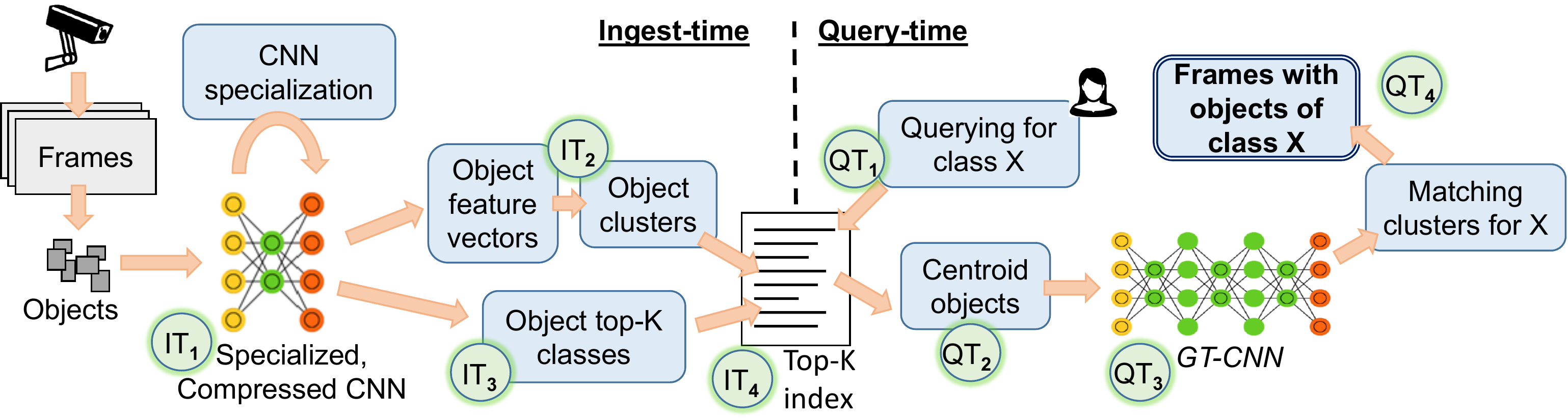}
\vspace{0.1in}
\caption{Overview of {\focus}.}
\vspace{0.1in}
\label{fig:arch}
\end{figure*}

%% High-level description
\noindent{\bf Architecture:} Figure~\ref{fig:arch} overviews the {\focus} design. \kt{{\focus} is a system that runs on centralized servers (such as data centers) where live video streams are continuously recorded and transmitted from cameras.} 
\begin{itemize}
\item{
At \emph{ingest-time} (left part of Figure~\ref{fig:arch}), {\focus} classifies objects in the incoming video frames and extracts their feature vectors.
For its \kff{ingest, {\focus}} uses highly compressed and specialized alternatives of the GT-\cnn model (IT$_\text{1}$ in Figure~\ref{fig:arch}). % to each video feed.
{\focus} then clusters objects based on their feature vectors (IT$_\text{2}$) and assigns to each cluster the \emph{top {\sf K}} most likely classes these objects belong to (based on classification confidence of the ingest \cnn) (IT$_\text{3}$). %; using top-K instead of just the top-most reduces false negatives.
It creates a \emph{top-{\sf K} index}, which maps each class to the set of object clusters (IT$_\text{4}$). The top-{\sf K} index is the output of {\focus}' ingest-time processing of videos.}% that likely contain objects of that class; each cluster then maps to the set of its objects and corresponding frames.
\item{
  At {\em query-time} (right part of Figure~\ref{fig:arch}), when the user queries for a certain class $X$ (QT$_\text{1}$), {\focus} retrieves the matching clusters from the top-{\sf K} index (QT$_\text{2}$), runs the \emph{centroids} of the clusters through GT-\cnn (QT$_\text{3}$), and returns all frames from the clusters whose centroids were classified by GT-\cnn as class $X$ (QT$_\text{4}$).} %Running the objects through the GT-\cnn removes the false positives contained in the top-K index. %\vspace{-.1in}
  
\end{itemize}

%For each cluster, we combine the top-K classes of each included object into a ranked list of top-K most likely classes for the cluster as a whole. This information is then stored in our \emph{top-K index}; this is \emph{logically} a table with the following columns: cluster ID, video stream ID, start and end time of included frames, encoded image of centroid object, and ranked list of top-K most likely classes.

\noindent{The top-{\sf K} ingest index is a mapping between the object classes and the clusters. \kff{In particular, we create a mapping from each object class to the clusters with top {\sf K} matching object classes. Separately, we store the mapping between clusters and their corresponding objects and frames. The structure of the index is:}

%\noindent{
\noindent\fbox{%
    \parbox{\columnwidth}{%
\texttt{object class $\rightarrow$ $\langle$cluster ID$\rangle$}
%} %and

%\noindent{\texttt{cluster ID $\rightarrow$ \[centroid object, cluster $\langle \text{objects}\rangle$, $\langle \text{frame IDs}\rangle$ of objects\]}}}
%\noindent{
\texttt{cluster ID $\rightarrow$ [centroid object, $\langle \texttt{objects}\rangle$ in cluster, $\langle \texttt{frame IDs}\rangle$ of objects]}
%}
    }%
}

We next explain how {\focus}' key techniques keep ingest cost and query latency low while also meeting the user-specified recall and precision targets. %We present the specifics of the techniques in \xref{sec:techniques}.

\noindent{\bf 1) Top-{\sf K} index via cheap ingesting:} {\focus} makes indexing at ingest-time cheap by  using compressed and specialized alternatives of the GT-\cnn for {\em each} video stream. {\em Compression} of \cnns~\cite{Simonyan15, pruning1,pruning2,lowrank}
uses fewer convolutional layers and other approximations (\xref{subsec:cnn}), while {\em specialization} of \cnns~\cite{mcdnn, DBLP:journals/corr/ShenHPK17} uses the observation that a specific video stream contains only a small number of object classes and their appearance is more constrained than in a generic video (\xref{subsubsec:limited}). Both optimizations are done automatically by {\focus} and together result in ingest-time \cnns that are up to $96\times$ cheaper than the GT-CNN.

%% top-K
%\noindent{\textbf{Achieving high accuracy:}}
%\noindent{\bf 2) Top-{\sf K} ingest index:} 
The cheap ingest-time \cnns are less accurate, i.e., their top-most results often do not match the top-most classifications of GT-\cnn. %, resulting in false negatives and false positives.
Therefore, to improve recall, % (i.e., not missing out on frames with the queried object class $X$),
{\focus} associates each object with the \emph{top-{\sf K}} classification results of the cheap \cnn, instead of only its top-most result.
Increasing {\sf K} increases recall because the top-most result of GT-\cnn often falls within the ingest-time \cnn's top-{\sf K} results. %, it also increases query time cost.
At query-time, {\focus} uses the GT-\cnn to remove objects in this larger
set that do not match the class, to regain the precision lost by including
the top-{\sf K}.
%At query-time, we need to classify the objects that match the queried class $X$ out of the cheap \cnn's ingest index with the GT-\cnn to ensure $100\%$ precision.% (i.e., avoid outputting frames that do not contain objects of the queried class $X$).

\noindent{\bf 2) Clustering similar objects.} A high value of {\sf K} at ingest-time increases the work \kff{done} at query time, thereby increasing query latency. To reduce this overhead, {\focus} clusters similar objects at ingest-time using feature vectors from the cheap ingest-time \cnn (\xref{subsubsec:features}). In each cluster, at query-time, we run only the cluster centroid through GT-\cnn and apply the classified result from the GT-\cnn to all objects in the cluster. Thus, a tight clustering of objects is crucial for \kff{high} precision and recall.
% introduces false positives in the results.

%We note that visually similar objects will be labeled as the same class by the GT-\cnn.
%Therefore, given a cluster of similar objects, we run only their centroid through GT-\cnn and apply the same label to the whole cluster.
%While we could cluster objects based on their pixel values,
%as explained in \xref{subsec:cnn}, the features of the penultimate layer of the \cnn are a good high-level descriptor of the object.
%As we demonstrate later, the \cnn features create much tighter clusters and significantly reduce the query-time overhead.

%- cluster_id
%- start/end time
%- i
%- class_name
%- centroid_feature_vector

%\pb{remove this para, since it almost completely repeats the second bullet above?}
%When a user queries {\focus} for a specific object class X,  we extract the cluster centroids that potentially match X using the top-K index (including constraints on video streams and time), run the centroid objects through the GT-\cnn model, and return all the frames from clusters that match the specified class X.

\noindent{\bf 3) Trading off ingest vs.~query costs.} {\focus} automatically chooses the ingest \cnn, its {\sf K}, and specialization and clustering parameters to achieve the desired precision and recall targets. These choices also help {\focus} trade off between the work done at ingest-time and query-time.
%\noindent\textbf{Balancing ingest and query cost:}
%Finally, {\focus} can trade work at ingest-time for more latency at query-time and allows the user to adjust this balance.
For instance, to save ingest work, {\focus} can select a cheaper ingest-time \cnn,
and then counteract the resultant loss in recall by using a higher {\sf K} and running the
expensive GT-\cnn on more objects at query time.
%For instance, the cheaper models we use at ingest-time, the less accurate they are; to achieve the precision and recall targets, we counteract it by running the expensive \cnn on more objects at query-time.
{\focus} chooses its parameters so as to offer a sharp improvement in one of the two costs for a small degradation in the other cost.
Because the desired trade-off point is application-dependent,
{\focus} provides users with options:
``ingest-optimized'', ``query-optimized'', and ``balanced'' (the default).
Figure~\ref{fig:result_trade_off}  presents an example
result.

\newcommand{\numClusters}{C}

\section{Video Ingest \& Querying Techniques}
\label{sec:techniques}

We describe the main techniques used in {\focus}: constructing approximate indexes with cheap \cnns at ingest-time (\xref{subsec:cheap_ingest}), specializing the \cnns to the specific videos (\xref{subsec:specialization}), and identifying similar objects and frames to save on redundant \cnn processing (\xref{subsec:redundancy}). % and query-time techniques to execute user queries. Finally, in~\xref{subsec:tuning} we describe how we adjust various parameters in the system to achieve low false positives and negatives while carefully trading off the between ingest- and query-time cost.
\xref{subsec:tuning} describes how {\focus} flexibly trades off ingest cost and query latency.% to achieve high precision and recall (relative to the GT-\cnn, see~\xref{sec:architecture}) and low ingest and query time costs.

%{\focus} takes two inputs: false negative rate (FNR) and false discovery rate (FDR~\cite{confusion_matrix}). False negative rate refers to the fraction of frames that contained the queried object class but was {\em missed by} {\focus}. False discovery rate refers to the fraction of frames in {\focus}'s output that {\em did not contain} the queried object class.\footnote{False discovery rate is a fractional quantification of false positives.} FNR and FDR are measured against the ground truth GT-\cnn's results. \xref{subsec:tuning} discusses the setting of {\focus}'s parameters to meet the FNR and FDR requirements.

%Let $M$ be a object classification model; applying $M$ to a video frame $f$ returns the rank of each of the $n$ classes, i.e., $M(f) = (r_1, \dots, r_n)$. Let $M_G$ be the ``ground-truth'' model, i.e., the expensive detection model that we want to imitate.

\subsection{Approximate Index via Cheap \kff{Ingest}}
\label{subsec:cheap_ingest}

%This means that when querying for video frames containing a specific object $q$, the list of returned frames $P(q)$ would contain frames that do not have object $q$ (according to our ground-truth model $M_q$), \emph{false positives}, and can also potentially miss frames that do have object $q$, \emph{false negatives}. At query time we filter out the false positives by running them through the ground-truth model $M_q$. However, the false negatives that the ingestion misses cannot be recovered. We therefore ensure that we bound the false negative rate to a small constant controlled by the user of the system.

% cheap ingest vs. fast query
{\focus} indexes the live videos at {\em ingest-time} to reduce the {\em query-time} latency. %We perform object detection on each frame, typically an inexpensive operation, and then 
We detect and classify the objects within the frames of the live videos using {\em ingest-time} \cnns that are far cheaper than the ground-truth GT-\cnn. We use these classifications to index objects by class.%While seeking to meet the precision and recall goals, we employ far cheaper alternatives for the ingest \cnns.
%our goal is to \emph{approximately index} video frames with objects that they contain at ingest-time using a cheap detection model.

% using cheap CNNs
\noindent{\bf Cheap ingest-time \cnn.}
As noted earlier, the user provides {\focus} with a GT-CNN.
Optionally, the user can also provide other \cnn architectures to be
used in {\focus}' search for cheap \cnns. Examples include object detector \cnns (which vary in their resource costs and accuracies) like YOLO \cite{DBLP:journals/corr/RedmonF16} and Faster RCNN \cite{DBLP:conf/nips/RenHGS15} that jointly detect the objects in a frame and classify them. Alternatively, objects can be detected separately using relatively inexpensive techniques like background subtraction~\cite{DBLP:conf/cvpr/BrutzerHH11}, which are well-suited for static cameras, \kff{as} in surveillance or traffic installations, and then the detected objects can be classified using object classification \cnn architectures such as 
ResNet \cite{DBLP:conf/cvpr/HeZRS16}, AlexNet \cite{DBLP:conf/nips/KrizhevskySH12} and VGG \cite{Simonyan15}.\footnote{{\focus} is agnostic to whether object detection and classification are done together or separately. In practice, the set of detected object bounding boxes (but not their classifications!) remain largely the same with different ingest \cnns, background subtraction, and the GT-\cnn.}

Starting from these user-provided CNNs, {\focus} applies various levels of compression, such as removing convolutional layers and reducing the input image resolution (\xref{subsec:cnn}).
This results in a large set of \cnn options for \kff{ingest}, \{CheapCNN$_1, \dots,$ CheapCNN$_n$\}, with a wide range of costs and accuracies, out of which {\focus} picks its ingest-time \cnn, CheapCNN$_\text{ingest}$.

\noindent{\bf Top-{\sf K} Ingest Index.} To keep recall high, {\focus} indexes each object using the {\em top {\sf K}} object classes from the output of CheapCNN$_\text{ingest}$, instead of using just the top-most class. Recall from \xref{subsec:cnn} that the output of the \cnn is a list of classes for each object in descending order of confidence. We make the following {\em empirical} observation: the top-most output of the expensive GT-\cnn for an object is often contained within the top-{\sf K} classes output by the cheap \cnn, for a small value of {\sf K}. % relative to the $1,000$ classes recognized by the \cnns).

Figure~\ref{fig:topK} \kff{demonstrates} the above observation by plotting the effect of {\sf K} on recall on one of our video streams from a static camera, \video{lausanne} (see \xref{sec:methdology}). \kf{We explore many cheaper ResNet18~\cite{DBLP:conf/cvpr/HeZRS16} models by removing one layer at a time with various input image sizes. The trend is the same among the \cnns we explore so we present three models for clarity: ResNet18, and ResNet18 with 4 and 6 layers removed; \kff{correspondingly to each model}, the input images were rescaled to 224, 112, and 56 pixels, respectively.}
%All models were retrained on their original training data (COCO \cite{DBLP:conf/eccv/LinMBHPRDZ14} dataset). 
These models were also {\em specialized} to the video stream (more in \xref{subsec:specialization}). We make two observations.

\begin{figure}[h]
  \centering
  \includegraphics[width=0.8\textwidth]{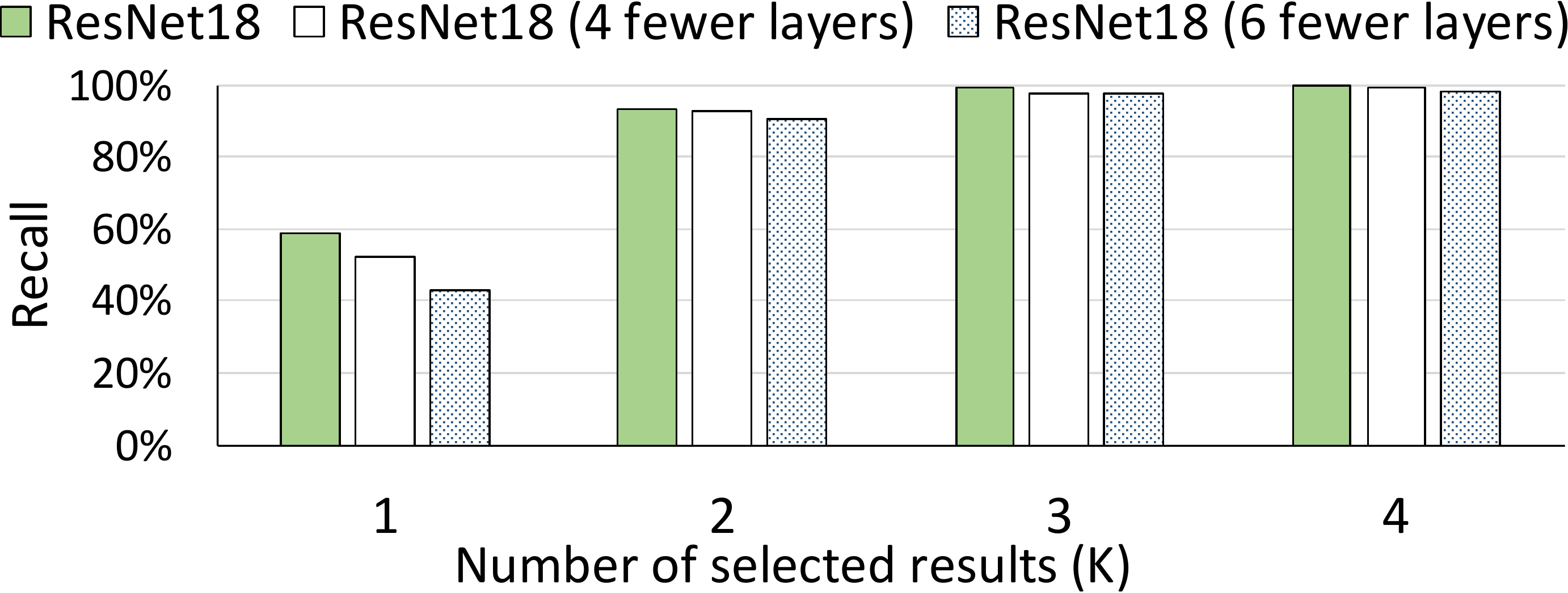}
  \caption[Effect of {\sf K} on the recall of cheap classifier \cnns to classify the detected objects]{Effect of {\sf K} on the recall of three cheap classifier \cnns to classify the detected objects. Recall is measured relative to the results of the GT-\cnn, YOLOv2 \cite{DBLP:journals/corr/RedmonF16}.}%The number within the parenthesis indicates how much cheaper the model is compared to our GT-\cnn, \gaa{YoloV2 \cite{yolo}} (\ga{ResNet compared to Yolo??} \sv{We can also remove the speedup from legend}).}
  \label{fig:topK}
\end{figure}

%Figure~\ref{fig:topK} plots the effect of {\sf K} on recall (for three sample cheap \cnns) on one of our video streams, \video{lausanne} (see \xref{sec:methdology}). As we discuss above, these cheap models were obtained by compressing the classification \cnns such as ResNet18~\cite{DBLP:conf/cvpr/HeZRS16} and AlexNet~\cite{DBLP:conf/nips/KrizhevskySH12}, and re-training them with their original training data (ImageNet~\cite{ILSVRC15}). We make two observations.

First, we observe steady increase in recall with increasing {\sf K}, for all three CheapCNNs.  As the figure shows, 
%CheapCNN$_1$, CheapCNN$_2$, and CheapCNN$_3$
all the cheap \cnns reach $\geq$ 99\% recall when {\sf K} $\geq$ 4. 
%60, {\sf K} $\geq$ 100, and {\sf K} $\geq$ 200, respectively. 
Note that all these models recognize 80 classes, so {\sf K} $=$ 4 represents only 5\% of the possible classes. %and the recall from these models is much higher than randomly select 20\% of the object classes, which is expected to be 20\%.
Second, there is a \emph{trade-off} between different models -- typically, the cheaper they are, the lower their recall with the same {\sf K}. %\ga{Still true. Check for K=3?}
\kf{However, we can compensate for the loss in recall in cheaper models using a larger {\sf K} to reach  \kff{a certain recall value}.}
Overall, we conclude that by selecting the appropriate model and {\sf K}, {\focus} can achieve the target recall.

%in the ingest-time CNN's outputs, we increase the chances of not missing objects of the queried class in our result.%, and it is much better than randomly selecting more results.

% approximation index
\noindent{\bf Achieving precision.} {\focus} creates the {\em top-{\sf K} index} from the top-{\sf K} classes output by CheapCNN$_\text{ingest}$ for every object at ingest-time. %This is an ``approximation'' index because it signifies that the object {\em likely} belongs to the class.
%Note that we do not differentiate between the different classes {\em within} the top-K.
While filtering for objects of the queried class $X$ using the top-{\sf K} index (with the appropriate {\sf K}) will have a high recall, this will lead to very low precision. Because we associate each object with {\sf K} classes (while it has only one true class), the average precision is only $1/{\sf K}$.
Thus, at query time, to improve precision, {\focus} determines the {\em actual} class of objects from the top-{\sf K} index using the expensive GT-\cnn and returns only the objects that match the queried class $X$.

\noindent{\bf Skipping GT-\cnn for high-confidence indexes.} {\focus} records the prediction confidence along with the top-{\sf K} results by CheapCNN$_\text{ingest}$. The system can skip invoking GT-\cnn for the indexes with prediction confidence higher than a chosen threshold (Skip$_\text{th}$). Not invoking GT-\cnn for these indexes can cause precision to fall if the threshold is too low. Hence, this parameter needs to be carefully selected to retain high precision. 

%on objects filtered by the top-K approximation index, we invoke the expensive GT-\cnn (like ResNet152), and return the frames corresponding to the objects that match the queried class.
%associated with the object that match the class queried by the user.

% picking M_i and K
\noindent{\bf Parameter selection.} The selection of the cheap ingest-time \cnn model (CheapCNN$_\text{ingest}$) and the {\sf K} value (for the top-{\sf K} results) has a significant influence on the recall of the output produced. Lower values of {\sf K} reduce recall, i.e., {\focus} will miss frames that contain the queried objects. At the same time, higher values of {\sf K} increase the number of objects to classify with GT-\cnn at query time, and hence adds to the latency. %Finally, the choice of CheapCNN$_i$ directly impacts the cost of ingesting the videos.
\kff{\xref{subsec:tuning} describes} how {\focus} sets these parameters because they have to be jointly set with other parameters \kff{described} in \xref{subsec:specialization} and \xref{subsec:redundancy}.

\subsection{Video-specific Specialization of Ingest \cnn}
\label{subsec:specialization}

%Recall from \xref{subsec:cheap_ingest} that {\focus} uses a cheap ingest-time \cnn, CheapCNN$_i$ to index object classes. 
%{\focus} {\em specializes} the ingest-time \cnn model to each video stream to reduce its cost. Model specialization benefits from two properties of objects. %in each video stream.
\kf{To further reduce the ingest cost, {\focus} {\em specializes} the ingest-time \cnn model to each video stream. As \xref{subsec:cnn} describes, model specialization~\cite{mcdnn} reduces cost by simplifying the task of \cnns. Specifically, model specialization takes advantage of two characteristics in real-world videos. First, most video streams have a limited set of object classes (\xref{subsubsec:limited}). Second, objects in a specific stream are often {\em visually more constrained} than objects in general (say, in the COCO \cite{DBLP:conf/eccv/LinMBHPRDZ14} dataset). The cars and buses that occur in a specific traffic camera have much less variability, e.g., they have very similar angle, distortion and size, compared to a generic set of vehicle images. Thus, classifying objects from a specific camera is a much simpler task than doing so from all cameras, resulting in cheaper ingest-time \cnns.}

While specializing \cnns to specific videos has been attempted in computer vision research (e.g., \cite{mcdnn, DBLP:journals/corr/ShenHPK17}), we explain its two key implications within {\focus}.

% high accuracy with considerably cheaper models. %, and can be achieved by a \cnn with fewer convolutional layers.% and much cheaper image features are enough.

%Specialization provides a cheaper model to build the ingest index for each video.

\noindent{\bf 1) Lower {\sf K} values.} Because the specialized \cnn classifies across fewer classes, they are more accurate, which enables {\focus} to achieved the desired recall with a much smaller {\sf K} (for the top-{\sf K} ingest index). We find that specialized models can usually use {\sf K} $\leq$ 4 (Figure~\ref{fig:topK}), much smaller compared to the typical {\sf K} needed for {\em generic} cheap \cnns. %TODO: Can we quantify this or remove the much smaller claim ?}. 
\kff{A smaller} {\sf K} translates to fewer objects that have to be classified by GT-\cnn at query time, thus reducing latency.

%Consequently, after filtering by the top-{\sf K} index and clustering, fewer objects have to be classified by GT-\cnn  at query time, which reduces query latency.%, which itself is faster because of the specialization. %We quantify these aspects in \xref{sec:evaluation}.

%We retrain the model for each video stream.
\noindent{\bf 2) Most frequent classes.} On each video stream, {\focus} periodically obtains a small sample of video frames and classifies their objects using GT-\cnn to estimate the ground truth of the distribution of object classes for the video (similar to Figure~\ref{fig:class_num_cdf}). From this distribution, {\focus} selects the most frequently occurring $L_s$ object classes to retrain new specialized models. % $M^s_{i,k}$. 
Because just a handful of classes often account for a dominant majority of the objects (\xref{subsubsec:limited}), low values of $L_s$ usually suffice.

%We target the \cnns on the objects detected in the specific video stream labeled by the GT-\cnn and retrain them to create specialized models.
%Specialization is also based off a family of \cnn architectures (such as ResNet \cite{DBLP:conf/cvpr/HeZRS16}, AlexNet \cite{DBLP:conf/nips/KrizhevskySH12}, and VGG \cite{Simonyan15}) with different number of convolution layers, %and other \cnn approximation techniques~\cite{mcdnn,lowrank,fitnets}, similar to \xref{subsec:cheap_ingest}. %Targeting the model to be video-specific often allows us to use lower resolutions of the objects that substantially reduces the resource usage. Specialization adds to the set of options available for ingest \cnns (\{CheapCNN$_1, ... ,$ CheapCNN$_n$\} in \xref{subsec:cheap_ingest}), and {\focus} picks the best model (CheapCNN$_i$) and the corresponding {\sf K} for the index.
%  We pick among these specialized models for ingest and query time \cnns,  and the corresponding {\sf K} for the top-{\sf K} index.

%We treat the GT-\cnn model's classifications as the ground truth and iteratively remove layers from the cheap ingest \cnn ($M_i$) until the error in the training exceeds a certain threshold.

%\noindent{\bf ``OTHER'' class:} 
While {\focus} specializes the \cnn towards the most frequently occurring $L_s$ classes, we also want to support querying of the {\em less} frequent classes. For this purpose, {\focus} includes an additional class called ``OTHER'' in the specialized model. Being classified as OTHER simply means not being one of the $L_s$ classes. At query time, if the queried class is among the OTHER classes of the ingest \cnn's index, {\focus} extracts all the clusters that match the OTHER class and classifies their centroids through the GT-\cnn model.\footnote{Specialized \cnns can be retrained quickly on a small dataset. Retraining is relatively infrequent and done once every few days. Also, because there will be considerably fewer objects in the video belonging to the OTHER class, we proportionally re-weight the training data to contain equal number of objects of all the classes.} %We treat a high incidence of the OTHER class as a signal to retrain the \cnn likely because the set of objects in the video have changed.

The parameter $L_s$ (for each \kff{video} stream) exposes the following trade-off. Using a small $L_s$ enables us to train a simpler model with cheaper ingest cost and lower query-time latency {\em for the popular classes}, \kff{but}, it also leads to a larger fraction of objects falling in the OTHER class. \kff{As a result, querying for the OTHER class} will be expensive because all those objects will have to be classified by the GT-\cnn. Using a larger value of $L_s$, on the other hand, leads to more expensive ingest and query-time models, but cheaper querying for the OTHER classes. We select $L_s$ in \xref{subsec:tuning}.%, jointly with the other parameters of \xref{subsec:cheap_ingest} and \xref{subsec:redundancy}.

\subsection{Redundant Object Elimination}
\label{subsec:redundancy}

% large K
At query time, {\focus} retrieves the objects likely matching the user-specified class from the top-{\sf K} index and infers their actual class using the GT-\cnn.
\kff{This ensures} precision of 100\%, but could cause significant latency at query time.
%The main latency cost at query time comes from classifying a large number of objects (filtered by the top-K approximation index of the cheap \cnn) with the GT-\cnn to remove the false positives.
Even if this inference were parallelized across many GPUs, it would incur a large cost.
{\focus} uses the following observation to reduce this cost: if two objects are visually similar, their feature vectors \kff{are also similar} and they would likely be classified as the same class (e.g., cars) by the GT-\cnn model (\xref{subsubsec:features}).

{\focus} {\em clusters} objects that are similar, invokes the expensive GT-\cnn only on the cluster centroids, and assigns the centroid's label to all objects in each cluster.
%Coupled with the observation in \xref{subsec:char} of considerable similarity between the frames in their objects,
Doing so dramatically reduces the work done by the GT-\cnn classifier at query time. {\focus} uses the feature vector output by the previous-to-last layer of the cheap ingest \cnn (see \xref{subsec:cnn}) for clustering. Note that {\focus} clusters the {\em objects} in the frames and not the frames as a whole.\footnote{\kff{Recall from \xref{subsec:cheap_ingest} that {\focus}' ingest process either (i) employs an object detector \cnn (e.g., YOLO) that jointly detects and classifies objects in a frame; or (ii) detects objects with background subtraction and then classifies objects with a classifier \cnn (e.g. ResNet). Regardless, we obtain the feature vector from the \cnns for {\em each object} in the frame.}}

%either object detection and classification done together (e.g., YOLO) or separately (e.g., background subtraction followed by ResNet). Regardless, we obtain the feature vector for {\em each object} in the frame from the \cnns, and the clustering is identical in both cases.} 

The key questions regarding clustering are \kff{\emph{how} we cluster % (algorithmic)
and \emph{when} we cluster. % (systemic).
We discuss both below}.

%As mentioned in \xref{subsec:cheap_ingest}, we run each frame through an object detector, and then cluster the extracted objects.

\noindent{\bf Clustering Heuristic.} We require two properties in our clustering technique. First, given the high volume of video data, it should be a single-pass algorithm to keep the overhead low, unlike most clustering algorithms, which are \emph{quadratic} complexity. Second, it should make no assumption on the number of clusters and adapt to outliers in data points on the fly. Given these requirements, we use the following simple approach for {\em incremental} clustering, which has been well-studied in the literature \cite{clustering1, clustering2}.

%% clustering technique
%%We adopt the following simple approach for incrementally adding new objects to clusters given a threshold $T$.
%At steady state, we will have a set of existing clusters $c_1, \dots, c_M$.\footnote{We obtain the initial clusters by clustering a randomly chosen set of objects.}

We put the first object into the first cluster $c_1$, \kt{and we make the first object as the centroid of $c_1$}.
To cluster a new object $i$ with a feature vector $f_i$,
we assign it to the closest cluster $c_j$ if \kt{the centroid of} $c_j$ is at most distance $T$ away from $f_i$,
where $T$ is a distance threshold.
%We measure distance using the dot product between the feature vector and cluster centroids.
However, if none of the clusters are within a distance $T$, we create a new cluster with centroid at
$f_i$.
We measure distance as the $L_2$ norm \cite{l2norm}
between the cluster centroid feature vector and the object feature vector $f_i$. To bound the time complexity for clustering, we keep the 
number of clusters actively being updated at a constant $\numClusters$. We do this by
\emph{sealing} the smallest cluster when the
number of clusters hits $\numClusters+1$, but we keep growing the popular clusters (such as similar cars). This
maintains the complexity as $O(\numClusters n)$, which is linear in $n$, the total number of objects. \kff{The value of $\numClusters$ has a very minor impact on our evaluation results, and we set $\numClusters$ as 100 in our evaluations.}
%and storing their data in the top-{\sf K} index. (I dont follow this. what happens to points in the smallest cluster?) }
%We periodically remove the smallest clusters from $c_1, \dots, c_M$, and fix the size of this list to a constant $M$.
%Using this algorithm, 

Clustering can reduce precision and recall depending on the parameter $T$.
If the centroid is classified by GT-\cnn as the queried class X but the cluster contains another object class, it reduces precision.
If the centroid is classified as a class different than X but the cluster has an object of class X, it reduces recall.
\xref{subsec:tuning} discuss setting $T$.

\noindent{\bf Clustering at Ingest vs. Query Time.} {\focus} clusters the objects at ingest-time rather than at query-time.
Clustering at query-time would involve \emph{storing} all feature vectors, \emph{loading} them for objects filtered from the ingest index and then clustering them.
%This also means that we need to \emph{store} all the feature vectors.
Instead, clustering {\em at ingest time} creates clusters right when the feature vectors are created and stores only the cluster centroids in the top-{\sf K} index.
This makes the query-time latency much lower and also reduces the size of the top-{\sf K} index.
We observe that the ordering of indexing and clustering operations is mostly \emph{commutative} in practice and has little impact on recall and precision.
We therefore use ingest-time clustering due to its latency and storage benefits.

\subsection{Trading off Ingest Cost and Query Latency}
\label{subsec:tuning}

{\focus}' goals of high recall/precision, low ingest cost and low query latency are \kff{affected} by its parameters: 
$(i)$ {\sf K}, the number of top results from the ingest-time \cnn to index an object;
$(ii)$ $L_s$, the number of popular object classes we use to create a specialized model;
$(iii)$ CheapCNN$_i$, the specialized ingest-time cheap \cnn;
$(iv)$ Skip$_\text{th}$, the confidence threshold to skip invoking GT-\cnn; and $(v)$ $T$, the distance threshold for clustering objects.
% and . In addition, we also select the cheap ingest model $M_i$ to use.
%All these together impact the FNR and FDR metrics as well as the ingest- and query-time costs.

\noindent{\bf Viable Parameter Choices.} {\focus} first prunes the parameter choices to only those that meet the desired precision and recall targets. 
%The effect of these four parameters is intertwined.
\kf{Among the five parameters, four parameters ({\sf K}, $L_s$, CheapCNN$_i$, and $T$) impact recall; only $T$ and Skip$_\text{th}$ impact precision.}
{\focus} samples a representative fraction of the video stream and classifies them using GT-\cnn for the ground truth. Next, for each combination of parameter values, {\focus} computes the precision and recall (relative to GT-\cnn's outputs) achievable for each of the object classes, and selects only those \kff{combinations} that meet the \kf{precision and recall} targets. 
%This is because, without clustering and applying the centroid's classification by GT-\cnn to all the objects in the cluster, we can achieve $100\%$ precision (at the expense of high query latency). If the clustering is not tight (i.e., high value of $T$), we lose
%%This is because we apply the cluster centroid's classification by GT-\cnn to all the objects in its cluster. Thus, if the clustering is not tight (i.e., high value of $T$), we lose precision.
%all false positives potentially introduced by a cheap model and top-K index will be removed by the GT-CNN, yet clustering using a large $T$ can introduce new false positives.

%\noindent{\bf Parameter Selection:} {\focus} selects parameter values {\em per video stream}. It samples a representative fraction of frames of the video stream and classifies them using GT-\cnn for the ground truth. For each combination of parameter values, {\focus} computes the expected precision and recall (using the ground truths generated by GT-\cnn) that would be achieved for each of the object classes. To navigate the combinatorial space of options for these parameters, we adopt a two-step approach. In the first step, {\focus} chooses CheapCNN$_i$, $L_s$ and $K$ using only the recall target. In the next step, {\focus} iterates through the values of $T$, the clustering distance threshold, and only select values that meet the precision target.

%\noindent{\bf Balancing Ingest Cost vs. Query Latency:} 
Among the viable parameter choices that meet the precision and recall targets, {\focus} {\em balances} ingest- and query-time costs. For example, picking a more accurate CheapCNN$_\text{ingest}$ will have higher ingest cost, but lower query cost because we can use a smaller {\sf K}. Using a less accurate CheapCNN$_\text{ingest}$ will have the opposite effect. % in ingest and query costs.

\noindent{\bf Pareto Boundary.} {\focus} identifies ``intelligent defaults'' that sharply improve one of the two costs for a small worsening of the other cost. 
Figure~\ref{fig:parameter_selection} illustrates the tradeoff between ingest cost and query latency for one of our video streams. The figure plots all the viable ``configurations'' (i.e., parameter choices that meet the precision and recall targets) based on their ingest cost (i.e., cost of CheapCNN$_\text{ingest}$) and query latency (i.e., the number of clusters that need to be checked at query time according to $K, L_s, T$ and Skip$_\text{th}$).

\begin{figure}[h]
  \centering
  \includegraphics[width=0.8\textwidth]{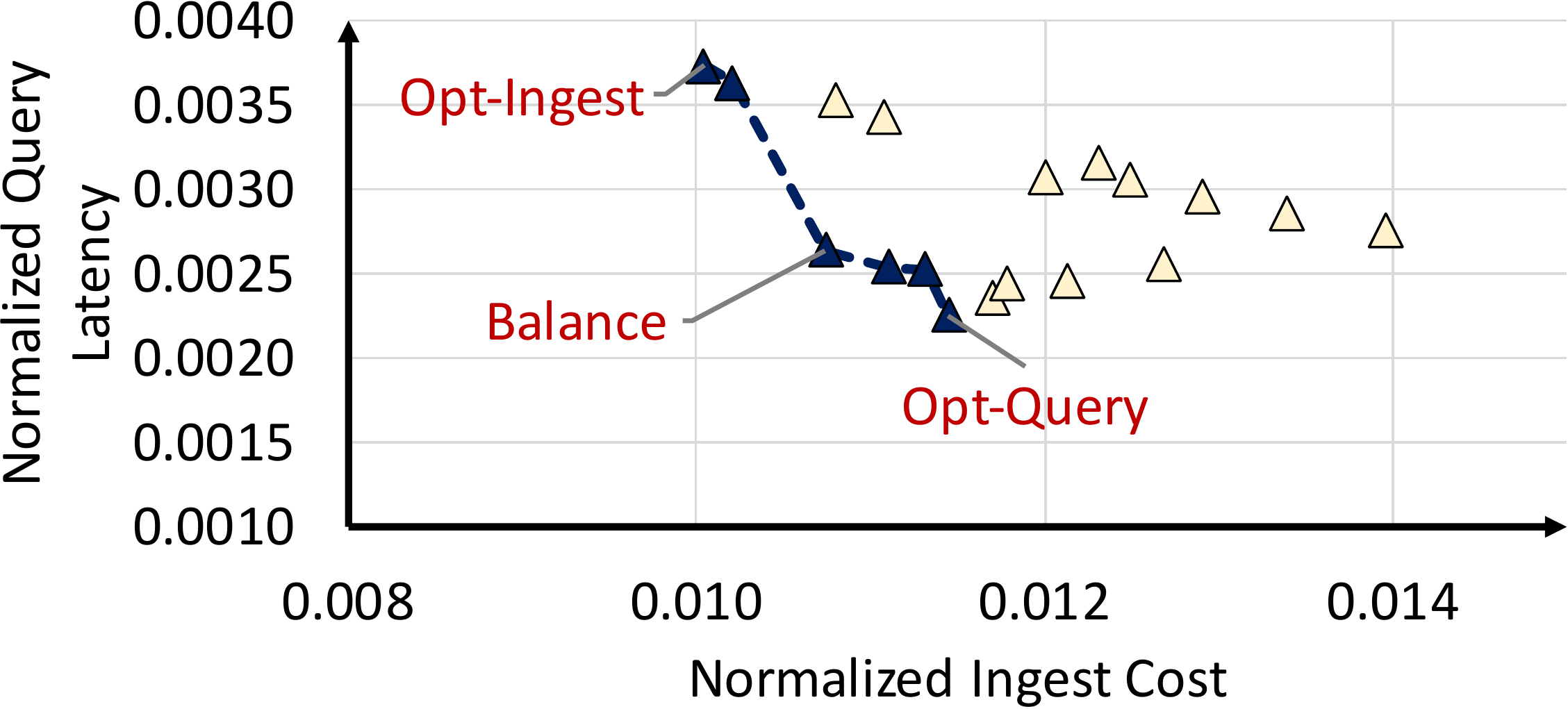}
  \caption[Parameter selection based on the ingest cost and query latency trade-off]{Parameter selection based on the ingest cost and query latency trade-off. The ingest cost is normalized to \kff{the cost of} ingesting all video frames with GT-\cnn (YOLOv2), while the query latency is normalized to \kff{the query latency using} {\noscope}. The dashed line is the Pareto boundary.}
  \label{fig:parameter_selection}
\end{figure}

We first extract the \emph{Pareto boundary}~\cite{pareto}, which is defined as the set of configurations among which we cannot improve one of the metrics without worsening the other. For example, in Figure~\ref{fig:parameter_selection}, the yellow triangles are not Pareto optimal when compared to the points on the dashed line. {\focus} can discard all \kff{non-Pareto} configurations because at least one point on the Pareto boundary is better than all non-Pareto points in {\em both} metrics.

% \ga{Worth elaborating? Maybe with an example?}

\noindent{\bf Tradeoff Policies.} {\focus} balances ingest cost and query latency (\sys{Balance} in Figure~\ref{fig:parameter_selection}) by selecting the configuration that minimizes the \emph{sum of ingest cost and query latency}. We measure ingest cost as the compute cycles taken to ingest the video and query latency as the average time (or cycles) required to query the video on the object classes that are recognizable by the ingest \cnn. By default, {\focus} chooses a \sys{Balance} policy that equally weighs ingest cost and query latency. Users can also provide any other weighted function to optimize their goal.
%A user can provide any other weighted function of ingest cost and query latency as the objective function to choose \sys{Balance} policy as per the user's optimization goal.  
%\ga{Details.} %Among the points on the Pareto boundary, the \emph{Balance} point minimizes the sum of ingest cost and query cost.

{\focus} also allows for other configurations based on the application's preferences and query rates. \sys{Opt-Ingest} minimizes the ingest cost and is applicable when the application expects most of the video streams to not get queried (such as surveillance cameras), as this policy minimizes the amount of wasted ingest work. On the other hand, \sys{Opt-Query} minimizes query latency \kff{but it} incurs a larger ingest cost. More complex policies can be easily implemented by changing how the query latency cost and ingest cost are weighted in our cost function. Such flexibility
enables {\focus} to fit a number of applications.

\section{Implementation}
\label{sec:implementation}

\kf{Because {\focus} targets large video datasets, a key requirement of {\focus'} implementation is the ability to scale and distribute computation across many machines. To this end, we implement {\focus} as three loosely-coupled modules which handle each of its three key tasks. Figure~\ref{fig:implementation} presents the architecture and the three key modules of {\focus}: the \emph{ingest processor} (M1), the \emph{stream tuner} (M2), and the \emph{query processor} (M3). These modules can be flexibly deployed on different machines based on the video dataset size and the available hardware resources (such as GPUs). We describe each module in turn.}

\begin{figure*}[h]
\centering
\includegraphics[width=0.9\textwidth]{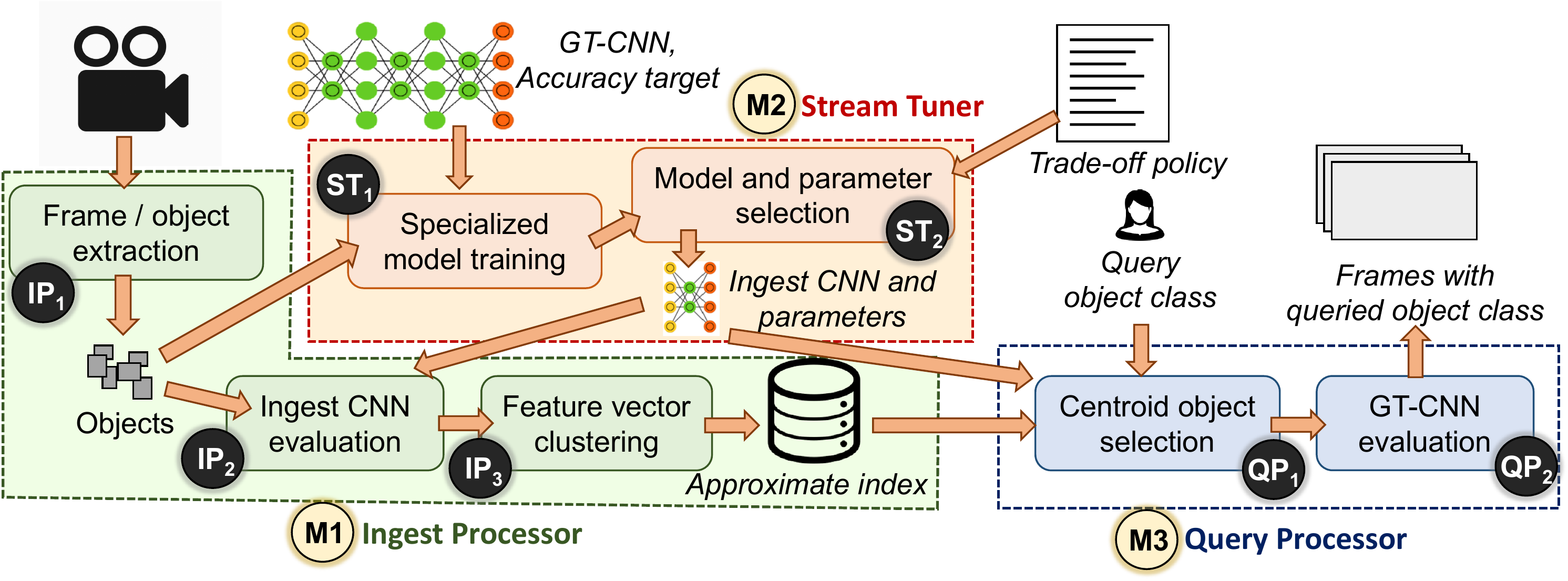}
\vspace{0.1in}
\caption{Key components of {\focus}.}
\vspace{0.1in}
\label{fig:implementation}
\end{figure*}

\subsection{Ingest Processor}
\label{subsec:ingest_processor}

\kf{{\focus}' ingest processor (M1) generates the approximate index (\xref{subsec:cheap_ingest}) for the input video stream. The work is distributed across many machines, with each machine running one worker process for \emph{each} video stream's ingestion. An ingest processor handles its input video stream with a four-stage pipeline: (i) extracting the moving objects from the video frames (IP$_1$ in Figure~\ref{fig:implementation}), (ii) inferring the top-{\sf K} indexes and the feature vectors of all detected objects with the ingest-time \cnn (IP$_2$ in Figure~\ref{fig:implementation}, \xref{subsec:cheap_ingest}), (iii) using the feature vector to cluster objects (\kff{IP$_3$ in Figure~\ref{fig:implementation}, \xref{subsec:redundancy}}), and (iv) storing the top-{\sf K} indexes of centroid objects in a database for efficient retrieval at query time.}

\kf{An ingest processor is configured differently for static (fixed-angle) and moving cameras. For static cameras, we extract object boxes by subtracting each video frame from the \emph{background frame}, which is obtained by averaging the frames in each hour of the video. \ko{We then index each object box with an ingest-time \emph{object classifier} \cnn. We accelerate the background subtraction with GPUs~\cite{opencv-gpu}. We use background subtraction for static cameras because running background subtraction with a cheap object classifier is much faster than running an ingest-time \emph{object detector} \cnn, and we find that both approaches have almost the same accuracy in detecting objects in static cameras. Hence, we choose the cheaper ingest option. }}

\kf{For moving cameras, we use a cheap, ingest-time \emph{object detector} \cnn (e.g., Tiny YOLO~\cite{DBLP:journals/corr/RedmonF16}) to generate the approximate indexes. We choose the object detection threshold (the threshold to determine if a box has an object) for the object detector \cnn such that we do not miss objects in GT-CNN while minimizing spurious objects.}

\subsection{Stream Tuner}
\label{subsec:stream_tuner}

\kf{The stream tuner (M2) determines the ingest-time CNN and \kff{{\focus}'} parameters for each video stream (\xref{subsec:tuning}). It takes four inputs: the sampled frames/objects, the GT-CNN, the desired accuracy relative to the GT-CNN, and the tradeoff policy between ingest cost and query latency (\xref{subsec:tuning}). Whenever executed, the stream tuner: (i) generates the ground truth of the sampled frames/objects with the GT-CNN; (ii) trains specialized ingest-time \cnns based on the ground truth (ST$_1$ in Figure~\ref{fig:implementation}); and (iii) selects the \kff{ingest-time \cnn and {\focus}' parameters} (ST$_2$ in Figure~\ref{fig:implementation}).}

\kf{{\focus} executes the stream tuner for each video stream \emph{before} launching the corresponding ingest processor. As the characteristics of video streams may change over time, {\focus} periodically launches the stream tuner to validate the accuracy of the selected parameters on sampled frames. The ingest-time \cnn and the system parameters are re-tuned if necessary to meet the accuracy targets (e.g., \cite{DBLP:conf/sigcomm/JiangABSS18}).}

\subsection{Query Processor}
\label{subsec:query_processor}

\kf{The task of the query processor is to return the video frames \kff{that contain the} user's queried object class. In response to a user query for class $X$, the query processor first retrieves the centroid objects with matching approximate indexes (QP$_1$ in Figure~\ref{fig:implementation}), and then uses the GT-CNN to determine the \kff{frames that do contain object class $X$} (QP$_2$ in Figure~\ref{fig:implementation}, \xref{subsec:cheap_ingest}). The GT-CNN evaluation can be easily distributed across many machines, if needed.}

\kf{We employ two optimizations to reduce the overhead of GT-CNN evaluation. First, we skip the GT-CNN evaluation for high-confidence indexes (\xref{subsec:cheap_ingest}). Second, we apply a query-specialized binary classifier~\cite{DBLP:journals/pvldb/KangEABZ17} on the frames that need to be checked before invoking the GT-CNN. These two optimizations make the query processor more efficient by \emph{not} running GT-CNN on all candidate centroid objects.}

\ignore{ % Old version
We describe the key aspects in {\focus}' implementation.

\noindent\textbf{Worker Processes.}
{\focus}' ingest-time work is distributed across many machines, with each machine running one {\em worker} process for {\em each video stream's} ingestion. The ingest worker receives the live video stream, %uses filter to remove stationary frames,
extracts the moving objects and
%(using background subtraction~\cite{bgs}); it is extensible to plug in any other object detector.
the detected objects are sent to the ingest-time \cnn to infer the top-K classes and the feature vectors. The ingest worker uses the features to cluster objects in its video stream and stores the top-K index in MongoDB~\cite{mongodb} for efficient retrieval at query-time.

Worker processes also serve queries by fetching the relevant frames off the top-{\sf K} index database and classifying the objects with GT-\cnn. We parallelize a query's work across many worker processes if resources are idle.

%\noindent{\bf Top-{\sf K} Index Database.}
%We use MongoDB~\cite{mongodb} to index the columns of the top-K index for efficient retrieval at query-time. Since the time complexity of clustering a new object is proportional to the number of clusters, we periodically ``seal'' some of the smallest clusters to disk but persist them in the top-K index. This ensures that even as we create new (potentially small) clusters, we keep the cost of clustering a new object to be bounded. %At the same time, popular clusters (of oft-occurring objects) will keep growing, thus directly reducing the query-time latency.

\noindent{\bf GPUs for \cnn classification.} The cheap \cnns and GT-CNN execute on GPUs (or other hardware accelerators for \cnns) which could either be local on the same machine as the worker processes or ``disaggregated'' on a remote cluster. This detail is abstracted away from our worker process and it seamlessly works with both designs.\sv{Can get rid of this if we dont have space}

\noindent\textbf{Dynamically adjusting $K$ at query-time.}
%Our top-K index stores, for each cluster, the $K$ most likely classes contained in that cluster.
%When querying for class $X$, we extract all clusters that contain $X$ among the top-K classes.
As an enhanced technique, we can select a new $K_x \le K$ at query time and only extract clusters where class $X$ appears among the top-$K_x$ classes; this will result in fewer clusters and thus also lower query-time latency.
This technique is useful in two scenarios:
1) some classes might be very accurately classified by the cheap \cnn; using a lower $K_x$ will still meet the user-specified recall and precision, yet will result in much lower latency;
2) if we want to retrieve only \emph{some objects of class X}, we can use very low $K_x$ to quickly retrieve them. If more objects are required, we can increase $K_x$ to extract a new batch of results.

%\noindent\textbf{Persisting results of GT-\cnn.}
%{\focus} balances the cost of video analytics across ingest and query time; this ensures that when the video is queried infrequently, we do not waste a lot of ingest cost.
%However, if the video \emph{is queried very frequently}, the query-time cost would accumulate and potentially increase dramatically.
%To overcome this, after applying the expensive GT-\cnn to a cluster centroid, we store the results in the top-K index for the specific cluster.
%This ensures, that the worst-case cost of {\focus} is proportional to the number of object clusters and does not depend on the query frequency.

%
%training specialized models
%
%storing the index
%
%querying: loading the index, clustering, calling ground-truth model 
}

\section{Evaluation}
\label{sec:evaluation}

We evaluate our {\focus} prototype with more than 160 hours of videos
from 14 real video streams that span traffic cameras,
surveillance cameras, and news channels. Our main results are:
%\sv{Are numbers in highlights updated ?}

%\squishlist
\begin{enumerate}
  \item {{\focus} is simultaneously 48$\times$ cheaper on average (up to
    92$\times$) than the \sys{Ingest-heavy} baseline in
    processing videos and 125$\times$ faster on average (up to 607$\times$)  than
    \sys{NoScope}~\cite{DBLP:journals/pvldb/KangEABZ17} in query
    latency {\textemdash} all the while achieving at least 99\% precision and
    recall (\xref{sec:overall_performance},
    \xref{sec:performance_by_technique}).}
    % We also break down the contribution of each technique towards these improvements    (\xref{sec:performance_by_technique}).

%\item {\focus} strikes a very good balance between ingest cost and
%  query latency. By investing a very small cost at ingest time (up to
%  98$\times$ cheaper than the \sys{Ingest-all} baseline), we can make
%  video queries up to 57$\times$ faster, while retaining the 95\%
%  accuracy target (\xref{sec:overall_performance}).
%\item The core techniques of {\focus} (model compression, per-stream
%  model specialization, and redundancy elimination) all contribute
%  significantly to cheaper ingest and smaller query latency
%  (\xref{sec:performance_by_technique}).
\item {{\focus} provides a rich trade-off space between ingest cost and
  query latency. If a user wants to optimize for \emph{ingest cost},
  {\focus} is 65$\times$ cheaper on average (up to 96$\times$) than the
  \sys{Ingest-heavy} baseline, while reducing query latency by
  100$\times$ on average. If the goal is to optimize for \emph{query
    latency}, {\focus} can achieve 202$\times$ (up to 698$\times$)
  faster queries than {\noscope} with 53$\times$ cheaper ingest. (\xref{sec:ingest_query_trade_off}).}

%  On average, the ingest cost is 65$\times$ cheaper
%  than the \sys{Ingest-heavy} baseline (and reduces query latency by
%  100$\times$) if optimizing for low-cost ingest. The query latency is
%  reduced by 202$\times$ (with 53$\times$ cheaper ingest) if
%  optimizing for query latency (\xref{sec:ingest_query_trade_off}).}
%\item {{\focus} is effective under broad conditions such as (2) various recall/precision targets, and (3) different query rates.}

  %high recall/precision targets (on average 15$\times$ faster than \sys{NoScope} and 57$\times$ cheaper than \sys{Ingest-all} for 99\% recall/precision, \xref{sec:result_accuracy_target}) and various frame sampling rates (30 fps--1 fps, \xref{sec:result_frame_sampling}).}
%\item {\kv{{\focus} is both faster and cheaper than a recent prior work, \sys{NoScope}~\cite{DBLP:journals/pvldb/KangEABZ17}. Across various recall/precision targets (95-99\%), {\focus} is 1.8 to 34.9$\times$ faster than \sys{NoScope} in query latency, and 1.4 to 33.5$\times$ cheaper in total cost (\xref{sec:noscope}).}}

  %\squishend
\end{enumerate}
%\item The overall cost (ingest + query) of {\focus} is lower than
%  either baselines even if the queries cover all the classes,
%  or if only a tiny fraction of video gets queried
%  (\xref{sec:result_query_rate}).
%\end{enumerate}

\subsection{Methodology}
\label{sec:methdology}

\noindent{\bf Software Tools.}
We use OpenCV 3.4.0~\cite{opencv} to decode the videos into frames, and we feed the frames to our evaluated systems, {\focus} and {\noscope}. {\focus} runs and trains \cnns with Microsoft Cognitive Toolkit
2.4~\cite{cntk}, an open-source deep learning system. \kf{Our ingest processor (\xref{subsec:ingest_processor}) stores the approximate index in MongoDB~\cite{mongodb} for efficient retrieval at query time.}

%and then use the built-in background subtraction
%algorithm~\cite{DBLP:conf/avbs/KaewTraKulPongB15} in OpenCV to extract
%moving objects from video frames. We use background subtraction
%instead of object detector \cnns (e.g.,
%YOLOv2~\cite{DBLP:journals/corr/RedmonF16} or Faster
%R-CNN~\cite{DBLP:conf/nips/RenHGS15}) to detect objects because: (1)
%running background subtraction is orders of magnitude faster than
%running these \cnns, and (2) background subtraction can detect moving
%objects more reliably, while object detector CNNs usually have difficulties on
%small objects~\cite{DBLP:conf/eccv/LiuAESRFB16}. Nonetheless, our system can seamlessly use object detector CNNs as well.
%

%We exclude the time of extracting frames and
%objects from our results because it is very minor comparing to running
%classifier \cnns, and it is the same amount of time in {\focus} and in
%the baselines.

\noindent{\bf Video Datasets.}  We evaluate 14 video streams that
span across traffic cameras, surveillance cameras, and news
channels. We record each video stream for 12 hours to cover both day
time and night time. Table~\ref{table:video_dataset} summarizes the
video characteristics. We strengthen our evaluation by including down
sampling (or frame skipping), one of the most straightforward
approaches to reduce ingest cost and query latency, into our
evaluation \emph{baseline}. \kff{Specifically, as the vast majority of
  objects show up for at least one second in our evaluated videos, we
  evaluate each video at 1 fps instead of 30 fps. We find that the object
  detection results at these two frame rates are almost the same.}
Each video is split evenly into a \emph{training set} and a \emph{test
  set}. The training set is used to train video-specialized \cnns and
select system parameters. We then evaluate the systems with the test
set. In some figures, we show results for only eight representative videos
to improve legibility.

\setlength\tabcolsep{3pt}

\begin{table}[h!]
\small \centering 
%\scriptsize
\begin{tabular}{|l|l|l|l|} \hline

Type & Camera & Name & Description \\ \hline

    & & auburn\_c & \scell{A commercial area intersection in the City of Auburn~\cite{auburn_c}} \\ \cline{3-4}
\multirow{7}{*}{\scell{\\Traffic}} & \multirow{7}{*}{\scell{\\Static}}
    & auburn\_r  & \scell{A residential area intersection in the City of Auburn~\cite{auburn_r}} \\ \cline{3-4}
    & & bellevue\_d & \scell{A downtown intersection in
  the City of Bellevue. The video\\ streams are obtained from city
    traffic cameras.} \\ \cline{3-4}
    & & bellevue\_r  & \scell{A residential area intersection in the City of Bellevue} \\ \cline{3-4}
    
    & & bend & \scell{A road-side camera in the City of Bend~\cite{bend}} \\ \cline{3-4}
    
    & & jackson\_h &  \scell{A busy intersection in Jackson Hole~\cite{jacksonh}} \\ \cline{3-4}

    & & jackson\_ts &  \scell{A night street in Jackson Hole. The video is downloaded from\\ the NoScope project website~\cite{noscope_repo}.} \\ \hline

\multirow{4}{*}{\scell{\\Surveillance}} & \multirow{4}{*}{\scell{\\Static}}
    & \scell{coral} & \scell{An aquarium video downloaded from the NoScope\\ project website~\cite{noscope_repo}} \\ \cline{3-4}
    & & \scell{lausanne} & \scell{A pedestrian plaza (Place de la Palud) in Lausanne~\cite{lausanne}} \\ \cline{3-4}
    & & \scell{oxford} & \scell{A bookshop street in the University of Oxford~\cite{oxford}} \\ \cline{3-4}
    & & sittard & \scell{A market square in Sittard~\cite{sittard}} \\  \hline

\multirow{3}{*}{\scell{News}} & \multirow{3}{*}{\scell{Moving}}
    & cnn & \scell{News channel} \\ \cline{3-4}
    & & foxnews & \scell{News channel} \\ \cline{3-4}
    & & msnbc & \scell{News channel} \\ \hline

\end{tabular}
\caption{Video dataset characteristics}
\label{table:video_dataset}
\end{table}

\ignore{
\noindent{\bf Static/Moving Cameras.} We configure {\focus} differently
for static and moving cameras. For static cameras, we extract object
boxes by subtracting each video frame from the \emph{background
  frame}, which is obtained by averaging the frames in each hour of
the video. \ko{We then index each object box with an ingest-time
  object classifier \cnn (chosen from various compressed versions of
  ResNet-18~\cite{DBLP:conf/cvpr/HeZRS16}). We accelerate the
  background subtraction with OpenCV GPU APIs~\cite{opencv-gpu}. We
  use background subtraction for static cameras because running
  background subtraction with an object classifier such as ResNet-18
  is an order of magnitude faster than running an ingest-time object
  detector \cnn, and we find that both approaches have almost the same
  accuracy in detecting objects in static cameras. Hence, we choose
  the cheaper ingest option for static cameras.}

  \ko{For moving cameras, we use an ingest-time object detector \cnn
    (Tiny YOLO~\cite{DBLP:journals/corr/RedmonF16}) to generate the
    approximate indexes. We choose the object detection threshold (the
    threshold to determine if a box has an object) for Tiny YOLO such
    that we do not miss objects in GT-CNN while not netting in
    spurious objects. For both static and moving cameras, {\focus}
    applies \sys{NoScope}'s query-specialized binary classifiers on
    the frames that need to be checked before invoking GT-CNN at query
    time.}
}
%In most of our
%evaluation, we only show results with static cameras because: (i) our
%video streams are mostly static cameras; and (ii) our query baseline,
%{\noscope}, is designed to optimize for static cameras.\sv{Why is is noscope optimized for static
%cameras. Why is our stream mostly static cameras ?}

\noindent{\bf Accuracy Target.}  We use
YOLOv2~\cite{DBLP:journals/corr/RedmonF16}, a state-of-the-art object
detector \cnn, as our ground-truth \cnn (GT-CNN): all
objects detected by GT-CNN are considered to be the correct answers.\footnote{We do not
  use the latest YOLOv3 or other object detector CNN such as
  FPN~\cite{DBLP:conf/cvpr/LinDGHHB17} as our GT-CNN because one of
  our baseline systems, {\noscope}, comes with the YOLOv2 code. Fundamentally,
  there is no restriction on the selection of GT-CNN for {\focus}.} For
each query, our default accuracy target is 99\% recall and
precision. To avoid overfitting, we use the \emph{training set} of each video to explore
system parameters with various recall/precision targets (i.e.,
100\%--95\% with a 0.5\% step), and we report the best system
parameters that can \emph{actually} achieve the \kff{recall/precision} target on the
\emph{test set}. We also evaluate other recall/precision targets such
as 97\% and 95\% (\xref{sec:result_accuracy_target}).

\ignore{
\footnote{Note that using recall and
  precision is a more informative metric than using \emph{accuracy},
  the major metric reported in the \sys{NoScope} paper. Accuracy is the
  ratio of the correct positive/negative classifications to the total
  number of samples. However, high accuracy does not always indicate
  good performance because of a phenomena called \emph{accuracy
    paradox}~\cite{accuracy_paradox, zhu2007knowledge,
    DBLP:conf/dmkdttt/Thomas008}. For example, if 95\% of the samples
  are ``negative'', a system that \emph{always} predicts ``negative''
  will have 95\% accuracy, however its recall and precision will be
  0\% when querying for ``positive''.}}
}

%  Note that in most practical
%  cases, only one of the two metrics (recall or precision) needs to be
%  high. For example, an investigator cares about high recall, and
%  looking through some irrelevant results is an acceptable trade-off.

%  By setting both targets high,
%  we are lower bounding the performance improvements that {\focus} can achieve.
  
%On the other hand, a user looking for images of a specific type of
%cars wants high precision, while missing some cars in the video is fine.

%Nonetheless, we set
%conservative targets to support different scenarios.

\noindent{\bf Baselines.}  \kf{We use baselines at two ends of the design spectrum: (1) \sys{Ingest-heavy}, the baseline system that uses GT-CNN to analyze all frames at ingest time, and stores the \kff{results as an index} for query; and (2) \sys{NoScope}, a recent state-of-the-art querying system~\cite{DBLP:journals/pvldb/KangEABZ17} that analyzes frames for the queried object class at query time. We also use a third baseline, \sys{Ingest-NoScope} that uses \sys{NoScope}'s techniques at ingest time. Specifically, \sys{Ingest-NoScope} runs the binary classifiers of \sys{NoScope} for all possible classes at \emph{ingest time}, invokes GT-CNN if any of the binary classifiers cannot produce a high-confidence result, and stores \kff{the results as an index} for query. To further strengthen the baselines, we augment all baseline systems with background subtraction, thus eliminating frames with no motion. \kff{As {\focus} is in the middle of the design spectrum, we compare {\focus}' ingest cost with \sys{Ingest-heavy} and \sys{Ingest-NoScope}, and we compare {\focus}' query latency with \sys{NoScope}.} }

%We strengthen both baselines
%with basic motion detection (background subtraction). Therefore, the
%baselines \emph{do not} run any GT-CNN on the frames that have no
%moving objects. Note that not running GT-CNN on idle frames is one of
%the core techniques in recent work~\cite{DBLP:conf/sensys/ChenRDBB15, DBLP:journals/pvldb/KangEABZ17}.

\noindent{\bf Metrics.} We use two performance metrics. The first
metric is \emph{ingest cost}, the end-to-end machine time to ingest
each video.  The second metric is \emph{query latency}, the end-to-end
latency for an object class query. Specifically, for each video
stream, we evaluate the object classes that collectively make
up 95\% of the detected objects in GT-CNN. We report the average query
latency on these object classes. We do not evaluate the bottom 5\%
classes because they are often random erroneous results in GT-CNN
\ko{(e.g., ``broccoli'' or ``orange'' in a traffic camera)}.
%and (ii) they would skew the results as there are far more
%such classes.

\ko{Both metrics include the time spent on all
  processing stages, such as detecting objects with background
  subtraction, running \cnns, clustering, reading and writing to the
  approximate index, etc.} Similar to prior
work~\cite{DBLP:journals/corr/RedmonF16,
  DBLP:journals/pvldb/KangEABZ17}, we report the end-to-end execution
time of each system while excluding the video decoding time, as the
decoding time can be easily accelerated with GPUs or
accelerators.

%Both metrics include only GPU time spent
%classifying images and exclude other (CPU) time spent decoding video
%frames, detecting moving objects, recording and loading video, and
%reading and writing to the top-{\sf K} index.  We focus solely on GPU
%time because when the GPU is involved, it is the bottleneck resource.
%\pg{I changed what was here before--is this new wording of the previous sentence accurate?}
%The query latency of \sys{Ingest-heavy} is 0 and the ingest cost of
%\sys{NoScope} is 0.
%Some of these are used by both the baselines and {\focus} and/or their cost is negligible compared to the GPU cost.

\noindent{\bf Experimental Platform.}  We run the experiments on
\sys{Standard\_NC6s\_v2} instances on the Azure cloud. Each instance is
equipped with a high-end GPU (NVIDIA Tesla P100), 6-core Intel Xeon
CPU (E5-2690), 112 GB RAM, a 10 GbE NIC, and runs 64-bit Ubuntu 16.04
LTS.

\ignore{ % Outline
\begin{itemize}
\item experiment setup (software / hardware)
\item dataset (video characteristics, length, frame rate)
\item CNN models (Expensive and cheap CNN models)
\item default accuracy target (10\% or 5\% FNR and FPR)
\item default ingest/query trade-off point (middle point of the Pareto
  boundary)
\end{itemize}
}

%\vspace{2ex}
\subsection{End-to-End Performance}
\label{sec:overall_performance}

{\bf Static Cameras.} We first show the end-to-end performance of {\focus} on static cameras
\ignore{by showing its ingest cost and query latency }when {\focus} aims to
balance these two metrics (\xref{subsec:tuning}).
Figure~\ref{fig:overall_results} compares the ingest cost of {\focus} and \sys{Ingest-NoScope}
with \sys{Ingest-heavy} and the query latency of {\focus} with
\sys{NoScope}.
%We compare the
%ingest cost of {\focus} with \sys{Ingest-heavy} in
%Figure~\ref{fig:overall_ingest}, and the query latency of {\focus} with
%\sys{NoScope} in Figure~\ref{fig:overall_query}.
We make three main observations.

\begin{figure}[h]
  \centering
  \includegraphics[width=0.8\textwidth]{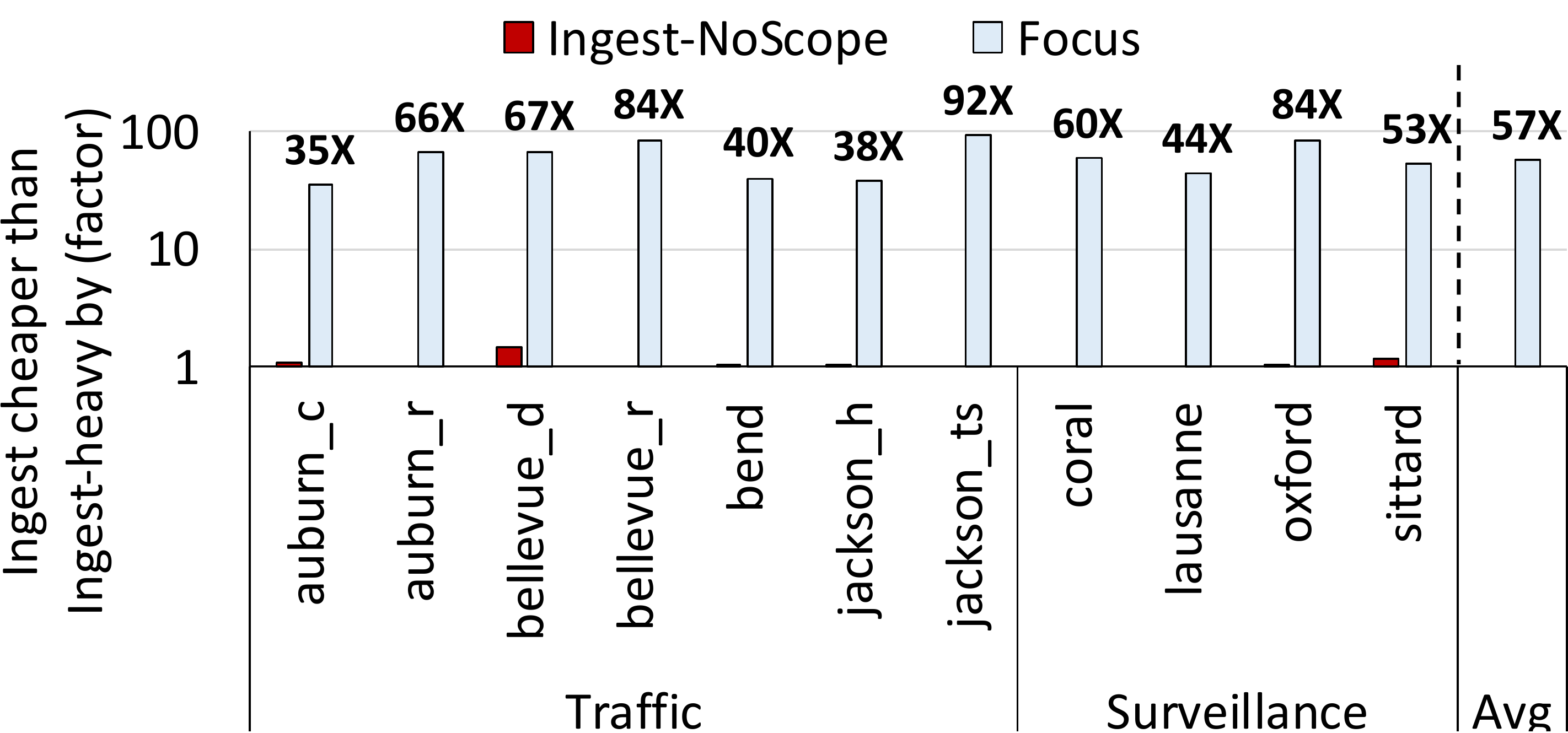}
%  \caption{{\focus} ingest cost compared to \sys{Ingest-heavy}}
%  \label{fig:overall_ingest}
%\end{figure}
%
%\begin{figure}[t!]
%  \centering
%\mbox{}
  \includegraphics[width=0.8\textwidth]{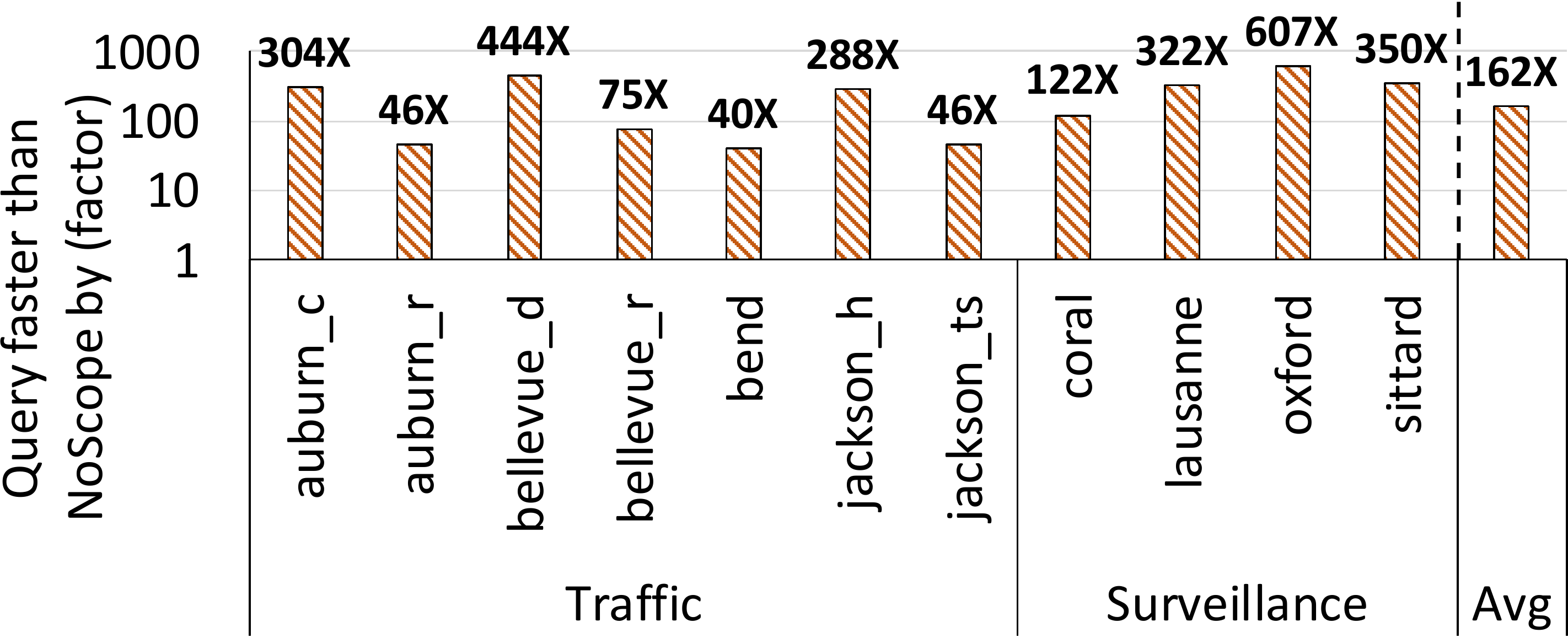}
  \caption[{\focus} ingest cost and query latency compared to baseline systems]{(Top) {\focus} ingest cost compared to \sys{Ingest-heavy}.
    (Bottom) {\focus} query latency compared to \sys{NoScope}.}
  %  \label{fig:overall_query}
    \label{fig:overall_results}
\end{figure}

First, {\focus} significantly improves query latency with a very small
cost at ingest time. {\focus} achieves 162$\times$
speedup \kff{(on average)} in query latency over \sys{NoScope} with a very small ingest
cost (57$\times$ cheaper than \sys{Ingest-heavy}\kff{, on average}), all
the while retaining 99\% recall and precision \kff{(not shown)}. {\focus} achieves two
orders of magnitude speedup over {\noscope} because: (i) the
ingest-time approximate indexing drastically narrows down the frames
that need to be checked at query time; and (ii) the feature-based
clustering further reduces the redundant work. In contrast, {\noscope}
needs to go through \emph{all} the frames at query time, which is
especially inefficient for the object classes that \kff{appear}
infrequently. We conclude that {\focus}' architecture provides a
valuable trade-off between ingest cost and query latency.

\ignore{On a 10-GPU
cluster, the query latency on a 12-hour video goes down from 30 minutes
to 45 seconds. The processing cost of each video stream
also goes down from \$250/month to \$4/month. This shows that
{\focus} strikes a very good balance between these two competing
goals effectively.}

\ko{Second, directly applying {\noscope}'s techniques at ingest time
  (\sys{Ingest-NoScope}) does not save much cost over
  \sys{Ingest-heavy}. There are two reasons for this: (1) While each
  binary classifier is relatively cheap, running multiple instances of
  binary classifiers (for all possible object classes) imposes
  non-trivial cost. (2) The system needs to invoke GT-CNN when any one
  of the binary classifiers cannot derive the correct answer. As a
  result, GT-CNN is invoked for most frames. Hence, the ingest cost of
  {\focus} is much cheaper than both, \sys{Ingest-heavy} and
  \sys{Ingest-NoScope}. This is because {\focus}' architecture
  only needs to construct the approximate index at ingest time which
  can be \kff{done cheaply} with an ingest-time \cnn.}

Third, {\focus} is effective across videos with
varying characteristics. It makes queries 46$\times$ to 622$\times$
faster \kff{than \sys{NoScope}} with a very small ingest cost (35$\times$ to 92$\times$
cheaper \kff{than \sys{Ingest-heavy}}) among busy intersections (\video{auburn\_c},
\video{bellevue\_d} and \video{jackson\_h}), normal intersections
(\video{auburn\_r}, \video{bellevue\_r}, \video{bend}), a night street
(\video{jackson\_ts}), busy plazas (\video{lausanne} and
\video{sittard}), a university street (\video{oxford}), and an
aquarium (\video{coral}). The gains in query latency are smaller for
some videos (\video{auburn\_r}, \video{bellevue\_r}, \video{bend},
and \video{jackson\_ts}). This is because {\focus}'
ingest \cnn is less accurate on these videos, and {\focus}
selects more conservative parameters (e.g., a larger {\sf K} such as 4--5 instead of 1--2) to \kff{attain}
the recall/precision targets. As a result, there is more work at
query time for these videos. Nonetheless, {\focus} still achieves at
least 40$\times$ speedup \kff{over \sys{NoScope}} in query latency. We conclude that the core
techniques of {\focus} are general and effective on a variety of
real-world videos.%\sv{Having no moving camera stuff here looks odd ?}

\noindent{}\ko{{\bf Moving Cameras.} We evaluate the applicability of {\focus} on moving cameras using three news channel video streams. These news videos \kff{were recorded with moving cameras and they change scenes between different news segments}. For moving cameras, we use a cheap object detector (Tiny YOLO, \kf{which is 5$\times$ faster than YOLOv2 for the same input image size}) as our ingest-time \cnn. \ignore{We generate the approximate index with a cheap object detector \cnn by lowering the confidence threshold of correct predictions (i.e., indexing more possible objects), so the approximate index can retain high recall. }Figure~\ref{fig:moving_cam} shows the end-to-end performance of {\focus} on moving cameras. }

\begin{figure}[h]
 \centering
 \begin{subfigure}[t]{0.50\linewidth}
 \centering
 \includegraphics[width=1.0\textwidth]{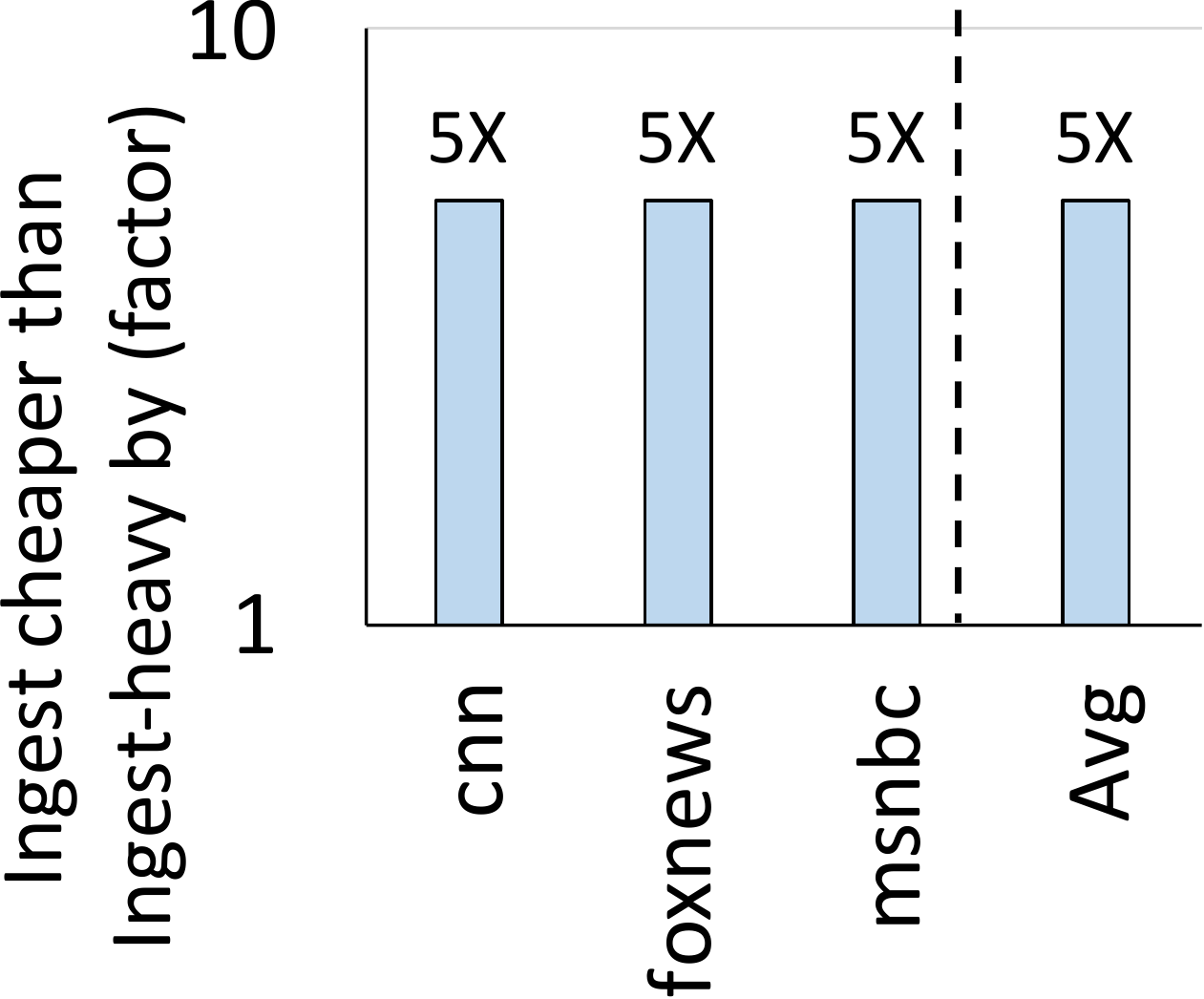}
% \caption{Ingest cost compared to \sys{Ingest-heavy}}
 \label{fig:moving_cam_ingest}
 \end{subfigure}
 \begin{subfigure}[t]{0.48\linewidth}
 \centering
 \includegraphics[width=1.0\textwidth]{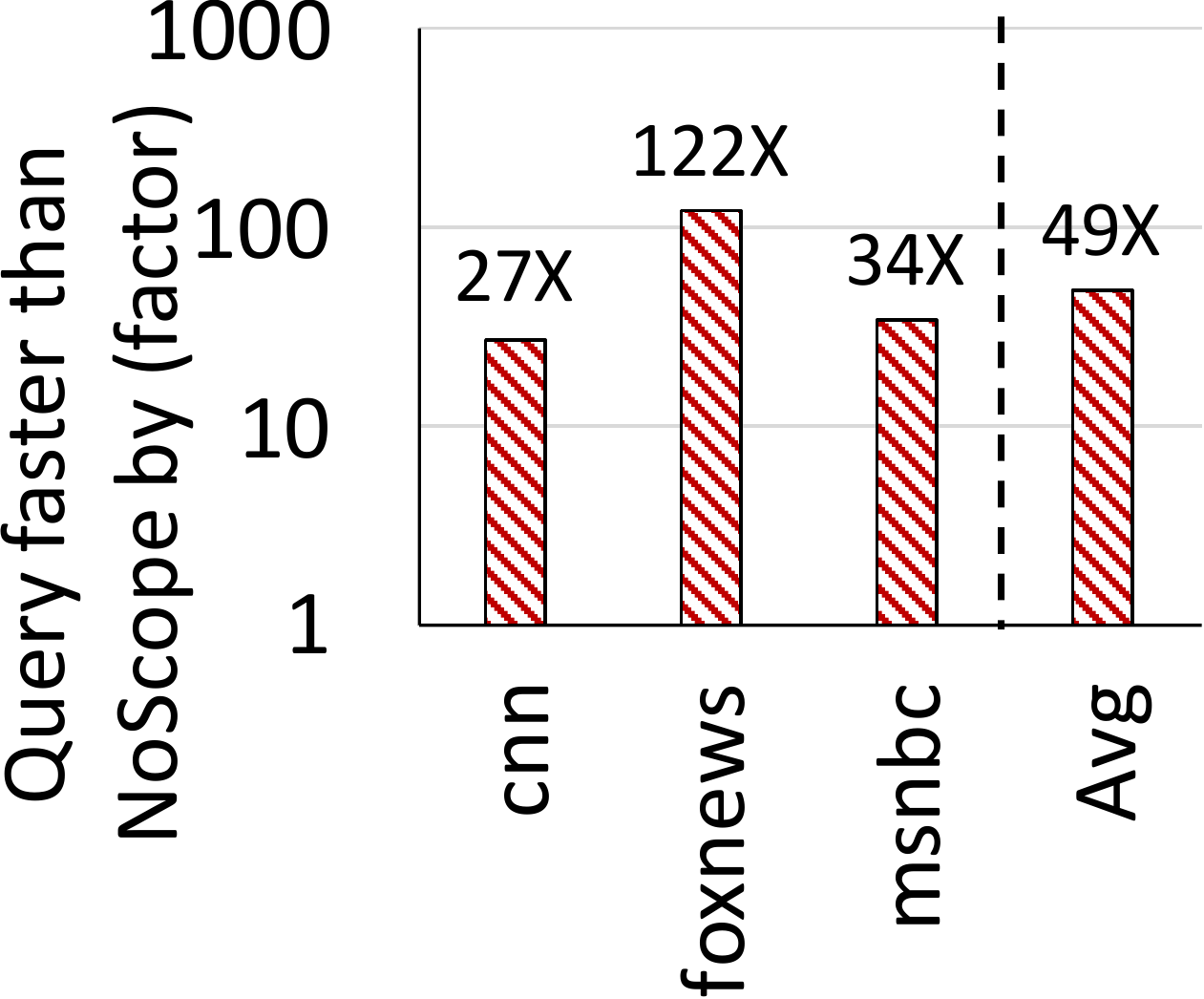}
% \caption{Query latency compared to \sys{NoScope}}
 \label{fig:moving_cam_query}
 \end{subfigure}
 \caption[{\focus} performance on moving cameras]{{\focus} performance on moving cameras. (Left) {\focus} ingest cost compared to \sys{Ingest-heavy}.
    (Right) {\focus} query latency compared to \sys{NoScope}.}
 \label{fig:moving_cam}
\end{figure}

\ko{As the figure shows, {\focus} is effective in reducing query latency with only a modest ingest cost. {\focus} achieves a 49$\times$ speedup in query latency on average over {\noscope}, with ingest cost that is 5$\times$ cheaper than \sys{Ingest-heavy}. We make two main observations. First, the ingest cost improvements on moving cameras \kff{(5$\times$)} is lower than the ones on static cameras (57$\times$). \kf{This is because moving cameras require a detector \cnn to detect objects, and it is more costly to run a cheap object detector (like Tiny YOLO) as opposed to using background subtraction to detect the objects and then classifying them using a cheap classifier \cnn (like compressed ResNet18). Our design, however, does not preclude using much cheaper detectors than Tiny YOLO, \kff{and we can further reduce the ingest cost of moving cameras by exploring even cheaper object detector \cnns.}} Second, {\focus}' techniques are very effective in reducing query latency on moving cameras. The approximate index generated by a cheap detector \cnn significantly narrows down the frames that need to be checked at query time. We conclude that the techniques of {\focus} are general and can be applied to a wide range of object detection \cnns and camera types. }

%This is because moving cameras require a detector \cnn to detect objects, which is much more expensive than doing so with background subtraction in static cameras. We can reduce the ingest cost of moving cameras by exploring even cheaper object detector \cnns. 

Averaging over both static and moving cameras, {\focus}' ingest cost is 
48$\times$ cheaper than \sys{Ingest-heavy}
and its queries are 125$\times$ faster than \sys{NoScope}.

\ko{We now take a deeper
 look at {\focus}' performance using representative static cameras.}

\ignore{Among these videos, the gains in
query latency are smaller for relatively less busy videos
(\video{bend}, \video{jacksonh}, \video{lausanne}, and
\video{oxford}). This is because these videos are dominated by fewer
object classes, and {\focus} has more work (i.e., analysis using
GT-CNN) to do at query time for these classes. We conclude that the
core techniques of {\focus} are general and effective on a variety of
real-world videos.}

\ignore{ % Outline
\begin{itemize}
  \item We strike a good balance between ingest cost and query
    latency. By investing very small cost at ingest time (up to 208X
    cheaper than the ingest-all baseline), we can save query latency
    by up to 69X
  \item We show that such good balance can be achieved across various
    videos. We can get good results even when the videos have many
    different classes of objects (CNN / Fox News) or very busy
    (Jackson Hole)
\end{itemize}
}

\subsection{Effect of Different {\focusSec} Components}
\label{sec:performance_by_technique}

Figure~\ref{fig:breakdown_by_technique} shows the breakdown of query
latency gains for two core techniques of {\focus}: (1) \sys{Approximate
  indexing}, which indexes each object with the top-{\sf K} results of
the ingest-time \cnn, and (2) \sys{Approximate indexing + Clustering},
which adds feature-based clustering at ingest time to reduce redundant
work at query time. We show the results that achieve at least 99\%
recall and precision. We make two observations.

\begin{figure}[h]
  \centering
  \includegraphics[width=0.8\textwidth]{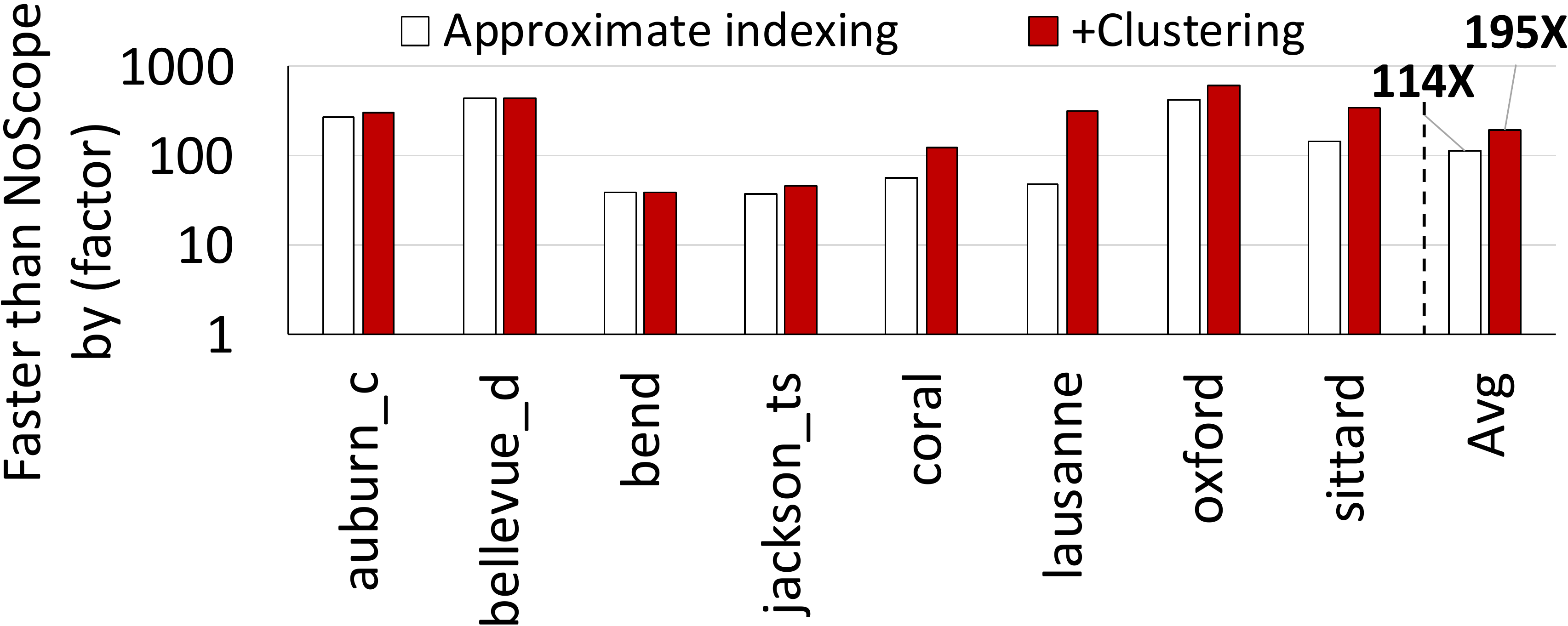}
  \caption{Effect of different {\focus} components on query latency
    reduction}
  \label{fig:breakdown_by_technique}
\end{figure}

First, approximate indexing is the major source of query latency
improvement. This is because approximate indexing effectively \kff{eliminates}
irrelevant objects for each query and bypasses the query-time
verification for high-confidence ingest predictions. As a result, only
a small fraction of frames need to be resolved at query time. On
average, approximate indexing alone is 114$\times$ faster than
{\noscope} in query latency.

%First, while pixel differencing can help both ingest cost and query
%latency, it is not the major source of improvement. This is because
%our baselines run the GT-CNN model \emph{only} on moving objects, and
%\emph{do not} spend time on the uneventful frames. As a result, one of the
%major benefits of pixel differencing, which is to eliminate idle
%frames, is already included in our baselines.

%First, generic compressed models provide benefits for both ingest
%cost and query latency, but they are not the major source of
%improvement. This is because the accuracy of a generic compressed model
%degrades significantly when we remove convolutional layers. In
%order to retain the recall/precision target, we need to choose relatively
%expensive compressed models (CheapCNN$_i$) and a larger {\sf K}, which incur higher
%ingest cost and query latency.%, respectively.
%
%Second, specializing the model (in addition to compressing it) greatly
%reduces ingest cost and query latency. Because of fewer convolutional
%layers and smaller input resolution, our specialized models are
%7$\times$ to 105$\times$ cheaper than the GT-CNN, while retaining the
%recall/precision target for each video streams. Running a specialized model at ingest
%time speeds up query latency by 12$\times$ to 32$\times$
%(Figure~\ref{fig:query_breakdown_by_technique}).

Second, clustering is a very effective technique to further reduce
query latency. Using clustering (on top of approximate indexing)
reduces the query latency by 195$\times$, significantly better than
approximate indexing alone. \ko{We see that clustering is especially
  effective on surveillance \kff{videos} (e.g., \video{coral},
  \video{lausanne}, and \video{oxford}) because objects in these
  videos tend to stay longer in the camera (e.g., ``person'' on a
  plaza compared to ``car'' in traffic videos), and hence there is
  more redundancy in these videos.}  This gain comes with a negligible
cost \kff{because we run our clustering algorithm
  (\xref{subsec:redundancy}) on the otherwise idle CPUs of the ingest machine
  while the GPUs run the ingest-time \cnn model.}

%\begin{figure}[t!]
% \centering
% \begin{subfigure}[t]{0.50\linewidth}
% \centering
% \includegraphics[width=1.0\textwidth]{focus/Figures/breakdown_ingest.pdf}
% \caption{Ingest cost}
% \label{fig:ingest_breakdown_by_technique}
% \end{subfigure}
% \begin{subfigure}[t]{0.48\linewidth}
% \centering
% \includegraphics[width=1.0\textwidth]{focus/Figures/breakdown_query.pdf}
% \caption{Query latency}
% \label{fig:query_breakdown_by_technique}
% \end{subfigure}
% \caption{Effect of different {\focus} components}
% \label{fig:breakdown_by_technique}
%\end{figure}

\ignore{ %outline
We will show two figures (ingest and query improvement) to break it
down:

For both figures, we show the results for
\begin{enumerate}
\item baseline
\item pixel-diff only
\item pixel-diff + generic cheap model + clustering
\item pixel-diff + specialized cheap model
\item pixel-diff + specialized cheap model + clustering
\end{enumerate}

Key takeaways:
\begin{itemize}
  \item Pixel-diff can help both ingest and query, but it's not the
    major improvement source. It is different from prior work because our
    baseline has background subtraction already, so frames without any
    object are not considered even for the baseline.
  \item Specialization is a very important technique. It is because
    specialized models only need to distinguish the dominant classes
    in a specific video stream, so it can achieve similar accuracy
    with fewer layers and smaller input size.
  \item Clustering is a very useful technique to remove the
    redundancy, and it is relatively cheap to do.
\end{itemize}

}

\subsection{Ingest Cost vs. Query Latency Trade-off}
\label{sec:ingest_query_trade_off}

%%%%%%
One of the \kff{important} features of {\focus} is the flexibility to tune
its system parameters to achieve different application goals
(\xref{subsec:tuning}). Figure~\ref{fig:result_trade_off_zoom} (the
zoom-in region of Figure~\ref{fig:result_trade_off}) depicts three
alternative settings for {\focus} that illustrate the trade-off space
between ingest cost and query latency, using the \video{oxford} video
stream: (1) \sys{{\focusTxt}-Opt-Query}, which optimizes for query
latency by increasing ingest cost, (2) \sys{{\focusTxt}-Balance}, which
is the default option that balances these two metrics
(\xref{subsec:tuning}), and (3): \sys{{\focusTxt}-Opt-Ingest}, which is
the opposite of \sys{{\focusTxt}-Opt-Query}.  The results are shown
relative to the \sys{Ingest-heavy} and
{\noscope} baselines. Each data label $(I, Q)$ indicates its ingest cost is
$I\times$ cheaper than \sys{Ingest-heavy}, while its query latency is
$Q\times$ faster than \sys{NoScope}.%\sv{Bring back just the zoomed in
%figure ?}
%%%%%%

\begin{figure}[h]
  \centering
  \includegraphics[width=0.70\textwidth]{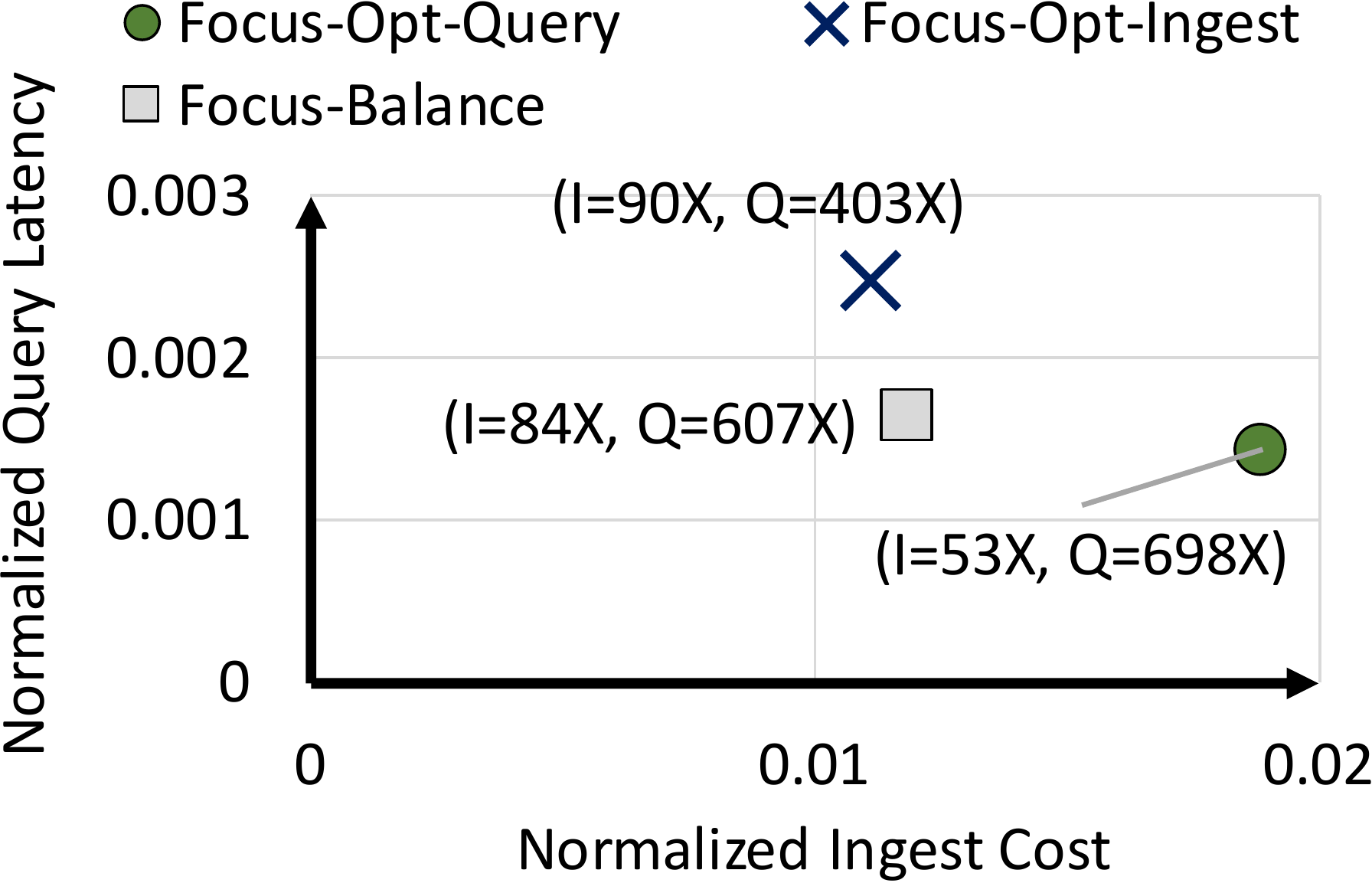}
  \caption{{\focus}' trade-off policies on an example video}
  \label{fig:result_trade_off_zoom}
\end{figure}

%%%%%% Text from before moving figure to the Intro
%%% One of the interesting features of {\focus} is the flexibility to tune
%%% its system parameters to achieve different application goals (\xref{subsec:tuning}).
%% To illustrate the trade-off space between ingest cost and query latency,
%% Figure~\ref{fig:result_trade_off} shows such trade-offs for one of our
%% video streams, \video{auburn\_c}. We plot three configurations of {\focus}:
%% (1) \sys{{\focusTxt}-Opt-Query}, which optimizes for query latency by
%% increasing ingest cost, (2) \sys{{\focusTxt}-Balance}, which is the
%% default option that balances these two metrics (\xref{subsec:tuning}),
%% and (3): \sys{{\focusTxt}-Opt-Ingest}, which is the opposite of
%% \sys{{\focusTxt}-Opt-Query}. We also plot the two baselines for
%% comparison. The chart at the right is the zoomed-in region that covers
%% the three configurations of {\focus}, and each data label $(I, Q)$ indicates
%% its ingest cost is $I\times$ cheaper than \sys{Ingest-heavy}, while its
%% query latency is $Q\times$ faster than \sys{NoScope}.

%% \begin{figure}[h]
%%   \centering
%%   \includegraphics[width=0.44\textwidth]{focus/Figures/result_trade_off.pdf}
%%   \caption{Trade-offs between query latency and ingest cost for
%%     \video{auburn\_c}}
%%   \label{fig:result_trade_off}
%% \end{figure}
%%%%%%%%%%%%%%%%%%%%%%%%%%%%

As Figure~\ref{fig:result_trade_off_zoom} shows, {\focus} offers very
good options in the trade-off space between ingest cost and query
latency. \sys{{\focusTxt}-Opt-Ingest} is 90$\times$ cheaper than
\sys{Ingest-heavy}, and makes the query 403$\times$ faster than a
query-optimized system (\sys{NoScope}). On the other hand,
\sys{{\focusTxt}-Opt-Query} reduces query latency \kff{even more (by
  698$\times$)} but it is still 53$\times$ cheaper than
\sys{Ingest-heavy}. As \kff{these points in the design space} are all good options compared to the
baselines, such flexibility enables a user to tailor {\focus} for
different contexts. For example, a camera that requires fast
turnaround time for queries can use \sys{{\focusTxt}-Opt-Query}, while
a video stream that will be queried rarely would choose
\sys{{\focusTxt}-Opt-Ingest} to reduce the amount of wasted ingest cost
in exchange for longer query latencies.

% if the vast majority of the videos
%are unlikely to be queried.

Figure~\ref{fig:result_trade_off_all} shows the $(I, Q)$ values for
both \sys{{\focusTxt}-Opt-Ingest} (\sys{Opt-I}) and
\sys{{\focusTxt}-Opt-Query} (\sys{Opt-Q}) for the representative
videos. As the figure shows, the flexibility to make different
trade-offs exists in most other videos. On average,
\sys{{\focusTxt}-Opt-Ingest} is 65$\times$ (up to 96$\times$) cheaper
than \sys{Ingest-heavy} in ingest cost while providing 100$\times$ (up
to 443$\times$) faster queries. \sys{{\focusTxt}-Opt-Query} makes
queries 202$\times$ (up to 698$\times$) faster with a higher ingest
cost (53$\times$ cheaper than \sys{Ingest-heavy}). \kf{Note that there is
  no fundamental limitation on the spread between
  \sys{{\focusTxt}-Opt-Query} and \sys{{\focusTxt}-Opt-Ingest} as we can
  expand the search space for ingest-time CNNs to further optimize
  ingest cost at the expense of query latency (or vice versa).} We conclude that
{\focus} enables flexibly optimizing for ingest cost or query latency for application's needs.
%and this makes {\focus} versatile \kff{in handling different videos}.
%\sv{Still no news videos ?}

\begin{figure}[h]
  \centering
  \includegraphics[width=0.9\textwidth]{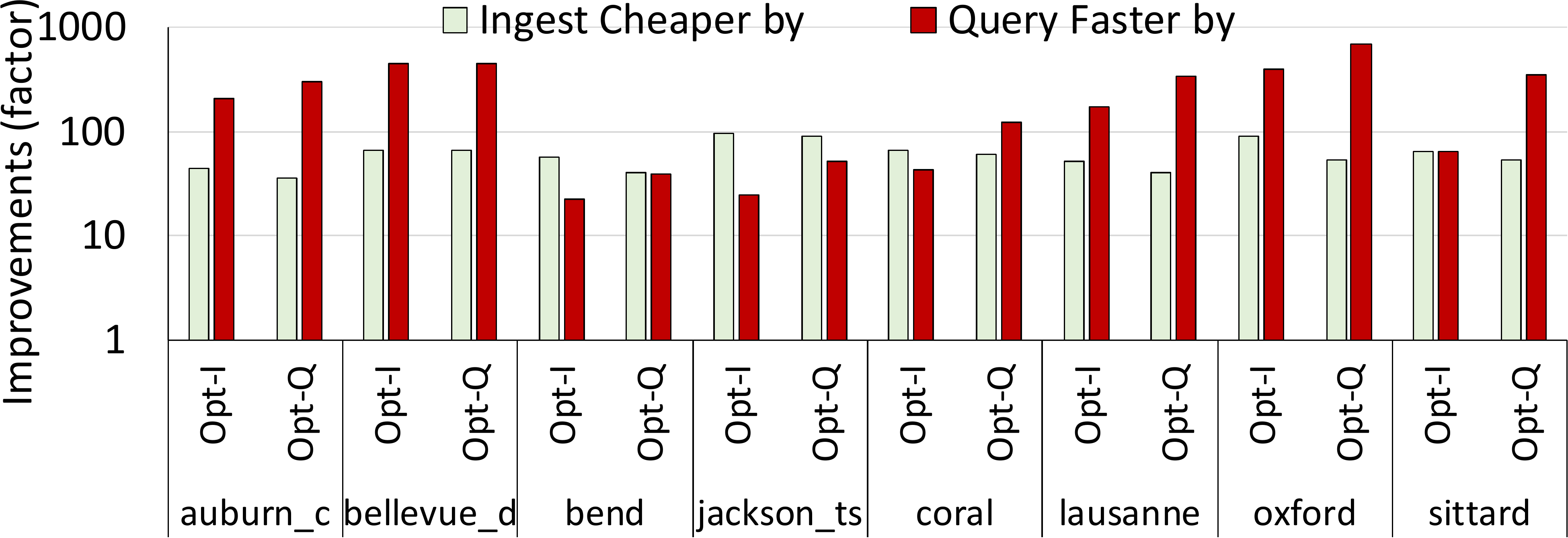}
  \caption{Ingest cost vs. query latency trade-off}
  \label{fig:result_trade_off_all}
\end{figure}

\ko{It is worth noting that the fraction of videos that get queried can
affect the applicability of {\focus}, especially in the case
where only a tiny fraction of videos gets queried. While \kff{\sys{{\focusTxt}-Opt-Ingest}} can
save the ingest cost by up to 96$\times$, it can be more costly than any purely query-time-only solution
if the fraction of videos that gets queried is less than $
\frac{1}{96} \approx 1\%$. In such a case, a user can still use
     {\focus} to significantly reduce query latency, but the
     cost of {\focus} can be higher than \kff{query-time-only solutions}.}

\ignore{
we can choose to do nothing at
ingest time and run all the techniques of {\focus} only at query time
when we know the fraction of videos that will get queried. While this
approach increases query latency, it is still faster than
\sys{NoScope} by a average of 23$\times$ (up to 40$\times$) in our
evaluation. We conclude that {\focus} is still better than both
baselines even under extreme query rates.\sv{We need more explanation
  here as to what we mean when we say we apply all our techniques at
  query time ?}  } %ignore

%\vspace{-10pt}

%
%\begin{figure}[t!]
% \centering
% \begin{subfigure}[t]{0.50\linewidth}
% \centering
% \includegraphics[width=1.0\textwidth]{focus/Figures/result_trade_off_opt_ingest.pdf}
% \caption{\sys{{\focusTxt}-Opt-Ingest}}
% \label{fig:result_tradeoff_opt_ingest}
% \end{subfigure}
% \begin{subfigure}[t]{0.48\linewidth}
% \centering
% \includegraphics[width=1.0\textwidth]{focus/Figures/result_trade_off_opt_query.pdf}
% \caption{\sys{{\focusTxt}-Opt-Query}}
% \label{fig:result_tradeoff_opt_query}
% \end{subfigure}
% \caption{Trade-offs between ingest cost and query latency for all videos}
% \label{fig:result_trade_off_all}
%\end{figure}
%

%outline
%  Show the trade-off space between ingest and query improvement like
%Figure~\ref{fig:ingest_query_tradeoff}. Discuss a few policies
%(optimize for ingest, balanced, optimized for query) and what's the
%results for each policy.}

\subsection{Sensitivity to Recall/Precision Target}
\label{sec:result_accuracy_target}

Figure~\ref{fig:result_accuracy} illustrates \kff{{\focus}' reduction in query
latency} compared to the baselines under different
recall/precision targets. Other than the default 99\% recall and
precision target, we evaluate both {\focus} and {\noscope} with two
lower targets, 97\% and 95\%.

%and
%compare the results with {\noscope} with respect to the same
%recall/precision targets.

\begin{figure}[h]
  \centering
  \includegraphics[width=0.8\textwidth]{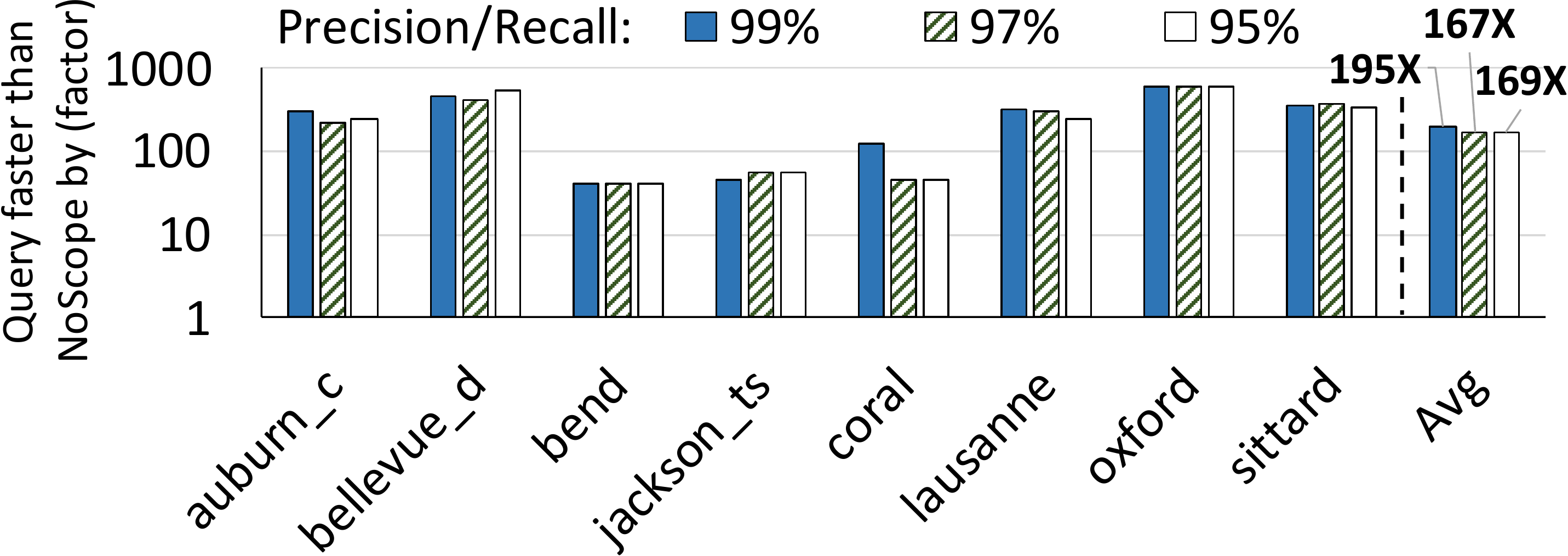}
  \caption{Sensitivity of query latency reduction to recall/precision target}
  \label{fig:result_accuracy}
\end{figure}

We observe that with lower accuracy targets, the \kff{query
latency improvement decreases slightly} for most videos, while the ingest cost
improvement does not change much (not graphed). The ingest cost is not
sensitive to the accuracy target because {\focus} still runs similar
ingest \cnns. {\noscope} can however apply more aggressive query-time optimization to reduce
query latency given lower accuracy targets. This decreases {\focus}' improvement over {\noscope} for several
videos. On average, {\focus} is faster than
\sys{NoScope} in query latency by 195$\times$, 167$\times$, and 169$\times$ with recall/precision of 
99\%, 97\%, and 95\%, respectively. We conclude that {\focus}'
techniques can achieve significant improvements on query
latency, irrespective of recall/precision targets.
%\sv{Does NoScope take precision recall targets ? Are
%these speedsups wrt to same baseline or different baselines ?}

%\begin{figure}[h]
%  \centering
%  \includegraphics[width=0.48\textwidth]{focus/Figures/result_accuracy_ingest.pdf}
%  \caption{Ingest cost sensitivity to accuracy target}

%  \label{fig:result_accuracy_ingest}
%\end{figure}

%\begin{figure}[h]
%  \centering
%  \mbox{}\\

\ignore{
\subsection{Sensitivity to Frame Sampling}
\label{sec:result_frame_sampling}

A common approach to reduce the video processing time is to use frame
sampling (i.e., periodically select a frame to process). However, not
all applications can use frame sampling because it can miss objects
that show up and disappear within a frame sampling window. As the
frame sampling rate is an application dependent choice, we study the
sensitivity of {\focus}' performance to different frame
rates. Figure~\ref{fig:result_frame_sample} shows the ingest cost and query
latency of {\focus} at different frame rates (i.e., 30 fps, 10 fps, 5 fps,
and 1 fps) compared to \sys{Ingest-heavy} and \sys{NoScope},
respectively. We make two observations.

\begin{figure}[h]
  \centering
  \includegraphics[width=0.48\textwidth]{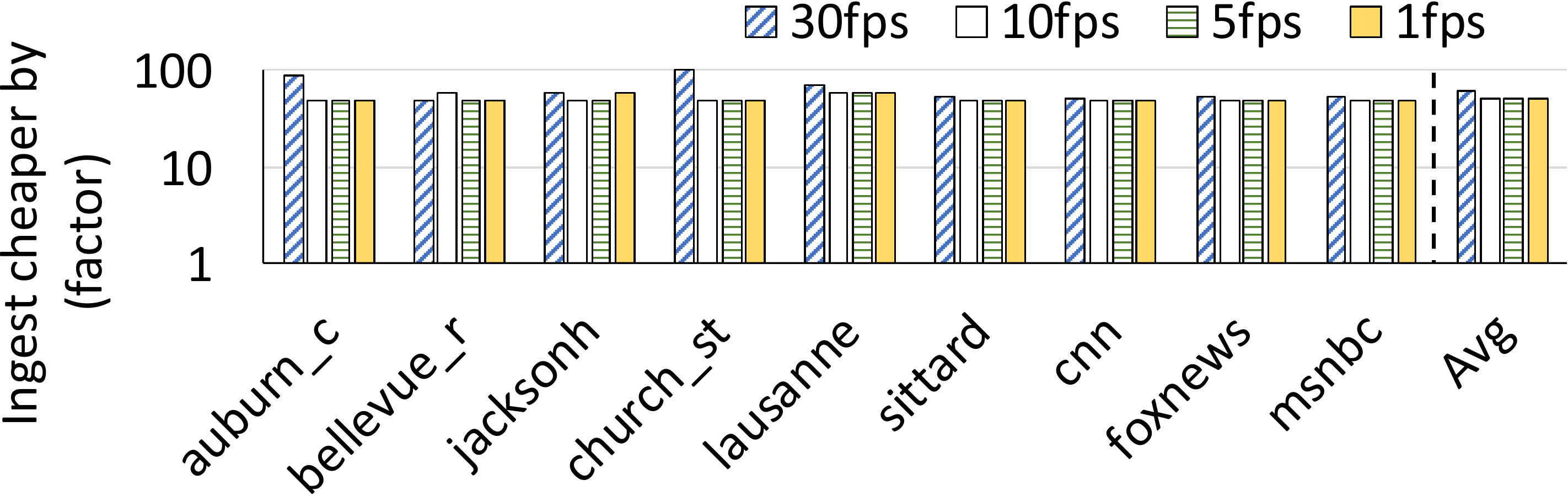}
%  \caption{Ingest cost sensitivity to frame sampling}
%  \label{fig:result_frame_sample_ingest}
%\end{figure}

%\vspace{-4ex}

%\begin{figure}[h]
%  \mbox{}\\
  \centering
  \includegraphics[width=0.48\textwidth]{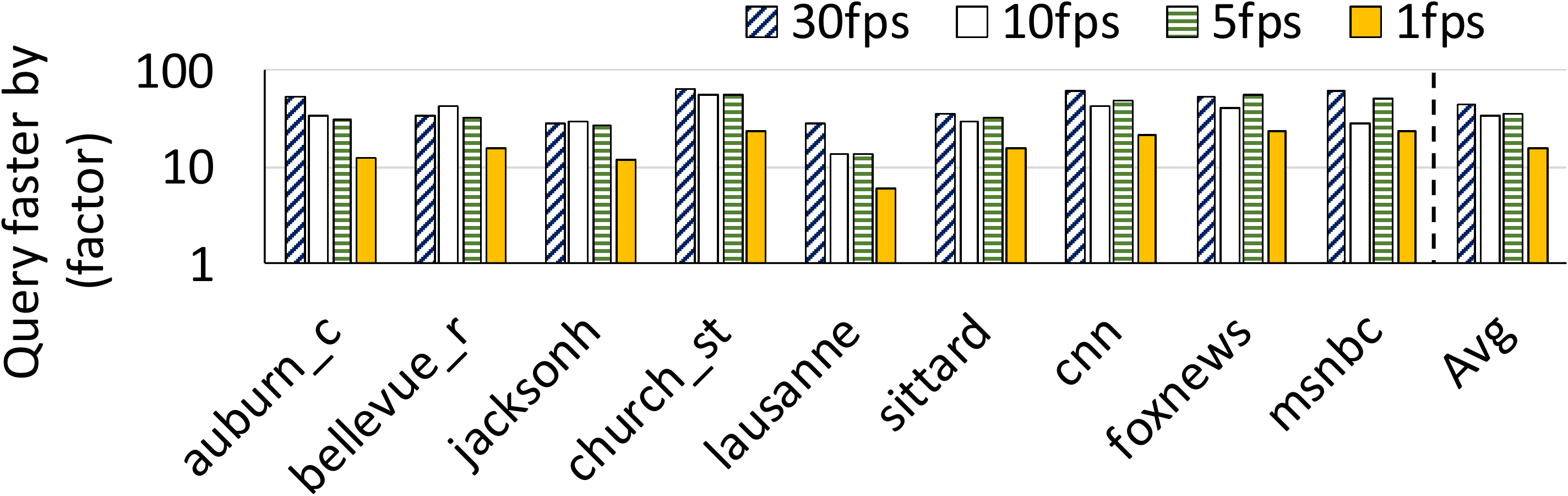}
  \caption{Ingest cost (top) and query latency (bottom) sensitivity to frame sampling}
  \label{fig:result_frame_sample}
\end{figure}

First, the ingest cost reduction is roughly the same across the different
frame rates. On average, the ingest cost of {\focus} is 63$\times$
cheaper than \sys{Ingest-heavy} at 30 fps, and it is 51$\times$ to
52$\times$ cheaper at lower frame rates. This is because the major
ingest cost saving comes from the specialized cheap \cnn
models (\xref{sec:performance_by_technique}), which are orthogonal to
frame rates.

Second, the query latency improvement of {\focus} degrades with lower
frame rates. This is expected because one of our key techniques to
reduce query latency is redundancy elimination, especially clustering
similar objects using \cnn feature vectors. At lower frame rates, the
benefit of this technique reduces because there are fewer
redundancies. Nonetheless, on average, {\focus} is still 16$\times$ faster than \sys{NoScope} at a very low frame rate (1 fps).

} %ignore

%\ignore{ %outline
%Show the results with different frame sampling rate (30fps, 10fps,
%5fps, 1fps, 0.5fps). It is an
%application choice and we show that our technique can improve
%performance across various frame rates.
%}

\ignore{
\subsection{Comparison Against NoScope}
\label{sec:noscope}

\kv{We also compare {\focus} to a recent work, \sys{NoScope}~\cite{DBLP:journals/pvldb/KangEABZ17}, which optimizes \cnn-based video queries. \sys{NoScope} employs a pixel difference detector to eliminate redundant frames and a query-specific binary classifier to short-circuit full-blown \cnn evaluation if the binary classifier is sufficiently confident. We download the \sys{NoScope} code and videos~\cite{noscope_repo} from their repository, and run both \sys{NoScope} and {\focus} on the same machine in our cluster (\xref{sec:methdology}). We evaluate both systems on the two videos and classes provided by \sys{NoScope} (``person'' for \video{coral-reef} and ``car'' for \video{jackson-town-square}). We run both systems at 1 fps as we observe that the \sys{NoScope} code is always configured to run at 1 fps. We compare both systems by picking their best configurations such that \emph{both} recall and precision meet various targets (i.e., 99\%, 97\%, and 95\%).\footnote{Note that using recall and precision is a more informative metric than using \emph{accuracy}, the major metric used in the \sys{NoScope} paper. Accuracy is the ratio of the correct positive/negative classifications to the total number of samples, and is often not very useful in real systems because of a phenomena called \emph{accuracy paradox}~\cite{accuracy_paradox, zhu2007knowledge, DBLP:conf/dmkdttt/Thomas008}. For example, if 95\% of the samples are ``negative'', a system that \emph{always} predicts ``negative'' will have 95\% accuracy, however its recall and precision will be 0\% when querying for ``positive''.}} We measure that {\focus} is faster and cheaper than \sys{NoScope}, with gains in latency up to $200\times$ for querying rarer object classes.

%We exclude fixed latency and cost (e.g., training stream-specific CNNs) from both systems as they are minor on long videos. \ga{Not necessary to bring it up?} 

\begin{figure}[t]
 \centering
 \begin{subfigure}[t]{0.52\linewidth}
 \centering
 \includegraphics[width=1.0\textwidth]{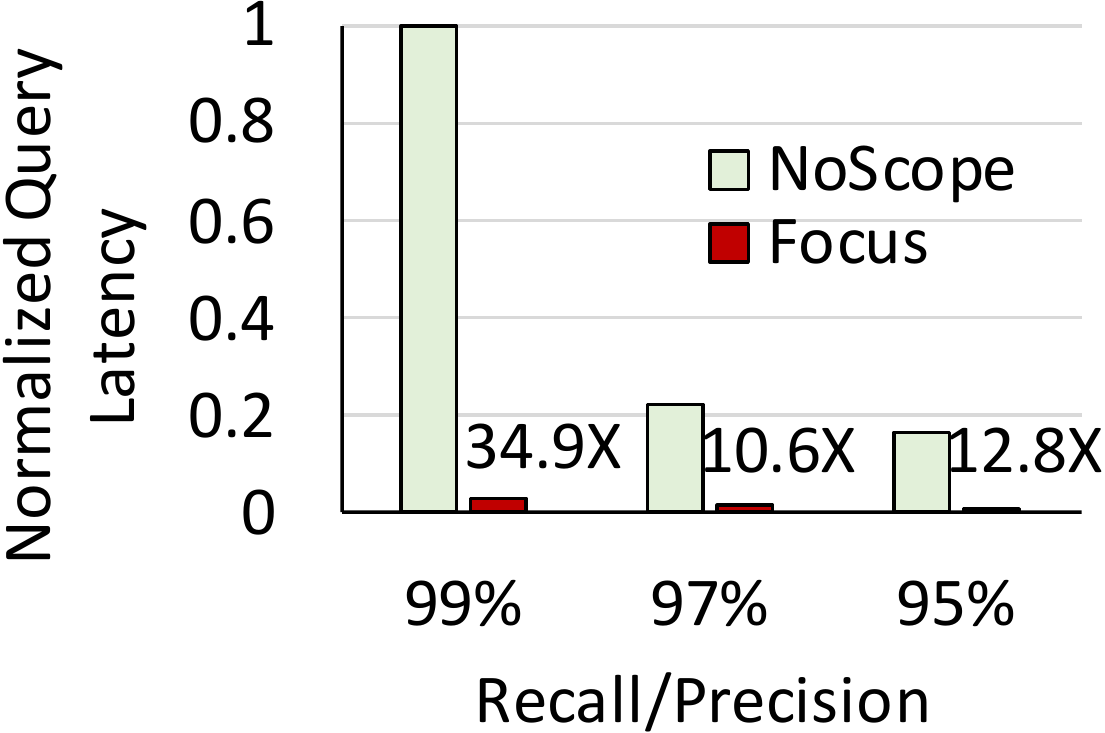}
 \caption{\video{jackson-town-square}}
 \label{fig:noscope_latency_jackson}
 \end{subfigure}
 \begin{subfigure}[t]{0.45\linewidth}
 \centering
 \includegraphics[width=1.0\textwidth]{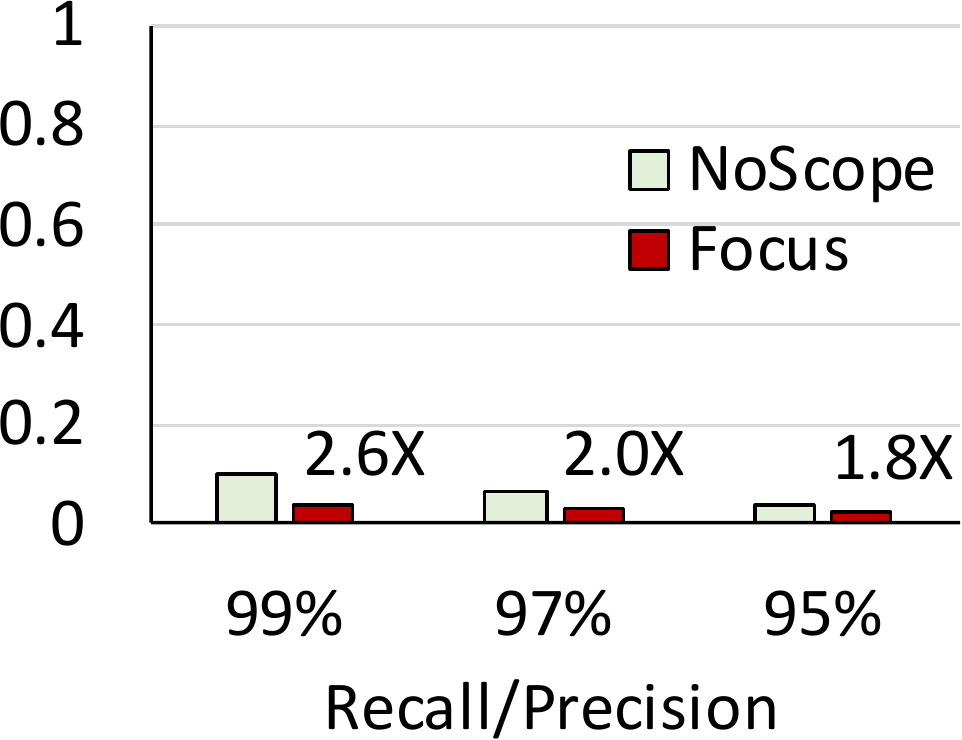}
 \caption{\video{coral-reef}}
 \label{fig:noscope_latency_coral}
 \end{subfigure}
  \caption{Query latency comparison between \sys{NoScope} and {\focus}. The data labels show the factor by which {\focus} is faster than \sys{NoScope} for the \emph{respective} recall/precision.}
 \label{fig:noscope_latency}
\end{figure}
\begin{figure}[t]
 \begin{subfigure}[t]{0.50\linewidth}
 \centering
 \includegraphics[width=1.0\textwidth]{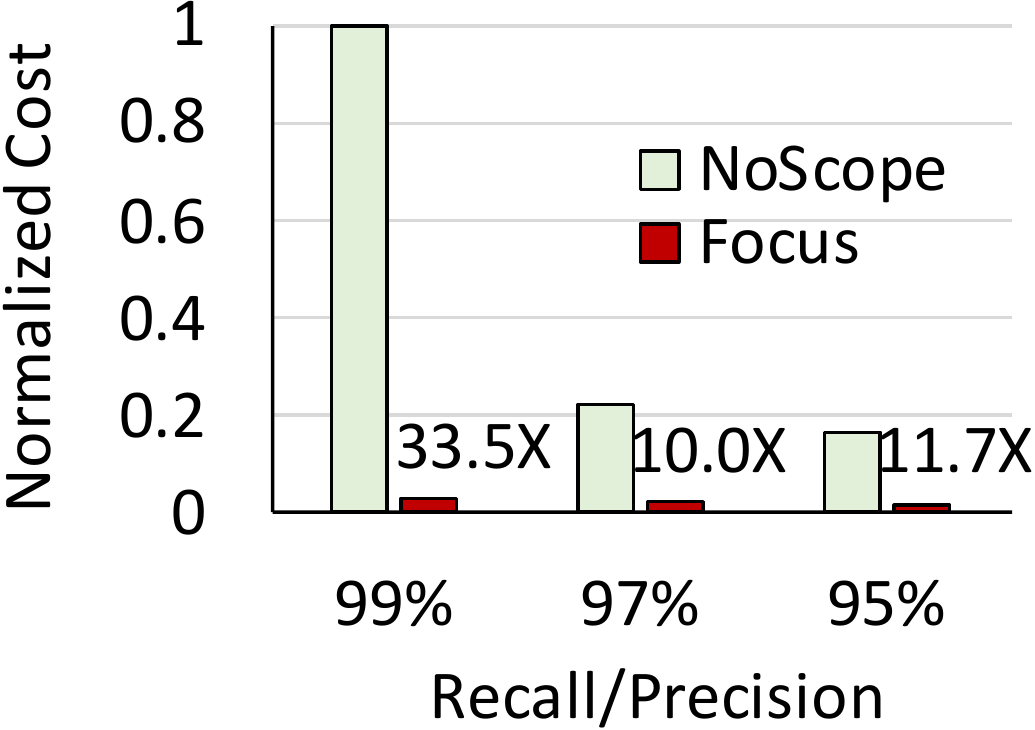}
 \caption{\video{jackson-town-square}}
 \label{fig:noscope_cost_jackson}
 \end{subfigure}
 \begin{subfigure}[t]{0.47\linewidth}
 \centering
 \includegraphics[width=1.0\textwidth]{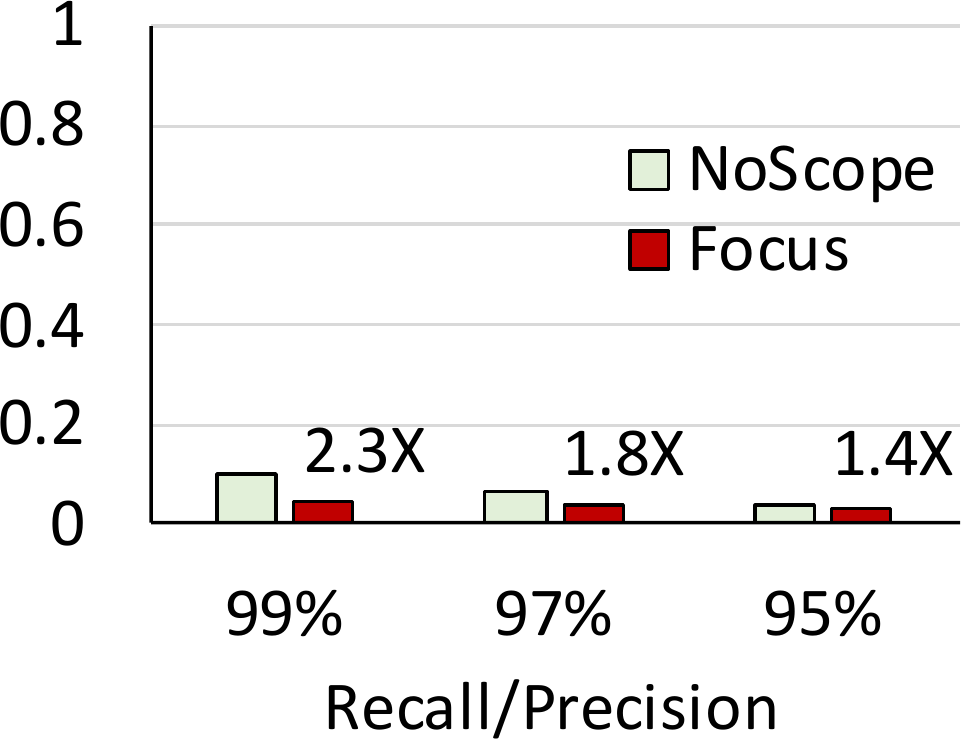}
 \caption{\video{coral-reef}}
 \label{fig:noscope_cost_coral}
 \end{subfigure}

 \caption{Total cost comparison between \sys{NoScope} and {\focus}. The data labels show the factor by which {\focus} is cheaper than \sys{NoScope} for the \emph{respective} recall/precision.}
 \label{fig:noscope_cost}
\end{figure}

%\begin{figure}[t]
% \centering
% \caption{Total cost comparison between \sys{NoScope} and {\focus}. The data labels show the factor by which {\focus} is cheaper than \sys{NoScope} for the \emph{respective} false negative/positive rates.}
% \label{fig:noscope_cost}
%\end{figure}

\kv{{\bf Query Latency:} Figure~\ref{fig:noscope_latency} depicts the query latency of the two systems, normalized to the latency of running YOLO~\cite{DBLP:journals/corr/RedmonF16} (the GT-CNN of \sys{NoScope}). Two observations are in order. }

\noindent{}\kv{1) {\focus} achieves 1.8$\times$ to 34.9$\times$ faster queries than \sys{NoScope} across various recall/precision targets. This is because {\focus}' motion estimation and ingest processing filter out many uneventful frames and irrelevant objects at ingest time, while \sys{NoScope} needs to process \emph{all} the frames at query time. This demonstrates the value of {\focus}' architecture of splitting work between ingest and query time.}

\noindent{}\kv{2) The gap between these two systems is much larger for \video{jackson-town-square} than for \video{coral-reef}. This is because there is only one class (``person'') showing up in \video{coral-reef}, which means querying on this video is essentially a binary classification problem. \sys{NoScope} is very effective on this video as its key techniques are designed for binary classification. In contrast, there are multiple classes (cars, trucks, and buses) in \video{jackson-town-square}, and \sys{NoScope} needs to invoke YOLO much more often to achieve the target recall/precision. We see that at 99\% recall/precision, \sys{NoScope} is almost as slow as always running YOLO. In comparison, {\focus} uses the top-{\sf K} index to ensure high recall at ingest time, while invoking GT-CNN at query time \emph{only} on the objects with matching indexes to achieve high precision. As a result, {\focus} outperforms \sys{NoScope} in query latency, especially for complicated videos with many object classes and higher precision/recall targets.}

  %\ga{Shouldn't this sentence be in the paragraph above that talks about the split arch.?} 

  {\bf Total Cost:} We also compare the total cost (machine cycles) of these two systems that includes all the work done by \sys{NoScope} and {\focus} (both ingest as well as query cost). Figure~\ref{fig:noscope_cost} shows the total cost for one query, normalized to the cost of running YOLO. We see that {\focus} is 1.4$\times$ to 33.5$\times$ cheaper than \sys{NoScope} across various recall/precision on both videos. This shows the ingest processing of {\focus} is very low cost (4\% to 17\% of {\focus}' total cost), which makes {\focus} competitive also at the overall resource cost. The cost gap will only increase if there are multiple distinct queries, since {\focus}' ingest cost is a one-time cost for all object classes. Note that the gap between {\focus} and \sys{NoScope} in query latency and cost will only grow at a higher frame rates, where {\focus}' object clustering is more effective  (\xref{sec:result_frame_sampling})
  % A key point to note is
  %\ga{Should we show the ingest/query split in cost?} \kh{I don't think it's required, but we can do that}
  %\ga{Define ``cost''?} 

\begin{figure}[t]
 \centering
 \begin{subfigure}[t]{0.52\linewidth}
 \centering
 \includegraphics[width=1.0\textwidth]{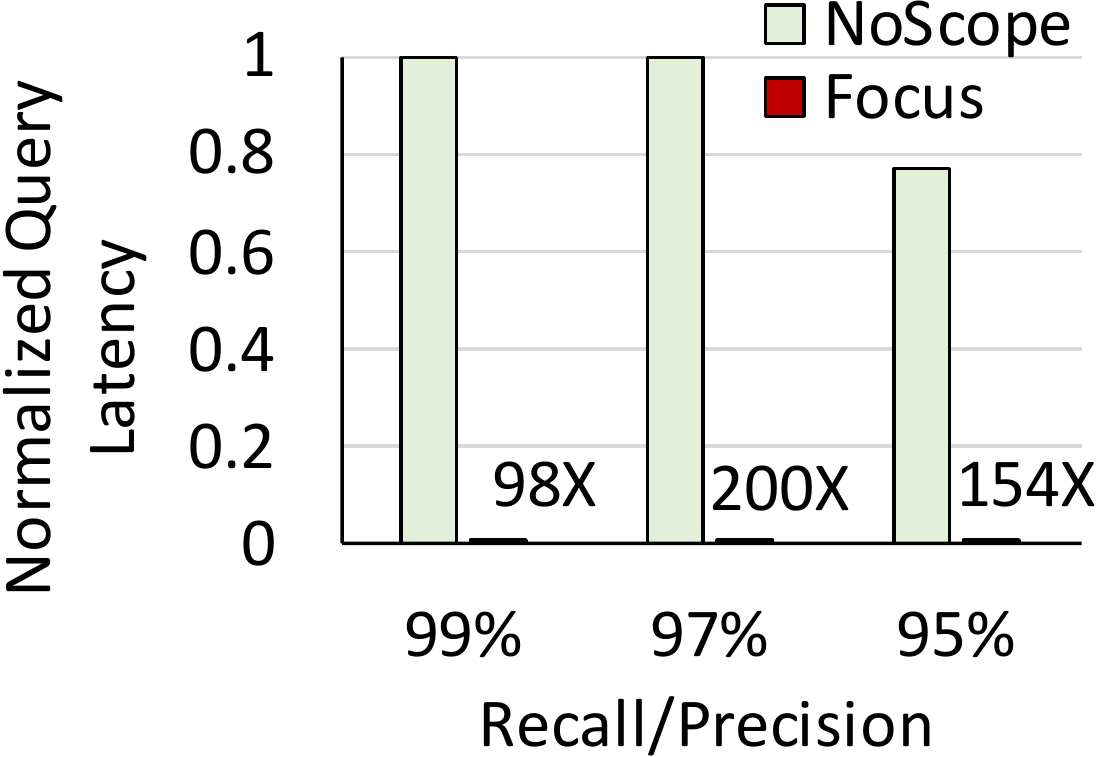}
 \caption{bus}
 \label{fig:noscope_latency_bus}
 \end{subfigure}
 \begin{subfigure}[t]{0.45\linewidth}
 \centering
 \includegraphics[width=1.0\textwidth]{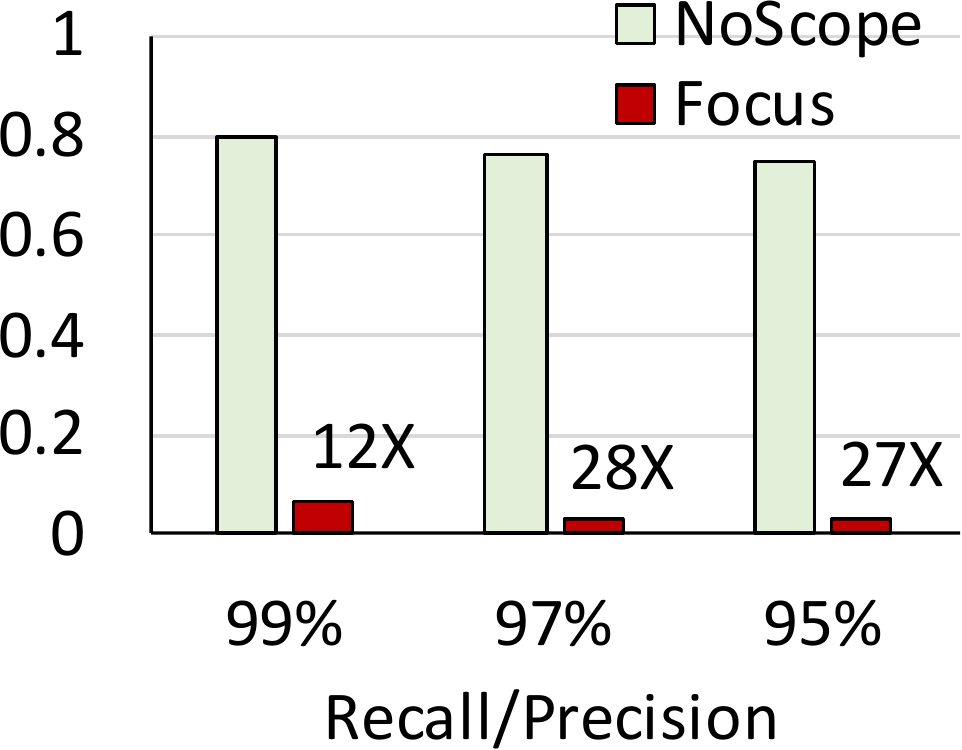}
 \caption{truck}
 \label{fig:noscope_latency_truck}
 \end{subfigure}
 \caption{Query latency comparison between \sys{NoScope} and {\focus} on less frequent classes in \video{jackson-town-square} }
 \label{fig:noscope_latency_bus_truck}
\end{figure}

%<<<<<<< HEAD
%\kv{{\bf Rare Object Classes:} A key architectural difference between \sys{NoScope} and {\focus} is that the query-specialized CNN of \sys{NoScope} does \emph{binary} classification at \emph{frame} level (e.g., if this frame has a car), while the ingest CNN of {\focus} indexes \emph{multiple} classes at \emph{object} level. This makes a large difference when a query in looking for less frequent object classes. Looking up less frequent classes is potentially a more useful application of video querying systems, e.g., finding something rare like a ``parachute'' in a traffic video is more interesting than finding something commonplace like ``car'' which will be most of the frames anyway.}
%=======
\kv{{\bf Rarer Object Classes:} A key architectural difference between \sys{NoScope} and {\focus} is that the query-specialized CNN of \sys{NoScope} does \emph{binary} classification at \emph{frame} level (e.g., if this frame has a car), while the ingest CNN of {\focus} indexes \emph{multiple} classes at \emph{object} level. This makes a large difference when a query is looking for a less frequent object class. Looking up less frequent classes is potentially a more useful application of video querying systems; e.g., finding something rare like an ``ambulance'' in a traffic video is more interesting than finding typical objects like ``car'' which will be present in most of the frames anyway.}
%>>>>>>> 2ce7f0ec08ef9d6cfa1e4fee389a3a625bb61767

\kv{Figure~\ref{fig:noscope_latency_bus_truck} shows the query latency on two less frequent classes (``bus'' and ``truck'') in \video{jackson-town-square}. As we see, even at 95\% recall/precision, the query latency of \sys{NoScope} does not improve much over the baseline (always run YOLO). This is because distinguishing similar classes is difficult at frame level. For less frequent classes, the confidence of \sys{NoScope}'s binary classifier is lower. As a result, \sys{NoScope} needs to invoke its GT-CNN much more often. In contrast, {\focus} performs more consistently across different classes, and achieves 12$\times$ to 200$\times$ speedup over \sys{NoScope}. We conclude that the architecture of {\focus} is much more effective than \sys{NoScope} in querying less frequent classes.%, which is potentially a more useful application than querying the most dominant class in a video.
}
} %ignore

%\ga{Fewer training samples is always a weak motivation...} 

%\subsection{Applicability with Moving Cameras}
%\label{sec:result_moving_cam}

\ignore{
\subsection{Applicability with Different Query Rate}
\label{sec:result_query_rate}

There are two factors that can affect the applicability of {\focus}: 1)
the number of classes that get queried over time and 2) the fraction
of videos that get queried. In the first extreme case where all the
classes and all the videos are queried, \sys{Ingest-heavy} could be a
good option because its cost is amortized among all the queries. In
our study, even in such an extreme case, the overall cost of {\focus}
is still 2$\times$ cheaper than \sys{Ingest-heavy} on average (up to
3$\times$ cheaper) because we run a very cheap \cnn at ingest time,
and we run GT-\cnn \emph{per object cluster} only \emph{once}
(\xref{sec:implementation}), so the overall cost is still cheaper than
\sys{Ingest-heavy}.

The second extreme case is only a tiny fraction of videos gets
queried. While {\focus} can save the ingest cost by up to 193$\times$
(\xref{sec:ingest_query_trade_off}), it can be more costly than
\sys{NoScope} if the fraction of videos that gets queried is less than $
\frac{1}{193} = 0.5\%$. In such a case, we can choose to do nothing at
ingest time and run all the techniques of {\focus} only at query time
when we know the fraction of videos that will get queried. While this
approach increases query latency, it is still faster than \sys{NoScope} by
a average of 23$\times$ (up to 40$\times$) in our evaluation. We
conclude that {\focus} is still better than both baselines even under
extreme query rates.\sv{We need more explanation here as to what we mean
when we say we apply all our techniques at query time ?}
} %ignore

\ignore{Show two graphs here. One is what if the number of query increases and
cover more classes. Are we still better than the ingest-all baseline?
(Because it can amortize the cost). The other one is what if the query
only happens to a very small fraction of videos. At which point our
technique would be worse than the query-all baseline? }

\subsection{Sensitivity to Object Class Numbers}
\label{sec:result_class_num}

\kf{We use the 1000 object classes in the ImageNet dataset~\cite{ILSVRC15} to study the sensitivity of {\focus}' performance to the number of object classes (compared to the 80 default object classes in the COCO~\cite{DBLP:conf/eccv/LinMBHPRDZ14} dataset). Our result shows that {\focus} is 15$\times$ faster (on average) in query latency and 57$\times$ cheaper (on average) in ingest cost than the baseline systems, while achieving 99\% recall and precision. We observe that the query latency improvements with 1000 object classes is lower than the ones with 80 object classes. The reason is that ingest-time \cnns are less accurate on more object classes, and we need to select a larger {\sf K} to achieve the target recall. Nonetheless, the improvements of {\focus} are robust with more object classes as {\focus} is over one order of magnitude faster than the baseline systems when differentiating 1000 object classes.}

\ignore{
\subsection{Discussions (not part of the outline)}
when computing average improvement, do we assume that each class is
equally likely to show up in a query? what if there's a different
distribution? what gains do we get then?

\kh{yes, we assume each class is equal. we can show the variation
  among all classes}

\subsection{Misc}
\begin{itemize}
\item show that we can properly deal with the ``other'' class
\item show distribution ``across'' classes, not just average. is it
  possible that some classes do much worse than others? how to fix
  this? \kh{It's possible, but I only observe this in 5-min videos. In
  1-hour videos, it seems almost all classes have similar improvements}
\item show that by tuning the various knobs we can achieve other tradeoffs in the design space
\item design space relative to ingest cost and query rate
\item show that feature clustering is much better than pixel-level clustering/dedup
\end{itemize}
}

\section{Other Applications}
\label{sec:other_use}

\kf{Applications that leverage \cnns to process large and continuously growing data share similar challenges as {\focus}. Examples of such applications are:}

%By flexibly and effectively dividing the work between ingest time and query time, we show that {\focus}'s architecture provides low-cost and low-latency queries on large video datasets. 

%similar ideas can also be applied to applications that . We briefly discuss two such opportunities.}

\noindent\kf{{\bf 1) Video and audio.} Other than querying for objects, many emerging video applications are also based on \cnns, such as event detection (e.g.,~\cite{DBLP:conf/cvpr/XuYH15}), emotion recognition (e.g.,~\cite{DBLP:conf/icmi/KanouPBFGMVCBFMJCDBAZLRDPWTSBKW13}), video classification (e.g.,~\cite{DBLP:conf/cvpr/KarpathyTSLSF14}), and face recognition (e.g.,~\cite{DBLP:conf/cvpr/SchroffKP15}). Audio applications such as speech recognition (e.g.,~\cite{DBLP:conf/icassp/Abdel-HamidMJP12}) are also based on \cnns.}

%As video and audio data keep growing, using {\focus}'s principle to generate approximate indexes and eliminate redundancy at ingest time is a promising approach to build an efficient query system for these applications.}

\noindent\kf{{\bf 2) Bioinformatics and geoinformatics.} Many bioinformatics and geoinformatics systems leverage \cnns to process a large dataset, such as anomaly classification in biomedical imaging (e.g.,~\cite{DBLP:conf/icarcv/LiCWZFC14, DBLP:journals/tmi/RothLLYSCKS16}), \kff{information decoding} in biomedical signal \kff{recordings} (e.g.,~\cite{DBLP:conf/nips/StoberCG14}), and pattern recognition in satellite imagery (e.g.,~\cite{DBLP:conf/kdd/AlbertKG17, DBLP:journals/tgrs/ChengWXWXP17}).} 

%These applications share similar challenges and characteristics with {\focus}, and thus building cheap and approximate indexes can potentially make these applications much more effective on large-scale data.}

\kf{Naturally, these applications need to answer user-specified queries, such as ``find all brain signal recordings with a particular perception'' or ``find all audio recordings with a particular keyword''. Supporting these queries faces similar challenges to {\focus}, as a system either: (i) generates a precise index at ingest time, which incurs \emph{high cost}; or (ii) does most of the heavy-lifting at query time, which results in \emph{high query latency}. Hence, {\focus}' architecture offers a low-cost and low-latency option: building an approximate index with cheap \cnns at ingest time and \kff{generating precise results based on the approximate index} at query time. While the indexing structure may need to be adapted to different applications, \kff{we believe} {\focus}' architecture and techniques can benefit many of these emerging applications.}

\section{Summary}
\label{sec:conclusion}

Answering queries of the form, {\em find me frames that contain objects of class X}, is an important workload on recorded video datasets. Such queries are used by analysts and investigators \kff{for various immediate purposes}, and it is crucial to answer them with low latency and low cost. This chapter presents {\focus}, a system that flexibly divides the query processing work between ingest time and query time. {\focus} performs low-cost ingest-time analytics on live video that later facilitates low-latency queries on the recorded videos.
%{\focus} uses compressed and specialized \cnns at ingest-time that substantially reduces cost. It also clusters similar objects to reduce the work done at query-time, and hence the latency. {\focus} selects the ingest-time \cnn and its parameters to smartly trade-off between the ingest-time cost and query-time latency.
At ingest time, {\focus} uses cheap \cnns to construct an \emph{approximate index} of all possible object classes in each frame to retain high recall. At query time, {\focus} leverages this approximate index to provide low latency, but compensates for the lower precision by judiciously using expensive \cnns. This architecture enables orders-of-magnitude faster queries with only a small investment at ingest time, and allows flexibly trading off ingest cost and query latency. 
Our evaluations using \kff{real-world videos} from traffic, surveillance, and news domains show that {\focus} reduces ingest cost on average by $48\times$ (up to $92\times$) and makes queries \kff{on average} $125\times$ (up to $607\times$) faster compared to state-of-the-art baselines. We conclude that {\focus}' architecture and techniques make it a highly practical and effective approach to querying large video datasets. \kf{We hope that the ideas and insights behind {\focus} can be applied to designing efficient systems for many other forms of querying on large and continuously-growing datasets in many domains, \kff{such as audio, bioinformatics, and geoinformatics.}}

%many other forms of video querying. %We conclude that {\focus} is a promising approach to querying large video datasets, and hope that it will enable future works on better determining the ingest-time and query-time trade-offs.% in video querying systems.
%Our next steps include training a specialized and highly accurate query-time \cnn for each stream and object to further reduce query latency. 

\chapter{ML Training over Geo-Distributed Data}
\label{ch:gaia}

As Chapter~\xref{ch:introduction} discusses, many ML applications
analyze massive amounts of data from user activities, pictures,
videos, etc., which are generated at very \khi{rapid rates, \emph{all
    over the world}}.  Many large organizations, such as
Google~\cite{google_dc}, Microsoft~\cite{microsoft_dc}, and
Amazon~\cite{amazon_dc}, operate tens of data centers globally to
minimize their service latency to end-users, and store massive
quantities of data all over the
globe~\cite{DBLP:conf/nsdi/VulimiriCGJPV15,
  DBLP:conf/nsdi/RabkinASPF14, DBLP:conf/sigcomm/PuABKABS15,
  DBLP:conf/sosp/WuBPKM13, clarinet,
  DBLP:journals/pvldb/GuptaYGKCLWDKABHCSJSGVA14,
  DBLP:conf/hpdc/HeintzCS15, DBLP:conf/cloud/HungGY15,
  DBLP:journals/pvldb/KloudasRPM15, DBLP:conf/cidr/VulimiriCGKV15}.

A \khi{commonly-used} approach to run an ML application over \khi{such
  rapidly generated data is to \emph{centralize} all data} into
\emph{one data center} over wide-area networks \khi{(WANs)} before
running the ML application~\cite{DBLP:journals/corr/CanoWMCF16,
  DBLP:conf/sigmod/ThusooSABJSML10, DBLP:journals/pvldb/LeeLLLR12,
  DBLP:conf/icde/AuradkarBDMFGGGGHKKKLNNPPQQRSSSSSSSSTTVWWZZ12}. However,
this approach can be prohibitively difficult because: (1) WAN
bandwidth is a scarce resource, and hence moving \khi{all data} can be
extremely slow~\cite{DBLP:conf/sigcomm/PuABKABS15,
  DBLP:journals/corr/CanoWMCF16}. Furthermore, the fast growing rate
of image and video generation will eventually saturate the total WAN
bandwidth, whose growth has been decelerating for many
years~\cite{GIG, DBLP:conf/nsdi/VulimiriCGJPV15}. (2) Privacy and data
sovereignty laws in some countries prohibit transmission of \emph{raw
  data} across national or continental
borders~\cite{DBLP:conf/nsdi/VulimiriCGJPV15,
  DBLP:conf/cidr/VulimiriCGKV15, DBLP:journals/corr/CanoWMCF16}.

This motivates the need to \emph{distribute} an ML system across
\emph{multiple data centers}, globally. In such a system, \khi{large
amounts} of raw data are stored locally in different data centers, and
the ML algorithms running over the distributed data communicate between
data centers using WANs.
%Many large-scale distributed ML systems exist
%today that can coordinate and communicate across many machines to run
%an ML algorithm over the network~\cite{DBLP:conf/osdi/ChilimbiSAK14,
%  XingHDKWLZXKY15, DBLP:journals/corr/MengBYSVLFTAOXX15, LowGKBGH12,
%  tensorflow2015-whitepaper, DBLP:conf/osdi/LiAPSAJLSS14}. However,
%all these systems assume that the distribution happens only
%\emph{within} a single data center over a local-area network (LAN),
%and hence are not designed to tolerate the scarce WAN bandwidth.
Unfortunately, existing large-scale distributed ML
systems~\cite{DBLP:conf/osdi/ChilimbiSAK14, XingHDKWLZXKY15,
  DBLP:journals/corr/MengBYSVLFTAOXX15, LowGKBGH12,
  tensorflow2015-whitepaper, DBLP:conf/osdi/LiAPSAJLSS14} are
\khi{suitable} only for data residing \emph{within} a single data
center.  Our experiments using three state-of-the-art distributed ML
systems (B\"{o}sen~\cite{DBLP:conf/cloud/WeiDQHCGGGX15},
IterStore~\cite{CuiTWXDHHGGGX14}, and
GeePS~\cite{DBLP:conf/eurosys/CuiZGGX16}) show that operating these
systems across as few as two data centers (over WANs) can cause a
slowdown of 1.8--53.7$\times$ (see
Section~\ref{subsec:ml_system_on_wan} and
Section~\ref{sec:evaluation}) relative to their performance within a
data center (over LANs).  Existing systems that do address challenges
in geo-distributed data
analytics~\cite{DBLP:conf/nsdi/VulimiriCGJPV15,
  DBLP:conf/nsdi/RabkinASPF14, DBLP:conf/sigcomm/PuABKABS15, clarinet,
  DBLP:conf/hpdc/HeintzCS15, DBLP:conf/cloud/HungGY15,
  DBLP:journals/pvldb/KloudasRPM15, DBLP:journals/corr/CanoWMCF16,
  DBLP:conf/cidr/VulimiriCGKV15} do \emph{not} consider the broad
class of important, sophisticated ML algorithms commonly run on ML
systems \khi{--- they focus instead on other types of computation,
  e.g., map-reduce or SQL}.

%Several prior works have proposed data analytics frameworks that span
%globally-distributed data centers to address the above challenges in moving raw
%data to a centralized location~\cite{DBLP:conf/nsdi/VulimiriCGJPV15,
%  DBLP:conf/nsdi/RabkinASPF14, DBLP:conf/sigcomm/PuABKABS15,
% clarinet,DBLP:journals/corr/CanoWMCF16}, but~\cite{DBLP:journals/corr/CanoWMCF16} is the only work to tackle
%  running an \emph{ML system} on geographically
%distributed data. They propose a new approach, Geo-Distributed Machine
%Learning (GDML), that leverages a communication-efficient
%algorithm~\cite{DBLP:journals/corr/MahajanKSB13} for logistic
%regression models. This solution, however, is highly specific to a certain class
%of ML algorithms (linear models), and requires the ML programmer to alter the
%algorithm itself to make it WAN efficient. Thus, it is important to explore new
%solutions that are \emph{general}, and hence applicable, to a wide variety of ML algorithms,
%without requiring any change to the ML algorithms or program.

\textbf{Our goal} in this chapter is to develop a geo-distributed ML
system that (1) minimizes communication over \khi{WANs}, so \khi{that
  the system is not} bottlenecked by the scarce WAN bandwidth; and (2)
is general enough to be applicable to a wide variety of ML algorithms,
without \khi{requiring any} changes to the algorithms themselves.

To achieve these goals, such a system needs to address two key
challenges. First, to efficiently utilize the limited (and
heterogeneous) WAN bandwidth, we need to find an effective
communication model that minimizes communication over WANs but still
retains the \khii{correctness guarantee} for an ML algorithm. This is
difficult because ML algorithms typically require extensive
communication to exchange updates that keep the global ML model
sufficiently consistent across data centers.  These updates are
required to be timely, irrespective of the available network
bandwidth, to ensure algorithm correctness. Second, we need to design
a \emph{general} system\ignore{with a communication model} that
effectively handles WAN communication for ML algorithms without
requiring any \khii{algorithm changes}.  This is challenging because
the communication patterns vary significantly across different ML
algorithms~\cite{DBLP:conf/icml/TakacBRS13,
  DBLP:conf/nips/TsianosLR12, DBLP:journals/corr/RichtarikT13,
  DBLP:conf/nips/JaggiSTTKHJ14, NewmanASW09, SmolaN10}.  Altering the
communication across systems can lead to different tradeoffs and
consequences for different
algorithms~\cite{DBLP:conf/nips/ZinkevichSL09}.

In this chapter, we introduce \emph{{\gaia}}, a new general,
geo-distributed ML system that is designed to efficiently operate over
a collection of data centers. {\gaia} builds on the widely used
\emph{parameter server} architecture (e.g.,~\cite{HoCCLKGGGX13,
  CuiTWXDHHGGGX14, XingHDKWLZXKY15, CuiCHKLKWDGGGX14,
  DBLP:conf/osdi/LiAPSAJLSS14, DBLP:conf/cloud/WeiDQHCGGGX15,
  DBLP:conf/osdi/ChilimbiSAK14, tensorflow2015-whitepaper,
  AhmedAGNS12, DeanCMCDLMRSTYN12}) that provides ML worker machines
with a distributed global shared memory abstraction for the ML model
parameters \khi{they collectively train \khii{until}
  \emph{convergence} to fit the input data}. The key idea of {\gaia} is
to maintain an \khi{approximately-correct copy} of the global ML model
\emph{within} each data center, and dynamically eliminate any
unnecessary communication \emph{between} data centers. {\gaia} enables
this by \khi{\emph{decoupling}} the \synchronization (i.e.,
communication/consistency) model within \khi{a} data center from the
\synchronization model between \khi{different} data centers.  This
differentiation allows {\gaia} to run a conventional \synchronization
model~\cite{dai2015analysis, DBLP:conf/cloud/WeiDQHCGGGX15,
  HoCCLKGGGX13} that maximizes utilization of the
\khii{more-freely-available} LAN bandwidth
% among worker machines and parameter servers
\emph{within} a data center. At the same time, \khi{across different
  data centers}, {\gaia} employs a new \synchronization model, called
\emph{\protocol (\protoabbrv)}, \khi{which} makes more efficient use
of the scarce and heterogeneous WAN bandwidth.  By ensuring \khi{that}
each ML model copy in different data centers is \emph{approximately
  correct} \khi{based on a precise notion} defined by \protoabbrv, we
guarantee ML algorithm convergence.

%the globally distributed worker machines make
%progress towards algorithm convergence in parallel with having little
%communication overhead across WANs.

\protoabbrv is based on \khii{a key finding} that the vast majority of
updates to the global ML model parameters from each ML worker machine
\khi{are \emph{insignificant}}. For example, our study \khi{of} three
classes of ML algorithms shows that more than 95\% of the updates
produce less than a 1\% change to the parameter value.  With
\protoabbrv, these insignificant updates to the same parameter within
a data center are \emph{aggregated} (and thus not communicated to
other data centers) until the aggregated updates are significant
enough. \protoabbrv allows the ML programmer to specify the
\emph{function} and the \emph{threshold} to determine the significance
of updates for each ML algorithm, while providing default
configurations for unmodified ML programs. For example, the programmer
can specify that all updates that produce more than a 1\% change are
significant.  \protoabbrv ensures all significant updates are
synchronized across all model copies in a timely manner.  It
dynamically \khii{adapts communication} to the available WAN bandwidth
between pairs of data centers and uses special \emph{selective
  barrier} and \emph{mirror clock} control messages to ensure
\khi{algorithm convergence} even during a \khi{period of sudden
  fall (negative spike)} in available WAN bandwidth.

In contrast to a state-of-the-art communication-efficient
\synchronization model, Stale Synchronous Parallel
(SSP)~\cite{HoCCLKGGGX13}, \khi{which} bounds how \emph{stale (i.e.,
  old)} a parameter can be, \protoabbrv bounds how \emph{inaccurate} a
parameter can be, in comparison to the \khi{most up-to-date} value.
\khi{Hence, it provides high flexibility in performing (or not
  performing)} updates, as the server can delay synchronization
\emph{indefinitely} as long as the aggregated update is insignificant.

We build \khi{two} prototypes of {\gaia} on top of two state-of-the-art
parameter server systems, one \khi{specialized} for
CPUs~\cite{CuiTWXDHHGGGX14} and another \khi{specialized} for
GPUs~\cite{DBLP:conf/eurosys/CuiZGGX16}. We deploy {\gaia} across 11
regions on Amazon EC2, and on a local cluster that emulates the WAN
bandwidth across different Amazon EC2 regions. Our evaluation with
three popular classes of ML algorithms shows that, compared to
\khi{two} state-of-the-art parameter server
systems\khi{~\cite{CuiTWXDHHGGGX14, DBLP:conf/eurosys/CuiZGGX16}}
deployed on WANs, {\gaia}: (1) significantly improves performance, by
1.8--53.5$\times$, (2) \khii{has performance within 0.94--1.40$\times$
  of running the same ML algorithm on a LAN in a single data center},
and (3) significantly reduces the monetary cost of running the same ML
algorithm on WANs, by 2.6--59.0$\times$.

\section{Motivation} 

To further motivate our work, we discuss WAN bandwidth constraints and
study the performance implications of running two state-of-the-art ML
systems over WANs.

\subsection{WAN Network Bandwidth and Cost}
\label{subsec:bandwidth_wan}

WAN bandwidth is a very
scarce resource~\cite{DBLP:conf/sigcomm/LaoutarisSYR11,
  DBLP:conf/nsdi/RabkinASPF14, DBLP:conf/nsdi/VulimiriCGJPV15} relative to LAN bandwidth.
%Phil: I think we want to keep this discussion brief since NSDI folks know more about this than we do!
%The major reason for the scarcity of WAN bandwidth is the cost---a
%submarine cable in an ocean is way more costly than a network cable in a data center.
Moreover, the high \khi{cost} of adding network bandwidth has resulted in
a deceleration of WAN bandwidth growth. The Internet capacity growth
has fallen steadily for many years, and the annual growth rates have
lately settled \khi{into} the low-30 percent range~\cite{GIG}.
%For ML systems (such as Figure~\ref{fig:parameter_server_wan}) running on a
%WAN, the scarce network bandwidth poses a difficult challenge for
%timely communication between worker machines and parameter servers.

To quantify the scarcity of WAN bandwidth between data centers, we
measure the network bandwidth between all pairs of Amazon EC2 sites in
11 different regions (Virginia, California, Oregon, Ireland,
Frankfurt, Tokyo, Seoul, Singapore, Sydney, Mumbai, and S\~{a}o
Paulo). We use \texttt{iperf3}~\cite{iperf} to measure the network
bandwidth of each pair of different regions for five rounds, and then
calculate the average
bandwidth. Figure~\ref{fig:ec2_network_bandwidth} shows the average
network bandwidth between each pair of different regions. We
make two observations.

\begin{figure}[h]
  \centering
  \includegraphics[width=0.8\textwidth]{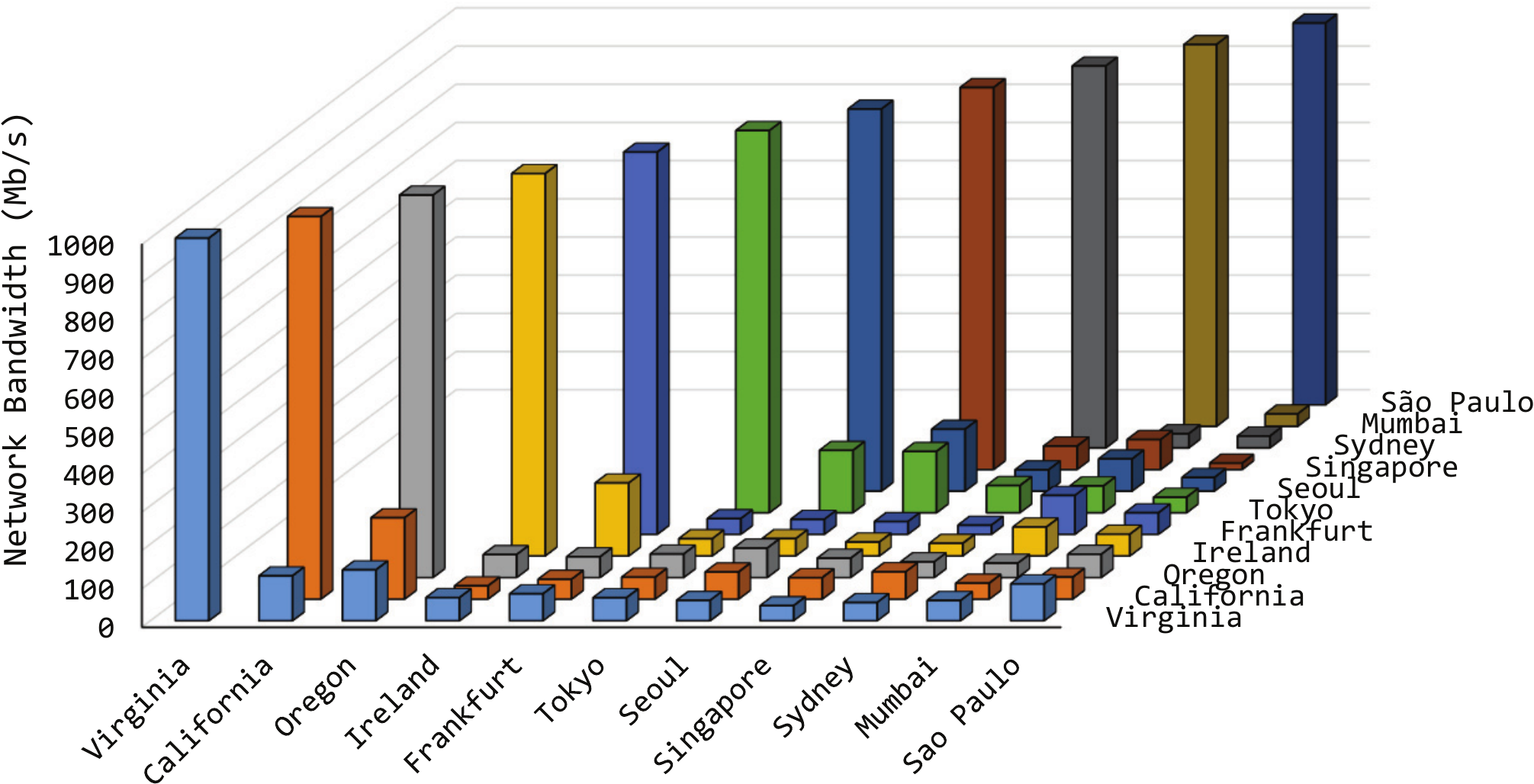}
  \caption{Measured network bandwidth between Amazon EC2 sites in 11 different regions}
  \label{fig:ec2_network_bandwidth}
\end{figure}

First, the WAN bandwidth between data centers is 15$\times$ smaller
than the LAN bandwidth within a data center on average, and up to
60$\times$ smaller \khii{in the worst case} (for Singapore \ding{214}
S\~{a}o Paulo).  Second, the WAN bandwidth \emph{varies significantly}
between different regions. The \khii{WAN} bandwidth between
\khi{geographically-close} regions (e.g., Oregon \ding{214} California
or Tokyo \ding{214} Seoul) is up to 12$\times$ of the bandwidth
between distant regions (e.g., Singapore \ding{214} S\~{a}o Paulo). As
Section~\ref{subsec:ml_system_on_wan} shows, the scarcity and
variation of the WAN bandwidth can significantly degrade the
performance of state-of-the-art \khi{ML} systems.

Another important challenge \khii{imposed by} WANs is the
\emph{\khi{monetary} cost} of communication. In data centers, the cost
of \khi{WANs} far exceeds the cost of \khii{a LAN} and \khi{makes up a
  significant fraction of the overall
  cost}~\cite{DBLP:journals/ccr/GreenbergHMP09}. Cloud service
providers, such as Amazon EC2, charge an extra fee for WAN
communication while \khi{providing} LAN communication free of
charge. The cost of WAN communication can be much higher than the cost
of \khii{the} machines themselves. For example, the cost of two
machines in Amazon EC2 communicating at the rate of the average WAN
bandwidth between data centers is up to 38$\times$ of the cost of
renting these two machines~\cite{amazon_ec2_price}. These costs make
running ML algorithms on \khii{WANs} much more expensive than running
them on a LAN.

\vspace{5pt}
\subsection{ML System Performance on WANs}
\label{subsec:ml_system_on_wan}

We study the performance \khi{implications} of deploying distributed
ML systems on WANs using two state-of-the-art parameter server
systems, IterStore~\cite{CuiTWXDHHGGGX14} and
B\"{o}sen~\cite{DBLP:conf/cloud/WeiDQHCGGGX15}. Our experiments are
conducted on our local 22-node cluster that emulates the WAN bandwidth
between Amazon EC2 data centers, the accuracy of which is validated
against a real Amazon EC2 deployment (see Section~\ref{subsec:setup}
for details). We run the same ML application, \emph{Matrix
  Factorization}~\cite{DBLP:conf/kdd/GemullaNHS11}
(Section~\ref{subsec:workload}), on both systems.

For each system, we evaluate both BSP and SSP as \khi{the}
\synchronization model (Section~\ref{subsec:ml_training_system}), with four
deployment settings: (1) \emph{LAN}, deployment within a single data
center, (2) \emph{EC2-ALL}, deployment across 11 aforementioned EC2
regions, (3) \emph{V/C WAN}, deployment across two data centers
\khi{that have the same} WAN bandwidth \khii{as that} between Virginia
and California \khi{(Figure~\ref{fig:ec2_network_bandwidth})},
representing a distributed ML setting within \khii{a} continent, and
(4) \emph{S/S WAN}, deployment across two data centers \khi{that have
  the same} WAN bandwidth \khii{as that} between Singapore and S\~{a}o
Paulo, \khi{representing the \khii{lowest} WAN bandwidth between any
  two Amazon EC2 regions}.

Figure~\ref{fig:performance_mf_baseline} shows the normalized
execution time \khi{until} algorithm convergence across the four
deployment settings. All results are normalized to IterStore using BSP
on a LAN. The data label on each bar represents how much slower the
WAN setting is than its \khi{\emph{respective}} LAN setting for the
given system, e.g., B\"{o}sen-BSP on EC2-ALL is 5.9$\times$ slower
than B\"{o}sen-BSP on LAN.

\begin{figure}[h]
  \centering
  \includegraphics[width=0.8\textwidth]{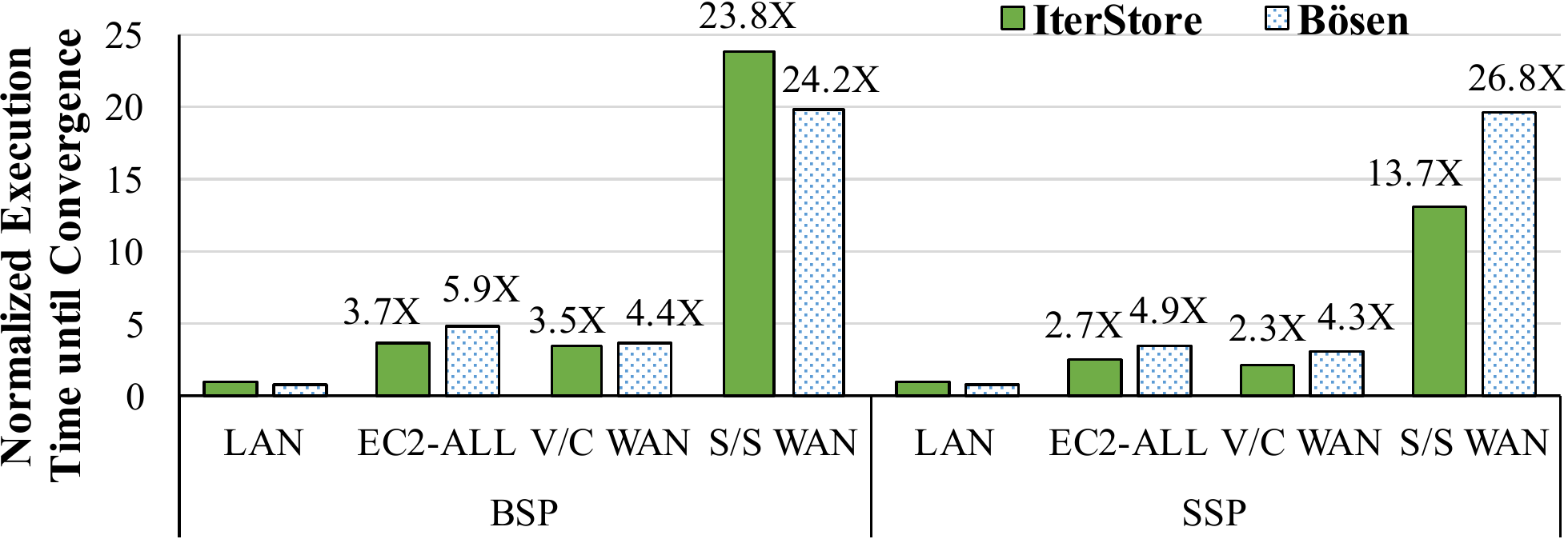}
  \caption{Normalized execution time until ML algorithm convergence
    when deploying two state-of-the-art distributed ML systems on a
    LAN and WANs}
  \label{fig:performance_mf_baseline}
\end{figure}

As we see, both systems suffer significant performance degradation
when \khi{deployed} across multiple data centers. When using BSP,
IterStore is 3.5$\times$ to 23.8$\times$ slower on WANs than it is on
a LAN, and B\"{o}sen is 4.4$\times$ to 24.2$\times$ slower. While
using SSP can \khi{reduce overall execution times of both systems},
both systems still show significant slowdown \khi{when run on WANs}
(2.3$\times$ to 13.7$\times$ for IterStore, and 4.3$\times$ to
26.8$\times$ for B\"{o}sen). We conclude that simply running
state-of-the-art distributed ML systems on WANs can seriously slow
down ML applications, and \khii{thus} we need a \khi{new} distributed
ML system that can be effectively deployed on WANs.

\ignore{

\subsection{Our Goal}

As we discuss above, simply running state-of-the-art distributed ML
systems on WAN can seriously slow down the ML applications. Another
approach, centralizing all input data into one data center and then
running ML systems locally, can also be prohibitively difficult due to
the limited WAN bandwidth and privacy laws. In this work, our goal is
to propose an \emph{effective and general} ML system that can be
deployed on WAN \emph{without} centralizing the input
data. Specifically, the system needs to: (1) utilize the scarce WAN
bandwidth to provide competitive performance, (2) minimize the costly
communication over WAN while guaranteeing ML algorithm convergence,
and (3) be general enough for a wide variety of ML algorithms.

}

\section{Our Approach: {\gaia}}

We introduce {\gaia}, a general ML system that can be effectively
deployed on WANs to address the increasing need to run ML applications
\emph{directly} on geo-distributed data. We identify two key
challenges in designing such a system
(Section~\ref{subsec:key_challenge}). We then introduce the system
architecture of {\gaia}, which differentiates the communication
\khi{\emph{within}} a data center from the communication
\khi{\emph{between}} different centers
(Section~\ref{subsec:system_overview}). Our approach is based on the
key empirical finding that the vast
%To enable effective
%communication on WANs, we make the key empiricalobservation that the vast
majority of communication within an ML system \khi{results in}
\emph{insignificant} changes to the state of the global model
(Section~\ref{subsec:key_observation}).  In light of this finding, we
design a new ML \synchronization model, called \emph{\protocol
  (\protoabbrv)}, \khi{which} can eliminate the insignificant
communication while ensuring the \khii{convergence and accuracy} of ML
algorithms. We describe \protoabbrv in detail in
Section~\ref{subsec:protocol}. Finally,
Section~\ref{subsec:convergence_proof} summarizes our theoretical
analysis \khi{of} how \protoabbrv guarantees algorithm convergence for
a widely-used ML algorithm, stochastic gradient descent (SGD) (the
full proof is \khi{in Appendix~\ref{appendix:proof}}).

\subsection{Key Challenges}
\label{subsec:key_challenge}

There are two key challenges in designing a general and effective ML
system on WANs.

\textbf{Challenge 1.} \emph{How to effectively communicate over WANs
  while retaining algorithm \khi{convergence and accuracy}?} As we see
above, state-of-the-art distributed ML systems can overwhelm the
scarce WAN bandwidth, \khi{causing significant slowdowns}. We need a
\khii{mechanism that significantly reduces the communication between data
centers} so \khi{that} the system can provide competitive
performance. However, reducing communication \khii{can affect} the
accuracy of an ML algorithm. A poor choice of \synchronization model
in a distributed ML system can \khi{prevent the ML algorithm from
  converging to the \khii{optimal} point (i.e., the best model to
  explain or fit the input data) that one can} achieve when using a
proper synchronization model~\cite{DBLP:conf/icml/BradleyKBG11,
  DBLP:conf/nips/RechtRWN11}. Thus, we need a mechanism that can
reduce communication intensity while ensuring \khi{that} the
communication occurs in a \khi{\emph{timely}} manner, even when the
network bandwidth is extremely stringent. \khi{This mechanism should
  provably guarantee algorithm convergence \emph{irrespective} of the
  network conditions}.

\ignore{This problem is more likely to happen on WANs because the
  communication happens much slower than LANs. }

\textbf{Challenge 2.} \emph{How to make the system generic and work
  for ML algorithms without requiring modification?} Developing an
effective ML algorithm takes \khi{significant} effort and experience,
making it a \khii{large} burden for the ML algorithm developers to change the
algorithm when deploying it on WANs. Our system should work across a
wide variety of ML algorithms, preferably \emph{without any change} to
the algorithms themselves. This is challenging because different ML
algorithms have different communication patterns, and the implication
of reducing communication can vary significantly among
them~\cite{DBLP:conf/icml/TakacBRS13, DBLP:conf/nips/TsianosLR12,
  DBLP:journals/corr/RichtarikT13, DBLP:conf/nips/JaggiSTTKHJ14,
  NewmanASW09, SmolaN10, DBLP:conf/nips/ZinkevichSL09}.

%KEVIN: Old two challenges, we follow the new structure now
\ignore{
\textbf{Challenge 1.} \emph{How to effectively communicate over WANs
  for a wide variety of ML algorithms?} As we see above, state-of-the-art
distributed ML systems require intensive communication that can
overwhelm the scarce WAN bandwidth and seriously slow down the
system. We need a mechanism to reduce the communication intensity
across data center boundaries so the system can provide competitive
performance. At the same time, this mechanism should work for many
different ML algorithms, and preferably \emph{without any change} to
the ML programs. This is a difficult challenge because different ML
algorithms have different communication patterns, and the implication
of reducing communication can vary significantly among
them~\cite{DBLP:conf/icml/TakacBRS13, DBLP:conf/nips/TsianosLR12,
  DBLP:journals/corr/RichtarikT13, DBLP:conf/nips/JaggiSTTKHJ14,
  NewmanASW09, SmolaN10, DBLP:conf/nips/ZinkevichSL09}.

\textbf{Challenge 2.} \emph{How to guarantee the algorithm convergence
  under extremely stringent WAN condition?} As we discuss in
Section~\ref{subsec:ml_training_system}, the goal of an ML algorithm is to find
the best model to fit the input data. The progress toward this goal is
usually measured by an explicit \emph{objective function}, which
reaches the \khii{optimal} value when the ML algorithm converges. Poorly
designed synchronization model in a distributed ML system can cause
the objective function never reaches the \khii{optimal} value that could have
been achieved by a proper synchronization
model~\cite{DBLP:conf/icml/BradleyKBG11,
  DBLP:conf/nips/RechtRWN11}. This problem is more likely to happen on
WANs because the communication happens much slower than LANs. Thus, we
need a mechanism to ensure the communication occurs in a
\emph{timeliness} manner, even when the network bandwidth is extremely
stringent. Furthermore, we need to prove this mechanism can guarantee
algorithm convergence \emph{irrespective} to the network condition.
}

\subsection{{\gaia} System Overview}
\label{subsec:system_overview}

We propose a new ML system, {\gaia}, that addresses the two key
challenges in designing a general and effective ML system on
WANs. {\gaia} is built on top the popular parameter server
architecture, which is proven to be effective on a wide variety
of ML algorithms (e.g.,~\cite{HoCCLKGGGX13, CuiTWXDHHGGGX14,
  XingHDKWLZXKY15, CuiCHKLKWDGGGX14, DBLP:conf/osdi/LiAPSAJLSS14,
  DBLP:conf/cloud/WeiDQHCGGGX15, DBLP:conf/osdi/ChilimbiSAK14,
  tensorflow2015-whitepaper, AhmedAGNS12, DeanCMCDLMRSTYN12}). As
discussed in Section~\ref{subsec:ml_training_system}, in \khi{the} parameter
server architecture, \emph{all} \khi{worker machines synchronize with
  each other through parameter servers} to ensure that the global model
state is \khi{up-to-date}. While this architecture guarantees
algorithm convergence, it also requires \khi{substantial}
communication between worker machines and parameter servers. To make
{\gaia} effective on WANs while fully utilizing the abundant LAN
bandwidth, we design a new system architecture to \emph{decouple} the
synchronization within a data center (LANs) from the synchronization
across different data centers (WANs).

Figure~\ref{fig:system_overview} shows an overview of {\gaia}. In
{\gaia}, each data center has some worker machines and parameter
servers. Each worker machine \khii{processes} a shard of the input data stored
in its data center to achieve data parallelism
(Section~\ref{subsec:ml_training_system}). The parameter servers in each data
center collectively maintain a version of \khi{the} \emph{global model
  copy} (\ding{182}), and each parameter server handles a shard of
this global model copy. \khi{A worker machine} \emph{only}
\texttt{READs} and \texttt{UPDATEs} the global model copy in its data
center.

\begin{figure}[h]
  \centering
  \includegraphics[width=0.8\textwidth]{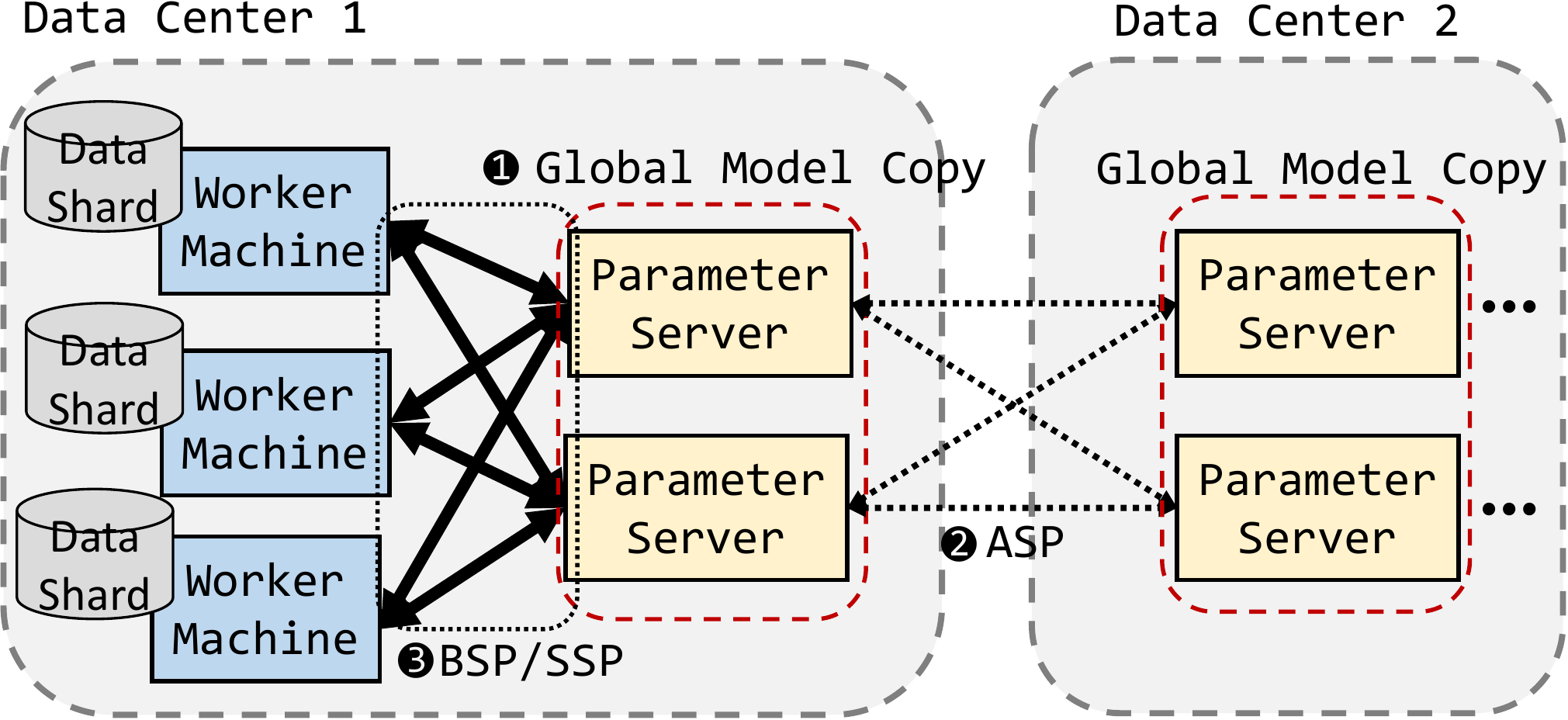}
  \caption{{\gaia} system overview}
  \label{fig:system_overview}
\end{figure}

To reduce the communication overhead over WANs, the global model copy
in each data center is only \emph{approximately correct}. This design
enables us to eliminate the insignificant, and thus unnecessary,
communication across different data centers. We design a new
\synchronization model, called \protocol (\protoabbrv~\ding{183}),
between parameter servers \khi{\emph{across} different data centers to
  ensure that} each global model copy is approximately correct even
with \khi{very low WAN bandwidth}. Section~\ref{subsec:protocol}
describes the details of ASP. On the other hand, worker machines and
parameter servers \emph{within} a data center synchronize with each
other using \khi{the} conventional BSP (Bulk Synchronous Parallel) or
SSP (Stale Synchronous Parallel) models (\ding{184}). \khi{These
  models allow} worker machines to \khi{quickly observe fresh updates
  that happen \emph{within}} a data center. \khi{Furthermore, worker
  machines and parameter servers within a data center can employ} more
aggressive communication schemes such as sending updates early and
often~\cite{dai2015analysis, DBLP:conf/cloud/WeiDQHCGGGX15} to fully
utilize the abundant (and free) network bandwidth on \khi{a LAN}.

\ignore{
As discussed in Section~\ref{subsec:key_observation}, the vast majority of
updates from the worker machines are very insignificant. In light of
this, a parameter server in {\gaia} \emph{aggregates} the updates from
the local worker machines and send the update to other data centers
\emph{only when} the aggregated update is \emph{significant}
enough. This design ensures each global model copy is approximately
correct while allowing the parameter servers to dynamically eliminate
the insignificant, and thus unnecessary, communication across
different data centers.
}

\subsection{Study of Update Significance}
\label{subsec:key_observation}

As discussed above, {\gaia} reduces the communication overhead over
WANs by eliminating insignificant communication. To understand the
benefit of our approach, we study the \emph{significance} of the
updates sent from worker machines to parameter servers. \khii{We study
  three classes of popular ML algorithms: \emph{Matrix Factorization
    (MF)}~\cite{DBLP:conf/kdd/GemullaNHS11}, \emph{Topic Modeling
    (TM)}~\cite{BleiNJ03}, and \emph{Image Classification
    (IC)}~\cite{DBLP:journals/neco/LeCunBDHHHJ89} (see
  Section~\ref{subsec:workload} for descriptions).} We run all the
algorithms \khii{until} convergence, analyze all the updates sent from worker
machines to parameter servers, and compare \khi{the change they cause
  on} the parameter value when the servers receive them. \khi{We
  define an update to be \emph{significant} if it causes $S$\% change
  on the parameter value, and we vary $S$, the significance threshold,
  between 0.01 and 10. Figure~\ref{fig:update_distribution} shows the
  percentage of insignificant updates among all updates, \khii{for} different
  values of $S$.}

\ignore{the ratio of the
total number of updates to the number of significant updates with
different significance thresholds. }

\begin{figure}[h]
  \centering
  \includegraphics[width=0.8\textwidth]{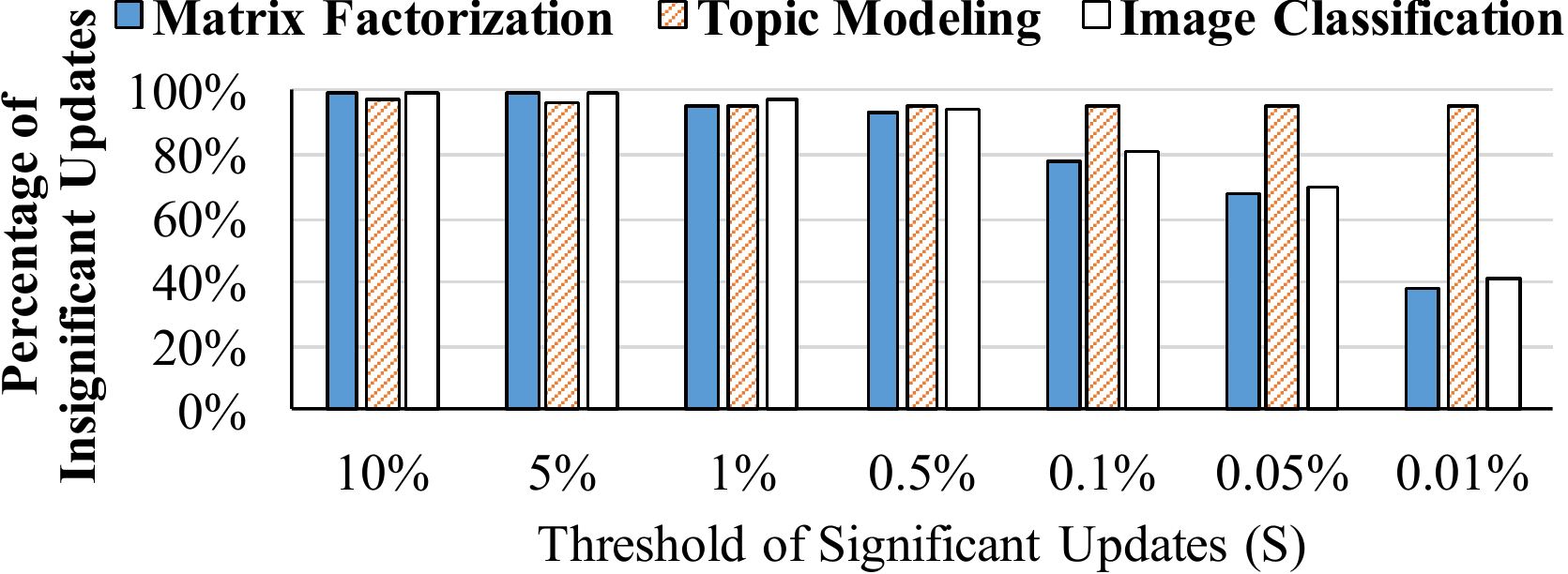}
  \caption{Percentage of insignificant updates}
  \label{fig:update_distribution}
\end{figure}

\ignore{ % We don't need both figures
\begin{figure}[ht!]
 \centering
 \begin{subfigure}[t]{0.98\linewidth}
 \centering
 \includegraphics[width=1.0\textwidth]{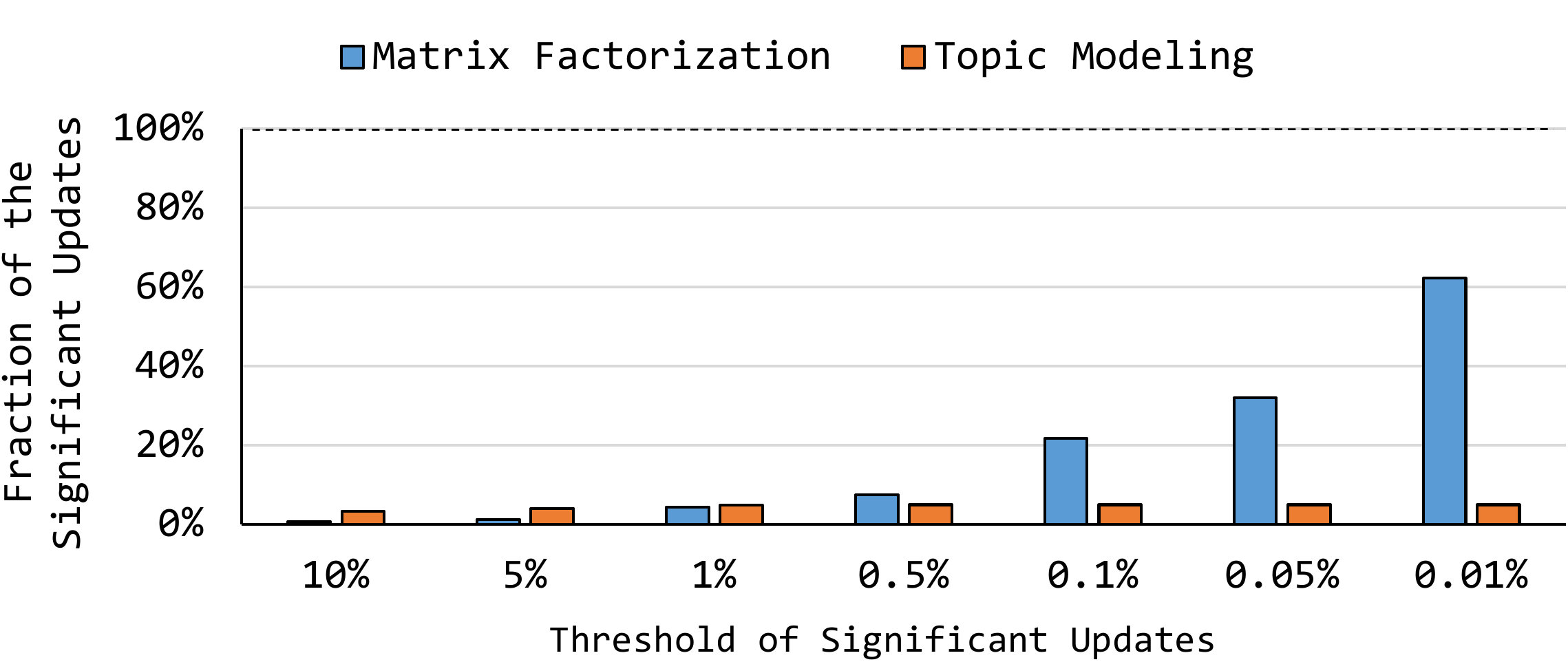}
 \caption{\emph{Fraction of the significant updates}} 
 \label{fig:update_distribution_fraction}
 \end{subfigure}
 \begin{subfigure}[t]{0.98\linewidth}
 \centering
 \includegraphics[width=1.0\textwidth]{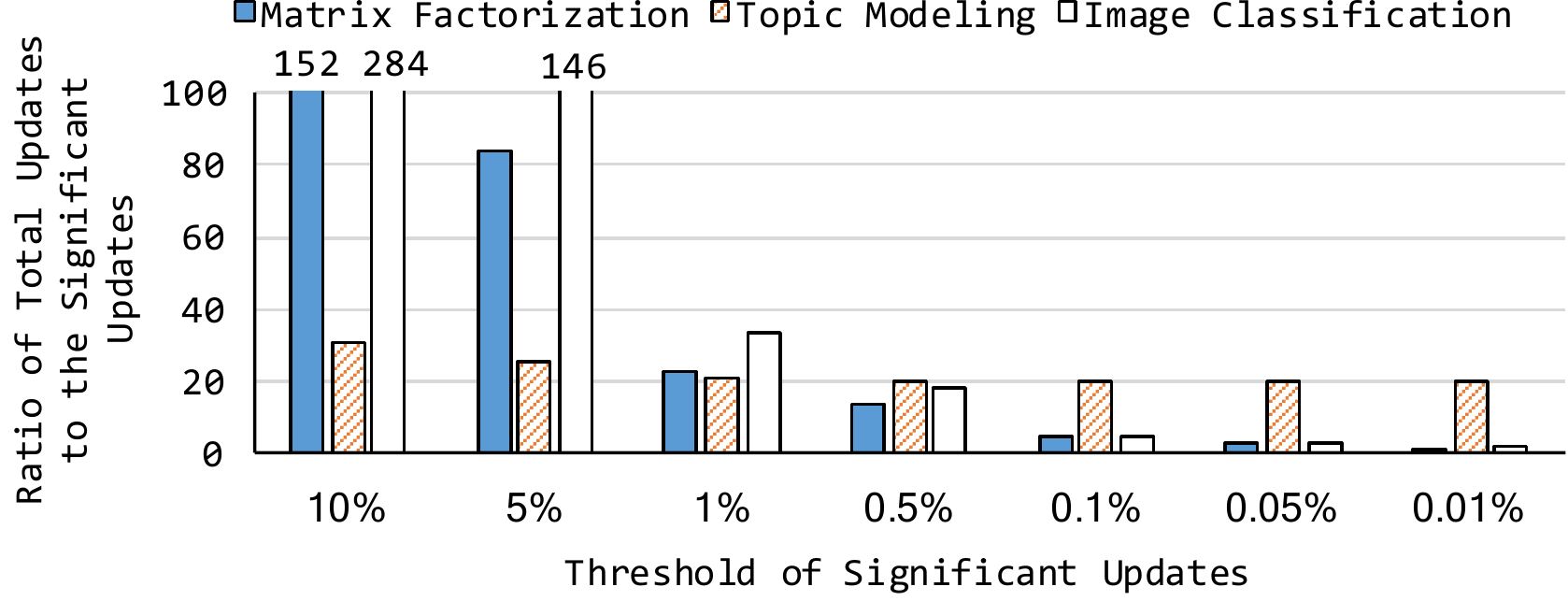} 
 \caption{\emph{Ratio of total updates to the significant updates}}
 \label{fig:update_distribution_ratio}
 \end{subfigure}
 \caption{Total updates vs. significant updates with different
   significance threshold}
 \label{fig:update_distribution}
\end{figure} 
}

\khi{As we see, the vast majority of updates in these algorithms are
  \emph{insignificant}. Assuming the significance threshold is 1\%,
  95.2\% / 95.6\% / 97.0\% of all updates are insignificant for
  \emph{MF} / \emph{TM} / \emph{IC}. When we relax the significance
  threshold to 5\%, 98.8\% / 96.1\% / 99.3\% of all updates are
  insignificant. Thus, most of the communication changes the ML model
  state only very slightly.}

\ignore{Second, the sensitivity to the significance threshold varies between
algorithms. When choosing different significance thresholds, \emph{MF}
and \emph{IC} show very high variation while \emph{TM} is less
sensitive to it. This gives us another opportunity to optimize the
communication: We can choose a very small threshold for the less
sensitive algorithms for safety, and we can dynamically change the
threshold for the more sensitive algorithms to make a trade-off
between communication overhead and accuracy.}

It is worth noting that our finding is consistent with the findings of
prior work~\cite{lars, DBLP:conf/uai/ElidanMK06, LowGKBGH12,
  DBLP:conf/cloud/ZhangGGW11, DBLP:conf/eurosys/KimHLZDGX16} on other
ML algorithms, such as PageRank and Lasso. These works observe that in
these ML algorithms, not all model parameters converge to their
\khii{optimal} value within the same number iterations --- a property
called \emph{non-uniform
  convergence}~\cite{DBLP:journals/corr/XingHXD15}. Instead of
examining the convergence rate, we \emph{quantify} the
\emph{significance} of updates \khi{with various significance
  thresholds}, which provides a unique opportunity to reduce the
communication over WANs.

% Move to related work
\ignore{Several prior
works~\cite{LowGKBGH12, DBLP:conf/cloud/ZhangGGW11,
  DBLP:conf/cloud/WeiDQHCGGGX15, DBLP:conf/eurosys/KimHLZDGX16} show
how to exploit this property to achieve faster algorithm convergence
with better computation schedule (such as computing important
parameters more often) or communication schedule (such as
communicating important parameters earlier). However, none of these
works quantify the \emph{significance} of updates and their
distribution, which provides a unique opportunity to reduce the
communication overhead over WANs.}

\subsection{\protocol}
\label{subsec:protocol}

The goal of our new \synchronization model, \protocol (\protoabbrv),
is to ensure \khii{that} the global model copy in each data center is
approximately correct. In this model, a parameter server \khi{shares
  only the significant updates with} other data centers, and
\protoabbrv ensures \khi{that these updates} can be seen by all data
centers in a timely fashion. \protoabbrv achieves this goal by using
three techniques: (1) the significance filter, (2) \protoabbrv
selective barrier, and (3) \protoabbrv mirror clock. We
\khii{describe} them in order.

\textbf{The significance filter.}  \protoabbrv takes two inputs from
an ML programmer to determine \khi{whether or not} an update is
significant. \khi{They} are: (1) a \emph{significance function} and
(2) an \emph{initial significance threshold}. The significance
function returns the significance of each \khi{update. We} define an
update as significant if its significance is larger than the
threshold. \khi{For example, an ML programmer can define the
  significance function as the update's magnitude relative to the
  current value ($|\frac{Update}{Value}|$), and set the initial
  significance threshold to 1\%.} The significance function can be
more sophisticated if the impact of parameter changes to the model is
not linear, or the importance of parameters is non-uniform (see
Section~\ref{subsec:sig_func}). A parameter server aggregates updates
from the local worker machines and shares the aggregated updates
\khi{with} other data centers when \khi{the aggregated updates} become
significant. To ensure \khi{that} the algorithm can converge to the
optimal point, \protoabbrv automatically \khi{reduces} the
significance threshold over time (specifically, if the original
threshold is $v$, then the threshold at iteration $t$ of the ML
algorithm is $v/\sqrt{t}$).

\textbf{\protoabbrv selective barrier.}  While we can greatly reduce
the communication overhead over WANs by sending \khi{only the}
significant updates, the WAN bandwidth might still be insufficient for
\khi{such} updates. In \khi{such a case}, the significant updates can
arrive too late, and we \khi{might not be able to} \khii{bound the
  deviation between different global model copies}. \protoabbrv
handles this case with \khi{the \emph{\protoabbrv selective barrier}}
(Figure~\ref{fig:protocol_barrier}) control message. When a parameter
server receives the significant updates (\ding{182}) at a rate that is
higher than the WAN bandwidth can \khi{support}, the parameter server
first sends the indexes of these significant updates (\khii{as opposed
to sending both the indexes and the update values together}) via an
\protoabbrv selective barrier (\ding{183}) to the other data
centers. The receiver of an \protoabbrv selective barrier blocks its
local worker from reading the specified parameters until it receives
the significant updates from the sender of the barrier. This technique
ensures that all worker machines in each data center \khi{are} aware
of the significant updates after a bounded network latency,
\ignore{(i.e., there is no extra queuing latency due to insufficient WAN
  bandwidth)} and \khi{they wait \emph{only}} for these updates. The
worker machines can \khi{make} progress as long as they do not depend
on any of these parameters.

\begin{figure}[ht!]
 \centering
 \begin{subfigure}[t]{0.49\linewidth}
 \centering
 \includegraphics[width=1.0\textwidth]{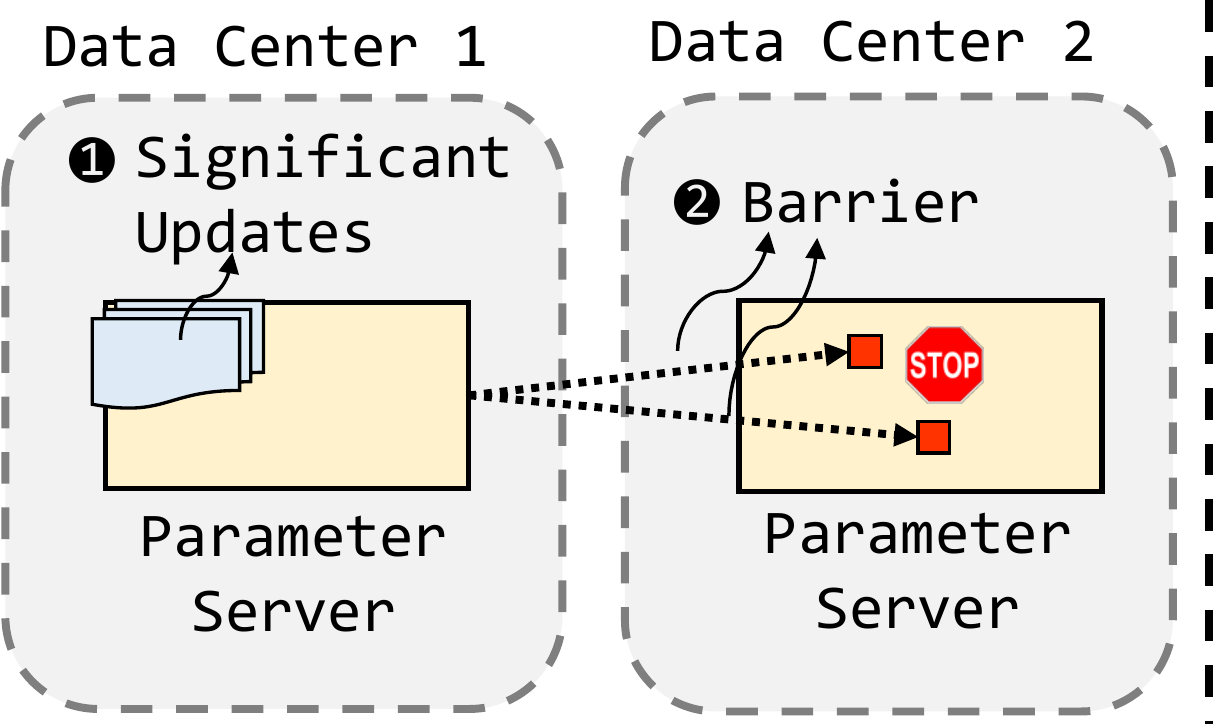}
 \caption{\emph{\protoabbrv selective barrier}} 
 \label{fig:protocol_barrier}
 \end{subfigure}
 \begin{subfigure}[t]{0.49\linewidth}
 \centering
 \includegraphics[width=1.0\textwidth]{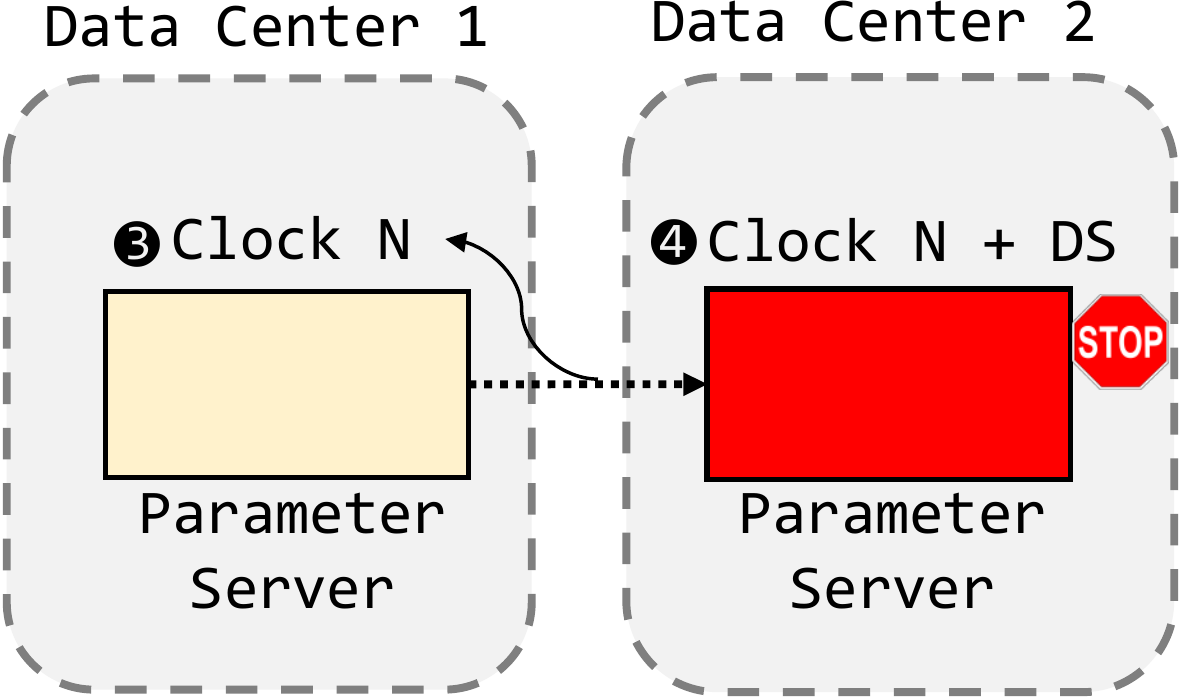} 
 \caption{\emph{Mirror clock}}
 \label{fig:protocol_mirror_clock}
 \end{subfigure}
 \caption{The \synchronization mechanisms of \protoabbrv}
 \label{fig:protocol_timeliness}
\end{figure} 

\textbf{Mirror clock.} \khi{The \protoabbrv select barrier ensures
  that the latency of the significant updates is no more than the
  network latency. \khii{However, it assumes that 1) the underlying
    WAN bandwidth and latency are fixed so that the network latency
    can be bounded, and 2) such} latency is short enough so that other
  data centers can be aware of them in time. In practice, WAN
  bandwidth can fluctuate over
  time~\cite{DBLP:conf/sigcomm/HongKMZGNW13}, and the WAN latency can
  be \khii{intolerably} high for some ML algorithms. \kt{Worse still,
    the ASP selective barrier messages could experience long delay
    when the network packets are dropped on WAN.} We need a mechanism to
  \khii{guarantee that} the worker machines are aware of the
  significant updates in time, irrespective of the WAN bandwidth or
  latency.}

We use the \emph{mirror clock} (Figure~\ref{fig:protocol_mirror_clock})
\khi{to provide this guarantee}. When each parameter server receives
all the updates from its local worker machines at the end of a clock
(e.g., an iteration), it reports its clock to the servers that are in
charge of the same parameters in the other data centers.  When a
server detects its clock is ahead of the slowest server that shares
the same parameters by a predefined \khi{threshold $DS$ (data center
  staleness)}, the server blocks its local worker machines from
reading its parameters until the slowest mirror server catches up. In
the example of Figure~\ref{fig:protocol_mirror_clock}, the server
clock in Data Center 1 is $N$, while the server clock in Data Center 2
is $(N + DS)$. As their difference reaches the predefined limit, the
server in Data Center 2 blocks its local worker from reading its
parameters. This mechanism is similar to the concept of
SSP~\cite{HoCCLKGGGX13}, but we use it only as the last resort to
\khi{guarantee algorithm convergence}. \ignore{In our experiments, it is
\khi{rarely} activated.}

\ignore{As discussed above, the mechanism of
\protoabbrv selective barrier needs to use the WAN bandwidth to determine
whether to send the barrier or not. However, WAN bandwidth can
fluctuate over time~\cite{DBLP:conf/sigcomm/HongKMZGNW13}. In a
pathological case where a parameter server does not send the
\protoabbrv selective barrier but the WAN bandwidth drops, while the server is
sending the significant updates, these updates can still arrive too
late.}

\subsection{Summary of Convergence Proof}
\label{subsec:convergence_proof}

In this section, we summarize our proof showing that a popular, broad
class of ML algorithms are guaranteed to converge 
under our new \protoabbrv \synchronization model.
The class we consider are ML algorithms expressed as convex optimization
problems that are solved using distributed stochastic gradient descent.

The proof follows the outline of prior work on
SSP~\cite{HoCCLKGGGX13}, with \khi{a new challenge, i.e.,} our new
\protoabbrv \synchronization model \khii{allows the synchronization of
  insignificant updates to be delayed indefinitely}. \khi{To} prove
algorithm convergence, \khi{our goal} is to show that the distributed
execution of an ML algorithm results in \khi{a set of} parameter
values that are very close (practically identical) to the values that
would be obtained under a serialized execution.  \ignore{We can think
  of an algorithm execution as a set of updates that are to be
  computed. What differentiates a serial and a distributed execution
  is simply the order by which these updates are seen and the sequence
  of the values that the parameters goes through.}

\ignore{
By algorithm convergence we mean that the distributed execution of an
iterative-convergent ML algorithm results in parameter values that are
very close (practically identical) to the values that would be obtained
under a serializable execution. We can think of an algorithm execution as
a set of updates that are to be computed. What then really differentiates
a serial and a distributed execution, becomes the the order by which these
updates are seen and the sequence of the values that the parameters go
through.
}

\khi{\khii{Let $f$ denote the objective function of an optimization
    problem, whose goal is to minimize $f$.} Let $\tilde{\bm{x}_t}$
  denote the sequence of noisy (i.e., inaccurate) \emph{views} of the
  parameters, where $t = 1, 2,...,T$ is the index of each view over
  time. Let $\bm{x}^{*}$ denote the value that minimizes
  $f$. Intuitively, we would like $f_t(\tilde{\bm{x}_t})$ to approach
  $f(\bm{x}^{*})$ as $t \rightarrow \infty$. We call the difference
  between $f_t(\tilde{\bm{x}_t})$ and $f(\bm{x}^{*})$
  \emph{regret}. We can prove $f_t(\tilde{\bm{x}_t})$ approaches
  $f(\bm{x}^{*})$ as $t \rightarrow \infty$ by proving that the
  \emph{average regret}, $\frac{R[X]}{T} \rightarrow 0$ as $T
  \rightarrow \infty$.}

\ignore{In {\gaia}, we can guarantee this goal since the system does not allow
each worker's view of the parameters to be very far from each
other. Within a data center, the bounded staleness (if not zero)
ensures that updates from faster threads are eventually seen by slow
workers after a bounded number of clocks.  Across data centers, the
fixed mirror clock difference (Section~\ref{subsec:protocol})
prohibits situations where workers in a data center proceed very far
ahead while there are significant updates need to be read from the
parameter servers in other data centers. We ensure that all updates
are \emph{eventually} seen by all workers as the threshold decreases
over time. }

\ignore{
{\gaia} postpones the delivery of insignificant updates till the moment
that the aggregate exceeds a threshold. So it both saves communication
bandwidth by sending an aggregate of many insignificant updates as an
aggregate relatively infrequently, and ensures that all updates are
\emph{eventually} seen by all workers, as while the threshold
decreases over time, there will be a clock at which the updates will
overcome the threshold and thus be sent through the WANs.
}
 
\khi{Mathematically, the above intuition is formulated with
Theorem~\ref{thm:asp_sgd}. The details of the proof and the notations
are in Appendix~\ref{appendix:proof}.}

{
  
\vspace{-5pt}  
\begin{theorem}
\textbf{(Convergence of SGD under ASP)}. \khi{Suppose
  that, in order to compute the minimizer $x^{*}$ of a convex function
  $f(\bm{x}) = \sum_{t=1}^{T}f_t(\bm{x})$, with $f_t, t=1,2,\ldots,T$,
  convex, we use stochastic gradient descent on one component
  $\nabla f_t$ at a time.  Suppose also that 1) the algorithm is
  distributed in $D$ data centers, each of which uses $P$ machines, 2)
  within each data center, the SSP protocol is used, with a fixed
  staleness of $s$, and 3) a fixed mirror clock difference $\ds{}$ is
  allowed between any two data centers.  Let $\bm{u}_t = -\eta_t \nabla
  f_t{(\tilde{\bm{x}_t})}$, where the step size $\eta_t$ decreases as
  $\eta_t = \frac{\eta}{\sqrt{t}}$ and the significance threshold
  $v_t$ decreases as $v_t = \frac{v}{\sqrt{t}}$. If we further assume
  that: $\lVert \nabla f_t(\bm{x}) \rVert \leq L,$ $ \forall \bm{x}
  \in dom(f_t)$ and $\max(D(\bm{x},\bm{x}')) \leq \Delta^{2}, \forall
  \bm{x}, \bm{x}' \in dom(f_t)$. Then, as $T \rightarrow \infty$,
  the regret $R[X] = \sum_{t=1}^{T}f_t(\tilde{\bm{x}_t})-f(\bm{x}^{*}) =
  O(\sqrt{T})$ and therefore $\lim_{T\to\infty} \frac{R[X]}{T}
  \rightarrow 0$.}
\end{theorem}
\addtocounter{theorem}{-1}
\vspace{-5pt}  
}

\section{Implementation}

\khi{We introduce the key components of {\gaia} in
  Section~\ref{subsec:system_component}, and discuss the operation and
  design of individual components in the remaining sections.}

\subsection{{\gaia} System Key Components}
\label{subsec:system_component}

Figure~\ref{fig:server_design} presents the key components of
{\gaia}. All of the key components are implemented in the parameter
servers, and can be transparent to the ML programs and the worker
machines. As we discuss above, we decouple the synchronization within
a data center (LANs) from the synchronization across different data
centers (WANs). The \emph{local server} (\ding{182}) in each parameter
server handles the synchronization between the worker machines in the
same data center using the conventional BSP or SSP \khi{models}. On
the other hand, the \emph{mirror server} (\ding{183}) and the
\emph{mirror client} (\ding{184}) \khi{handle} the synchronization with
other data centers using our \protoabbrv model. Each of these three
components \khi{runs} as an individual thread. \ignore{We first
  explain how these key components handle major system operations and
  communicate with their counterparts, and then discuss their system
  optimizations.}

\ignore{Our implementation of the \protoabbrv
model is inspired by the concept of \emph{control plane and data
  plane} in networks. }

\begin{figure}[h]
  \centering
  \includegraphics[width=0.8\textwidth]{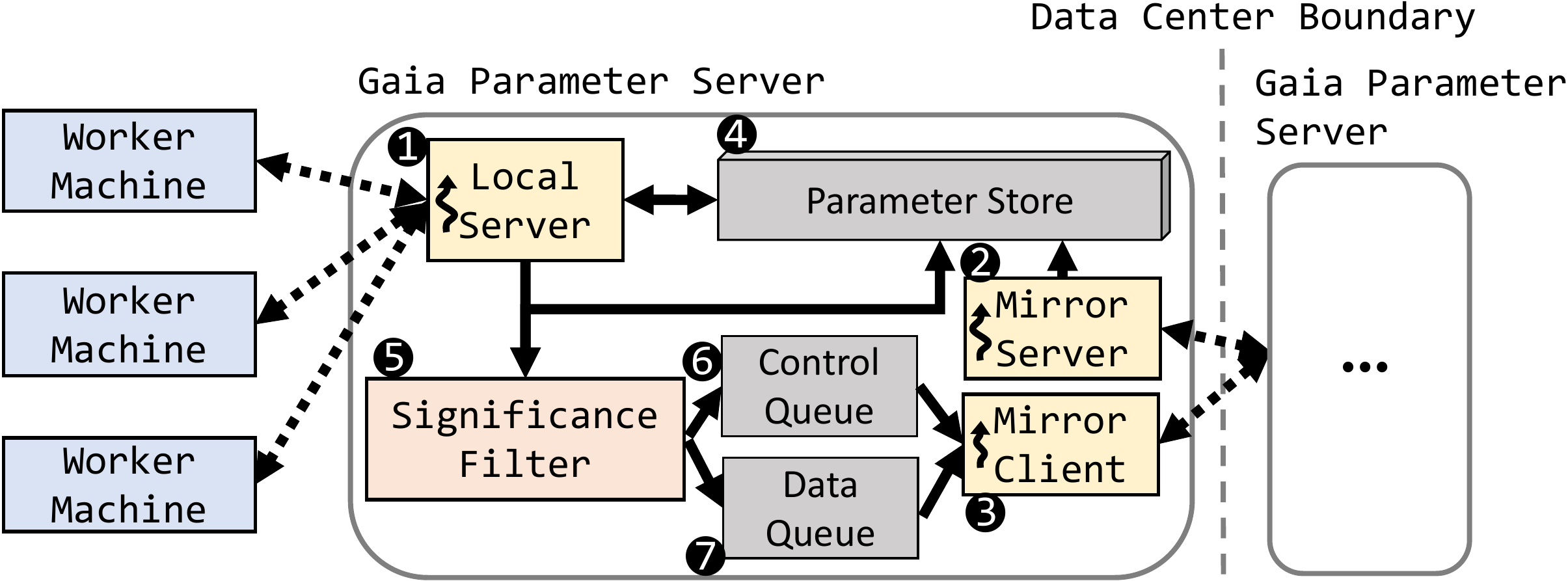}
  \caption{Key components of {\gaia}}
  \label{fig:server_design}
\end{figure}

\subsection{System Operations and Communication}

We present a walkthrough of major system operations and communication.

\textbf{\texttt{UPDATE} from a worker machine.}  When a \emph{local
  server} (\ding{182}) receives a parameter update from a worker
machine, it updates the parameter in its \emph{parameter store}
(\ding{185}), which maintains the parameter value and its accumulated
update. The local server then invokes the \emph{significance filter}
(\ding{186}) to determine whether \khi{or not} the accumulated update
of this parameter is significant. \khi{If it is,} the significance
filter sends a \texttt{MIRROR UPDATE} request to the mirror client
(\ding{184}) and resets the accumulated update for this parameter.

\textbf{Messages from the significance filter.}  \khi{The significance
  filter sends out three types of messages. First, as discussed above,
  it sends a \texttt{MIRROR UPDATE} request to the mirror client
  through the data queue (\ding{188}). Second, when the significance
  filter detects that the arrival rate of significant updates is
  higher than the underlying WAN bandwidth \khi{that it monitors at
    every iteration}, it first sends an \texttt{\protoabbrv Barrier}
  (Section~\ref{subsec:protocol}) to the control queue (\ding{187})
  before sending the \texttt{MIRROR UPDATE}. The mirror client
  (\ding{184}) prioritizes the control queue over the data queue,
  \khi{so that the barrier is} sent out earlier than the
  update. Third, to maintain the mirror clock
  (Section~\ref{subsec:protocol}), the \khi{significance} filter also
  sends a \texttt{MIRROR CLOCK} request to the control queue at the
  end of each clock in the local server.}

\textbf{Operations in the mirror client.}  The mirror client thread
wakes up when there is a request from the control queue or the data
queue.\ignore{The major function of mirror client is to coalesce the
  messages to the same destination.} \khi{Upon} waking up, the mirror
client walks through the queues, packs \khii{together} the messages to
the same destination, and sends them.

\ignore{The mirror client then sends the packed message to the
destination. , with a limit on the size of each packed
message. }

\textbf{Operations in the mirror server.} The mirror server handles
above messages (\texttt{MIRROR UPDATE}, \texttt{\protoabbrv BARRIER},
and \texttt{MIRROR CLOCK}) according to our \protoabbrv model. For
\texttt{MIRROR UPDATE}, it applies the update to the corresponding
parameter in the parameter store. \khi{For} \texttt{\protoabbrv
  BARRIER}, it sets a flag in the parameter store to block the
corresponding parameter from being read until it receives the
corresponding \texttt{MIRROR UPDATE}. \khi{For} \texttt{MIRROR CLOCK},
the mirror server updates its local mirror clock state for each
parameter server in other data centers, and enforces the predefined
clock difference \khi{threshold} $DS$ (Section~\ref{subsec:protocol}).

\ignore{the local server always ensures that the local clock is not ahead
of the slowest mirror clock by a predefined limit.}

\subsection{Advanced Significance Functions}
\label{subsec:sig_func}

As we discuss in Section~\ref{subsec:protocol}, the significance
filter allows the ML programmer to specify a custom \emph{significance
  function} to calculate the significance of each update. By providing
an advanced significance function, {\gaia} can be more effective
\khi{at} eliminating the insignificant communication. If several
parameters are always referenced together to calculate the next
update, the significance function can take into account the values of
all these parameters. For example, if three parameters $a$, $b$, and
$c$ are always used as $a \cdot b \cdot c$ in an ML algorithm, the
significance of $a$, $b$, and $c$ can be calculated as the change on
$a \cdot b \cdot c$. If one of them is 0, \khi{any change in another
  parameter, however large it may be, is insignificant}. Similar
principles can be applied to model parameters that are non-linear or
non-uniform. For unmodified ML programs, the system applies default
significance functions, such as the relative magnitude of an update
for each parameter.

\subsection{Tuning of Significance Thresholds}

\khi{The user of {\gaia} can specify two different goals for {\gaia}:
  (1) speed up algorithm convergence by fully utilizing the available
  WAN bandwidth and (2) minimize the communication cost on WANs. In
  order to achieve \khii{either of these goals}, the significance
  filter maintains two significance thresholds and dynamically tunes
  these thresholds. The first threshold is the \emph{hard}
  significance threshold. The purpose of this threshold is to
  guarantee ML algorithm convergence. As we discuss in our theoretical
  analysis (Section~\ref{subsec:convergence_proof}), the initial
  threshold is provided by the ML programmer or a default system
  setting, and the significance filter reduces it over
  time. \kt{Specifically, the significance threshold decreases whenever
    the learning rate decreases.} Every update whose significance is
  above the hard threshold is guaranteed to be sent to other data
  centers. The second threshold is the \emph{soft} significance
  threshold. \khii{The purpose of it is to use underutilized WAN
    bandwidth to speed up convergence. This threshold} is tuned based
  on the arrival rate of the significant updates and the underlying
  WAN bandwidth. When the user chooses to optimize the first goal
  (speed up algorithm convergence), the system lowers the soft
  significance threshold whenever there is underutilized WAN
  bandwidth. The updates whose significance is larger than the soft
  significance threshold are sent in a best-effort manner. On the
  other hand, if the goal of the system is to minimize the WAN
  communication costs, the soft significance threshold is not
  activated.}

% Kevin: Old version
\ignore{The significance filter maintains two significance thresholds
  and dynamically tunes these thresholds based on the goal of the
  system. The first threshold is the \emph{hard} significance
  threshold. It is the threshold that needs to be always followed to
  guarantee ML algorithm convergence. As we discuss in our theoretical
  analysis (Section~\ref{subsec:convergence_proof}), the initial
  threshold is provided by the ML programmer or a default system
  setting, and the significance filter shrinks it over time. All the
  updates whose significance lies above the hard threshold are
  guaranteed to be sent to other data centers. The second threshold is
  the \emph{soft} significance threshold, which is tuned based on the
  arrival rate of significant updates and the underlying WAN
  bandwidth. When the goal of the system is to fully utilize the WAN
  bandwidth to speedup algorithm convergence, the system can lower the
  soft significance threshold if there is underutilized WAN
  bandwidth. The updates whose significance is larger than the soft
  significance threshold are sent in a best-effort manner. The soft
  significance threshold is not activated if the goal of the system is
  to reduce the communication cost on WAN.}

While the configuration of the initial \khi{hard} threshold depends on
how error tolerant each ML algorithm is, a simple and conservative
threshold (such as 1\%--2\%) \khi{is} likely to work in most
cases. \khii{This is because most ML algorithms initialize their
  parameters with random values,} and make \khi{large} changes to
their model parameters at early phases. Thus, they \khi{are} more
error tolerant at the beginning. As {\gaia} \khi{reduces} the
threshold over time, \khi{its} accuracy loss is
limited. \kt{Typically, the ML programmer selects the initial hard
  threshold by running {\gaia} locally with several values, and then
  selects the threshold value that can achieve target model accuracy
  (e.g., the accuracy of BSP) while minimizing network communication.}
An ML expert can choose a more aggressive threshold based on
\khi{domain} knowledge of the ML algorithm.

\subsection{Overlay Network and Hub}
\label{subsec:overlay_network}

While {\gaia} can \khii{eliminate the insignificant updates}, each data
center needs to \emph{broadcast} the significant updates to all the
other data centers. \khi{This broadcast-based communication could
  limit the scalability of {\gaia}} when we deploy {\gaia} to many data
centers. To make {\gaia} more scalable with \khi{more} data centers, we
use the concept of overlay
networks~\cite{DBLP:journals/comsur/LuaCPSL05}.

As we discuss in Section~\ref{subsec:bandwidth_wan}, the WAN bandwidth
between \khi{geographically-close} regions is much higher than
\khi{that} between distant regions. In light of this, \khii{{\gaia} supports
having geographically-close data centers form} a \emph{data
  center group}. \khi{Servers in a data center group send their}
significant updates \khi{only to the other servers in the same
  group}. Each group has \emph{hub} data centers that are in charge of
aggregating all the significant updates within the group, and
\khi{sending to the hubs} of the other groups. Similarly, \khi{a hub
  data center broadcasts} the aggregated significant updates from
other groups to the other data centers within \khi{its} group. Each
data center group can designate different hubs for communication with
different data center groups, so the system can utilize more links
within a data center group. For example, the data centers in Virginia,
California, and Oregon can form a data center group and assign the
data center in Virginia as the hub to communicate with the data
centers in Europe and the data center in Oregon as the hub to
communicate with the data centers is Asia. This design allows {\gaia}
to broadcast the significant updates with lower communication cost.

\section{Methodology}

\subsection{Experiment Platforms}
\label{subsec:setup}
We use three different platforms for our evaluation. 

\ignore{to measure the effectiveness of {\gaia}.}

\textbf{Amazon-EC2.} We deploy {\gaia} to 22 machines spread across 11
EC2 regions as we show in Figure~\ref{fig:ec2_network_bandwidth}.  In
each EC2 region we start two instances of type \texttt{c4.4xlarge} or
\texttt{m4.4xlarge}\khi{~\cite{amazon_ec2_price}}, depending on their
availability. Both types of instances have 16 CPU cores and at least
30GB RAM, running 64-bit Ubuntu 14.04 LTS (HVM). In all, our
deployment uses 352 CPU cores and 1204 GB RAM.

\textbf{Emulation-EC2.} As the \khi{monetary} cost of running all
experiments on EC2 is too high, we run some experiments on our local
cluster that emulates the computation power and WAN bandwidth of
EC2. We use the same number of machines (22) in our local cluster.
Each machine is equipped with a 16-core Intel Xeon CPU (E5-2698),
\khi{an} NVIDIA Titan X GPU, 64GB RAM, a 40GbE NIC, \khii{and runs}
the same OS \khi{as above}. The computation power and the LAN speeds
of our machines are higher than the ones we get from EC2, so we slow
down the CPU and LAN speeds to \khi{match} the speeds on EC2. We model
the measured EC2 WAN bandwidth
(Figure~\ref{fig:ec2_network_bandwidth}) with \khi{the} Linux Traffic
Control tool~\cite{TC}. As Section~\ref{subsec:performance_ec2} shows,
our emulation platform gives very similar results to the results from
\khi{our} real EC2 deployment.

\textbf{Emulation-Full-Speed.} We run some of our experiments on our
local cluster that emulates the WAN bandwidth of EC2 at \emph{full
  speed}. \khi{We use} the same settings as \textbf{Emulation-EC2}
except we do not slow down the CPUs and the LAN. We use this platform
to show the results of deployments with more powerful nodes.

\subsection{Applications}
\label{subsec:workload}
We evaluate {\gaia} with three popular ML applications.  \ignore{We use
  the three popular ML applications to test the effectiveness of
  {\gaia}.}

\textbf{Matrix Factorization (MF)} is a technique commonly used in
\khi{recommender} systems, e.g., \khi{\khii{systems} that recommend}
movies to users on Netflix (a.k.a.~collaborative
filtering)~\cite{DBLP:conf/kdd/GemullaNHS11}.  Its goal is to discover
latent interactions between two entities, such as users and movies,
via matrix factorization. For example, input data can be a partially
filled matrix $X$, where every entry is a user's rating for a movie,
each row corresponding to a user, and each column \khi{corresponding}
to a specific movie.  Matrix factorization factorizes $X$ into factor
matrices $L$ and $R$ such that their product approximates $X$ (i.e.,
$X\approx LR$).  Like other systems~\cite{harlap2016straggler,
  DBLP:conf/nips/ZinkevichSL09, CuiTWXDHHGGGX14}, we implement
\emph{MF} using the stochastic gradient descent (SGD) algorithm. Each
worker is assigned a portion of the \khi{known} entries in $X$. The
$L$ matrix is stored locally in each worker, and the $R$ matrix is
stored in parameter servers. Our experiments use the \emph{Netflix}
dataset, a 480K-by-18K sparse matrix with 100M known entries. They are
configured to factor the matrix into the product of two matrices\khi{,
  each} with rank 500.

\textbf{Topic Modeling (TM)} is an unsupervised method for discovering
hidden semantic structures (\emph{topics}) in an unstructured
collection of \emph{documents}, each consisting of a bag (multi-set)
of \emph{words}~\cite{BleiNJ03}.  \emph{TM} discovers the topics via
word co-occurrence. For example, ``policy'' is more likely to co-occur
with ``government'' than ``bacteria'', and thus ``policy'' and
``government'' are categorized to the same topic associated with
political terms\ignore{, and ``bacteria'' to another topic associated
  with disease terms}. Further, a document with many instances of
``policy'' would be assigned a topic distribution that peaks for the
\khi{politics-related} topics. \emph{TM} learns the hidden topics and
the documents' associations with those topics jointly.  Common
applications for \emph{TM} include community detection in social
networks and news categorizations\ignore{, and groupings from genetics
  data}. We implement our \emph{TM} solver using collapsed Gibbs
sampling~\cite{griffiths2004finding}. We use the \emph{Nytimes}
dataset~\cite{nytimes-web}, which has 100M words in 300K documents
with a vocabulary size of 100K. \khi{Our experiments
classify words and documents into 500 topics.}

\ignore{During every iteration,
  each worker goes through its subset of documents and adjusts the
  topic assignment of the documents and the words.} 

{\bf Image Classification (IC)} is a task to classify images into
categories, and the state-of-the-art approach is using deep learning
and \khi{convolutional} neural networks
(CNNs)~\cite{DBLP:journals/neco/LeCunBDHHHJ89}. Given a set of images
with known categories (training data), the ML algorithm trains a CNN
to learn the relationship between the image features and their
categories. The trained CNN is then used to predict the categories of
another set of images (test data). We use
GoogLeNet~\cite{DBLP:conf/cvpr/SzegedyLJSRAEVR15}, one of the
state-of-the-art CNNs as our model. We train GoogLeNet using
stochastic gradient descent with
back propagation~\cite{rumelhart1988learning}. As training a CNN with
a large number of images requires \khi{substantial} computation, doing
so on CPUs can take hundreds of machines over a
week~\cite{DBLP:conf/osdi/ChilimbiSAK14}. Instead, we use distributed
GPUs with a popular deep learning framework,
Caffe~\cite{DBLP:journals/corr/JiaSDKLGGD14}, which is hosted by a
state-of-the-art GPU-specialized parameter server system,
GeePS~\cite{DBLP:conf/eurosys/CuiZGGX16}. Our experiments use the
ImageNet Large Scale Visual Recognition Challenge 2012
(ILSVRC12)~\cite{ILSVRC15} dataset, which \khi{consists of 1.3M
  training images and 50K test images}. Each image is labeled as
  one of the 1,000 pre-defined categories.

\subsection{Performance Metrics and Algorithm Convergence Criteria}
We use two performance metrics to evaluate the effectiveness of a
globally distributed ML system. The first metric is the
\emph{execution time \khi{until} algorithm convergence}. We use the
following algorithm convergence criterion, based on guidance from our
ML experts: if the value of the objective function \khi{(the
  \emph{objective value})} in an algorithm changes \khi{by} less than
2\% over the course of 10 iterations, \khi{we declare that the
  algorithm has converged~\cite{harlap2016straggler}}. \khii{In order
  to ensure that each algorithm \emph{accurately} converges to the
  optimal point, we first run each algorithm on our local cluster
  until it converges, and we record the absolute objective value.} The
execution time of each setting is the time it takes to converge to
this absolute objective value. \ignore{Because the objective value
  calculation is relatively expensive, we calculate it offline using
  our system checkpoints.}

The second metric is the \emph{cost of
  algorithm convergence}. We calculate the cost based on the cost
model of Amazon EC2~\cite{amazon_ec2_price}, including the cost of the
server time and the cost of data transfer on WANs. 
%\khii{We provide the details of the cost model in Appendix~\ref{appendix:cost_model}.}
We use the on-demand pricing of Amazon EC2 published for January 2017
as our cost model~\cite{amazon_ec2_price}. As the pricing might change
over time, we provide the details of the cost model in
Table~\ref{table:cost_model}. The
CPU instance is \texttt{c4.4xlarge} or \texttt{m4.4xlarge},
depending on the availability in each EC2 region. The GPU instance is
\texttt{g2.8xlarge}. The low-cost instance (\texttt{m4.xlarge}) is the
one used for centralizing input data. All the instance costs are shown
in USD per hour. All WAN data transfer costs are shown in USD per GB.

\setlength{\tabcolsep}{0.5em} % for the horizontal padding
\begin{table}[h]
\small \centering 
\begin{tabular}{|c||c|c|c|c|c|} \hline

Region & \scell{CPU \\ Instance} & \scell{GPU \\ Instance}
& \scell{Low-cost \\ Instance} & \scell{Send to \\
  WANs} & \scell{Recv. from \\ WANs} \\ \hhline{|=#=|=|=|=|=|}

Virginia & \$0.86 & \$2.60 & \$0.22 & \$0.02 & \$0.01 \\ \hline 

California & \$1.01 & \$2.81 & \$0.22 & \$0.02 & \$0.01 \\ \hline

Oregon & \$0.86 & \$2.60 & \$0.22 & \$0.02 & \$0.01 \\ \hline

Ireland & \$0.95 & \$2.81 & \$0.24 & \$0.02 & \$0.01 \\ \hline

Frankfurt & \$1.03 & \$3.09 & \$0.26 & \$0.02 & \$0.01 \\ \hline

Tokyo & \$1.11 & \$3.59 & \$0.27 & \$0.09 & \$0.01 \\ \hline

Seoul & \$1.06 & \$3.59 & \$0.28 & \$0.08 & \$0.01 \\ \hline

Singapore & \$1.07 & \$4.00 & \$0.27 & \$0.09 & \$0.01 \\ \hline

Sydney & \$1.08 & \$3.59 & \$0.27 & \$0.14 & \$0.01 \\ \hline

Mumbai & \$1.05 & \$3.59 & \$0.26 & \$0.09 & \$0.01 \\ \hline

S\~{a}o Paulo & \$1.37 & \$4.00 & \$0.34 & \$0.16 & \$0.01 \\ \hline

\end{tabular}
\caption{Cost model details}
\label{table:cost_model}
\end{table}

\section{Evaluation Results}
\label{sec:evaluation}

We evaluate the effectiveness of {\gaiasys} \khi{by evaluating three
  types of systems/deployments}: (1) \baselinesys, two
state-of-the-art parameter server systems
(IterStore~\cite{CuiTWXDHHGGGX14} for \emph{MF} and \emph{TM},
GeePS~\cite{DBLP:conf/eurosys/CuiZGGX16} for \emph{IC}) that are
deployed across multiple data centers. Every worker machine handles
the data in its data center, while the parameter servers are
distributed evenly across all the data centers; (2) {\gaiasys}, our
prototype systems based on IterStore and GeePS, deployed across
multiple data centers; and (3) \khii{\lansys, the baseline parameter
  servers (IterStore and GeePS) that are deployed within a single data
  center (also on 22 machines) that already hold all the data,
  representing the ideal case of all communication on a LAN}. For each
\khi{system}, we evaluate two ML synchronization models: \emph{BSP}
and \emph{SSP} (Section~\ref{subsec:ml_training_system}). For
\baselinesys and \lansys, \khi{BSP and SSP are used} among all worker
machines, whereas for {\gaiasys}, they are used only within each data
center. For better readability, we present the results for BSP first and 
show the results for SSP in Section~\ref{subsec:ssp}.

\subsection{Performance on EC2 Deployment}
\label{subsec:performance_ec2}

We first present the performance of {\gaiasys} and \baselinesys when
they are deployed across 11 EC2 data
centers. Figure~\ref{fig:performance_11_regions} shows the normalized
execution time \khi{until} convergence for our ML applications,
normalized to \baselinesys on EC2. The data label on each bar is the
speedup over \baselinesys for \emph{the respective deployment}. As
Section~\ref{subsec:setup} discusses, we run only \emph{MF} on EC2
due to the high \khi{monetary} cost of WAN data transfer. Thus, we
present the results of \emph{MF} on all three platforms, while we show
the results of \emph{TM} and \emph{IC} \khi{only} on our emulation
platforms. As Figure~\ref{fig:performance_mf_11_regions} shows, our
emulation platform (\emph{Emulation-EC2}) matches the execution time
of \khi{our} real EC2 deployment (\emph{Amazon-EC2}) very well. We
make two major observations.

\begin{figure*}[h!]
 \centering
 \begin{subfigure}[t]{0.7\linewidth}
 \centering
 \includegraphics[width=1.0\textwidth]{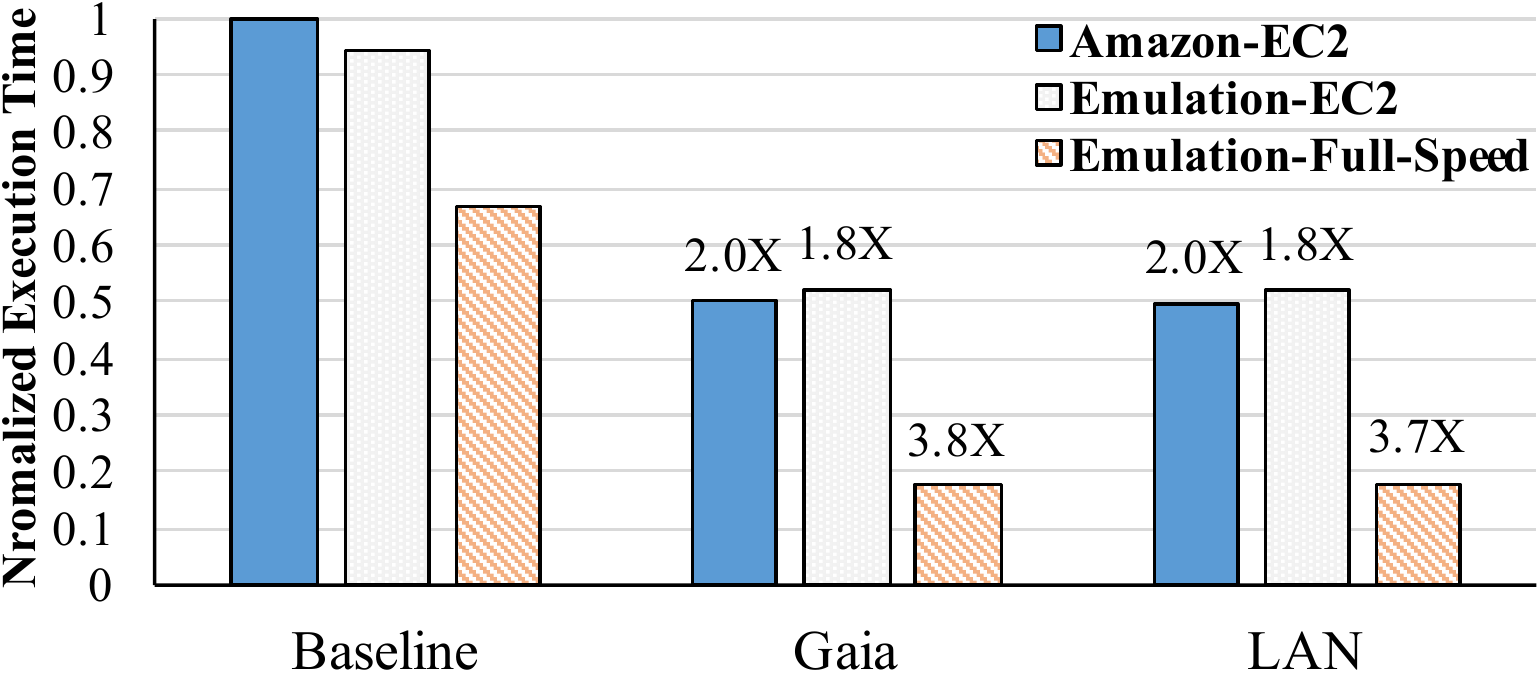}
 \caption{\emph{Matrix Factorization (MF)}} 
 \label{fig:performance_mf_11_regions}
 \end{subfigure}
 \begin{subfigure}[t]{0.7\linewidth}
 \centering
 \includegraphics[width=1.0\textwidth]{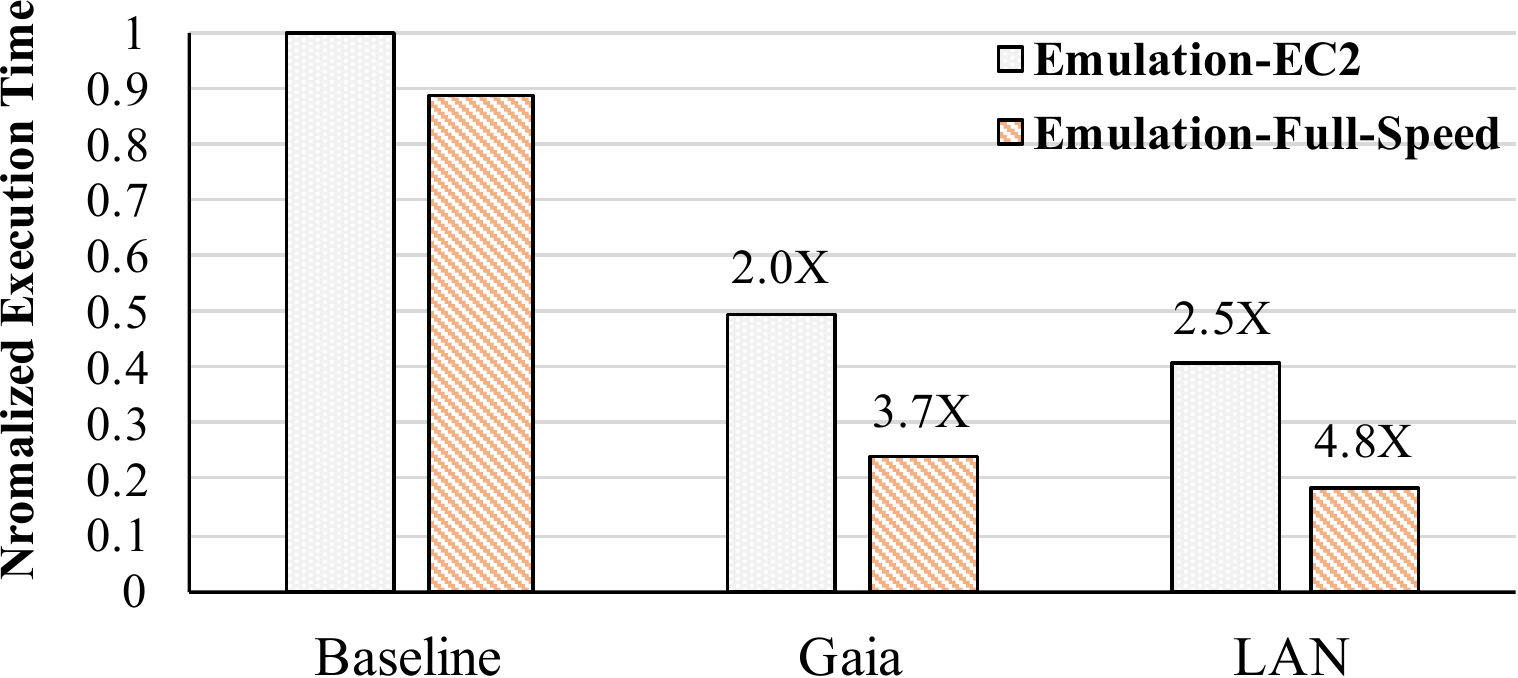} 
 \caption{\emph{Topic Modeling (TM)}}
 \label{fig:performance_tm_11_regions}
 \end{subfigure}
 \begin{subfigure}[t]{0.7\linewidth}
 \centering
 \includegraphics[width=1.0\textwidth]{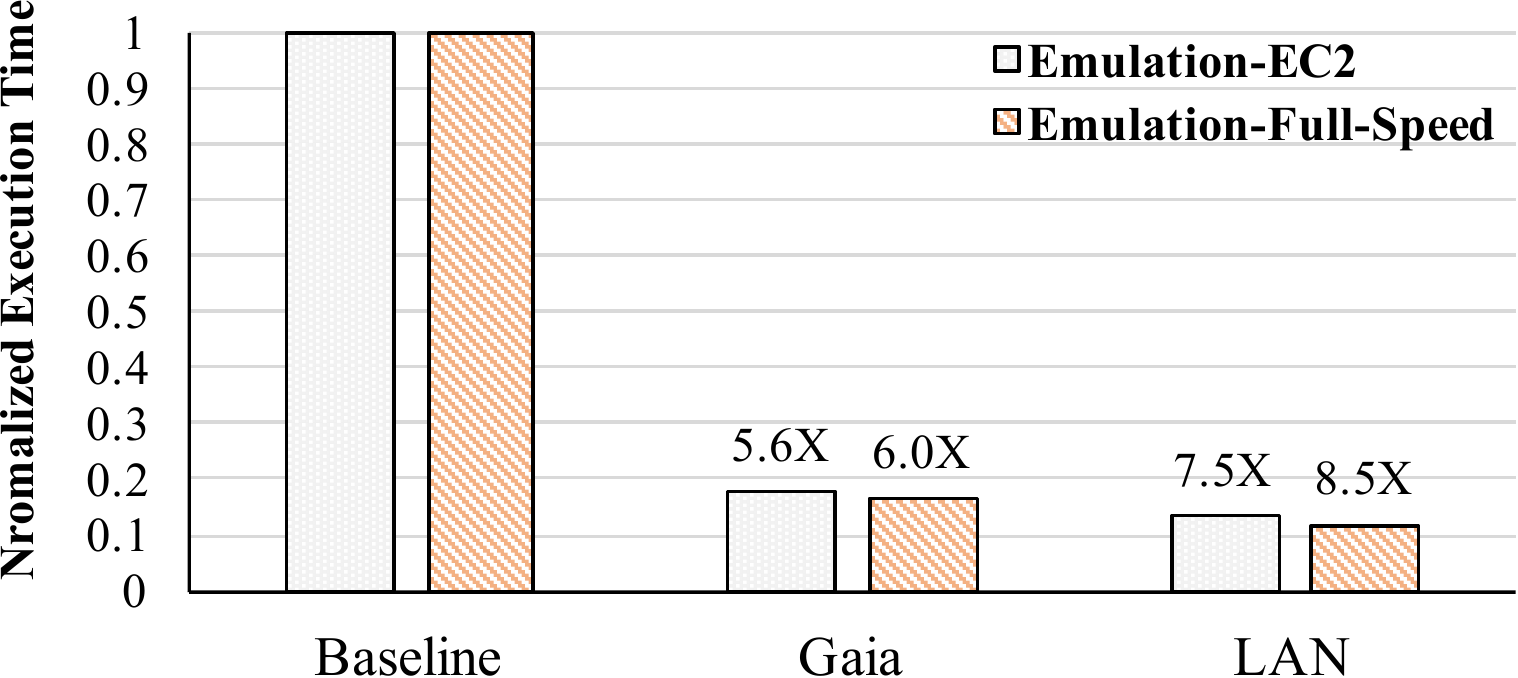} 
 \caption{\emph{Image Classification (IC)}}
 \label{fig:performance_ic_11_regions}
 \end{subfigure}
 \caption{Normalized execution time until convergence when deployed
   across 11 EC2 regions and our emulation cluster}
 \label{fig:performance_11_regions}
 \vspace{10pt}
\end{figure*} 

First, we find \khi{that} {\gaiasys} significantly improves the
performance of \baselinesys when deployed globally across many EC2
data centers. For \emph{MF}, {\gaiasys} provides a speedup of
2.0$\times$ over \baselinesys. Furthermore, the performance of
{\gaiasys} is very similar to the performance of \khi{\lansys},
\khi{indicating that} {\gaiasys} almost \khii{attains} the performance
\emph{upper bound} with the given computation resources. For
\emph{TM}, {\gaiasys} delivers a similar speedup (2.0$\times$) and is
within 1.25$\times$ of the ideal speed of \lansys. \khi{For}
\emph{IC}, {\gaiasys} provides a speedup of 5.6$\times$ over
\baselinesys, which is within 1.32$\times$ of the \lansys speed,
\khi{indicating} that {\gaiasys} is also effective on a GPU-based ML
system. The gap between \baselinesys and \lansys is larger for
\emph{IC} than \khi{for} the other two applications. This is because
the GPU-based ML system generates parameter updates at a higher rate
than the CPU-based one, \khi{and therefore} the limited WAN bandwidth
slows it down more significantly.

Second, {\gaiasys} provides a higher performance gain when deployed on
a more powerful platform. As Figure~\ref{fig:performance_11_regions}
shows, the performance gap between \khi{\baselinesys and \lansys}
significantly increases on \emph{Emulation-Full-Speed} compared to the
slower platform \emph{Emulation-EC2}. This is expected because the WAN
bandwidth becomes a more critical bottleneck when the computation time
reduces and the LAN bandwidth increases. {\gaiasys} successfully
mitigates the WAN bottleneck in this more challenging
\khii{\emph{Emulation-Full-Speed}} setting, and improves the system
performance by 3.8$\times$ for \emph{MF}, 3.7$\times$ for \emph{TM},
and 6.0$\times$ for \emph{IC} over \baselinesys, \khi{ approaching the
  speedups provided by \lansys}.

\ignore{Third, SSP is not necessarily the better local synchronization model
for {\gaiasys}. As the results of \emph{TM} shows, {\gaiasys} with BSP
outperforms {\gaiasys} with SSP. The reason is that SSP allows using stale
values to get the benefit of more efficient communication. However,
compared to \baselinesys, the benefit of employing SSP to reduce
communication overhead is much smaller because {\gaiasys} only uses it to
synchronize a few machines in a data center. Thus, the cost of
inaccuracy outweighs the benefit for SSP in this case. Fortunately,
{\gaiasys} decouples the synchronization model within a data center from
the synchronization model across different data centers, so we can
freely choose the one that works better in {\gaiasys}.}

\subsection{Performance and WAN Bandwidth}
\label{subsec:performance_wan_bandwidth}

To understand how {\gaiasys} performs under different \khi{amounts of}
WAN bandwidth, we evaluate \khi{two} settings where \baselinesys and
{\gaiasys} are deployed across two data centers with two WAN bandwidth
configurations: (1) \emph{V/C WAN}, \khi{which} emulates the WAN
bandwidth between Virginia and California, representing a setting
within the same continent; and (2) \emph{S/S WAN}, \khi{which}
emulates the WAN bandwidth between Singapore and S\~{a}o Paulo,
representing the \khi{lowest} WAN bandwidth \khi{between any two}
Amazon EC2 sites. All the experiments are conducted on our emulation
platform at full speed. \khi{Figures~\ref{fig:performance_vc_wan}
  and~\ref{fig:performance_ss_wan} show the results.} Three
observations are in order.

\begin{figure}[h!]
 \centering
 \begin{subfigure}[t]{0.34\linewidth}
 \centering
 \includegraphics[width=1.0\textwidth]{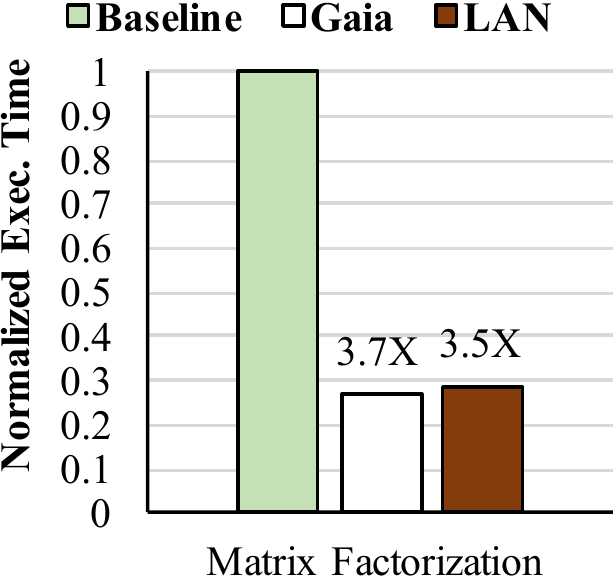}
 \end{subfigure}
 \begin{subfigure}[t]{0.312\linewidth}
 \centering
 \includegraphics[width=1.0\textwidth]{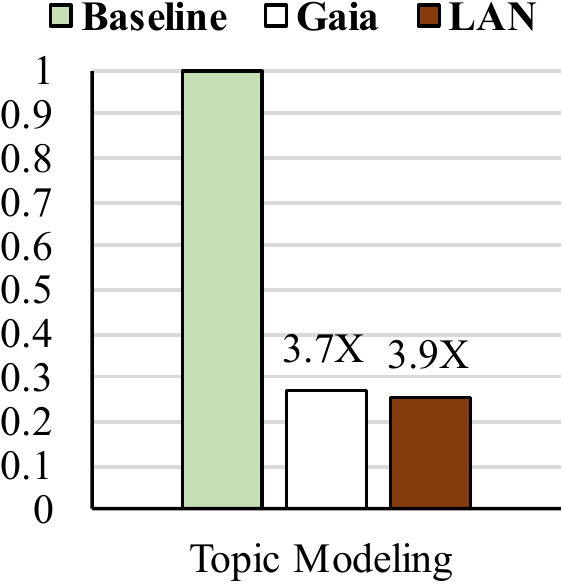} 
 \end{subfigure}
 \begin{subfigure}[t]{0.312\linewidth}
 \centering
 \includegraphics[width=1.0\textwidth]{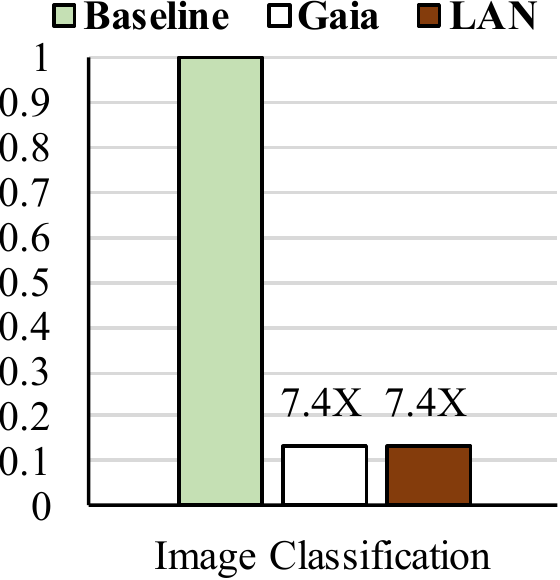} 
 \end{subfigure}
 \caption{Normalized execution time \khi{until} convergence with the
   WAN bandwidth between Virginia and California}
 \label{fig:performance_vc_wan}
\end{figure}

\begin{figure}[h!]
 \centering
 \begin{subfigure}[t]{0.34\linewidth}
 \centering
 \includegraphics[width=1.0\textwidth]{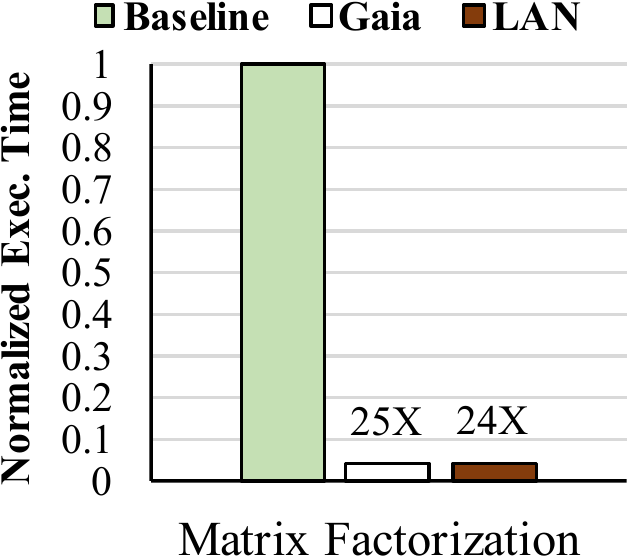}
 \end{subfigure}
 \begin{subfigure}[t]{0.315\linewidth}
 \centering
 \includegraphics[width=1.0\textwidth]{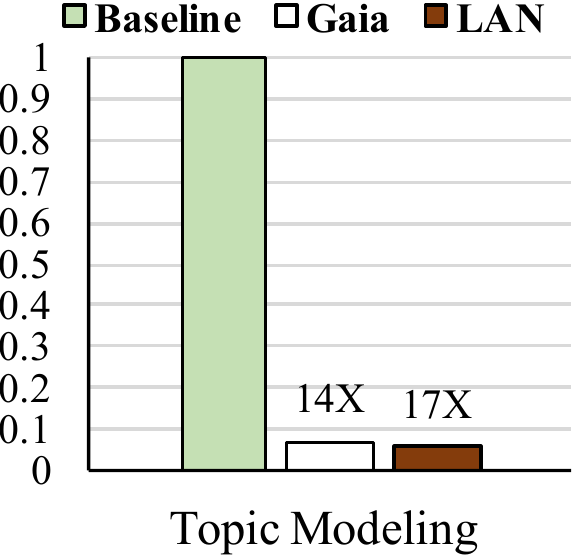} 
 \end{subfigure}
 \begin{subfigure}[t]{0.315\linewidth}
 \centering
 \includegraphics[width=1.0\textwidth]{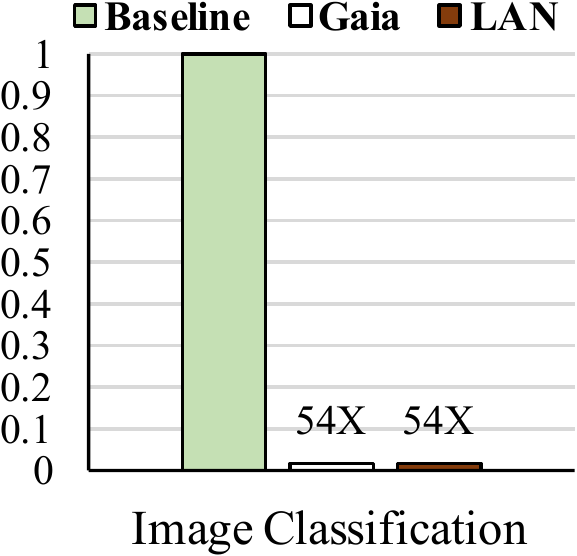} 
 \end{subfigure}
 \caption{Normalized execution time \khi{until} convergence with the WAN
   bandwidth between Singapore and S\~{a}o Paulo}
 \label{fig:performance_ss_wan}
\end{figure} 

First, {\gaiasys} successfully matches the performance of \lansys \khi{when
  WAN bandwidth is high (\emph{V/C WAN})}. As
Figure~\ref{fig:performance_vc_wan} shows, {\gaiasys} achieves a speedup
of 3.7$\times$ for \emph{MF}, 3.7$\times$ for \emph{TM}, and
7.4$\times$ for \emph{IC}. For all three ML applications, the
performance of {\gaiasys} on \khi{WANs} is almost the same as \khi{\lansys
  performance}.

\ignore{For \emph{TM}, {\gaiasys} achieves a
higher speedup than the one on 11 data centers
(Section~\ref{subsec:performance_ec2}). The reason is that {\gaiasys}
needs to broadcast the significant updates to all the other data
centers, so the communication overhead is higher as the number of data
centers increases. While we employ the concept of network overlay
(Section~\ref{subsec:overlay_network}) to mitigate this effect, there
is still more overhead comparing to the setting of two data centers.}

Second, {\gaiasys} still performs very well \khi{when WAN bandwidth is
  low (\emph{S/S WAN}, Figure~\ref{fig:performance_ss_wan}):} {\gaiasys}
provides a speedup of 25.4$\times$ for \emph{MF}, 14.1$\times$ for
\emph{TM}, and 53.5$\times$ for \emph{IC}, and successfully approaches
\lansys \khii{performance}. \khi{These results show that} our design is
robust for both CPU-based and GPU-based ML systems, and \khi{it
  can deliver high performance even under scarce WAN bandwidth}.

Third, for \emph{MF}, the performance of {\gaiasys} (on \khi{WANs}) is
slightly better than \khii{\lansys} performance. \khi{This} is
because we run ASP between different data centers, and the workers in
each data center need to synchronize \khi{only} with each other
locally in each iteration. As long as the mirror updates on WANs
\khi{are} timely, each iteration of {\gaiasys} can be faster than
\khi{that} of \lansys, \khi{which needs to synchronize across all
  workers. While {\gaiasys} needs more iterations than \lansys
  due to the accuracy loss, {\gaiasys} can still outperform \lansys due
  to the faster iterations.}

\subsection{Cost Analysis}
\label{subsec:cost}

Figure~\ref{fig:cost} shows the monetary cost of running ML
applications \khii{until} convergence based on the Amazon EC2 cost
model, normalized to the cost of \baselinesys on 11 EC2
regions. \khi{Cost is divided into three components: (1) the cost of
  machine time spent on computation, (2) the cost of machine time spent
  on waiting for networks, and (3) the cost of data transfer across
  different data centers.} As we discuss in
Section~\ref{subsec:bandwidth_wan}, there is no cost for data transfer
within a \khi{single} data center in \khi{Amazon EC2}. The data label
on each bar shows \khi{the factor by which} the cost of {\gaiasys} is
cheaper than the cost of \emph{each respective} \baselinesys. We
evaluate all three deployment setups that we discuss in
Sections~\ref{subsec:performance_ec2}
and~\ref{subsec:performance_wan_bandwidth}. We make \khi{two} major
observations.

\begin{figure*}[h!]
 \centering
 \begin{subfigure}[t]{0.6\linewidth}
 \centering
 \includegraphics[width=1.0\textwidth]{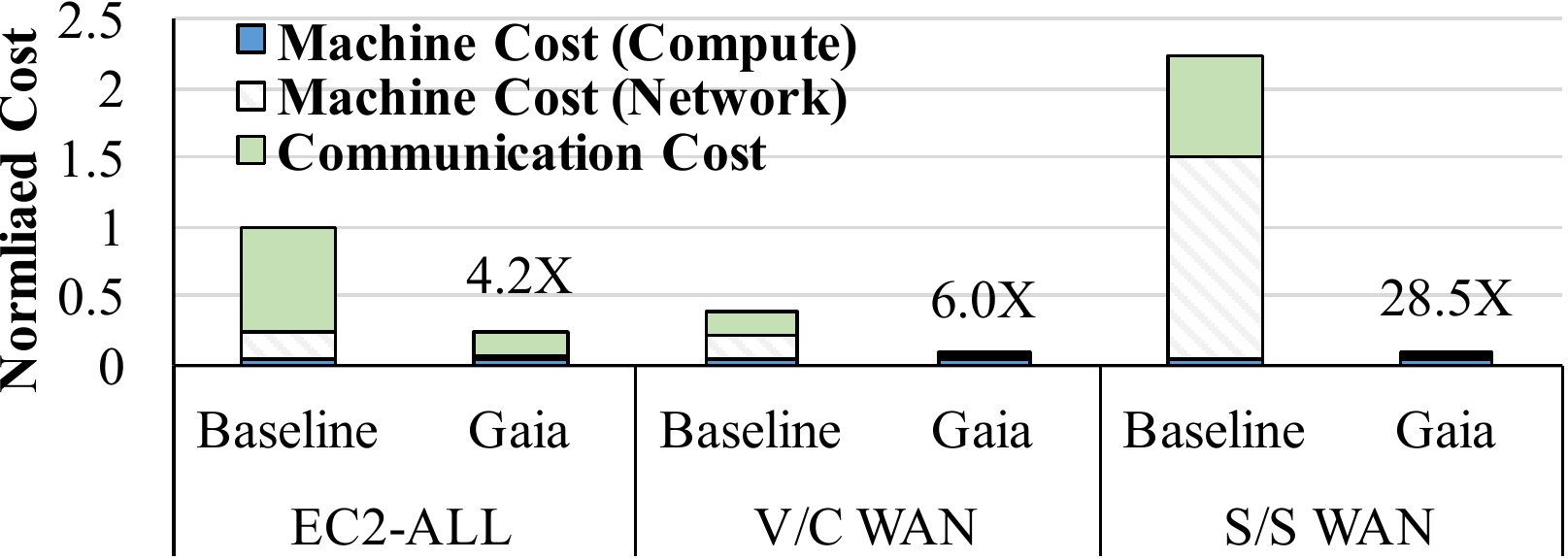}
 \caption{\emph{Matrix Factorization (MF)}} 
 \label{fig:cost_mf}
 \end{subfigure}
 \begin{subfigure}[t]{0.6\linewidth}
 \centering
 \includegraphics[width=1.0\textwidth]{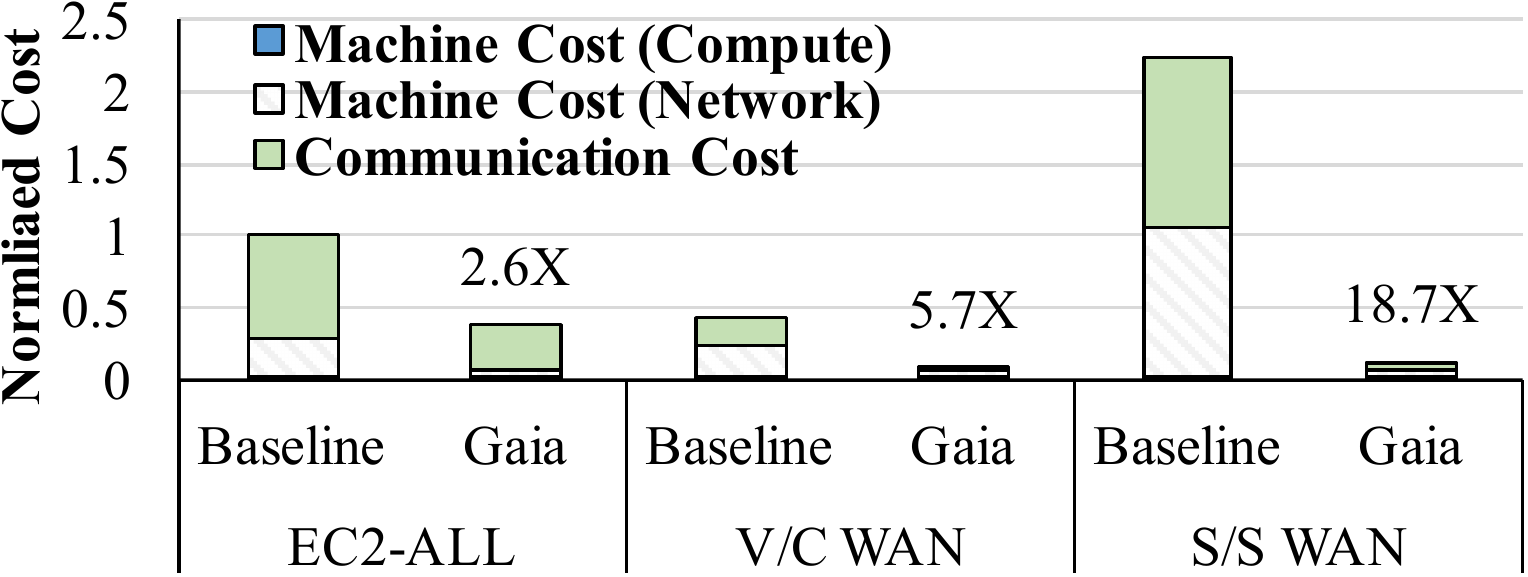} 
 \caption{\emph{Topic Modeling (TM)}}
 \label{fig:cost_tm}
 \end{subfigure}
 \begin{subfigure}[t]{0.6\linewidth}
 \centering
 \includegraphics[width=1.0\textwidth]{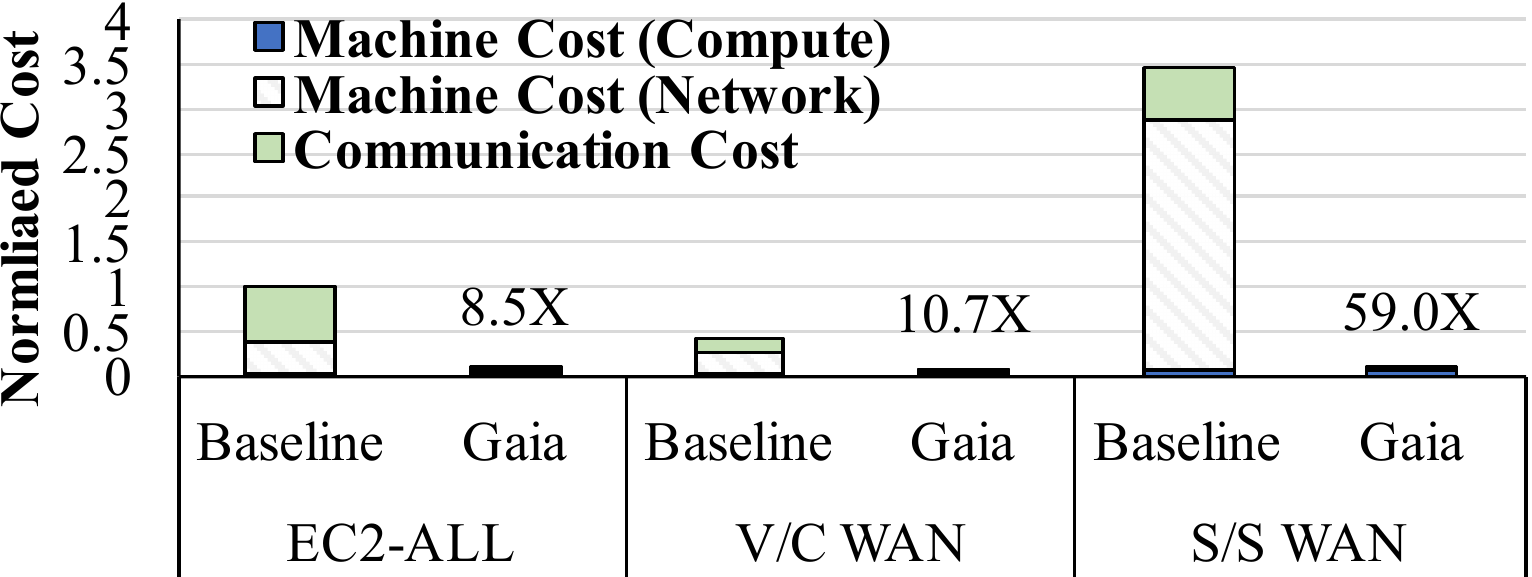} 
 \caption{\emph{Image Classification (IC)}}
 \label{fig:cost_ic}
 \end{subfigure}
 \caption{\khii{Normalized monetary cost of {\gaiasys} vs. \baselinesys}}
 \label{fig:cost}
% \vspace{5pt}
\end{figure*} 

\ignore{First, the cost of data transfer on WANs and the cost of machines
waiting for networks are the dominant factors of the total
cost. Across the evaluated settings, \khi{the sum of these two cost
  sources} are responsible for \khi{86.8--98.9\%} of the total
cost. \khi{For} \baselinesys, the cost of data transfer for the
EC2-ALL setting is higher than the one for the V/C WAN setting. This
is because when we have more data centers, more parameter servers are
not in the same data center so more traffic goes to WANs. The reason
why data transfer in the S/S WAN setting is more costly than the V/C
WAN setting is because \khi{transferring a} byte between Singapore and
S\~{a}o Paulo is more expensive than \khi{doing so} between Virginia
and California.}

First, {\gaiasys} is very effective in \khii{reducing} the cost of
running a geo-distributed ML application. Across all the evaluated
settings, {\gaiasys} is 2.6$\times$ to 59.0$\times$ cheaper than
\baselinesys. Not surprisingly, the major cost saving comes from the
reduction of data transfer on WANs and the reduction of machine time
\khii{spent on} waiting for networks. For the S/S WAN setting, the
cost of waiting for networks is a more important factor than the other
two settings, because it takes more time to transfer the same amount
of data under \khi{low WAN bandwidth}. As {\gaiasys} significantly
improves system performance and reduces data communication overhead,
it significantly reduces both cost sources. We conclude that {\gaiasys}
is a cost-effective system for geo-distributed ML applications.

Second, {\gaiasys} reduces data transfer cost much more when deployed
on a \khi{smaller} number of data centers. The reason is that
{\gaiasys} needs to broadcast the significant updates to
\khi{\emph{all} data centers}, so communication cost is higher as the
number of data centers increases. While we employ \khi{network
  overlays} (Section~\ref{subsec:overlay_network}) to mitigate this
effect, there is still more overhead \khii{with more than two data
  centers.}  Nonetheless, the cost of {\gaiasys} is still much cheaper
(4.2$\times$/2.6$\times$/8.5$\times$) than \baselinesys \khi{even}
when deployed across 11 data centers.

\subsection{Comparisons with \khi{Centralized Data}}
\label{subsec:centralize}

\khii{{\gaia} obtains its good performance without moving any raw data,
  greatly reducing WAN costs and respecting privacy and data
  sovereignty laws that \emph{prohibit} raw data movement. For
  settings in which raw data movement is \emph{allowed},}
Table~\ref{table:centralize_compare} summarizes the performance and
cost comparisons between {\gaiasys} and the \khi{centralized} data
approach (\khi{\centralizesys}), which moves \khi{\emph{all}} the
geo-distributed data into \khi{a single} data center and then runs the
ML application over the data. We make \centralizesys very cost
efficient by moving the data into the \emph{cheapest} data center in
each setting, and we use low cost machines
(\texttt{m4.xlarge}\khi{~\cite{amazon_ec2_price}}) to move the
data. We make two major observations.

\begin{table}[h]
\centering \caption{Comparison between {\gaiasys} and \centralizesys}
\label{table:centralize_compare}	
\begin{tabular}{|c|c|c|c|} \hline
Application & Setting & \scell{{\gaiasys} Speedup \\ over \centralizesys} & \scell{{\gaiasys}
cost / \\ \centralizesys
cost} \\ \hline

\multirow{3}{*}{MF} & EC2-ALL & 1.11 & 3.54 \\ \cline{2-4}
                    & V/C WAN & 1.22 & 1.00 \\ \cline{2-4}
                    & S/S WAN & 2.13 & 1.17 \\  \hline

\multirow{3}{*}{TM} & EC2-ALL & 0.80 & 6.14 \\ \cline{2-4}
                    & V/C WAN & 1.02 & 1.26 \\ \cline{2-4}
                    & S/S WAN & 1.25 & 1.92 \\  \hline

\multirow{3}{*}{IC} & EC2-ALL & 0.76 & 3.33 \\ \cline{2-4}
                    & V/C WAN & 1.12 & 1.07 \\ \cline{2-4}
                    & S/S WAN & 1.86 & 1.08 \\  \hline
\end{tabular} 
\end{table}

First, {\gaiasys} outperforms \centralizesys for most settings, except
for \emph{TM} and \emph{IC} in the EC2-ALL setting. Other than these
two cases, {\gaiasys} provides a 1.02--2.13$\times$ speedup over
\centralizesys. \khi{This} is because {\gaiasys} does not need to wait
for data movement over WANs, and the performance of {\gaiasys} is very
close to \khi{that} of \lansys. On the other hand, \centralizesys
performs better when there is a performance gap between {\gaiasys} and
\lansys, especially in the setting of \khi{all 11 data centers for
  \emph{TM} and \emph{IC}}. \khi{The data movement overhead of
  \centralizesys is smaller in this setting} because each data center
has \khi{only} a small fraction of the data, and \khi{\centralizesys
  moves the data from all data centers in parallel.}

Second, \centralizesys is more cost-efficient than {\gaiasys}, but the
gap is small in the two data centers setting. This is because the
total WAN traffic of {\gaiasys} is still larger than the size of the
training data, even though {\gaiasys} significantly reduces the
communication overhead over \baselinesys. The cost gap is larger in
the setting of 11 data centers (3.33--6.14$\times$) than \khi{in} two
data centers (1.00--1.92$\times$), because the WAN traffic of
{\gaiasys} is \khi{positively correlated with} the number of data
centers (Section~\ref{subsec:overlay_network}).

\ignore{
It is worth noting that \centralizesys may not be possible due to
\khii{the constraints of privacy and data sovereignty laws}. {\gaiasys}
provides a high-performance and cost-efficient alternative to
\centralizesys, especially if the number of data centers is
small. When the number of data centers is large, the optimal option can
be \khi{to move} data to several data centers (within \khi{constraints
  of laws}) before \khi{running} {\gaiasys}. We leave the exploration of
\khi{such alternatives that combine {\gaiasys} with input data
  movement} to future work.
}

\subsection{Effect of Synchronization Mechanisms}
\label{subsec:async}

One of the major design \khi{considerations} of \protoabbrv is to
ensure \khi{that} the significant updates arrive in a \khi{timely}
manner to guarantee algorithm convergence. To understand the
effectiveness of our proposed synchronization mechanisms (\khi{i.e.,}
\protoabbrv selective barrier and mirror clock), we \khi{run}
\emph{MF} and \emph{TM} on {\gaiasys} with both mechanisms disabled
\khi{across} 11 EC2 regions. Figure~\ref{fig:async_effect} shows the
progress toward algorithm convergence with the synchronization
mechanisms enabled ({\gaiasys}) and disabled
(\texttt{Gaia\textunderscore Async}). For \emph{MF}, lower object
value is better, while for \emph{TM}, higher is better.

\ignore{The objective value for
\emph{MF} is the lower, the better, while the one for \emph{TM} is
the opposite.}

\begin{figure}[h!]
% \vspace{5pt}
 \centering
 \begin{subfigure}[t]{0.48\linewidth}
 \centering
 \includegraphics[width=1.0\textwidth]{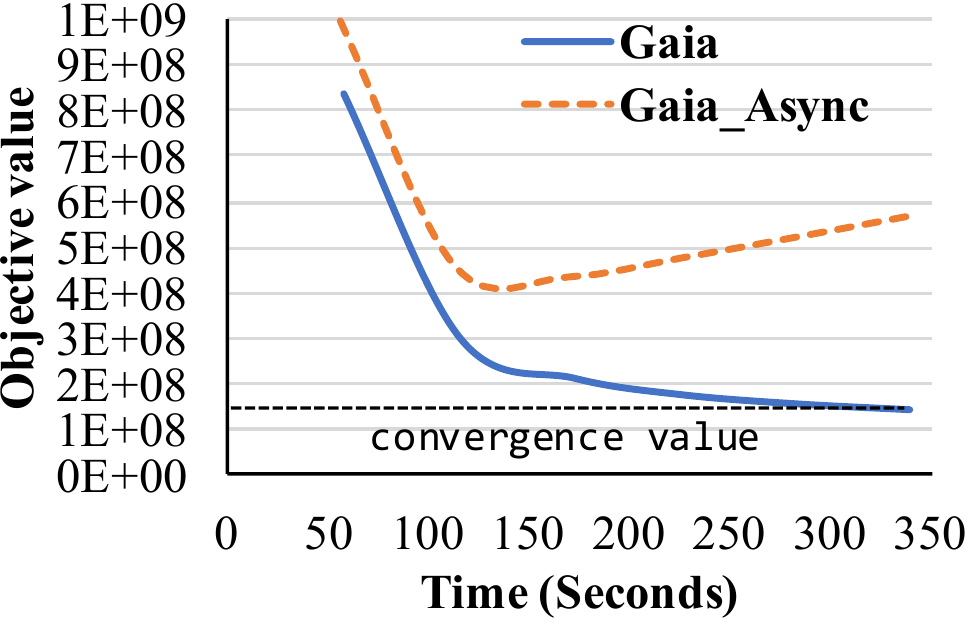}
 \caption{\emph{Matrix Factorization (MF)}} 
 \label{fig:mf_async}
 \end{subfigure}
 \begin{subfigure}[t]{0.48\linewidth}
 \centering
 \includegraphics[width=1.0\textwidth]{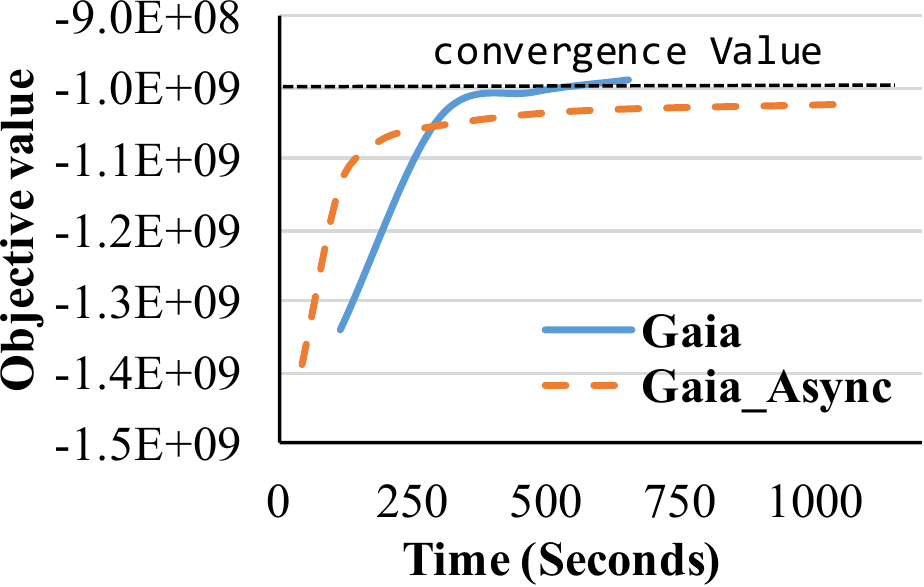} 
 \caption{\emph{Topic Modeling (TM)}}
 \label{fig:tm_async}
 \end{subfigure}
 \caption{Progress toward algorithm convergence \khi{with and without}
   {\gaia}'s synchronization mechanisms}
 \label{fig:async_effect}
\end{figure}

As \khi{Figure~\ref{fig:async_effect} shows}, {\gaiasys} steadily
reaches algorithm convergence for both applications. In contrast,
\texttt{Gaia\textunderscore Async} diverges from the optimum point at
\textasciitilde 100 seconds for \emph{MF}. \khi{For} \emph{TM},
\texttt{Gaia\textunderscore Async} \khi{looks like it makes} faster progress at
the beginning \khi{of execution because it eliminates} the
synchronization overhead. However, it makes very slow progress
\khi{after} \textasciitilde 200 seconds and \khi{does not reach the
  value that results in convergence until 1100 seconds}. It may take a
long time for \texttt{Gaia\textunderscore Async} to reach that point,
\khi{if ever}. \khi{Thus, the lack of synchronization} leads to worse
model quality than \khi{that achieved by using} proper
\synchronization mechanisms.  Both results demonstrate that the
\synchronization mechanisms we introduce in \protoabbrv \khi{are}
effective and vital for deploying ML algorithms on {\gaia} on WANs.

\subsection{Performance Results of SSP}
\label{subsec:ssp}

We present the performance results of SSP
  for \emph{MF (Matrix Factorization)} and \emph{TM (Topic Modeling)}
  here. We do not present the results of SSP for \emph{IC (Image
  Classification)} because SSP has worse performance than BSP
  for \emph{IC}~\cite{DBLP:conf/eurosys/CuiZGGX16}. In our evaluation,
  BSP and SSP are used among all worker machines for \baselinesys and
\lansys, whereas for {\gaiasys}, they are used only within each data
center. To show the performance difference between BSP and SSP, we
show both results together.

\subsubsection{SSP Performance on EC2 Deployment}

Similar to Section~\ref{subsec:performance_ec2},
Figures~\ref{fig:performance_mf_11_regions_ssp}
and~\ref{fig:performance_tm_11_regions_ssp} show the execution time
\khi{until} convergence for \emph{MF} and \emph{TM}, normalized to
\baselinesys with BSP on EC2. The data label above each bar shows the
speedup over \baselinesys for \emph{the respective deployment
  and \synchronization model}.

\begin{figure}[h]
  \centering
  \includegraphics[width=0.8\textwidth]{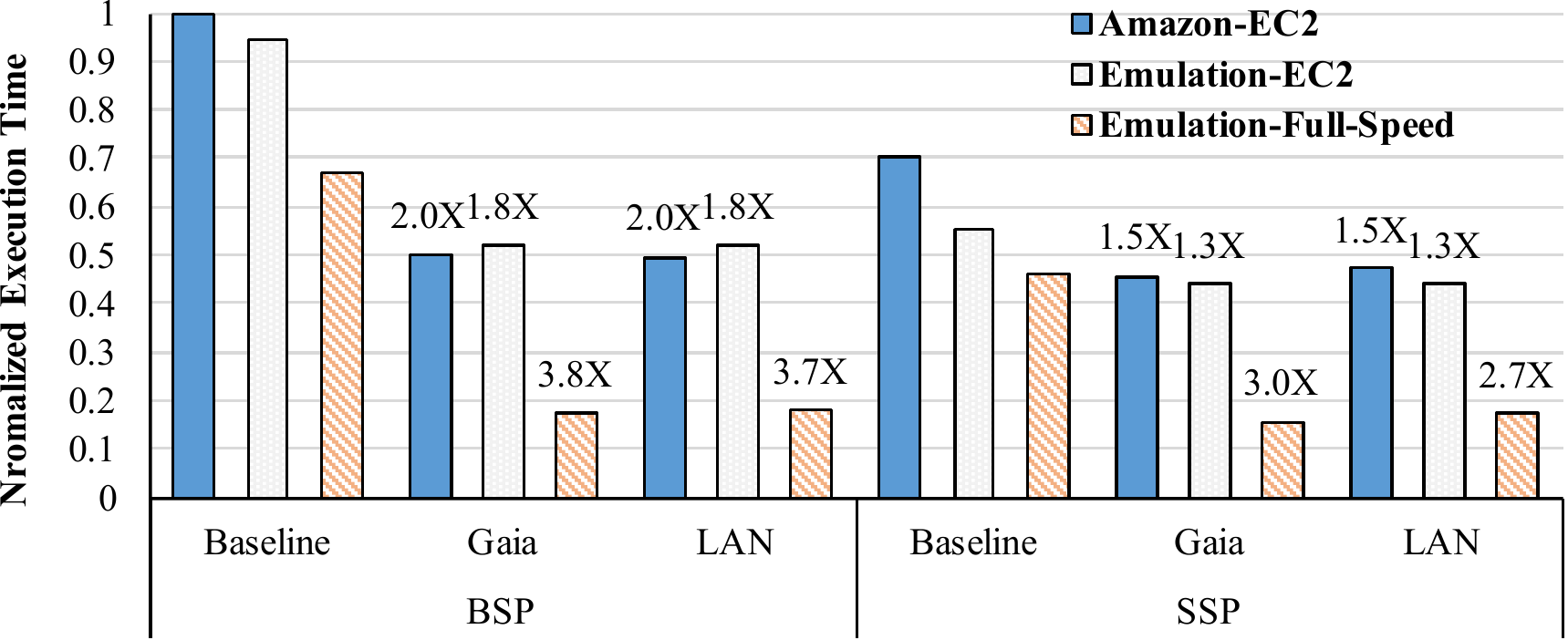}
  \caption{Normalized execution time of \emph{MF} until convergence
    when deployed across 11 EC2 regions}
  \label{fig:performance_mf_11_regions_ssp}
\end{figure}

\begin{figure}[h]
  \centering
  \includegraphics[width=0.8\textwidth]{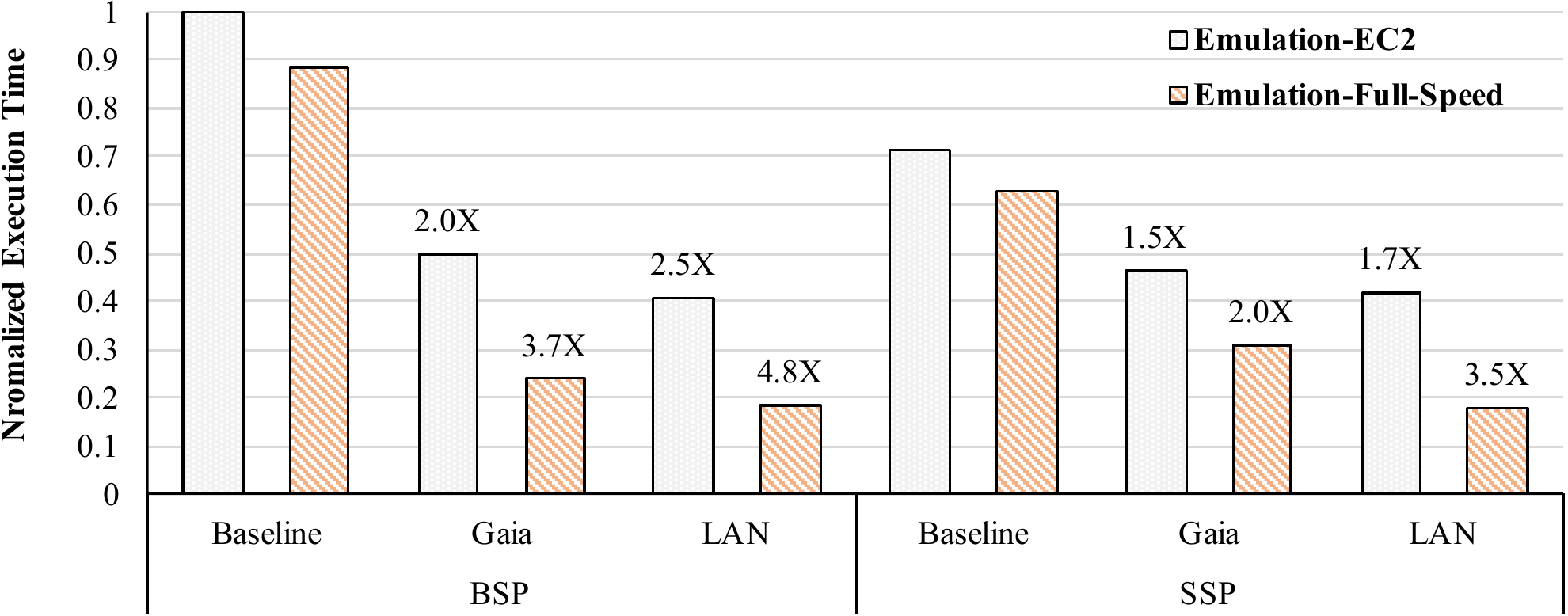}
  \caption{Normalized execution time of \emph{TM} until convergence
    when deployed across 11 EC2 regions}
  \label{fig:performance_tm_11_regions_ssp}
\end{figure}

We see that {\gaiasys} significantly improves the performance of
\baselinesys with SSP. For \emph{MF}, {\gaiasys} provides a speedup of
1.3--3.0$\times$ over \baselinesys with SSP, and successfully
approaches the speedups of \lansys with SSP. For \emph{TM}, {\gaiasys}
achieves speedups of 1.5--2.0$\times$ over \baselinesys. Note that for
\emph{TM}, {\gaiasys} with BSP outperforms {\gaiasys} with SSP. The
reason is that \khii{SSP allows using stale, and thus inaccurate, values in
order} to get the benefit of more efficient communication. However,
compared to \baselinesys, the benefit of employing SSP to reduce
communication overhead is much smaller for {\gaiasys} because it uses
SSP only to synchronize a small number of machines within a data
center. Thus, the cost of inaccuracy outweighs the benefit of SSP in
this case. Fortunately, {\gaiasys} decouples the synchronization model
within a data center from the synchronization model across different
data centers. Thus, we can freely choose the combination
of \synchronization models that works better
for {\gaiasys}.

\subsubsection{SSP Performance and WAN Bandwidth}

Similar to Section~\ref{subsec:performance_wan_bandwidth},
Figures~\ref{fig:performance_vc_wan_ssp}
and~\ref{fig:performance_ss_wan_ssp} show the normalized execution
time until convergence on two deployments: V/C WAN and S/S WAN. The
data label above each bar shows the speedup over \baselinesys for \emph{the
  respective deployment and \synchronization model}.

\begin{figure}[h]
 \centering
 \begin{subfigure}[h]{0.48\linewidth}
 \centering
 \includegraphics[width=1.0\textwidth]{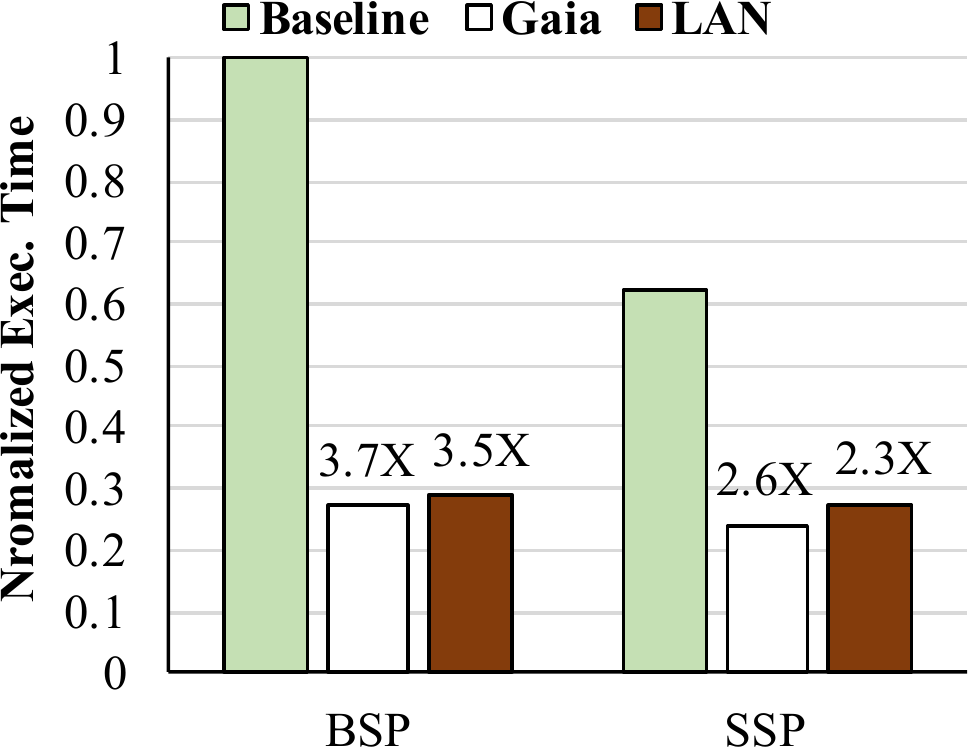}
 \caption{\emph{Matrix Factorization (MF)}} 
 \end{subfigure}
 \begin{subfigure}[h]{0.48\linewidth}
 \centering
 \includegraphics[width=1.0\textwidth]{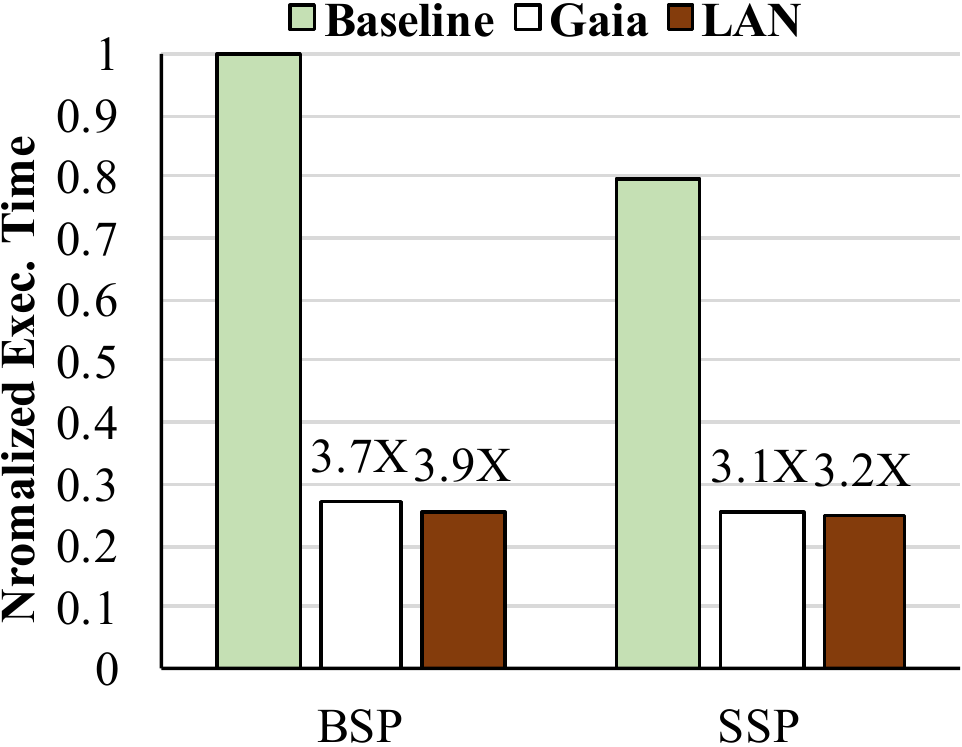}
 \caption{\emph{Topic Modeling (TM)}}
 \end{subfigure}
 \caption{Normalized execution time \khi{until} convergence with the
   WAN bandwidth between Virginia and California}
 \label{fig:performance_vc_wan_ssp}
\end{figure}

\vspace{5pt}
\begin{figure}[h]
 \centering
 \begin{subfigure}[h]{0.48\linewidth}
 \centering
 \includegraphics[width=1.0\textwidth]{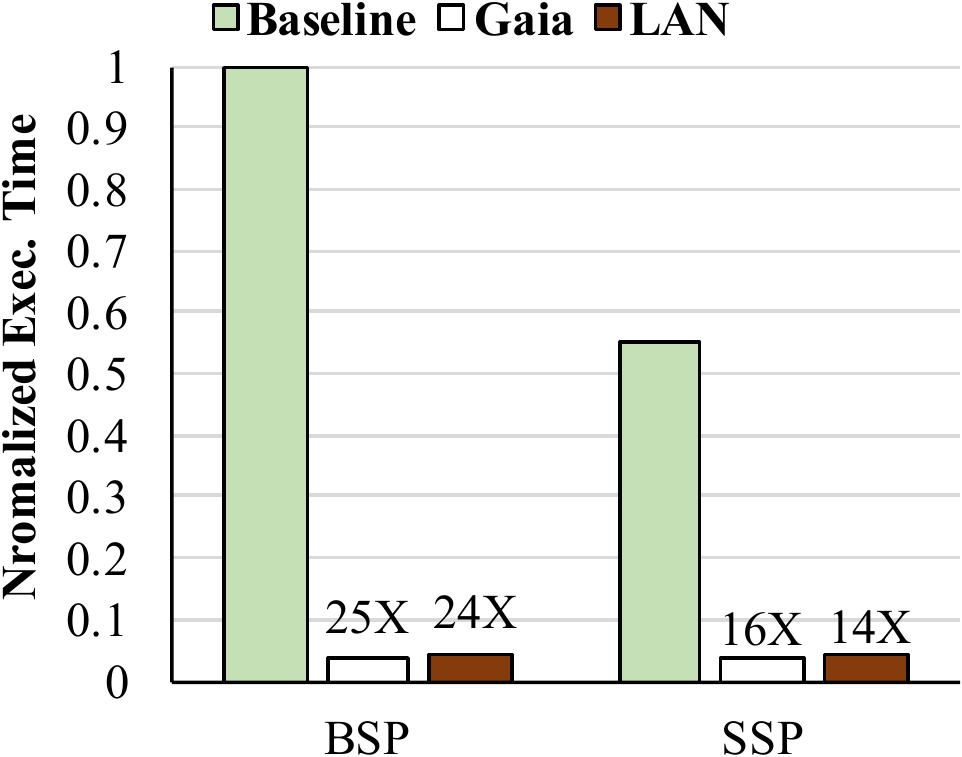}
 \caption{\emph{Matrix Factorization (MF)}} 
 \end{subfigure}
 \begin{subfigure}[h]{0.48\linewidth}
 \centering
 \includegraphics[width=1.0\textwidth]{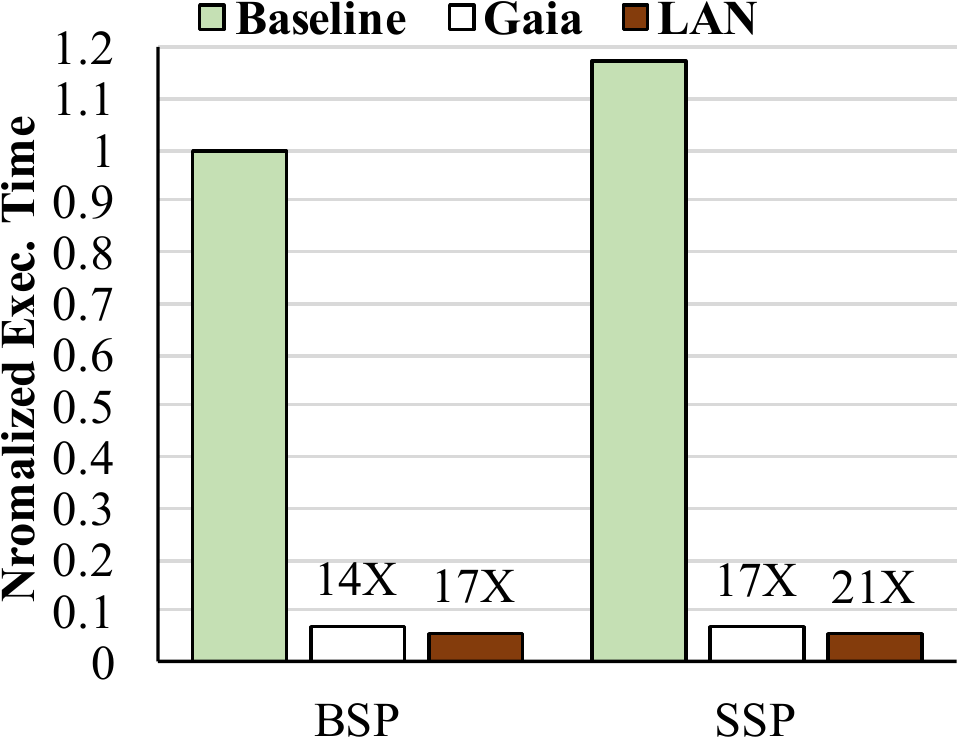}
 \caption{\emph{Topic Modeling (TM)}}
 \end{subfigure}
 \caption{Normalized execution time \khi{until} convergence with the
   WAN bandwidth between Singapore and S\~{a}o Paulo}
 \label{fig:performance_ss_wan_ssp}
\end{figure}
\vspace{-2pt}

We find that {\gaiasys} performs very well compared to \baselinesys
with SSP in both high WAN bandwidth (V/C WAN) and low WAN bandwidth
(S/S WAN) settings. For V/C WAN, {\gaiasys} achieves a speedup of
2.6$\times$ for \emph{MF} and 3.1$\times$ for \emph{TM} over
\baselinesys with SSP. For both applications, the performance of
{\gaiasys} is almost the same as the performance of \lansys. For S/S WAN,
{\gaiasys} provides a speedup of 15.7$\times$ / 16.8$\times$ for
\emph{MF} / \emph{TM} over \baselinesys with SSP, and successfully
approaches the \lansys speedups. We conclude that {\gaiasys} provides
significant performance improvement over \baselinesys,
irrespective of the \synchronization model used by \baselinesys.

\section{Summary}

In this chapter, we introduce {\gaia}, a new ML system that is
designed \ignore{to be deployed over WANs }to efficiently run ML
algorithms on globally-generated data over WANs, without any need to
change the ML algorithms. {\gaia} decouples the synchronization within
a data center (LANs) from the synchronization across different data
centers (WANs), enabling flexible and effective synchronization over
LANs and WANs. We introduce a new synchronization model, \protocol
(\protoabbrv), to efficiently utilize the scarce and heterogeneous WAN
bandwidth while ensuring \khi{convergence} of the ML algorithms with a
theoretical guarantee. Using \protoabbrv, {\gaia} dynamically
eliminates insignificant, and thus unnecessary, communication over
WANs. Our evaluation \ignore{with an 11-region Amazon EC2 deployment
  and a local cluster with emulated WAN bandwidth across EC2 regions
}shows that {\gaia} significantly outperforms two state-of-the-art
parameter server systems on WANs, and is within 0.94--1.40$\times$ of
\khi{the speed of running the same ML algorithm on a LAN}. {\gaia}
also significantly reduces the monetary cost of running the same ML
algorithm on WANs, by 2.6--59.0$\times$. We conclude that {\gaia} is a
practical and effective system to enable globally-distributed ML
applications, and \khi{we believe} the ideas behind {\gaia}'s system
design for communication across WANs can be applied to many other
large-scale distributed ML systems.

%\section{Appendix}
%\label{sec:appendix}

%\appendix
%\pagebreak
%\section*{Appendix}

\begin{subappendices}

\section{Convergence Proof of SGD under ASP}
\label{appendix:proof}

Stochastic Gradient Descent is a very popular algorithm, widely used
for finding the minimizer/maximizer of a criterion (sum of
differentiable functions) via iterative steps. The intuition behind the
algorithm is that we randomly select an initial point $\bm{x}_0$ and
keep moving toward the negative direction of the gradient, producing a
sequence of points $\bm{x}_i, i=1,...n$ until we detect that moving
further decreases (increases) the
minimization (maximization, respectively) criterion only negligibly.

Formally, step $t$ of the SGD algorithm is defined as:

\vspace{-10pt}
\begin{small}
\begin{equation}
  \bm{x}_t = \bm{x}_{t-1} - \eta_t \nabla f_t(\bm{x}_t) = \bm{x}_{t-1} - \eta_t \bm{g}_t = \bm{x}_{t-1} + \bm{u}_t 
\end{equation}
\end{small}
\vspace{-10pt}

\noindent
where $\eta_t$ is the step size at step $t$, $\nabla
f_t(\bm{x}_t)$ or $\bm{g}_t$ is the gradient at step $t$, and
$\bm{u}_t = \eta_t \bm{g}_t $ is the update of step $t$.

Let us define an order of the updates up to step $t$. Suppose that the
  algorithm is distributed in $D$ data centers, each of which uses $P$
  machines, and the logical clocks that mark progress start at 0.
  Then,

\vspace{-5pt}
\begin{small}
\begin{equation}
  \bm{u}_t= \bm{u}_{d,p,c} = \bm{u}_{\lfloor \frac{t}{P} \rfloor \bmod D, \ t \bmod P, \ \lfloor \frac{t}{DP} \rfloor}
\end{equation}
\end{small}
\vspace{-5pt}

\noindent
represents a mapping that loops through clocks
($c=\lfloor \frac{t}{DP} \rfloor$) and for each clock loops through
data centers ($d=\lfloor \frac{t}{P} \rfloor \bmod D$)
and for each data center loops through its workers ($p=t \bmod P$).

We now define a reference sequence of states that a single machine serial execution would
go through if the updates were observed under the above ordering:

\vspace{-5pt}
\begin{small}
\begin{equation}
  \bm{x}_t= \bm{x}_0 + \sum_{t'=1}^{t}\bm{u}_{t'}
\end{equation}
\end{small}

Let $\ds$ denote the threshold of mirror clock difference between
different data centers. At clock $c$, let $A_{d,c}$ denote the
$(c-\ds)$-width window of updates at data center $d$:
$A_{d,c}=[0,P-1] \times [0, c-\ds{}-1]$. Also, let $K_{d,c}$ denote the
subset of $A_{d,c}$ of significant updates (i.e., those
broadcast to other data centers) and $L_{d,c}$ denote the subset of
$A_{d,c}$ of the insignificant updates (not broadcast) from this data
center. Clearly, $K_{d,c}$ and $L_{d,c}$ are disjoint and their union
equals $A_{d,c}$.
 
Let $s$ denote a user-chosen staleness threshold for SSP. Let
  $\tilde{\bm{x}_t}$ denote the sequence of noisy (i.e.,
  inaccurate) \emph{views} of the parameters $\bm{x}_t$. Let $B_{d,c}$
  denote the $2s$-width window of updates at data center $d$:
  $B_{d,c}=[0,P-1] \times [c-s, c+s-1]$. A worker $p$ in data center $d$
  will definitely see its own updates and may or may not see updates
  from other workers that belong to this window. Then, $M_{d,c}$
  denotes the set of updates that are not seen in $\tilde{\bm{x_t}}$
  and are seen in $\bm{x}_t$, whereas $N_{d,c}$ denotes the updates
  that are seen in $\tilde{\bm{x_t}}$ and not seen in $\bm{x}_t$. The
  sets $M_{d,c}$ and $N_{d,c}$ are disjoint and their union equals the
  set $B_{d,c}$.

We define the noisy view $\tilde{\bm{x_t}}$ using the above mapping:
%(i.e., it corresponds to data center $d
%= \Big[ \lfloor \frac{t}{P} \rfloor \bmod D \Big]$, worker $p = \Big[t \bmod
%P\Big]$ and clock $c = \lfloor \frac{t}{DP} \rfloor)$.

\begin{small}
\begin{align}
  \bm{\tilde{x}}_{d,p,c}
                 &= \sum_{p'=0}^{P-1}\sum_{c'=0}^{c-s-1}\bm{u}_{d,p',c'}
                 +\sum_{c'=c-s}^{c-1}\bm{u}_{d, p, c'} \nonumber \\
                 &+\sum_{(p',c') \in B'_{d,c} \subset
                 B_{d,c}}\bm{u}_{d,p',c'} +\sum_{d'\neq
                 d} \Big[\sum_{(p',c') \in
                 K_{d',c'}}\bm{u}_{d',p',c'} \Big]
\end{align}
\end{small}

The difference between the reference view $\bm{x}_t$ and the noisy
view $\bm{\tilde{x}_t}$ becomes:

\vspace{-15pt}
\begin{small}
\begin{align}
\label{eq:dx}
   \tilde{\bm{x}_t}-\bm{x}_t &= \tilde{\bm{x}}_{d,p,c}-\bm{x}_t = \tilde{\bm{x}}_{\lfloor \frac{t}{P} \rfloor \bmod D, t \bmod P, \lfloor \frac{t}{DP} \rfloor} - \bm{x}_t = \nonumber \\
                             &-\sum_{i \in M_{d, c}} \bm{u}_i +\sum_{i \in N_{d, c}} \bm{u}_i
                             - \sum_{d' \neq d} \sum_{i \in L_{d',
                             c}} \bm{u}_i \nonumber \\ &+\sum_{d' \neq d} \Big[-\sum_{i \in M_{d', c}} \bm{u}_i +\sum_{i \in N_{d', c}} \bm{u}_i \Big] 
\end{align}
\end{small}

Finally, let $D(\bm{x}, \bm{x'})$ denote the distance between points
$\bm{x}, \bm{x'} \in \mathbb{R}^{n}$:
\vspace{-3pt}
\begin{small}
\begin{equation}
  D(\bm{x}, \bm{x'}) = \frac{1}{2} {\lVert \bm{x} - \bm{x'} \rVert}^{2}.
\end{equation}
\end{small}

We now prove the following lemma:

\vspace{-3pt}
\begin{lemma}
For any $\bm{x}^{*}, \tilde{\bm{x}_t} \in \mathbb{R}^{n}$,
\begin{small}
\begin{align}  
    \langle \tilde{\bm{x}_t}-\bm{x}^{*}, \tilde{\bm{g}_t} \rangle 
    &= \frac{1}{2}\eta_t {\lVert \tilde{\bm{g}_t} \rVert}^{2} + \frac{D(\bm{x}^{*}, \bm{x}_t) - D(\bm{x}^{*}, \bm{x}_{t+1})}{\eta_t} \nonumber \\
    &+ \Big[-\sum_{i \in M_{d,c}} \eta_i \langle \tilde{\bm{g}_i}, \tilde{\bm{g}_t} \rangle
             +\sum_{i \in N_{d,c}} \eta_i \langle \tilde{\bm{g}_i}, \tilde{\bm{g}_t} \rangle \Big] \nonumber \\
    &+ \sum_{d' \neq d} \Big[-\sum_{i \in L_{d',c}} \eta_i \langle \tilde{\bm{g}_i}, \tilde{\bm{g}_t} \rangle \Big] \nonumber \\
    &+ \sum_{d' \neq d} \Big[-\sum_{i \in M_{d',c}} \eta_i \langle \tilde{\bm{g}_i}, \tilde{\bm{g}_t} \rangle
                             +\sum_{i \in
    N_{d',c}} \eta_i \langle \tilde{\bm{g}_i}, \tilde{\bm{g}_t} \rangle \Big]
\end{align}
\end{small}

\end{lemma}

\vspace{-15pt}
\begin{proof}

\begin{small}
\begin{align}
  D(& \bm{x}^{*}, \bm{x}_{t+1}) - D(\bm{x}^{*}, \bm{x}_t)
     =  \frac{1}{2} {\lVert \bm{x}^{*} - \bm{x}_{t+1} \rVert}^{2} - \frac{1}{2} {\lVert \bm{x}^{*} - \bm{x}_t \rVert}^{2} \nonumber \\
    &=  \frac{1}{2} {\lVert \bm{x}^{*} - \bm{x}_t + \bm{x}_t - \bm{x}_{t+1} \rVert}^{2} - \frac{1}{2} {\lVert \bm{x}^{*} - \bm{x}_t \rVert}^{2} \nonumber \\
    &=  \frac{1}{2} {\lVert \bm{x}^{*} - \bm{x}_t + \eta_t \tilde{\bm{g}_t} \rVert}^{2} - \frac{1}{2} {\lVert \bm{x}^{*} - \bm{x}_t \rVert}^{2} \nonumber \\
    &=  \frac{1}{2} \langle \bm{x}^{*} - \bm{x}_t
     + \eta_t \tilde{\bm{g}_t}, \bm{x}^{*} - \bm{x}_t
     + \eta_t \tilde{\bm{g}_t} \rangle -\frac{1}{2} \langle \bm{x}^{*} - \bm{x}_t, \bm{x}^{*} - \bm{x}_t \rangle \nonumber \\
    &=  \frac{1}{2} \langle \bm{x}^{*} - \bm{x}_t, \bm{x}^{*} - \bm{x}_t \rangle 
       +\frac{1}{2} \langle \eta_t \tilde{\bm{g}_t}, \eta_t \tilde{\bm{g}_t} \rangle
       +\langle \bm{x}^{*} - \bm{x}_t, \eta_t \tilde{\bm{g}_t} \rangle \nonumber \\
    & \quad -\frac{1}{2} \langle \bm{x}^{*} - \bm{x}_t, \bm{x}^{*} - \bm{x}_t \rangle \nonumber \\
    &= \frac{1}{2} {\eta_t}^{2} {\lVert \tilde{\bm{g}_t} \rVert}^{2} +\eta_t \langle \bm{x}^{*}-\bm{x}_t, \tilde{\bm{g}_t} \rangle \nonumber \\
    &= \frac{1}{2} {\eta_t}^{2} {\lVert \tilde{\bm{g}_t} \rVert}^{2} -\eta_t \langle \bm{x}_t-\bm{x}^{*}, \tilde{\bm{g}_t} \rangle \nonumber \\
    &= \frac{1}{2} {\eta_t}^{2} {\lVert \tilde{\bm{g}_t} \rVert}^{2} -\eta_t \langle \bm{x}_t-\tilde{\bm{x}_t}+\tilde{\bm{x}_t}-\bm{x}^{*} \rangle \nonumber \\
    &= \frac{1}{2} {\eta_t}^{2} {\lVert \tilde{\bm{g}_t} \rVert}^{2} -\eta_t \langle \bm{x}_t -\tilde{\bm{x}_t}, \tilde{\bm{g}_t} \rangle
       -\eta_t \langle \tilde{\bm{x}_t}-\bm{x}^{*}, \tilde{\bm{g}_t} \rangle \implies \nonumber
\end{align}

\begin{align}
\label{eq:lemma2}
    \langle \tilde{\bm{x}_t}-\bm{x}^{*}, \tilde{\bm{g}_t} \rangle
    &= \frac{1}{2}\eta_t {\lVert \tilde{\bm{g}_t} \rVert}^{2}
       +\frac{D(\bm{x}^{*}, \bm{x}_t) - D(\bm{x}^{*}, \bm{x}_{t+1})}{\eta_t} \nonumber \\
    & -\langle \bm{x}_t -\tilde{\bm{x}_t}, \tilde{\bm{g}_t} \rangle
\end{align}
\end{small}

Substituting the RHS of Equation~\ref{eq:dx} into
Equation~\ref{eq:lemma2} completes the proof.
\end{proof}

\begin{theorem}\label{thm:asp_sgd}
\textbf{(Convergence of SGD under ASP)}.

Suppose
  that, in order to compute the minimizer $x^{*}$ of a convex function
  $f(\bm{x}) = \sum_{t=1}^{T}f_t(\bm{x})$, with $f_t,t=1,2,\ldots,T$,
  convex, we use stochastic gradient descent on one component
  $\nabla f_t$ at a time.  Suppose also that 1) the algorithm is
  distributed in $D$ data centers, each of which uses $P$ machines, 2)
  within each data center, the SSP protocol is used, with a fixed
  staleness of $s$, and 3) a fixed mirror clock difference $\ds{}$ is
  allowed between any two data centers.  Let $\bm{u}_t = -\eta_t \nabla
  f_t{(\tilde{\bm{x}_t})}$, where the step size $\eta_t$ decreases as
  $\eta_t = \frac{\eta}{\sqrt{t}}$ and the significance threshold
  $v_t$ decreases as $v_t = \frac{v}{\sqrt{t}}$. If we further assume
  that: $\lVert \nabla f_t(\bm{x}) \rVert \leq L,$ $\forall \bm{x}
  \in dom(f_t)$ and $\max(D(\bm{x},\bm{x}')) \leq \Delta^{2}, \forall
  \bm{x}, \bm{x}' \in dom(f_t)$. Then, as $T \rightarrow \infty$,

\ignore{Suppose that in order to compute
the minimizer $x^{*}$ of a convex function
$f(\bm{x}) = \sum_{t=1}^{T}f_t(\bm{x})$, with ($f_t, t=1,2 \dots T$ convex),
we use Stochastic Gradient Descent on one component $\nabla f_t$ at a time.
Suppose also that the algorithm is distributed in $D$ data centers, each of
which uses $P$ machines and within each data center the SSP protocol is used,
with a fixed staleness of $s$ and a fixed mirror clock difference $\ds{}$ is
allowed between any two data centers.
Let $\bm{u}_t = -\eta_t \nabla f_t{(\tilde{\bm{x}_t})}$, where the step size
$\eta_t$ decrases as $\eta_t = \frac{\eta}{\sqrt{t}}$ and the threshold
$v_t$ decreases as $v_t = \frac{v}{\sqrt{t}}$. If we further assume that:
$\lVert \nabla f_t(\bm{x}) \rVert \leq L$ $,
\forall \bm{x} \in dom(f_t)$, $\max(D(\bm{x},\bm{x}')) \leq \Delta^{2},
\forall \bm{x}, \bm{x}' \in dom(f_t)$. Then, as $T \rightarrow \infty$,}

\vspace{-10pt}
\begin{small}
\begin{align*}
  R[X] = \sum_{t=1}^{T}f_t(\tilde{\bm{x}_t})-f(\bm{x}^{*})=O(\sqrt{T}) 
\end{align*}
\end{small}

and therefore 

\vspace{-10pt}
\begin{small}
\begin{align*}
  \lim_{T\to\infty} \frac{R[X]}{T} \rightarrow 0 
\end{align*}
\end{small}

\end{theorem}

\begin{proof}
\begin{small}
\begin{align}
  R[X] &=    \sum_{t=1}^{T}f_t(\tilde{\bm{x}_t}) - f_t(\bm{x}^{*}) \nonumber \\
       &\leq \sum_{t=1}^{T} \langle \nabla f_t(\tilde{\bm{x}_t}), \tilde{\bm{x}_t} - \bm{x}^{*} \rangle \quad \text{(convexity of $f_t$)} \nonumber \\
       &=    \sum_{t=1}^{T} \langle \tilde{\bm{g}_t}, \tilde{\bm{x}_t}-\bm{x}^{*} \rangle \nonumber \\
       &=    \sum_{t=1}^{T} \Bigg[\frac{1}{2}\eta_t {\lVert \tilde{\bm{g}_t} \rVert}^{2} + \frac{D(\bm{x}^{*}, \bm{x}_t) - D(\bm{x}^{*}, \bm{x}_{t+1})}{\eta_t} \nonumber \\
       & \quad  +\sum_{d' \neq d} \Big[-\sum_{i \in L_{d',c}} \eta_i \langle \tilde{\bm{g}_i}, \tilde{\bm{g}_t} \rangle \Big]
             \nonumber \\ & \quad +\Big[-\sum_{i \in M_{d,c}} \eta_i \langle \tilde{\bm{g}_i}, \tilde{\bm{g}_t} \rangle
                   +\sum_{i \in N_{d,c}} \eta_i \langle \tilde{\bm{g}_i}, \tilde{\bm{g}_t} \rangle \Big] \nonumber \\
       & \quad  +\sum_{d' \neq d} \Big[-\sum_{i \in M_{d',c}} \eta_i \langle \tilde{\bm{g}_i}, \tilde{\bm{g}_t} \rangle
                   +\sum_{i \in N_{d',c}} \eta_i \langle \tilde{\bm{g}_i}, \tilde{\bm{g}_t} \rangle \Big] \Bigg]
\end{align}
\end{small}

We first bound the upper limit of the term:
$\sum\limits_{t=1}^{T} \frac{1}{2} \eta_t
{\lVert \tilde{\bm{g}_t} \rVert}^{2}$:

\begin{small}
\begin{align}
   \sum_{t=1}^{T} \frac{1}{2} \eta_t {\lVert \tilde{\bm{g}_t} \rVert}^{2} 
   &\leq \sum_{t=1}^{T} \frac{1}{2} \eta_t L^{2}  \hspace{20mm} (\lVert \nabla f_t(\bm{x}) \rVert \leq L) \nonumber \\ 
   &=    \sum_{t=1}^{T} \frac{1}{2} \frac{\eta}{\sqrt{t}} L^{2} \nonumber \\
   &=    \frac{1}{2} \eta L^{2} \sum_{t=1}^{T} \frac{1}{\sqrt{t}} \hspace{15mm} (\sum_{t=1}^{T} \frac{1}{\sqrt{t}} \leq 2\sqrt{T}) \nonumber \\
   &\leq \frac{1}{2} \eta L^{2} 2\sqrt{T} = \eta L^{2} \sqrt{T} 
\end{align}
\end{small}

Second, we bound the upper limit of the term: $\sum\limits_{t=1}^{T} \frac{D(x^{*}, x_t) - D(x^{*}, x_{t+1})}{\eta_t}$:

\begin{small}
\begin{align}
   \sum_{t=1}^{T} & \frac{D(\bm{x}^{*}, \bm{x}_t) - D(\bm{x}^{*}, \bm{x}_{t+1})}{\eta_t} \nonumber \\
   &=   \frac{D(\bm{x}^{*}, \bm{x}_1)}{\eta_1}
   - \frac{D(\bm{x}^{*}, \bm{x}_{T+1})}{\eta_T} 
   +\sum_{t=2}^{T} D(\bm{x}^{*}, \bm{x}_t)(\frac{1}{\eta_t}-\frac{1}{\eta_{t-1}}) \nonumber \\
   &\leq \frac{\Delta^{2}}{\eta} - 0
   + \frac{\Delta^{2}}{\eta} \sum_{t=2}^{T} [\sqrt{t}
   - \sqrt{t-1}] \nonumber \quad (\max(D(\bm{x},\bm{x}')) \leq \Delta^{2}) \nonumber \\
   &=   \frac{\Delta^{2}}{\eta} + \frac{\Delta^{2}}{\eta}[\sqrt{T}-1] \nonumber \\
   &=   \frac{\Delta^{2}}{\eta}\sqrt{T} 
\end{align}
\end{small}

Third, we bound the upper limit of the term: 
$\sum\limits_{t=1}^{T} \sum\limits_{d' \neq d}
  \Big[
    -\sum\limits_{i \in L_{d',c}} \eta_i \langle \tilde{\bm{g}_i}, \tilde{\bm{g}_t} \rangle 
  \Big]
$:

\begin{small}
\begin{align}
  \sum_{t=1}^{T}  & \sum_{d' \neq d} \Big[-\sum_{i \in L_{d',c}} \eta_i \langle \tilde{\bm{g}_i}, \tilde{\bm{g}_t} \rangle \Big] \nonumber \\
  &\leq \sum_{t=1}^{T} (D-1) \Big[-\sum_{i \in L_{d',c}} \eta_i \langle \tilde{\bm{g}_i}, \tilde{\bm{g}_t} \rangle \Big] \leq \sum_{t=1}^{T} (D-1) v_t \nonumber \\
  &=    (D-1)\sum_{t=1}^{T} \frac{v}{\sqrt{t-(s+\ds{}+1)P}} \nonumber \\
  &\leq (D-1)v\sum_{t=(s+\ds{}+1)P+1}^{T} \frac{1}{\sqrt{T-(s+\ds{}+1)P}} \nonumber \\
  &\leq 2(D-1)v \sqrt{T-(s+\ds{}+1)P} \nonumber \\ 
  &\leq 2(D-1)v \sqrt{T} \nonumber \\
  &\leq 2Dv \sqrt{T} 
\end{align}
\end{small}
\vspace{-10pt}

\noindent
where the fourth inequality follows from the fact that: \\ $\sum\limits_{t=(s+\ds{}+1)P+1}^{T} \frac{1}{\sqrt{T-(s+\ds{}+1)P}} \leq \sqrt{T-(s+\ds{}+1)P}$.

Fourth, we bound the upper limit of the term: 
$\sum\limits_{t=1}^{T}
  \Big[
    -\sum\limits_{i \in M_{d,c}} \eta_i \langle \tilde{\bm{g}_i}, \tilde{\bm{g}_t} \rangle
    +\sum\limits_{i \in N_{d,c}} \eta_i \langle \tilde{\bm{g}_i}, \tilde{\bm{g}_t} \rangle 
  \Big]
$:

\begin{small}
\begin{align}
  \sum_{t=1}^{T} & \Big[-\sum_{i \in M_{d,c}} \eta_i \langle \tilde{\bm{g}_i}, \tilde{\bm{g}_t} \rangle
                        +\sum_{i \in N_{d,c}} \eta_i \langle \tilde{\bm{g}_i}, \tilde{\bm{g}_t} \rangle \Big] \nonumber \\
  &\leq  \sum_{t=1}^{T} [\lvert M_{d,c} \rvert + \lvert N_{d,c} \mid] \eta_{max(1, t-(s+1)P)}L^2 \nonumber \\
  &=     L^{2} \Big[\sum_{t=1}^{(s+1)P} [\lvert M_{d,c} \rvert
                    + \lvert N_{d,c} \rvert] \eta \nonumber \\
                    &  \quad +\sum_{t=(s+1)P+1}^{T} [\lvert M_{d,c} \rvert + \lvert N_{d,c} \rvert] \eta_{t-(s+1)P} \Big] \nonumber\\
  &=     L^{2} \Big[\sum_{t=1}^{(s+1)P} [\lvert M_{d,c} \rvert + \lvert N_{d,c} \rvert] \eta \nonumber \\
  & \quad  +\sum_{t=(s+1)P+1}^{T} [\lvert M_{d,c} \rvert + \lvert N_{d,c} \rvert] \frac{\eta}{\sqrt{t-(s+1)P}} \Big] \nonumber \\ 
  &\leq  \eta L^{2} \Big[\sum_{t=1}^{(s+1)P} 2s(P-1) \nonumber \\
  & \quad + \sum_{t=(s+1)P+1}^{T} 2s(P-1) \frac{1}{\sqrt{t-(s+1)P}} \Big]  \nonumber \\   
  &=     2 \eta L^{2}s(P-1)\Big[(s+1)P + \sum_{t=(s+1)P+1}^{T} \frac{1}{\sqrt{T-(s+1)P}}\Big] \nonumber \\
  &\leq  2 \eta L^{2}s(P-1)\Big[(s+1)P + 2 \sqrt{T-(s+1)P}\Big] \nonumber \\
  &\leq  2 \eta L^{2}s(P-1)[(s+1)P + 2 \sqrt{T}] \nonumber \\
  &=     2 \eta L^{2}s(s+1)(P-1)P + 4 \eta L^{2}s(P-1) \sqrt{T} \nonumber\\
  &\leq  2 \eta L^{2}(s+1)(s+1)(P-1)P + 4 \eta L^{2}(s+1)(P-1) \sqrt{T} \nonumber \\
  &=     2 \eta L^{2}(s+1)^{2}(P-1)P + 4 \eta L^{2}(s+1)(P-1) \sqrt{T} \nonumber \\
  &\leq  2 \eta L^{2}(s+1)^{2}PP + 4 \eta L^{2}(s+1)P \sqrt{T} \nonumber \\
  &=     2 \eta L^{2}[(s+1)P]^{2} + 4 \eta L^{2}(s+1)P \sqrt{T}
\end{align}
\vspace{-15pt}
\end{small}

\noindent
where the first inequality follows from the fact that $\eta_{max(1, t-(s+1)P)} \geq \eta_t, t \in M_{d,t} \cup N_{d,t}$,
the second inequality follows from the fact that $\lvert M_{d,t} \rvert + \lvert N_{d,t} \rvert \leq 2s(P-1)$,
and the third inequality follows from the fact that
$\sum\limits_{t=(s+1)P+1}^{T} \Big[\frac{1}{\sqrt{T-(s+1)P}}\Big] \leq
2\sqrt{T-(s+1)P}$.

%\vspace{10pt}

Similarly, $\forall d' \in D'=D \setminus \left\{d \right\}$, we can prove that: 

\vspace{-15pt}
\begin{small}
\begin{align*}
 \sum_{t=1}^{T} & \Big[-\sum\limits_{i \in M_{d',c}} \eta_i \langle \tilde{\bm{g}_i}, \tilde{\bm{g}_t} \rangle
                       +\sum\limits_{i \in N_{d',c}} \eta_i \langle \tilde{\bm{g}_i}, \tilde{\bm{g}_t} \rangle \Big] \leq \nonumber \\
                & 2 \eta L^{2}[(s+\ds{}+1)P]^{2} + 4 \eta L^{2}(s+\ds{}+1)P \sqrt{T} \nonumber
\end{align*}
\vspace{-10pt}
\end{small}

\noindent
which implies:

\vspace{-15pt}
\begin{small}
\begin{align*}
  \sum_{t=1}^{T} & \sum\limits_{d' \neq d} \Big[ -\sum\limits_{i \in M_{d',c}} \bm{u}_i 
                                               +\sum_{i \in N_{d', c}} \bm{u}_i \Big] \leq \nonumber \\
                 & D\Big[2 \eta L^{2}[(s+\ds{}+1)P]^{2} + 4 \eta L^{2}(s+\ds{}+1)P \sqrt{T} \Big] \nonumber
\end{align*}
\end{small}
\vspace{-10pt}

By combining all the above upper bounds, we have:

\vspace{-10pt}
\begin{small}
\begin{align}
  R[X] &\leq \eta L^{2} \sqrt{T} + \frac{\Delta^{2}}{\eta}\sqrt{T} +2Dv \sqrt{T} + 2 \eta L^{2}[(s+1)P]^{2} \nonumber \\
       &+ 4 \eta L^{2}(s+1)P \sqrt{T} \nonumber \\
       & +D\Big[2 \eta L^{2}[(s+\ds{}+1)P]^{2} + 4 \eta L^{2}(s+\ds{}+1)P \sqrt{T} \Big] \nonumber \\
       &= O(\sqrt{T}) 
\end{align}
\end{small}

\noindent
and thus $\lim\limits_{T\to\infty} \frac{R[X]}{T} \rightarrow 0$.
\end{proof}
\vspace{-5pt}

%\section{EC2 Cost Model Details}
%\label{appendix:cost_model}

\end{subappendices}

%\chapter{Understanding The Non-IID Data Partition Problem for Decentralized ML}
\chapter{The Non-IID Data Partition Problem for Decentralized ML}
\label{ch:noniid}

%\comm{Kevin}{Status: Done: Section 1--7, 9; Working on Appendix}

%\comm{Phil}{Status: Done: Sections 1--7, 9; Not started: Section 8}

%\comm{Amar}{Status: Done: Sections 1--2; Working on: Section 3}

%\section{Introduction}
%\label{sec:intro}

As Chapter~\ref{ch:gaia} discusses, the advancement of machine learning (ML) is heavily dependent on the processing of massive amounts of data.
The most timely and relevant data, however, are often generated at different devices all over the world, e.g., data collected by mobile phones and video cameras.
Because of communication and privacy constraints, gathering all these data for centralized processing is impractical/infeasible, motivating the need for ML training over widely distributed data ({\bf\emph{decentralized learning}}).
For example, as Chapter~\ref{ch:gaia} describes, \emph{geo-distributed learning}~\cite{DBLP:conf/nsdi/HsiehHVKGGM17} trains a global ML model over data spread across geo-distributed data centers. 
Similarly, \emph{federated learning}~\cite{DBLP:conf/aistats/McMahanMRHA17} trains a centralized model over data from a large number of devices (mobile phones).

\textbf{Key Challenges in Decentralized Learning.} There are two key challenges in decentralized learning. 
First, training a model over decentralized data using traditional training approaches (i.e., those designed for centralized data) requires massive communication, which drastically slows down the training process because the communication is bottlenecked by the limited wide-area or mobile network bandwidth~\cite{DBLP:conf/nsdi/HsiehHVKGGM17, DBLP:conf/aistats/McMahanMRHA17}.
Second, decentralized data are typically generated at different contexts, which can lead to significant differences in the \emph{distribution} of the data across data partitions.
%({\bf\emph{non-IID data partitions}}).
For example, facial images collected by cameras will reflect the demographics of each camera's location, and images of kangaroos will be collected only from cameras in Australia or zoos.
Unfortunately, existing decentralized learning algorithms (e.g., ~\cite{DBLP:conf/nsdi/HsiehHVKGGM17, DBLP:conf/aistats/McMahanMRHA17, DBLP:conf/nips/SmithCST17, DBLP:conf/ICLR/LinHMWD18, DBLP:conf/icml/TangLYZL18}) mostly focus on reducing communication, as they either (i) assume the data partitions are independent and identically distributed (IID) or (ii) conduct only a very limited study on non-IID data partitions.  This leaves a key question mostly unanswered: \emph{What happens to different ML applications and decentralized learning algorithms when their training data partitions are not IID?}

\textbf{Our Goal and Key Findings.} In this work, we aim to answer the above key question by conducting the first detailed empirical study of the impact of non-IID data partitions on decentralized learning.
Our study covers various ML applications, ML models, training datasets, decentralized learning algorithms, and degrees of deviation from IID.
We focus on deep neural networks (DNNs) as they are the most relevant solutions for our applications. 
Our study reveals three key findings:

\begin{enumerate}
%\squishlist
\item Training over non-IID data partitions is a fundamental and pervasive problem for decentralized learning. 
All three decentralized learning algorithms in our study suffer from major model quality loss (or even divergence) when run on non-IID data partitions, across \emph{all} applications, models, and training datasets in our study.
\item DNNs with \emph{batch normalization}~\cite{DBLP:conf/icml/IoffeS15} are particularly vulnerable to non-IID data partitions, suffering significant model quality loss even under BSP, the most communication-heavy approach to decentralized learning.
\item The degree of deviation from IID (the \emph{skewness}) is a key determinant of the difficulty level of the problem.
%solution to address it: i.e., a highly non-IID dataset may require a vastly different mechanism compared to a less skewed dataset.
\end{enumerate}
%\squishlistend
\kt{These findings reveal that non-IID data is an important yet heavily understudied challenge in decentralized learning, worthy of extensive study.}%\footnote{To facilitate this, we will release our experimental infrastructure and datasets open source.}

\kt{\textbf{Solutions.}
As two initial steps towards addressing this vast challenge, we first 
show that among the many proposed alternatives to batch normalization, \emph{group normalization}~\cite{DBLP:conf/eccv/WuH18} avoids the skew-induced accuracy loss of batch normalization under BSP.
With this fix, all models in our study perform well under BSP for non-IID data, and the problem can be viewed as a trade-off between accuracy and communication frequency.
Intuitively, there is a tug-of-war among the different data partitions, with each partition pulling the model to reflect its data, and only frequent communication, tuned to the skew-induced accuracy loss, can save the overall model accuracy of the algorithms in our study.
Accordingly, we present \sscout, a system-level approach that adapts the communication frequency of decentralized learning algorithms to accuracy loss, by cross-validating model accuracy across data partitions (\emph{model traveling}).
Our experimental results show that \sscout's adaptive approach automatically reduces communication by 9.6$\times$ (under high skew) to 34.1$\times$ (under mild skew) while retaining the accuracy of BSP.}

%Our study focuses on \emph{label-based} partitioning of data, in which the distribution of labels varies across partitions.
%The paper concludes with a broader taxonomy of regimes of non-IID data, the study of which is left to future work.

\section{Background and Setup}
\label{sec:background}

We provide background on decentralized learning and popular algorithms for this learning setting (\xref{subsec:decentral}) and then describe our study's experimental setup (\xref{subsec:setup}).

\subsection{Decentralized Learning}
\label{subsec:decentral}

In a decentralized learning setting, we aim to train a ML model $w$ based on all the training data samples $(x_i, y_i)$ that are generated and stored in one of the $K$ partitions (denoted as $P_k$). The goal of the training is to fit $w$ to all data samples. Typically, most decentralized learning algorithms assume the data samples are independent and identically distributed (IID) among different $P_k$, and we refer to such a setting as the \emph{IID setting}. Conversely, we call it the \emph{Non-IID setting} if such an assumption does not hold.

We evaluate three popular decentralized learning algorithms to see how they perform on different applications over the IID and Non-IID settings. 
These algorithms can be used with a variety of stochastic gradient descent (SGD) approaches, and aim to reduce communication, either among data partitions ($P_k$) or between the data partitions and a centralized server. 
%The algorithms are:

\begin{itemize}
%\squishlist
 \item \gaia~\cite{DBLP:conf/nsdi/HsiehHVKGGM17}, a geo-distributed learning algorithm that dynamically eliminates insignificant communication among data partitions. 
 Each partition $P_k$ accumulates updates $\Delta w_j$ to each model weight $w_j$ locally, and communicates $\Delta w_j$ to all other data partitions only when its relative magnitude exceeds a predefined threshold (Algorithm~\ref{algo:gaia} in Appendix~\ref{appendix:decentral_algo})%\footnote{All Appendices are in the supplemental material.}).
 
 \item \fedavg~\cite{DBLP:conf/aistats/McMahanMRHA17}, a popular algorithm for federated learning that combines local SGD on each client with model averaging.
 Specifically, \fedavg selects a subset of the partitions $P_k$ in each epoch, runs a prespecified number of local SGD steps on each selected $P_k$, and communicates the resulting models back to a centralized server.
 %and computes the gradient of the loss over all the data held by each selected $P_k$. 
The server averages all these models and uses the averaged $w$ as the starting point for the next epoch (Algorithm~\ref{algo:fedavg} in Appendix~\ref{appendix:decentral_algo}).
   
 \item \dgc~\cite{DBLP:conf/ICLR/LinHMWD18}, a popular algorithm that communicates only a pre-specified amount of gradients each epoch, with various techniques to retain model quality such as momentum correction, gradient clipping~\cite{DBLP:conf/icml/PascanuMB13}, momentum factor masking, and warm-up training~\cite{DBLP:journals/corr/GoyalDGNWKTJH17} (Algorithm~\ref{algo:dgc} in Appendix~\ref{appendix:decentral_algo}).
%\squishlistend
\end{itemize}

In addition to these decentralized learning algorithms, we also show the results of using BSP (bulk synchronous parallel)~\cite{DBLP:journals/cacm/Valiant90} over the IID and Non-IID settings.
BSP is significantly slower than the above algorithms because it does not seek to reduce communication: All updates from each $P_k$ are accumulated and shared among all data partitions after each training iteration (epoch).
As noted earlier, for decentralized learning, there is a natural tension between the frequency of communication and the quality of the resulting model: differing distributions among the $P_k$ pull the model in different directions---more frequent communication helps mitigate this ``tug-of-war'' in order that the model well-represents all the data.
Thus, BSP, with its full communication every iteration, is used as the target baseline for model quality for each application.
%(For the IID setting, we use as the target baseline the published model quality for each application.)

As noted above, all three decentralized learning algorithms and BSP can use a variety of SGD algorithms to train ML models.
\kt{Throughout the study, we use a popular training algorithm, vanilla momentum SGD~\cite{DBLP:journals/nn/Qian99}, to train the DNNs models.}

\subsection{Experimental Setup}
\label{subsec:setup}

Our study consists of three dimensions: \emph{(i)} ML applications/models, \emph{(ii)} decentralized learning algorithms, and \emph{(iii)} degree of deviation from IID. 
We explore all three dimensions with rigorous experimental methodologies. 
In particular, we make sure the accuracy of our trained ML models on IID data matches the reported accuracy in corresponding papers. 
To our knowledge, this is the first detailed empirical study on ML over non-IID data partitions. 

\textbf{Applications.} We evaluate different deep learning applications, DNN model structures, and training datasets:
%to understand how pervasive the problem of non-IID data partitions is. Our applications are:

\begin{itemize}
%\squishlist
 \item \appimage with four DNN models (AlexNet~\cite{DBLP:conf/nips/KrizhevskySH12}, GoogLeNet~\cite{DBLP:conf/cvpr/SzegedyLJSRAEVR15}, LeNet~\cite{lecun1998gradient}, and ResNet~\cite{DBLP:conf/cvpr/HeZRS16}) over two datasets (CIFAR-10~\cite{krizhevsky2009learning} and ImageNet~\cite{ILSVRC15}). We use the validation data accuracy as the model quality metric.
 \item \appface with the center-loss face 
 %DNN 
 model \cite{DBLP:conf/eccv/WenZL016} over the CASIA-WebFace~\cite{DBLP:journals/corr/YiLLL14a} dataset. We use verification accuracy on the LFW dataset~\cite{LFWTech} as the model quality metric.
%\squishlistend
\end{itemize}

For all applications, we tune the training parameters (e.g., learning rate, minibatch size, number of epochs, etc.) such that the baseline model (BSP in the IID setting) achieves the model quality of the original paper. 
We then use these training parameters in all other settings. 
We further ensure that training/validation accuracy has stopped improving by the end of all our experiments.
\kt{It is worth noting that tuning the training parameters could result in different model quality results in both IID and Non-IID settings, but they are not the main focus on this study. We leave the exploration of more combinations of training parameters to future work.}
Appendix~\ref{appendix:training_parameter} lists all major training parameters in our study.

\textbf{Non-IID Data Partitions.} We create non-IID data partitions by partitioning datasets using the \emph{labels} on the data, i.e., using image class for image classification and person identities for face recognition.
This partitioning emulates real-world non-IID settings, which often involve highly unbalanced label distributions across different locations (e.g., kangaroos only in Australia or zoos, a person's face in only a few locations worldwide).
%or even exclusive labels in certain locations (e.g., most faces only exist in certain cameras or devices). 
We control the degree of deviation from IID by controlling the fraction of data that are non-IID. 
For example, 20\% non-IID indicates 20\% of the dataset are partitioned by labels, while the remaining 80\% are partitioned randomly. 

\textbf{Hyper-Parameters Selection.} 
%Decentralized learning can be used in a variety of real-world settings, with differing numbers of data partitions, differing compute resources at each partition, and differing communication bandwidths between partitions.
%Decentralized learning algorithms provide \emph{hyper-parameters} that can be used to adapt to different real-world settings, such as number of data partitions, compute resources, and available bandwidth.
%These hyper-parameters typically control the amount of communication between data partitions (or between the data partitions and a centralized server), so that practitioners can strike a good balance between model quality and communication overhead (and thus training time). 
The algorithms we study provide the following hyper-parameters (see Appendix~\ref{appendix:decentral_algo} for the detail of these algorithms) to control the amount of communication (and hence the training time):

\begin{itemize}
%\squishlist
\item \gaia uses $T_{0}$, the initial threshold to determine if a $\Delta w_j$ is significant. 
Starting with this initial $T_0$, the significance threshold decreases whenever the learning rate decreases. 

\item \fedavg uses $Iter_{Local}$ to control the number of local SGD steps on each selected $P_k$.

\item \dgc uses $s$ to control the sparsity of updates (update magnitudes in top $s$ percentile are exchanged). 
Following the original paper~\cite{DBLP:conf/ICLR/LinHMWD18}, $s$ follows a warm-up scheduling: 75\%, 93.75\%, 98.4375\%, 99.6\%, 99.9\%. 
We use a hyper-parameter $E_{warm}$, the number of epochs for each warm-up sparsity, to control the duration of the warm-up.
For example, if $E_{warm} = 4$, $s$ is 75\% in epochs 1--4, 93.75\% in epochs 5--8, 98.4375\% in epochs 9--12, 99.6\% in epochs 13--16, and 99.9\% in epochs 17+.
%\squishlistend
\end{itemize}

We select a hyper-parameter $\theta$ of each decentralized learning algorithms ($T_{0}$, $Iter_{Local}$, $E_{warm}$) so that \emph{(i)} $\theta$ achieves the same model quality as BSP in the IID setting and \emph{(ii)} $\theta$ achieves similar communication savings across the three decentralized learning algorithms. 
We study the sensitivity of our findings to the choice of $\theta$ in \xref{subsec:decentral_parameter}.

\section{Non-IID Study: Results Overview}
\label{sec:overview}

This chapter seeks to answer the question as to what happens to ML applications, ML models, and decentralized learning algorithms when their training data partitions are not IID.
In this section, we provide an overview of our findings, showing that
non-IID data partitions cause \emph{major model quality loss}, across all applications, models, and algorithms in our study.
We discuss the results for \appimage  in \xref{subsec:overview_cifar10} and \xref{subsec:overview_imagenet} and for \appface in \xref{subsec:overview_face}.

\subsection{Image Classification with CIFAR-10}
\label{subsec:overview_cifar10}

We first present the model quality with different decentralized learning algorithms over the IID and Non-IID settings for \appimage over the CIFAR-10 dataset. 
We use five partitions ($K = 5$) in this evaluation. 
As the CIFAR-10 dataset consists of ten object classes, each data partition has two object classes in the Non-IID setting. 
Figure~\ref{fig:overview_cifar10} shows the results with four popular DNNs (AlexNet, GoogLeNet, LeNet, and ResNet).
According to the hyper-parameter criteria in \xref{subsec:setup}, we select $T_0 = 10\%$ for \gaia, $Iter_{Local} = 20$ for \fedavg, and $E_{warm} = 8 $ for \dgc.
We make two major observations.

\begin{figure*}[t!]
\centering  
\includegraphics[width=1.0\textwidth]{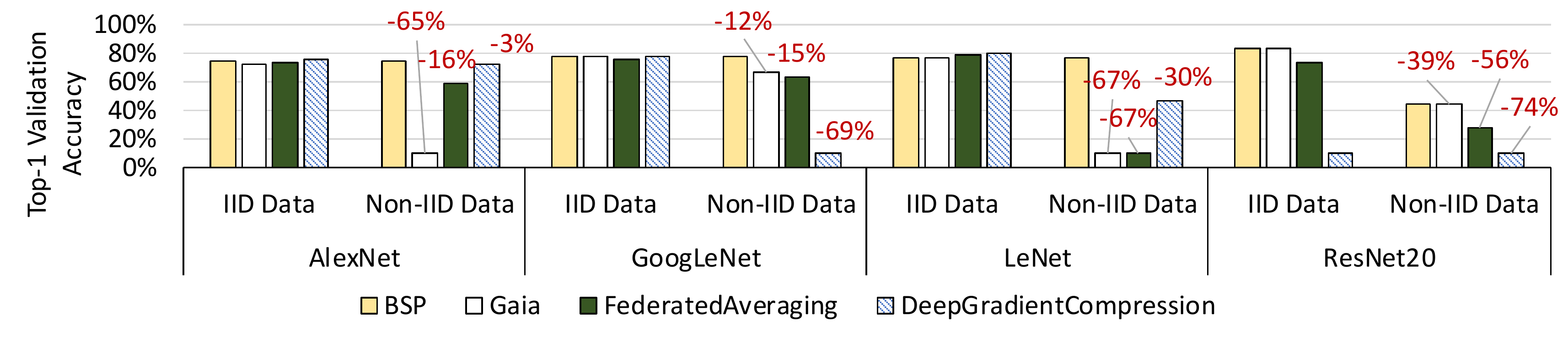}\vspace{-0.15in}
\caption[Top-1 validation accuracy for \appimage over the CIFAR-10 dataset]{Top-1 validation accuracy for \appimage over the CIFAR-10 dataset. 
Each ``-x\%'' label indicates the accuracy loss from BSP in the IID setting.}
%The number in the parentheses for each decentralized algorithm is the speedup of this algorithm over BSP.
\label{fig:overview_cifar10} 
\vspace{-0.02in}
\end{figure*}  

\textbf{1) It is a pervasive problem.} 
\emph{All} three decentralized learning algorithms lose significant model quality for \emph{all} four DNNs in the Non-IID setting. 
We see that while these algorithms retain the validation accuracy of BSP in the IID setting with 15$\times$--20$\times$ communication savings (agreeing with the results from the original papers for these algorithms), they lose 3\% to 74\% validation accuracy in the Non-IID setting.
Simply running these algorithms for more epochs would not help because the training/validation accuracy has already stopped improving.
Furthermore, the training is completely diverged in some cases, such as \dgc with GoogLeNet and ResNet20 (\dgc with ResNet20 also diverges in the IID setting). 
The pervasiveness of the problem is quite surprising, as we have a diverse set of decentralized learning algorithms and DNNs. 
\kt{While BSP can retain model quality for most of the DNNs (AlexNet, GoogLeNet, and LeNet), its heavy communication (an order of magnitude more than these decentralized learning algorithms) makes it impractical for most decentralized learning scenarios that are bottlenecked by communication.}
This result shows that Non-IID data is a pervasive and challenging problem for decentralized learning, and this problem has been heavily overlooked. 
\xref{sec:decentral} discusses the cause of this problem.

\textbf{2) Even BSP cannot completely solve this problem.} We see that even BSP, with its full communication every iteration, cannot retain model quality for some DNNs in the Non-IID setting. 
In particular, the validation accuracy of ResNet20 in the Non-IID setting is 39\% lower than that in the IID setting. 
This finding suggests that, for some DNNs, it \emph{may not be possible} to solve the Non-IID data challenge by communicating more frequently between data partitions ($P_k$). 
We find that this problem not only exists in ResNet20, but in all DNNs we study with batch normalization layers (ResNet10, BN-LeNet~\cite{DBLP:conf/icml/IoffeS15} and Inception-v3~\cite{DBLP:conf/cvpr/SzegedyVISW16}).
We discuss this problem and potential solutions in \xref{sec:batch_norm}.

\subsection{Image Classification with ImageNet}
\label{subsec:overview_imagenet}

We study \appimage over the ImageNet dataset~\cite{ILSVRC15} dataset (1,000 image classes) to see if the Non-IID data problem exists in different datasets. 
We use two partitions ($K = 2$) in this experiment so each partition gets 500 image classes. 
%We use well-established training hyper-parameters for GoogLeNet~\cite{googlenetmodel} and ResNet~\cite{simon2016cnnmodels}, and we make sure the validation accuracy in our setup matches the original papers~\cite{DBLP:conf/cvpr/HeZRS16, DBLP:conf/cvpr/SzegedyLJSRAEVR15}. 
According to the hyper-parameter criteria in \xref{subsec:setup}, we select $T_0 = 40\%$ for \gaia, $Iter_{Local} = 200$ for \fedavg, and $E_{warm} = 4$ for \dgc.

%\begin{wrapfigure}{R}{0.5\textwidth}
\begin{figure}[h]
\centering
\includegraphics[width=0.70\textwidth]{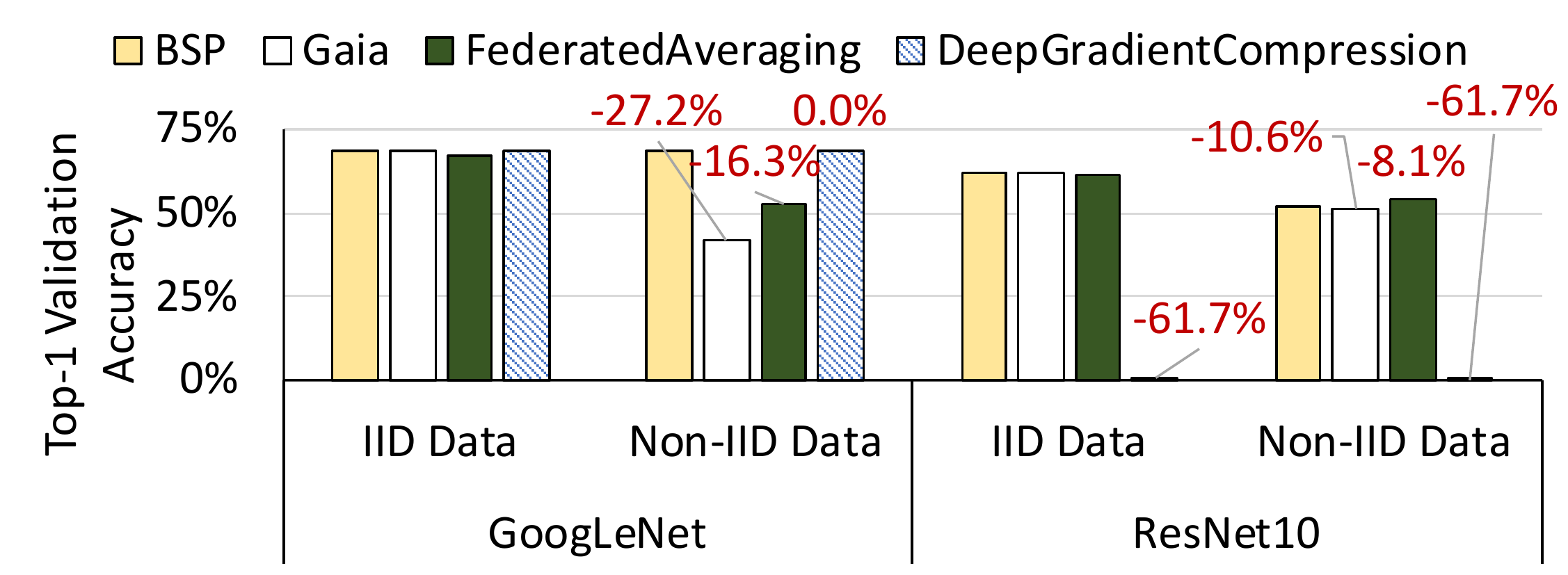}\vspace{-0.15in}
\caption[Top-1 validation accuracy for \appimage over the ImageNet dataset]{Top-1 validation accuracy for \appimage over the ImageNet dataset. Each ``-x\%'' label indicates the accuracy loss from BSP in the IID setting.} 
\label{fig:overview_imagenet}
\end{figure}
%\end{wrapfigure}  
    
\textbf{The same trend in different datasets.} Figure~\ref{fig:overview_imagenet} illustrates the validation accuracy in the IID and Non-IID  settings. 
Interestingly, we observe the same problems in the ImageNet dataset, whose number of classes is two orders of magnitude more than the CIFAR-10 dataset. 
First, we see that \gaia and \fedavg lose significant validation accuracy (8.1\% to 27.2\%) for both DNNs in the Non-IID setting. 
On the other hand, while \dgc is able to retain the validation accuracy for GoogLeNet in the Non-IID setting, it cannot converge to a useful model for ResNet10. 
Second, BSP also cannot retain the validation accuracy for ResNet10 in the Non-IID setting, which concurs with our observation in the CIFAR-10 study. 
These results show that the Non-IID data problem not only exists in various decentralized learning algorithms and DNNs, but also exists in different datasets.

\subsection{Face Recognition}
\label{subsec:overview_face}

We further examine another popular ML application, \appface, to see if the Non-IID data problem is a fundamental challenge across different applications. 
We again use two partitions ($K = 2$) in this evaluation, and we store different people's faces in different partitions in the Non-IID setting. 
According to the hyper-parameter criteria in \xref{subsec:setup}, we select $T_0 = 20\%$ for \gaia, $Iter_{Local} = 50$ for \fedavg, and $E_{warm} = 8$ for \dgc. 
It is worth noting that the verification process of \appface is fundamentally different from \appimage, as \appface does \emph{not} use the classification layer (and thus the training labels) at all in the verification process. 
Instead, for each pair of verification images, the trained DNN is used to compute a feature vector for each image, and the distance between these feature vectors is used to determine if the two images belong to the same person. 

\begin{figure}[h]
\centering
\includegraphics[width=0.70\textwidth]{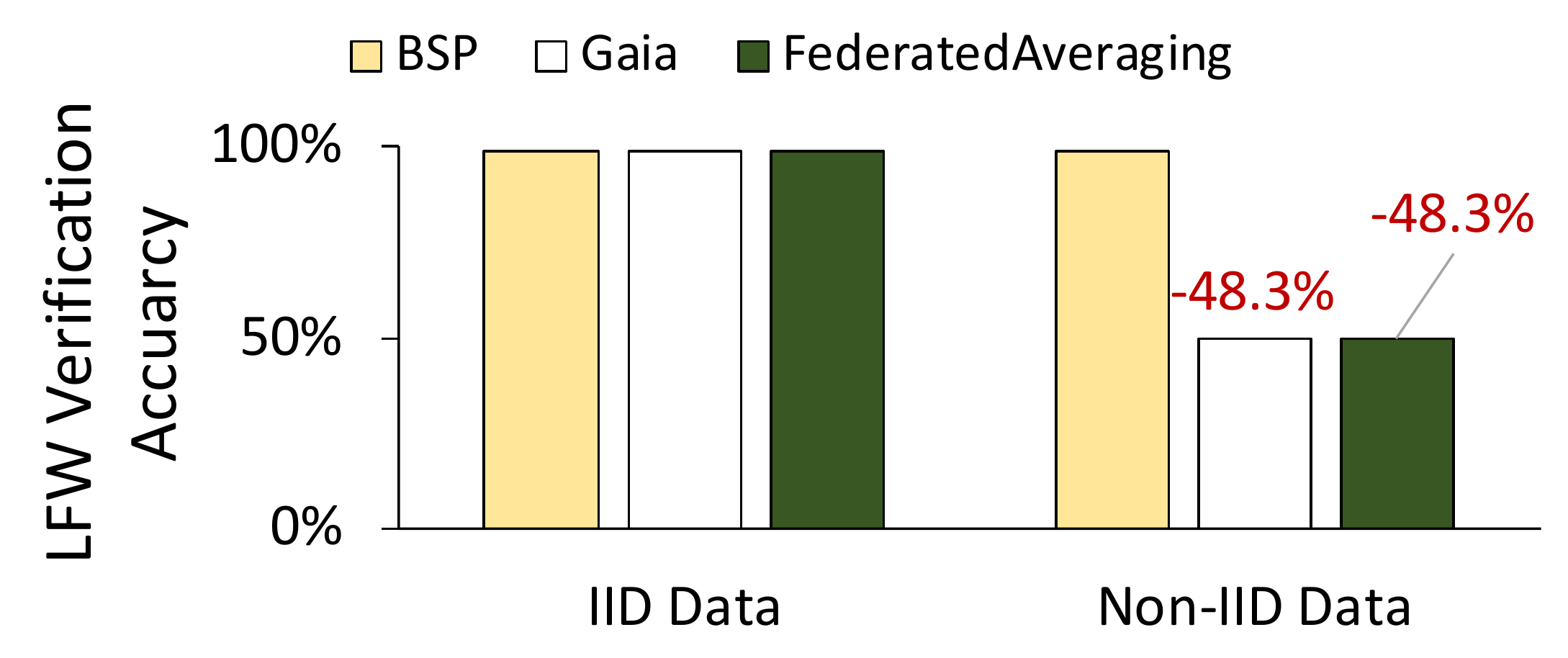}\vspace{-0.15in}
\caption[LFW verification accuracy for \appface]{LFW verification accuracy for \appface. Each ``-x\%'' label indicates the accuracy loss from BSP in the IID setting.} 
\label{fig:overview_face}
\vspace{-0.1in}
\end{figure}

\textbf{The same problem in different applications.} Figure~\ref{fig:overview_face} presents the LFW verification accuracy using different decentralized learning algorithms in the IID and Non-IID settings. Again, the same problem happens in this application: these decentralized learning algorithms work well in the IID setting, but they lose significant accuracy in the Non-IID setting. In fact, both \gaia and \fedavg cannot converge to a useful model in the Non-IID setting, and their 50\% accuracy is from random guessing (the verification process is a series of binary questions). This result is particularly noteworthy as \appface uses a vastly different verification process that does not rely on the training labels, which are used to create the Non-IID setting to begin with. We conclude that Non-IID data is a fundamental and pervasive problem across various applications, datasets, models, and decentralized learning algorithms.

\section{Problems of Decentralized Learning Algorithms}
\label{sec:decentral}

The results in \xref{sec:overview} show that three diverse decentralized learning algorithms all suffer drastic accuracy loss in the Non-IID setting.
We investigate the potential reasons for this (\xref{subsec:decentral_cause}) and the sensitivity to hyper-parameter choice (\xref{subsec:decentral_parameter}).

\subsection{Reasons for Model Quality Loss}
\label{subsec:decentral_cause}

%\subsubsection{Gaia} 
%\label{subsec:decentral_cause_gaia}

\textbf{Gaia.}
%In order to understand what results in model quality loss in the Non-IID setting, we 
We extract the \gaia-trained models from both partitions (denoted DC-0 and DC-1) for \appimage over the ImageNet dataset, and then evaluate the validation accuracy of each model based on the \emph{image classes} in each partition. 
As Figure~\ref{fig:cause_gaia} shows, the validation accuracy is pretty consistent among the two sets of image classes when training the model in the IID setting: the results for IID DC-0 Model are shown, and IID DC-1 Model is the same. 
However, the validation accuracy varies drastically under the Non-IID setting (Non-IID DC-0 Model and Non-IID DC-1 Model). 
Specifically, both models perform well for the image classes in their respective partition, but they perform very poorly for the image classes that are \emph{not} in their respective partition.
%, which explains the accuracy loss. 
This reveals that using \gaia in the Non-IID setting results in \emph{completely different} models among data partitions, and each model is only good for recognizing the image classes in its data partition.

%\begin{wrapfigure}{R}{0.5\textwidth}
\begin{figure}[h]
\centering
\includegraphics[width=0.70\textwidth]{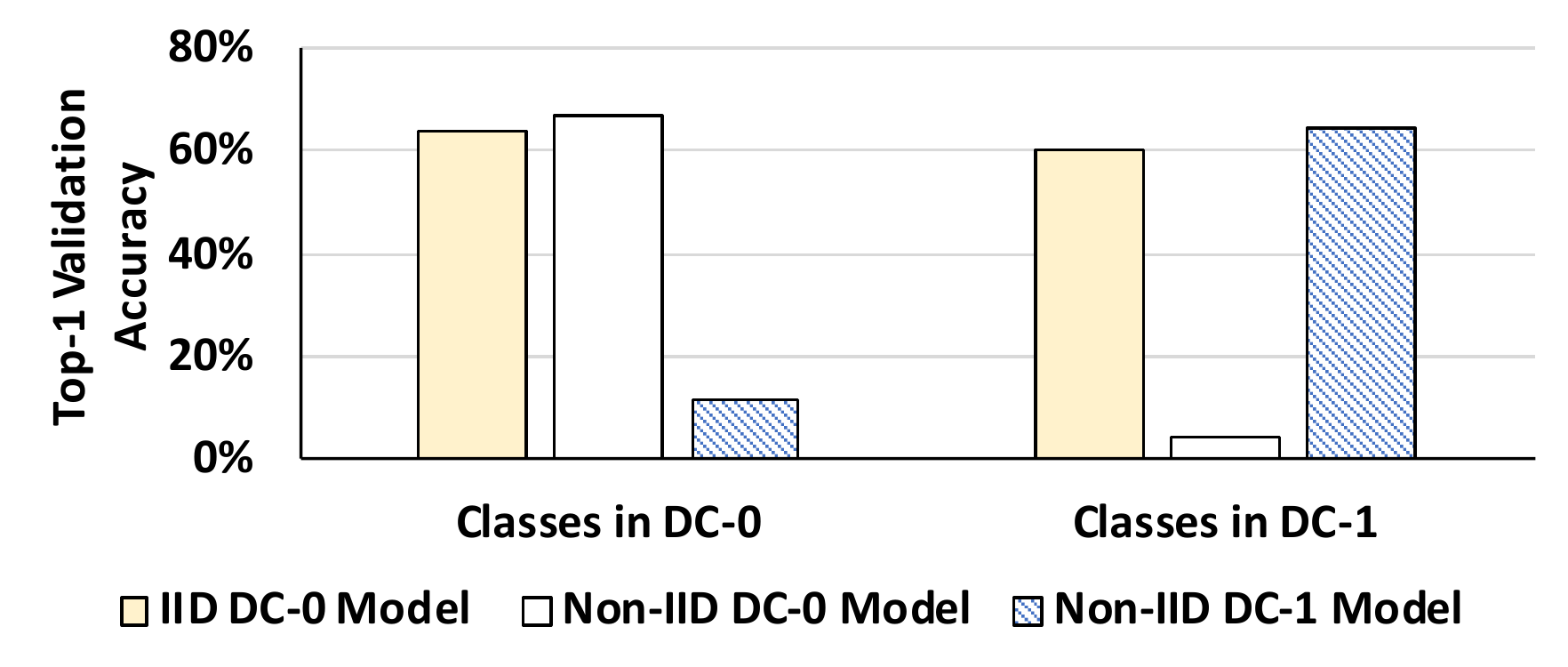}\vspace{-0.15in}
\caption{Top-1 validation accuracy (ImageNet) for models in different partitions.}
\label{fig:cause_gaia}
\vspace{-0.05in}
\end{figure}
%\end{wrapfigure}  

This raises the following question: How does \gaia produce completely different models in the Non-IID setting, given that \gaia synchronizes all significant updates ($\Delta w_j$) to ensure that the differences across models in each weight $w_j$ is insignificant (\xref{sec:background})? 
To answer this, we first compare each weight $w_j$ in the Non-IID DC-0 and DC-1 Models, and find that the average difference among all the weights is only 0.5\% (reflecting that the threshold for significance in the last epoch was 1\%).
However, we find that given the same input image, the \emph{neuron} values are vastly different (at an average difference of 173\%). 
This finding suggests that small model differences can result in completely different models.
Mathematically, this is because weights are both positive and negative: a small percentage difference in individual weights of a neuron can lead to a large percentage difference in its value.
As \gaia eliminates insignificant communication, it creates an opportunity for models in each data partition to specialize for the image classes in their respective data partition, at the expense of other classes.

\textbf{DeepGradientCompression.}
While local model specialization explains why \gaia performs poorly in the Non-IID setting, it is still unclear why other decentralized learning algorithms also exhibit the same problem. 
More specifically, \dgc and \fedavg always maintain \emph{one} global model, hence there is no room for local model specialization. 
To understand why these algorithms perform poorly, we study the average residual update delta ($||\Delta w_i / w_i||$) with \dgc. 
This number represents the magnitude of the gradients that have \emph{not} yet been exchanged among different $P_k$, due to its communicating only a fixed number of gradients in each epoch (\xref{sec:background}). 
Thus, it can be seen as the amount of gradient divergence among different $P_k$.

Figure~\ref{fig:update_delta_dgc} depicts the average residual update delta for the first 20 training epochs when training ResNet20 over the CIFAR-10 dataset. 
We show only the first 20 epochs because the training diverges after that in the Non-IID setting. 
As the figure shows, the average residual update delta is an order of magnitude higher in the Non-IID setting (283\%) than that in the IID setting (27\%). 
Hence, each $P_k$ generates large gradients in the Non-IID setting, which is not surprising as each $P_k$ sees vastly different training data in the Non-IID setting. 
However, these large gradients are not synchronized because \dgc sparsifies the gradients at a fixed rate. 
When they are finally synchronized, they may have diverged so much from the global model that they lead to the divergence of the whole model. 
Our experiments also support this proposition, as we see \dgc diverges much more often in the Non-IID setting. 

%\begin{wrapfigure}{R}{0.48\textwidth}
\begin{figure}[h]
\centering
\includegraphics[width=0.80\textwidth]{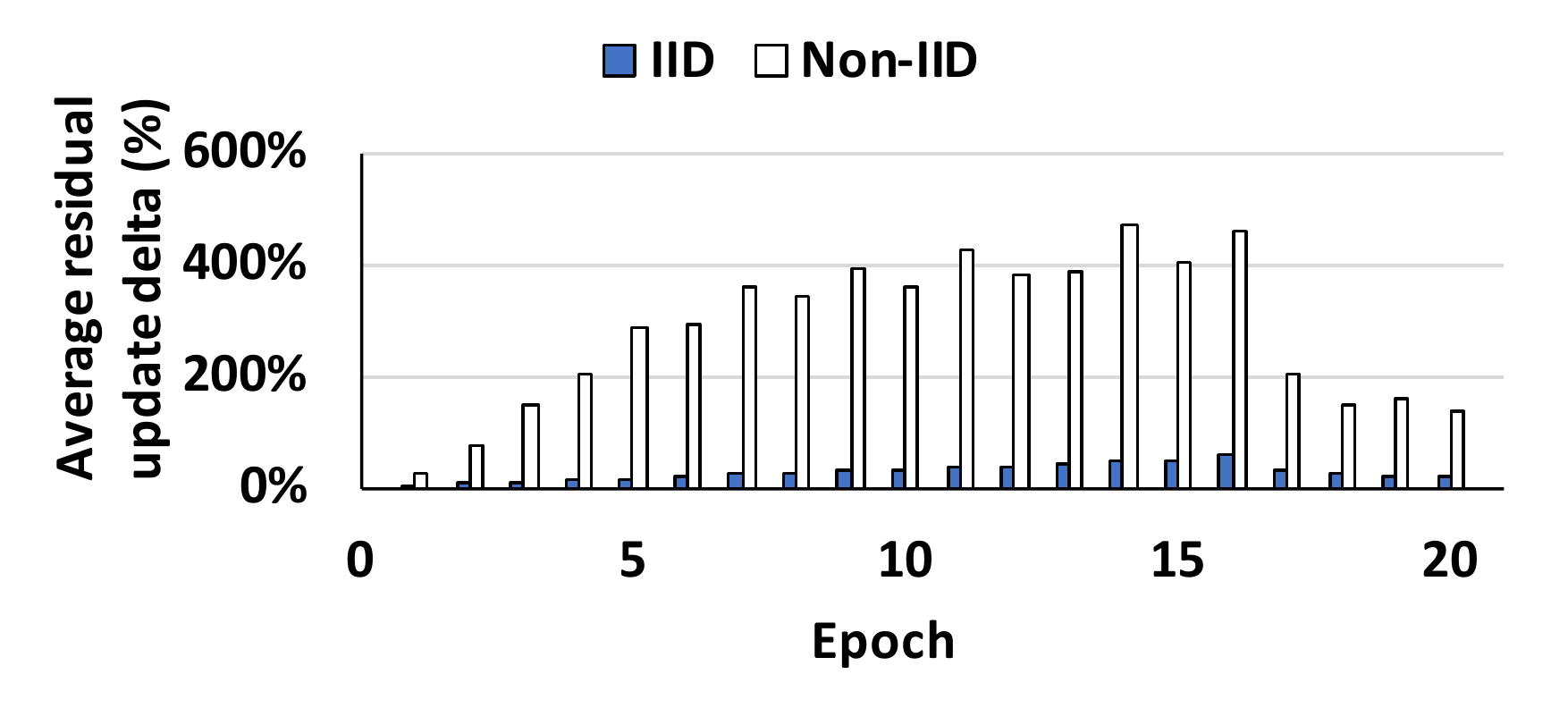}\vspace{-0.15in}
\caption{Average residual update delta (\%) for \dgc over the first 20 epochs.} 
\label{fig:update_delta_dgc}
\end{figure}
%\vspace{-30pt}
%\end{wrapfigure} 

%\subsubsection{FederatedAveraging}

\textbf{FederatedAveraging.}
The above analysis for \dgc can also apply to \fedavg, which delays communication from each $P_k$ by a fixed number of local iterations.
If the weights in different $P_k$ diverge too much, the synchronized global model can lose accuracy or completely diverge~\cite{DBLP:journals/corr/abs-1806-00582}. 
We validate this by plotting the average local weight update delta for \fedavg at each global synchronization ($||\Delta w_i / w_i||$, where $w_i$ is the averaged global model weight). 
Figure~\ref{fig:update_delta_fedavg} depicts this number for the first 25 training epochs when training AlexNet over the CIFAR-10 dataset. 
As the figure shows, the average local weight update delta in the Non-IID setting (48.5\%) is much higher than that in the IID setting (20.2\%), which explains why Non-IID data partitions lead to major accuracy loss for \fedavg.
The difference is less pronounced than with \dgc, so the impact on accuracy is smaller.

%\begin{wrapfigure}{R}{0.48\textwidth}
\begin{figure}[h]
\centering
\includegraphics[width=0.80\textwidth]{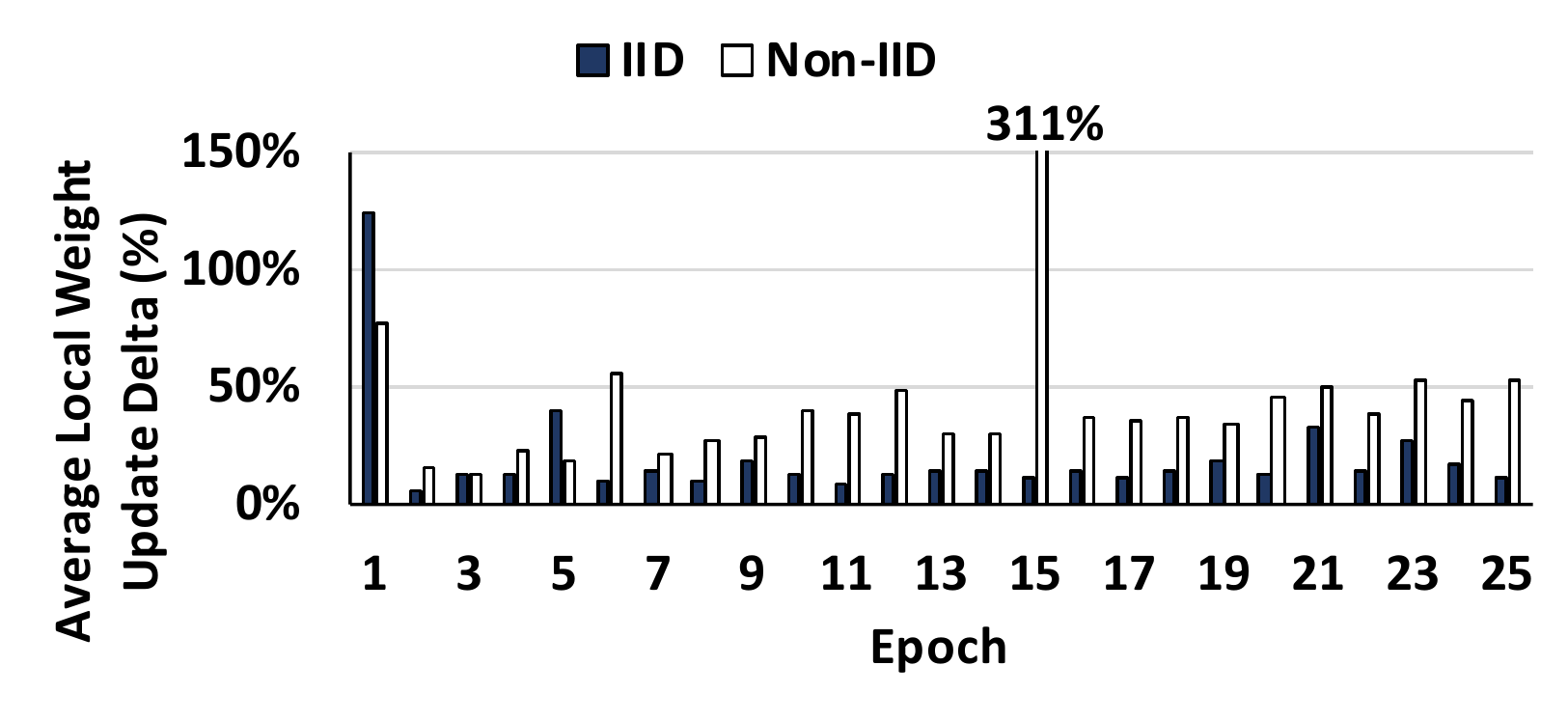}
\caption{Average local update delta (\%) for \fedavg over the first 25 epochs.} 
\label{fig:update_delta_fedavg}
\end{figure}
%\end{wrapfigure} 

\subsection{Algorithm Hyper-Parameters}
\label{subsec:decentral_parameter}

We now study the sensitivity of the non-IID problem to hyper-parameter choice.
%We consider a rich set of hyper-parameters for the decentralized algorithms to understand the effect of tuning these hyper-parameters in the Non-IID setting (\xref{subsec:setup}). 
Table~\ref{tbl:gaia_hyper_parameter} presents the \gaia results for varying its $T_0$ hyper-parameter (\xref{subsec:setup}) when training on CIFAR-10, and we leave the results for the other two algorithms to Appendix~\ref{appendix:decentral_parameter}. 
As the table shows, we study seven choices for $T_0$ and compare the results with BSP. 
We make two major observations.

First, almost all hyper-parameter settings lose significant accuracy in the Non-IID setting (relative to BSP in the IID setting).
Even with a relatively conservative hyper-parameter setting (e.g., $T_0 = 2\%$, the most communication-intensive of the choices shown), we still see 4.4\% to 35.7\% accuracy loss among these DNNs. 
On the other hand, the exact same hyper-parameter choice for \gaia in the IID setting can achieve close to BSP-level accuracy (within 1.1\%). 
We see the same trend with much more aggressive hyper-parameter settings as well (e.g., $T_0 = 40\%$). 
This shows that the problem of Non-IID data partitions is not specific to particular hyper-parameter settings, and that hyper-parameter settings that work well in the IID setting may perform poorly in the Non-IID setting.

Second, more conservative hyper-parameter settings (which implies more frequent communication among the $P_k$) often greatly decrease the accuracy loss in the Non-IID setting. 
For example, the validation accuracy with $T_0 = 2\%$ is significantly higher than the one with $T_0 = 30\%$. 
This suggests that we may be able to use more frequent communication among the $P_k$ for higher model quality in the Non-IID setting (mitigating the ``tug-of-war'' among the $P_k$  (\xref{subsec:decentral})). 
However, this trend is not monotonic, as several more conservative hyper-parameter settings result in worse models (e.g., GoogLeNet with $T_0 = 2\%$ vs.~$T_0 = 20\%$). 
We conclude that hyper-parameter tuning alone may not solve all the problems in the Non-IID setting, but it is a direction worth further exploration. 

\begin{table}[t]
  \centering
  \small
  \begin{tabular}{c cc cc cc cc}
    \toprule
    \multirow{2}{*}{Configuration} & \multicolumn{2}{c}{AlexNet} & \multicolumn{2}{c}{GoogLeNet} & \multicolumn{2}{c}{LeNet} & \multicolumn{2}{c}{ResNet20} \\
    \cmidrule(lr){2-3} \cmidrule(lr){4-5} \cmidrule(lr){6-7} \cmidrule(lr){8-9}
    & {IID} & {Non-IID} & {IID} & {Non-IID} & {IID} & {Non-IID} & {IID} & {Non-IID} \\ \midrule
    BSP &  74.9\% & 75.0\%	& 79.1\% & 78.9\% & 77.4\% & 76.6\% & 83.7\% & \bad{44.3\%} \\ \midrule
    $T_0 = 2\%$ & 73.8\% & \bad{70.5\%} & 78.4\% & \bad{56.5\%} & 76.9\% & \bad{52.6\%} & 83.1\% & \bad{48.0\%} \\ \midrule
    $T_0 = 5\%$ & 73.2\% & \bad{71.4\%} & 77.6\% & \bad{75.6\%} & \bad{74.6\%} & \bad{10.0\%} & 83.2\% & \bad{43.1\%} \\ \midrule
    $T_0 = 10\%$ & 73.0\% & \bad{10.0\%} & 78.4\% & \bad{68.0\%} & 76.7\% & \bad{10.0\%} & 84.0\% & \bad{45.1\%} \\ \midrule
    $T_0 = 20\%$ & \bad{72.5\%} & \bad{37.6\%} & 77.7\% & \bad{67.0\%} & 77.7\% & \bad{10.0\%} & 83.6\% & \bad{38.9\%} \\ \midrule
    $T_0 = 30\%$ & \bad{72.4\%} & \bad{26.0\%} & 77.5\% & \bad{23.9\%} & 78.6\% & \bad{10.0\%} & 81.3\% & \bad{39.4\%} \\ \midrule
    $T_0 = 40\%$ & \bad{71.4\%} & \bad{20.1\%} & 77.2\% & \bad{33.4\%} & 78.3\% & \bad{10.1\%} & 82.1\% & \bad{28.5\%} \\ \midrule
    $T_0 = 50\%$ & \bad{10.0\%} & \bad{22.2\%} & \bad{76.2\%} & \bad{26.7\%} & 78.0\% & \bad{10.0\%} & \bad{77.3\%} & \bad{28.4\%}  \\     
    \bottomrule \\
  \end{tabular}
  \caption[Top-1 validation accuracy (CIFAR-10) varying \gaia's $T_0$ hyper-parameter.]{Top-1 validation accuracy (CIFAR-10) varying \gaia's $T_0$ hyper-parameter.
  The configurations with more than 2\% accuracy loss from BSP in the IID setting are highlighted.  
  Note that larger settings for $T_0$ mean significantly greater communication savings.}
  %(we used $T_0=10\%$ in \xref{subsec:overview_cifar10}).}
  \label{tbl:gaia_hyper_parameter}
%  \vspace{-20pt}
\end{table}

\section{Batch Normalization: Problem and Solution}
\label{sec:batch_norm}

As \xref{sec:overview} discusses, even BSP cannot retain model quality in the Non-IID setting for DNNs with batch normalization layers. 
Given how popular batch normalization is, this is a problem that can have far-reaching ramifications. 
We first discuss why batch normalization is particularly vulnerable in the Non-IID setting (\xref{subsec:batch_norm_problem}) and then study alternative normalization techniques, including one---Group Normalization---that works better in this setting (\xref{subsec:batch_norm_solution}). 

%As \xref{sec:overview} discusses, even BSP cannot retain model quality in the Non-IID setting for DNNs with batch normalization layers. 
%We discuss more about problem of Batch Normalization and potential solutions.
%Given how popular batch normalization is, this is a problem that can have far-reaching ramifications. 
%We first discuss why batch normalization is particularly vulnerable in the Non-IID setting (\xref{subsec:batch_norm_problem}) and then study alternative normalization techniques, including one---Group Normalization---that works better in this setting (\xref{subsec:batch_norm_solution}). 

\subsection{The Problem of Batch Normalization in the Non-IID Setting}
\label{subsec:batch_norm_problem}

Batch normalization~\cite{DBLP:conf/icml/IoffeS15} (BatchNorm) is one of the most popular mechanisms in deep learning, and it has been employed by default in most deep learning models (more than 11,000 citations). 
BatchNorm enables faster and more stable DNN training because it enables larger learning rates, which in turn make convergence much faster and help avoid sharp local minimum (hence, the model generalizes better).

\textbf{How BatchNorm works.} 
BatchNorm aims to stabilize a DNN by normalizing the input distribution of selected layers such that the inputs $x_i$ on each channel $i$ of the layer have zero mean ($\mu_{x_i} = 0$) and unit variance ($\sigma_{x_i} = 1$). 
Because the global mean and variance is unattainable with stochastic training, BatchNorm uses \emph{minibatch mean and variance} as an estimate of the global mean and variance. 
Specifically, for each minibatch $\mathcal{B}$, BatchNorm calculates the minibatch mean $\mu_{\mathcal{B}}$ and variance $\sigma_{\mathcal{B}}$, and then uses $\mu_{\mathcal{B}}$ and $\sigma_{\mathcal{B}}$ to normalize each $x_i$ in $\mathcal{B}$~\cite{DBLP:conf/icml/IoffeS15}. 
Recent work shows that BatchNorm enables larger learning rates because: \emph{(i)} BatchNorm corrects large gradient updates that could result in divergence~\cite{DBLP:conf/nips/BjorckGSW18} and \emph{(ii)} BatchNorm makes the underlying problem's landscape significantly more smooth~\cite{DBLP:conf/nips/SanturkarTIM18}.

\textbf{BatchNorm and the Non-IID setting.} 
While BatchNorm is effective in practice, its dependence on minibatch mean and variance ($\mu_{\mathcal{B}}$ and $\sigma_{\mathcal{B}}$) is known to be problematic in certain settings. 
This is because BatchNorm uses $\mu_{\mathcal{B}}$ and $\sigma_{\mathcal{B}}$ for training, but it typically uses an estimated global mean and variance ($\mu$ and $\sigma$) for validation. 
If there is a major mismatch between these means and variances, the validation accuracy is going to be low because the input distribution during validation does not match the distribution during training. 
This can happen if the minibatch size is small or the sampling of minibatches is not IID~\cite{DBLP:conf/nips/Ioffe17}. 
The Non-IID setting in our study exacerbates this problem because each data partition $P_k$ sees very different training samples. 
Hence, the $\mu_{\mathcal{B}}$ and $\sigma_{\mathcal{B}}$ in each $P_k$ can vary significantly in the Non-IID setting, and the synchronized global model may not work for \emph{any} set of data.
Worse still, we cannot simply increase the minibatch size or do better minibatch sampling to solve this problem, because in the Non-IID setting the underlying training dataset in each $P_k$ does not represent the global training dataset.

%\begin{wrapfigure}{R}{0.50\textwidth}
\begin{figure}[h]
\centering
\includegraphics[width=0.70\textwidth]{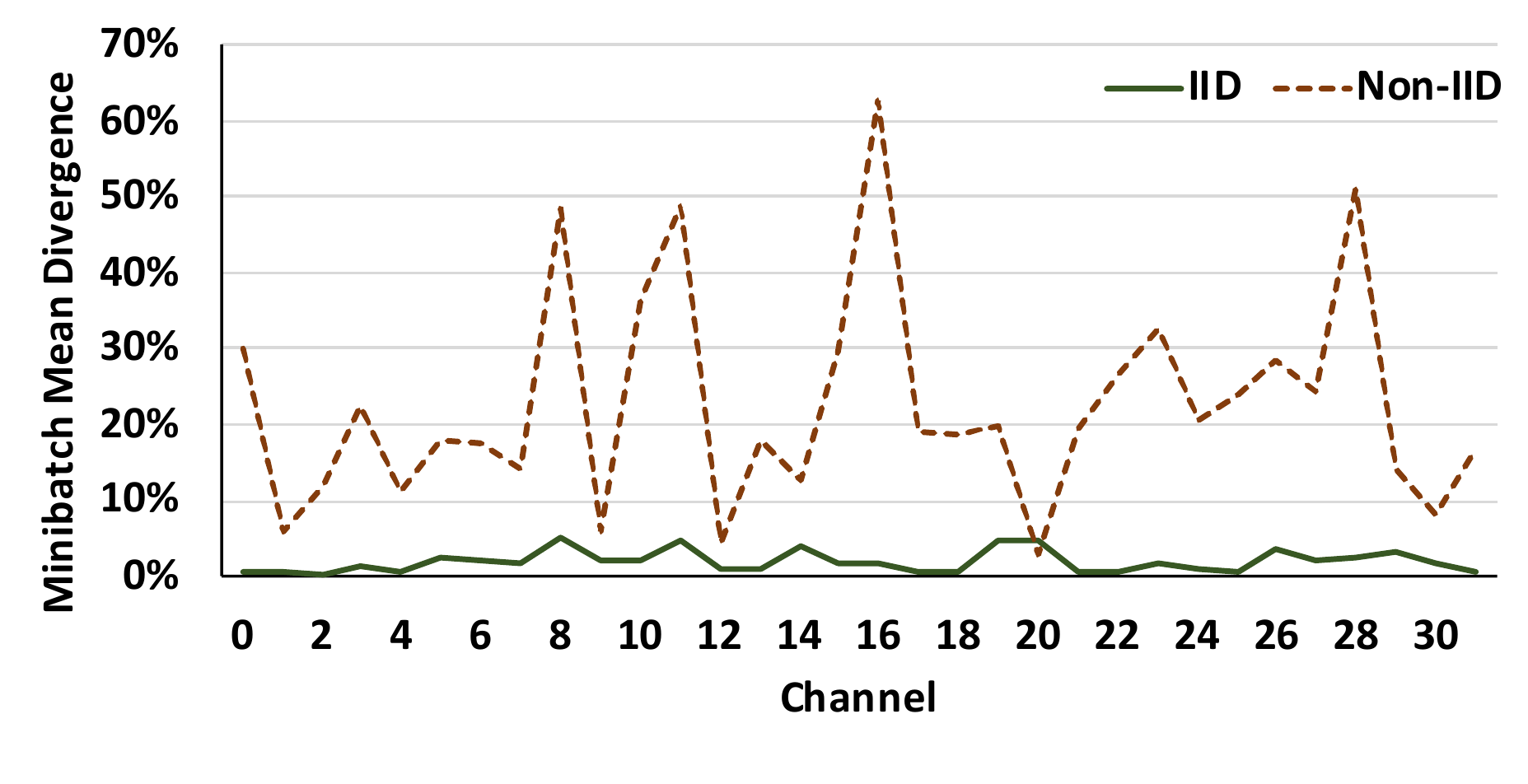}\vspace{-0.2in}
\caption{Minibatch mean divergence for the first layer of BN-LeNet over CIFAR-10 using two $P_k$.} 
\label{fig:batch_mean_divergence}
\end{figure}
%\vspace{-30pt}
%\end{wrapfigure} 

We validate if there is indeed major divergence in $\mu_{\mathcal{B}}$ and $\sigma_{\mathcal{B}}$ among different $P_k$ in the Non-IID setting. 
We calculate the divergence of $\mu_{\mathcal{B}}$ as the difference between $\mu_{\mathcal{B}}$ in different $P_k$ over the average $\mu_{\mathcal{B}}$ (i.e., it is $\frac{||\mu_{\mathcal{B}, P_0} - \mu_{\mathcal{B}, P_1}||}{||AVG(\mu_{\mathcal{B}, P_0},\ \mu_{\mathcal{B}, P_1})||}$ for two partitions $P_0$ and $P_1$). 
We use the average $\mu_{\mathcal{B}}$ over every 100 minibatches in each $P_k$ so that we get better estimation. 
Figure~\ref{fig:batch_mean_divergence} depicts the divergence of $\mu_{\mathcal{B}}$ for each channel of the first layer of BN-LeNet, which is constructed by inserting BatchNorm to LeNet after each convolutional layer. 
As we see, the divergence of $\mu_{\mathcal{B}}$ is significantly larger in the Non-IID setting (between 6\% to 51\%) than that in the IID setting (between 1\% to 5\%). 
We also observe the same trend in minibatch variances $\sigma_{\mathcal{B}}$ (not shown). 
As discussed earlier, this phenomenon is detrimental to training: Each $P_k$ uses very different $\mu_{\mathcal{B}}$ and $\sigma_{\mathcal{B}}$ to normalize its model, but the resultant global model can use only one $\mu$ and $\sigma$ which cannot match all of these diverse batch means and variances. 
As this problem has nothing to do with the frequency of communication among $P_k$, it explains why even BSP cannot retain model accuracy for BatchNorm in the Non-IID setting.

\subsection{Alternatives to Batch Normalization}
\label{subsec:batch_norm_solution}

As the problem of BatchNorm in the Non-IID setting is due to its dependence on minibatches, the natural solution is to replace BatchNorm with alternative normalization mechanisms that are \emph{not} dependent on minibatches. 
Unfortunately, most existing alternative normalization mechanisms have their own drawbacks. We first discuss the normalization mechanisms that have major shortcomings, and then we discuss one particular mechanism that may be used as a potential solution.

\textbf{Weight Normalization~\cite{DBLP:conf/nips/SalimansK16}.} 
Weight Normalization (WeightNorm) is a normalization scheme that normalizes the weights in a DNN as oppose to the neurons (which is what BatchNorm and most other normalization techniques do). 
WeightNorm is not dependent on minibatches as it is normalizing the weights. 
However, while WeightNorm can effectively control the variance of the neurons, it still needs a mean-only BatchNorm in many cases to achieve the model quality and training speeds of BatchNorm~\cite{DBLP:conf/nips/SalimansK16}. 
This mean-only BatchNorm makes WeightNorm vulnerable to the Non-IID setting again, because there is a large divergence in $\mu_{\mathcal{B}}$ among the $P_k$ in the Non-IID setting (\xref{subsec:batch_norm_problem}).

\textbf{Layer Normalization~\cite{DBLP:journals/corr/BaKH16}.} 
Layer Normalization (LayerNorm) is a technique that is inspired by BatchNorm. 
Instead of computing the mean and variance of a minibatch for each \emph{channel}, LayerNorm computes the mean and variance across all channels for each \emph{sample}. 
Specifically, if the inputs are four-dimensional vectors $\mathcal{B}\times\mathcal{C}\times\mathcal{W}\times\mathcal{H}$ (batch $\times$ channel $\times$ width $\times$ height), BatchNorm produces $\mathcal{C}$ means and variances along the $\mathcal{B} \times \mathcal{W} \times \mathcal{H}$ dimensions. 
On the other hand, LayerNorm produces $\mathcal{B}$ means and variances along the $\mathcal{C} \times \mathcal{W} \times \mathcal{H}$ dimensions (per-sample mean and variance). 
As the normalization is done on a per-sample basis, LayerNorm is not dependent on minibatches. 
However, LayerNorm makes a key assumption that all inputs make similar contributions to the final prediction. But this assumption does not hold for some models such as convolutional neural networks, where the activation of neurons should not be normalized with non-activated neurons. 
As a result, BatchNorm still outperforms LayerNorm for these models~\cite{DBLP:journals/corr/BaKH16}. 

\textbf{Batch Renormalization~\cite{DBLP:conf/nips/Ioffe17}.} 
Batch Renormalization (BatchReNorm) is an extension to BatchNorm that aims to alleviate the problem of small minibatches (or inaccurate minibatch mean, $\mu_{\mathcal{B}}$, and variance, $\sigma_{\mathcal{B}}$). 
BatchReNorm achieves this by incorporating the estimated global mean ($\mu$) and variance ($\sigma$) during \emph{training}, and introducing two hyper-parameters to contain the difference between ($\mu_{\mathcal{B}}$, $\sigma_{\mathcal{B}}$) and ($\mu$, $\sigma$).
These two hyper-parameters are gradually relaxed such that the earlier training phase is more like BatchNorm, and the later phase is more like BatchReNorm. 

%\begin{wraptable}{R}{0.48\textwidth}
\begin{table}[h]
  \centering
  \small
  \begin{tabular}{cc cc}
  \toprule
  \multicolumn{2}{c}{BatchNorm} & \multicolumn{2}{c}{BatchReNorm} \\
  \cmidrule(lr){1-2} \cmidrule(lr){3-4} 
  {IID} & {Non-IID} & {IID} & {Non-IID} \\ \midrule
  78.8\% & \bad{65.4\%} & 78.1\% & \bad{75.3\%} \\
  \bottomrule \\
  \end{tabular}
  \caption[Top-1 validation accuracy (CIFAR-10) with BatchNorm and BatchReNorm]{Top-1 validation accuracy (CIFAR-10) with BatchNorm and BatchReNorm for BN-LeNet, using BSP with $K=2$ partitions.}
  \label{tbl:batch_renorm_results}
\end{table}
%\end{wraptable} 

We evaluate BatchReNorm with BN-LeNet over CIFAR-10 to see if BatchReNorm can solve the problem of Non-IID data partitions. 
We replace all BatchNorm layers with BatchReNorm layers, and we carefully select the BatchReNorm hyper-parameters so that BatchReNorm achieves the highest validation accuracy in both the IID and Non-IID settings.
Table~\ref{tbl:batch_renorm_results} shows the Top-1 validation accuracy. 
We see that while BatchNorm and BatchReNorm achieve similar accuracy in the IID setting, they both perform worse in the Non-IID setting. 
In particular, while BatchReNorm performs much better than BatchNorm in the Non-IID setting (75.3\% vs. 65.4\%), BatchReNorm still loses $\sim\!\!3\%$ accuracy compared to the IID setting. 
This is not surprising, because BatchReNorm still relies on minibatches to certain degree, and prior work has shown that BatchReNorm's performance still degrades when the minibatch size is small~\cite{DBLP:conf/nips/Ioffe17}. 
Hence, BatchReNorm cannot completely solve the problem of Non-IID data partitions, which is a more challenging problem than small minibatches.

\textbf{Group Normalization~\cite{DBLP:conf/eccv/WuH18}.} 
Group Normalization (GroupNorm) is an alternative normalization mechanism that aims to overcome the shortcomings of BatchNorm and LayerNorm. 
GroupNorm divides adjacent channels into groups of a prespecified size $\mathcal{G}_{size}$, and computes the per-group mean and variance for \emph{each} input sample. 
Specifically, for a four-dimensional input $\mathcal{B}\times\mathcal{C}\times\mathcal{W}\times\mathcal{H}$, GroupNorm partitions the set of channels ($\mathcal{C}$) into multiple groups ($\mathcal{G}$) of size $\mathcal{G}_{size}$.
GroupNorm then computes $|\mathcal{B}| \cdot |\mathcal{G}|$ means and variances along the $\mathcal{G}_{size} \times \mathcal{W} \times \mathcal{H}$ dimension. 
Hence, GroupNorm does not depend on minibatches for normalization (the shortcoming of BatchNorm), and GroupNorm does not assume all channels make equal contributions (the shortcoming of LayerNorm). 

We evaluate GroupNorm with BN-LeNet over CIFAR-10 to see if we can use GroupNorm as an alternative to BatchNorm in the Non-IID setting.
We carefully select $\mathcal{G}_{size} = 2$, which works best with this DNN. 
Figure~\ref{fig:group_norm_results} shows the Top-1 validation accuracy with GroupNorm and BatchNorm across decentralized learning algorithms. 
We make two major observations.

%\begin{wrapfigure}{R}{0.60\textwidth}
\begin{figure}[h]
\centering
\includegraphics[width=0.80\textwidth]{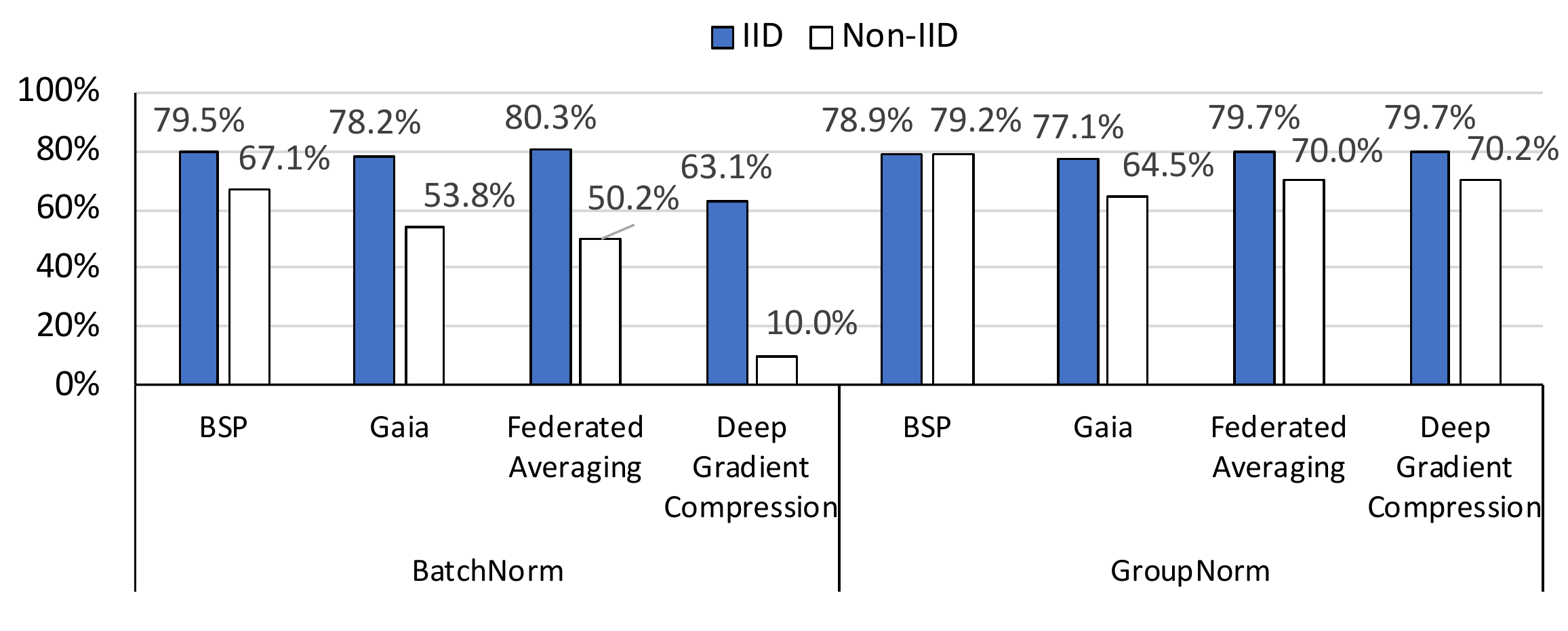}
\caption{Top-1 validation accuracy (CIFAR-10) with BatchNorm and GroupNorm for BN-LeNet with $K = 5$ partitions.} 
\label{fig:group_norm_results}
\end{figure}
%\vspace{-30pt}
%\end{wrapfigure} 

First, GroupNorm successfully recovers the accuracy loss of BatchNorm with BSP in the Non-IID setting. 
As the figure shows, GroupNorm with BSP achieves 79.2\% validation accuracy in the Non-IID setting, which is as good as the accuracy in the IID setting. 
This shows GroupNorm can be used as an alternative to BatchNorm to overcome the Non-IID data challenge for BSP. 
Second, GroupNorm dramatically helps the decentralized learning algorithms to improve model accuracy in the Non-IID setting as well.
We see that with GroupNorm, there is 14.4\%, 8.9\% and 8.7\% accuracy loss for \gaia, \fedavg and \dgc, respectively. 
While the accuracy losses are still significant, they are better than their BatchNorm counterparts by an additive 10.7\%, 19.8\% and 60.2\%, respectively. 
%This suggests GroupNorm can be a legitimate option to replace BatchNorm for decentralized learning settings in practice.
%\comm{Phil}{Just to doublecheck, all the Gaia results in this paper are for the original local-model variant of Gaia, right?}
%\comm{Kevin}{Yes}

\textbf{Summary.}
Overall, our study shows that GroupNorm~\cite{DBLP:conf/eccv/WuH18} can be a good alternative to BatchNorm in the Non-IID setting, especially for computer vision tasks. 
For BSP, it fixes the problem, while for decentralized learning algorithms, it greatly decreases the accuracy loss.
However, it is worth noting that BatchNorm is widely adopted in many DNNs, hence, more study should done to see if GroupNorm can always replace BatchNorm for different applications and DNN models. 
As for other tasks such as recurrent (e.g., LSTM~\cite{DBLP:journals/neco/HochreiterS97}) and generative (e.g., GAN~\cite{DBLP:conf/nips/GoodfellowPMXWOCB14}) models, other normalization techniques such as LayerNorm~\cite{DBLP:journals/corr/BaKH16} can be good options because \emph{(i)} they are shown to be effective in these tasks and \emph{(ii)} they are not dependent on minibatches, hence, they are unlikely to suffer the problems of BatchNorm in the Non-IID setting. 

\section{Degree of Deviation from IID}
\label{sec:skewness}

Our study in \xref{sec:overview}--\xref{sec:batch_norm} 
assumes a strict case of non-IID data partitions, where each training label only exists in a data partition \emph{exclusively}.
While this assumption may be a reasonable approximation for some applications (e.g., for \appface, a person's face image may exist only in a data partition for a geo-region in which the person lives), it could be an extreme case for other applications (e.g., \appimage).
Here, we study how the problem of non-IID data changes with the degree of deviation from IID (the skewness) by controlling the fraction of data that are non-IID (\xref{subsec:setup}). 
Figure~\ref{fig:noniid_degree} illustrates the CIFAR-10 Top-1 validation accuracy of AlexNet and GN-LeNet (our name for BN-LeNet with GroupNorm replacing BatchNorm, Figure~\ref{fig:group_norm_results}) in the 20\%, 40\%, 60\% and 80\% non-IID setting.
We make two key observations.

\begin{figure*}[h]
  \centering
  \begin{subfigure}[t]{0.48\linewidth}
    \centering
    \includegraphics[width=1.0\textwidth]{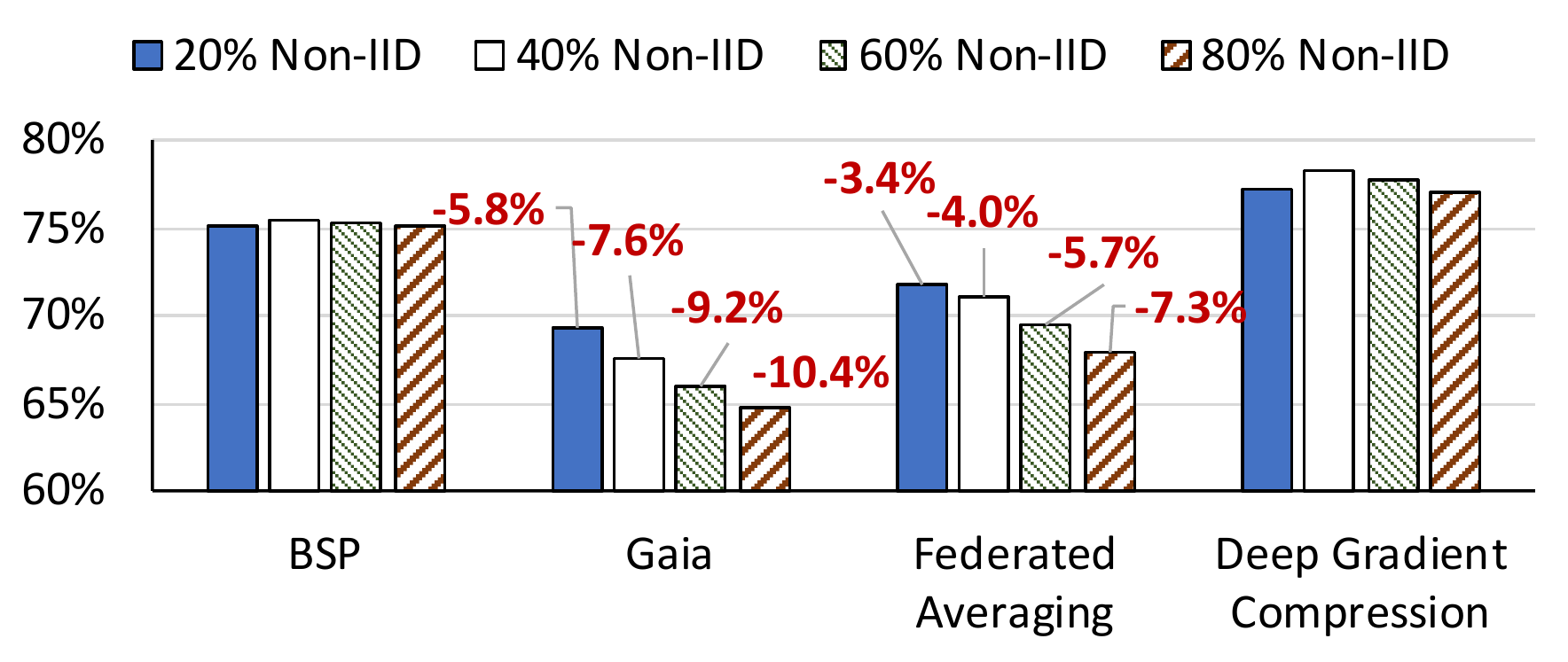}
    \caption{AlexNet} 
    \label{fig:noniid_degree_alexnet}
  \end{subfigure}
   \centering 
  \begin{subfigure}[t]{0.48\linewidth}
    \centering
    \includegraphics[width=1.0\textwidth]{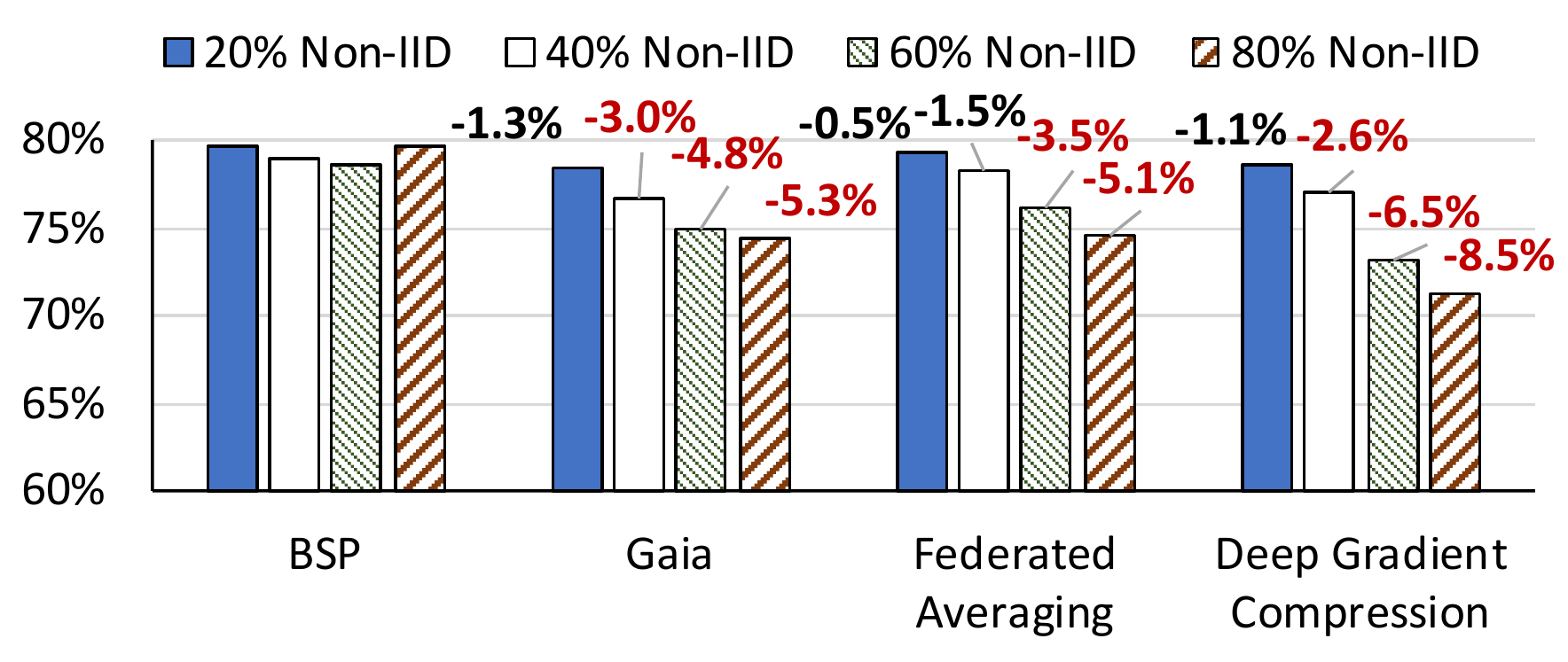}
    \caption{GN-LeNet} 
    \label{fig:noniid_degree_gnlenet}
  \end{subfigure}
  \caption[Top-1 validation accuracy (CIFAR-10) over various degrees of non-IID data]{Top-1 validation accuracy (CIFAR-10) over various degrees of non-IID data. We have zoomed in on 60\% accuracy and above. The ``-x\%'' label on each bar indicates the accuracy loss from BSP in the IID setting.}
  \label{fig:noniid_degree}
\end{figure*}

\textbf{1) Partial non-IID data is also problematic.} 
We see that for all three decentralized learning algorithms, partial non-IID data still cause major accuracy loss. 
Even with a small degree of non-IID data such as 20\%, we still see 5.8\% and 3.4\% accuracy loss for \gaia and \fedavg in AlexNet (Figure~\ref{fig:noniid_degree_alexnet}). 
The only exception is AlexNet with \dgc, which retains model accuracy in these partial non-IID settings. 
However, the same technique suffers significant accuracy loss for GN-LeNet in the partial non-IID settings (Figure~\ref{fig:noniid_degree_gnlenet}).
We conclude that the problem of non-IID data does not occur only with exclusive non-IID data partitioning, and hence, the problem exists in a vast majority of practical decentralized settings. 

\textbf{2) The degree of deviation from IID often determines the difficulty level of the problem.} 
We observe that the degree of skew changes the landscape of the problem significantly. 
In most cases, the model accuracy gets worse with higher degrees of skew, and the accuracy gap between 80\% and 20\% non-IID data can be as large as 7.4\% (GN-LeNet with \dgc). 
We see that while most decentralized learning algorithms can retain model quality with certain degree of non-IID data, there is usually a limit. 
For example, when training over 20\% non-IID data, all three decentralized learning algorithms stay within 1.3\% accuracy loss for GN-LeNet (Figure~\ref{fig:noniid_degree_gnlenet}). 
However, their accuracy losses become unacceptable when they are dealing with 40\% or higher non-IID data. 
%This observation suggests that: \emph{(i)} we may need different solutions for different degrees of non-IID data and \emph{(ii)} knowing the degree of deviation from IID may be quite useful in determining an appropriate solution. 

%\section{{\sscoutSec}: Decentralized Learning over Arbitrarily Non-IID Data Partitions}
{\ktb
  
\section{Our Approach: {\sscoutSec}}
\label{sec:solution}

%As \xref{sec:overview} shows, the problem of non-IID data partitions is a pervasive and fundamental problem for existing decentralized learning algorithms.

To address the problem of non-IID data partitions, we introduce {\sscout}, a generic, system-level approach that enables communication-efficient decentralized learning over \emph{arbitrarily} non-IID data partitions. We provide an overview of {\sscout} (\xref{subsec:solution_overview}), describe its key mechanisms (\xref{subsec:comm_control}), and present evaluation results (\xref{subsec:solution_results}).

\subsection{Overview of \sscout}
\label{subsec:solution_overview}

The goal of {\sscout} is a system-level solution that (i) enables high-accuracy, communication-efficient decentralized learning over arbitrarily non-IID data partitions; and (ii) is general enough to be applicable to a wide range of ML applications, ML systems, and decentralized learning algorithms. To this end, we design {\sscout} as a system-level module that can be integrated with various decentralized learning algorithms and ML systems. 

\begin{figure}[h!]
\centering
\includegraphics[width=0.8\textwidth]{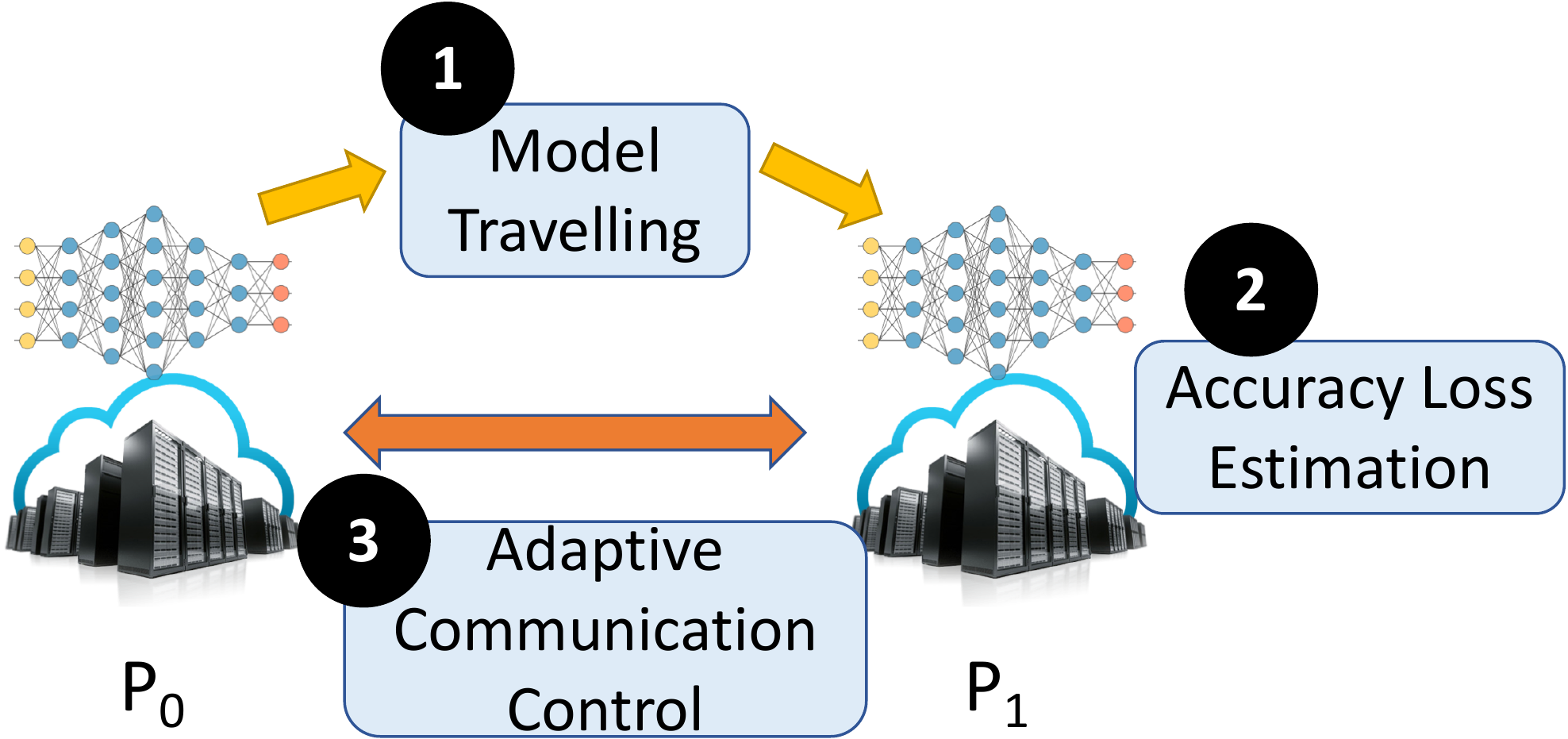}
\caption{Overview of {\sscout}} 
\label{fig:solution_overview}
\end{figure}

Figure~\ref{fig:solution_overview} overviews the {\sscout} design.

\begin{itemize}

\item \textbf{Estimate the degree of deviation from IID.} As \xref{sec:skewness} shows, knowing the degree of skew is very useful to determine an appropriate solution.
  To learn this key information, {\sscout} periodically moves the ML model from one data partition ($P_k$) to another during training (\emph{model traveling}, \ding{182} in Figure~\ref{fig:solution_overview}).
  {\sscout} then evaluates how well a model performs on a remote data partition by evaluating the model accuracy with a subset of training data on the remote node. 
  As we already know the training accuracy of this model in its originated data partition, we can infer the \emph{accuracy loss} in this remote data partition (\ding{183}).
The accuracy loss is essentially the performance gap for the same model over two different data partitions, which can be used as an approximation of the degree of skew.
For example, it is very likely that a remote data partition consists of very different data characteristics if the model in the local data partition has reached training accuracy 60\%, but the same model achieves only 30\% accuracy in the remote data partition.
More importantly, accuracy loss \emph{directly} captures the extent to which the model underperforms on the different data partition.

%\item \textbf{Estimate the degree of deviation from IID.} As \xref{sec:skewness} shows, knowing the degree of deviation from IID is very useful to determine an appropriate solution.
%  To learn this key information, {\sscout} periodically moves the ML model from one data partition ($P_k$) to another during training (\emph{model traveling}, \ding{182} in Figure~\ref{fig:solution_overview}).
%  {\sscout} then evaluates how well a model performs on a remote data partition by evaluating the model accuracy with a subset of training data on the remote node. As we already know the training accuracy of this model in its originated data partition, we can infer the \emph{accuracy loss} in this remote data partition (\ding{183}).
%  The accuracy loss is essentially the performance gap for the same model over two different data partitions, which can be used as an approximation of the degree of deviation from IID.
% For example, it is very likely that a remote data partition consists of very different data characteristics if the model in local data partition has reached training accuracy 30\%, but the same model only achieves 15\% accuracy in the remote data partition.

\item \textbf{Adaptive communication control (\ding{184}).} Based on the accuracy loss {\sscout} learns from model traveling, {\sscout} controls the tightness of communication among data partitions to retain model quality. {\sscout} controls the communication tightness by automatically tuning the hyper-parameters of the decentralized learning algorithm (\xref{subsec:decentral_parameter}). 
  This tuning process is essentially solving an optimization problem that aims to minimize communication among data partitions while keeping accuracy loss within a reasonable threshold (\xref{subsec:comm_control} provides more details).
%\squishlistend
\end{itemize}

{\sscout} handles non-IID data partitions in a manner that is transparent to ML applications and decentralized learning algorithms, and it controls communication based on the accuracy loss across partitions. Thus, we do not need to use the most conservative mechanism (e.g., BSP) all the time, and can adapt to whatever skew is present for the particular ML application and its training data partitions (\xref{sec:skewness}).

%{\sscout} handles non-IID data partitions in a manner that is transparent to ML applications and decentralized learning algorithms, and it controls communication based on the degree of deviation from IID. Thus, we do not need to use the most conservative mechanism (e.g., BSP) all the time as the degree of deviation from IID varies with ML applications and their data generation contexts (\xref{sec:skewness}).

\subsection{Mechanism Details}
\label{subsec:comm_control}

We now discuss the mechanisms of {\sscout} in detail.

\noindent \textbf{Accuracy Loss.} The accuracy loss between data partitions represents the degree of model divergence. As \xref{subsec:decentral_cause} discusses, ML models in different data partitions tend to specialize for their training data, especially when we use decentralized learning algorithms to relax communication. 
%As a result, accuracy drop is likely to be proportional to the degree of training data difference.

Figure~\ref{fig:accuracy_drop} demonstrates the above observation by plotting the accuracy loss between different data partitions when training GoogleNet over CIFAR-10 with \gaia. Two observations are in order. First, the accuracy drop changes drastically from the IID setting (0.4\% on average) to the Non-IID setting (39.6\% on average). This is expected as each data partition sees very different training data in the Non-IID setting, which leads to very different models in different data partitions. Second, more conservative hyper-parameters can lead to smaller accuracy drop in the Non-IID setting. For example, the accuracy drop for $T_0 = 2\%$ is significantly smaller than for larger settings of $T_0$.
%This is also intuitive as model divergence can be controlled by tightening communication between data partitions. 

\begin{figure}[h]
  \centering
  \begin{subfigure}[t]{0.9\linewidth}
    \centering
    \includegraphics[width=1.0\textwidth]{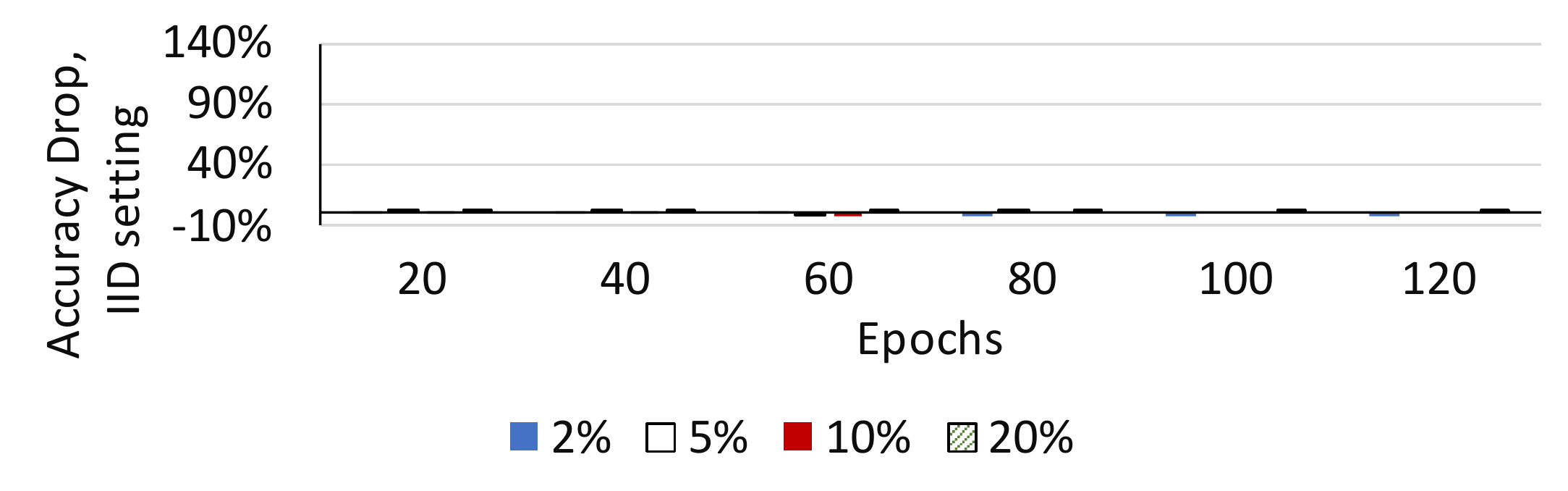}
  \end{subfigure}
   \centering 
  \begin{subfigure}[t]{0.9\linewidth}
    \centering
    \includegraphics[width=1.0\textwidth]{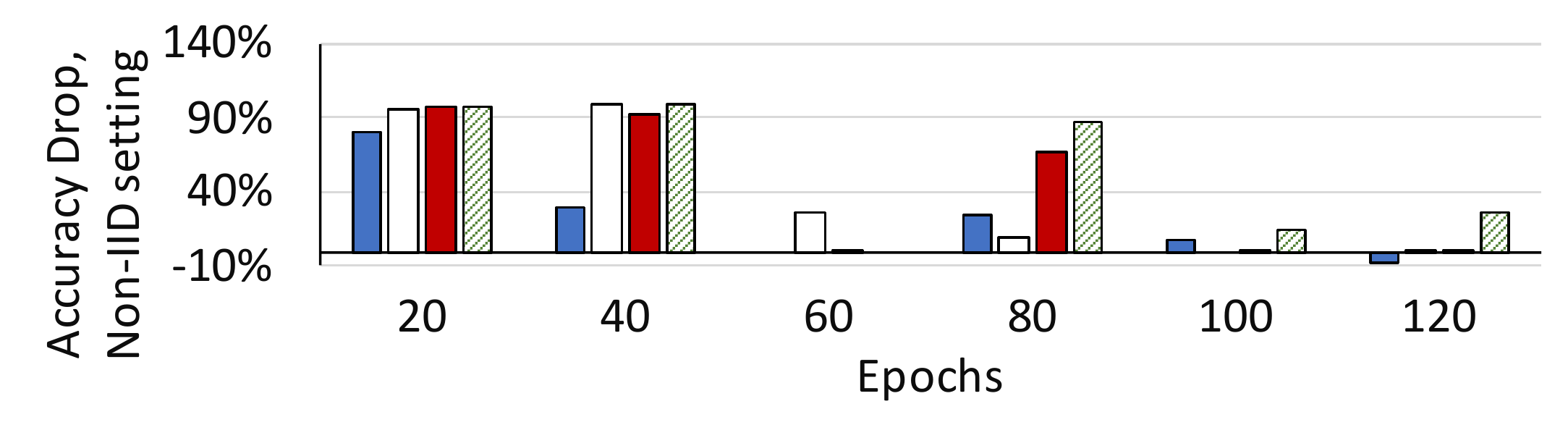}
  \end{subfigure}
  \vspace{-0.2in}
  \caption{Training accuracy drop between data partitions when training GoogleNet over CIFAR-10 with \gaia. Each bar represents a $T_0$ for \gaia}
  \label{fig:accuracy_drop}
\end{figure}

Based on the above observation, we can use accuracy loss (i) to estimate how much the models diverge from each other (reflecting training data differences); and (ii) to serve as an objective function for communication control. With accuracy loss, we do not need any domain-specific information from each ML application to learn and adapt to different degrees of deviation from IID, which makes {\sscout} much more widely applicable.

\noindent \textbf{Communication Control.} The goal of communication control is to retain model quality while minimizing communication among data partitions. Specifically, given a set of hyper-parameters $\theta_t$ for each iteration (or minibatch) $t$, the optimization problem for {\sscout} is to minimize the total amount of communication for a data partition $P_k$:

\ignore{
\begin{small}
  \begin{equation}
    \text{Comm} = \sum_{mt=0}^{\ceil*{\frac{T(\theta)}{MTP}}} \sum_{t=mt \cdot MTP}^{(mt + 1) \cdot MTP} C(\theta_t) + \sum_{mt=0}^{\ceil*{\frac{T(\theta)}{MTP}}} CM
  \end{equation}
\end{small}
}

\begin{small}
  \begin{equation}
    \label{eq:opt}
    \argmin_{\theta, MTP} \left( \sum_{mt=0}^{\ceil*{\frac{T(\theta)}{MTP}}} \sum_{t=mt \cdot MTP}^{(mt + 1) \cdot MTP} C(\theta_t) + \sum_{mt=0}^{\ceil*{\frac{T(\theta)}{MTP}}} CM \right)
  \end{equation}
\end{small}

where $T(\theta)$ is the total number of iterations to achieve the target model accuracy given all hyper-parameters $\theta$ throughout the training, $C(\theta_t)$ is the amount of communication given $\theta_t$, $MTP$ is the period size (in iterations) for model traveling, and $CM$ is the communication cost for the ML model (for model traveling).

%Larger $MTP$ saves communication overhead for model traveling, but this may lead to suboptimal $\theta$ as we have fewer opportunities to tune it. On the other hand, more conservative $\theta$ leads to higher communication ($C(\theta_t)$), but potentially smaller total iterations $T(\theta)$ to achieve target model quality, especially in the Non-IID setting.

In practice, however, it is impossible to optimize for Equation~\ref{eq:opt} with one-pass training because we cannot know $T(\theta)$ with different $\theta$ unless we train the model multiple times. 
%Also, it is unlikely that the target model accuracy is known in advance. In another word, we need to be able to select $\theta_t$ on the fly without knowing $T(\theta)$. 
We solve this problem by optimizing a proxy problem, which aims to minimize the communication while keeping the \emph{accuracy loss} to a small threshold $\sigma_{AL}$ so that we can control model divergence caused by non-IID data partitions. Specifically, our target function is:

\begin{small}
  \vspace{-10pt}
  \begin{equation}
  \label{eq:tune}
    \argmin_{\theta_t} \left( \lambda_{AL} \left( \texttt{max}(0, AL (\theta_{t}) - \sigma_{AL}) \right) + \lambda_{C} \frac{C(\theta_t)}{CM} \right)
  \end{equation}
  \vspace{-10pt}
\end{small}

where $AL (\theta_{t})$ is the accuracy loss based on the previously selected hyper-parameter (we memoize the most recent value for each $\theta$), and $\lambda_{AL}$, $\lambda_{C}$ are given parameters to determine the weight of accuracy loss and communication, respectively. We can employ various auto-tuning algorithms with Equation~\ref{eq:tune} to select $\theta_t$ such as hill climbing, stochastic hill climbing~\cite{russell2016artificial}, and simulated annealing~\cite{van1987simulated}. Note that we make $MTP$ not tunable here to further simplify the tuning.

\noindent \textbf{Model Traveling Overhead.} Using model traveling to learn accuracy loss can lead to heavy communication overhead if we need to do so for \emph{each pair} of data partitions, especially if we have a large number of data partitions. For broadcast-based decentralized learning settings (e.g., geo-distributed learning), we leverage the idea of overlay network in \gaia to reduce the communication overhead for model traveling. Specifically, we use hubs to combine and broadcast models~\cite{DBLP:conf/nsdi/HsiehHVKGGM17}. 
The extra hops incurred are fine since model traveling is not latency sensitive. As for server-client decentralized learning settings (e.g., federated learning), {\sscout} only needs to control the communication frequency between server and clients, and the overhead of model traveling can be combined with model downloading at the beginning of each communication round between the server and clients.

\subsection{Evaluation Results}
\label{subsec:solution_results}

We implement and evaluate {\sscout} in a GPU parameter server system~\cite{DBLP:conf/eurosys/CuiZGGX16} based on Caffe~\cite{DBLP:journals/corr/JiaSDKLGGD14}. We evaluate several aforementioned auto-tuning algorithms and we find that hill climbing provides the best results. We compare {\sscout} with two other baselines: (1) \sys{BSP}: the most communication-heavy approach that retains model quality in all Non-IID settings; and (2) \sys{Oracle}: the ideal, yet unrealistic, approach that selects the most communication-efficient $\theta$ that retains model quality by \emph{running all possible $\theta$} in each setting prior to measured execution. Figure~\ref{fig:solution_result} shows the communication savings over \sys{BSP} for both {\sscout} and \sys{Oracle} when training with \gaia. Note that all results achieve the same validation accuracy as \sys{BSP}. We make two observations.

\begin{figure}[h]
  \centering
  \begin{subfigure}[t]{0.48\linewidth}
    \centering
    \includegraphics[width=1.0\textwidth]{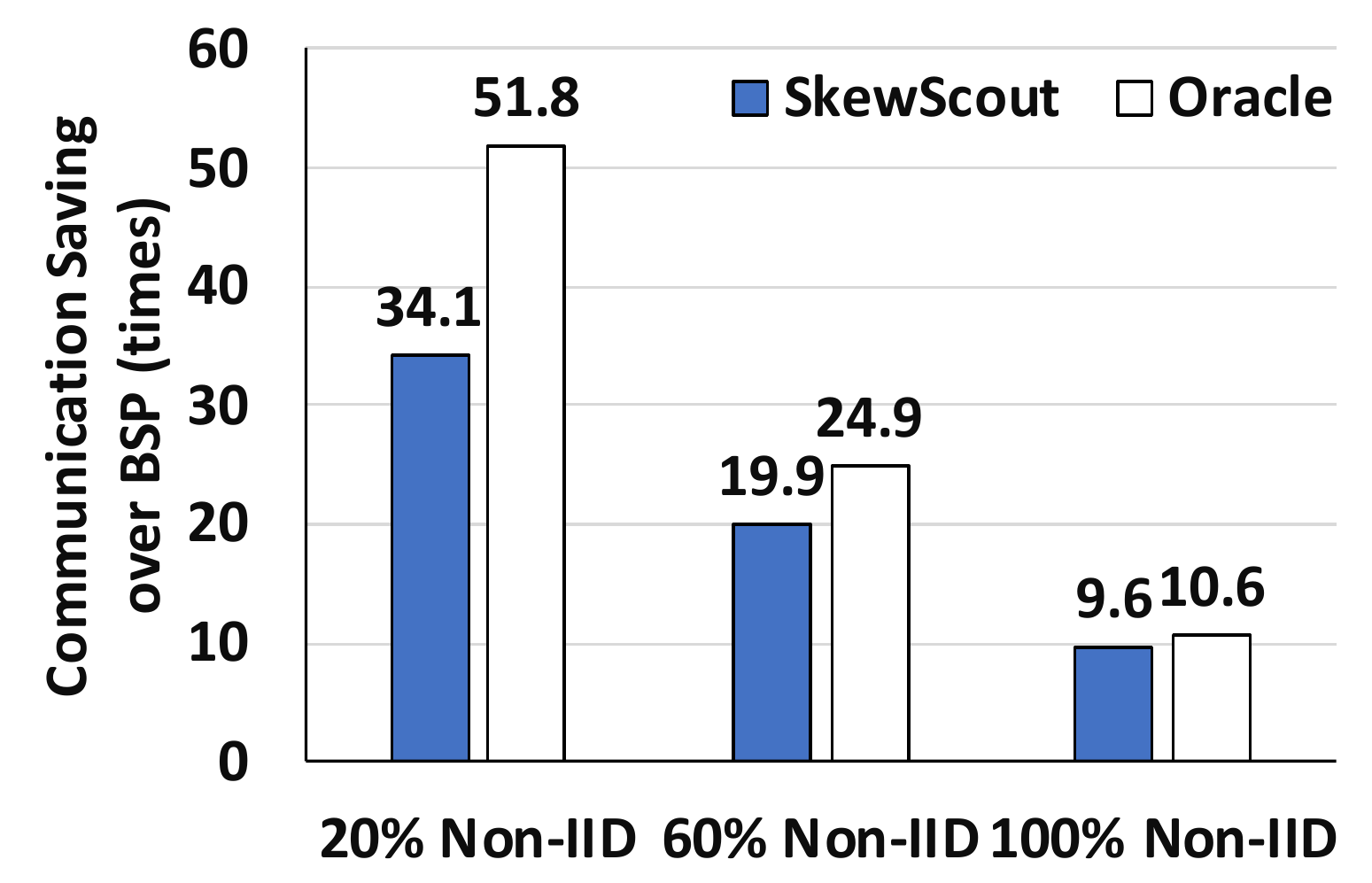}
    \caption{AlexNet}
  \end{subfigure}
   \centering 
  \begin{subfigure}[t]{0.48\linewidth}
    \centering
    \includegraphics[width=1.0\textwidth]{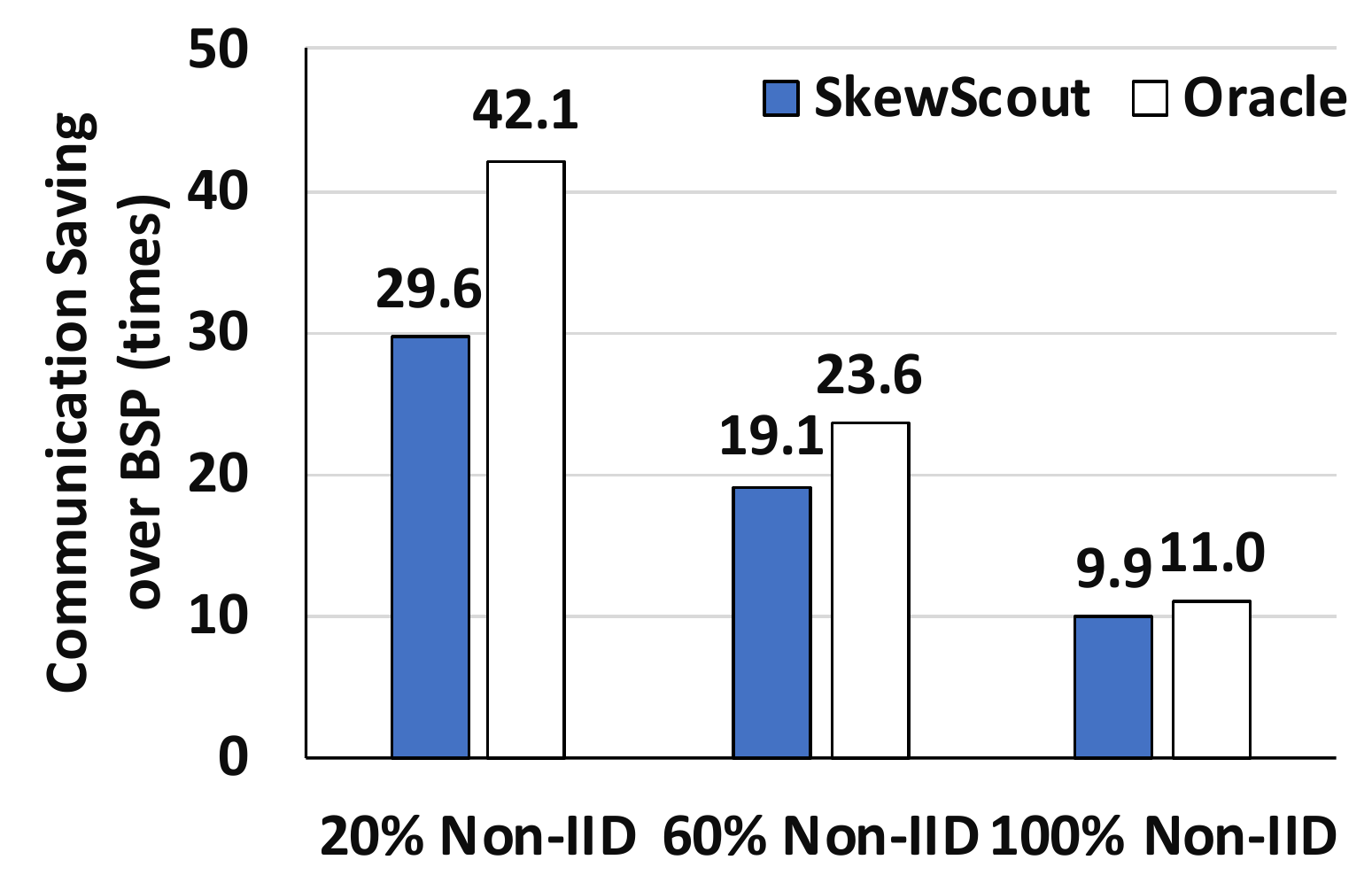}
    \caption{GoogLeNet}
  \end{subfigure}
  \vspace{-0.1in}
  \caption{Communication savings over BSP with {\sscout} and \sys{Oracle} for training over CIFAR-10.}
  \label{fig:solution_result}
\end{figure}

First, {\sscout} is much more effective than \sys{BSP} in handling Non-IID settings. Overall, {\sscout} achieves 9.6--34.1$\times$ communication savings over \sys{BSP} in various Non-IID settings without sacrificing model accuracy. As expected, {\sscout} saves more communication with less skewed data because {\sscout} can loosen communication in these settings (\xref{sec:skewness}). 

Second, {\sscout} is not far from the ideal \sys{Oracle} baseline. Overall, {\sscout} only requires 1.1--1.5$\times$ more communication than \sys{Oracle} to achieve the same model accuracy. {\sscout} cannot match the communication savings of \sys{Oracle} because: (i) {\sscout} needs to do model traveling periodically, which leads to some extra overheads; and (ii) for some $\theta$, high accuracy loss at the beginning can still end up with a high quality model, which {\sscout} cannot foresee. As \sys{Oracle} requires many runs in practice, we conclude that {\sscout} is an effective, one-pass solution for decentralized learning over non-IID data partitions. 

}

\section{Summary}

As most timely and relevant ML data are generated at different places, decentralized learning provides an important path for ML applications to leverage these decentralized data. 
However, decentralized data are often generated at different contexts, which leads to a heavily understudied problem: \emph{non-IID training data partitions}. 
We conduct a detailed empirical study of this problem, revealing three key findings. 
First, we show that training over non-IID data partitions is a fundamental and pervasive problem for decentralized learning, as all decentralized learning algorithms in our study suffer major accuracy loss in the Non-IID setting. 
Second, we find that DNNs with batch normalization are particularly vulnerable in the Non-IID setting, with even the most communication-heavy approach being unable to retain model quality. 
We further discuss the cause and potential solution to this problem. 
Third, we show that the difficulty level of the non-IID data problem varies greatly with the degree of deviation from IID. 
\kt{Based on these findings, we present \sscout, a system-level approach to
minimizing communication while retaining model quality even under non-IID data.
We hope that the findings and insights in this chapter, as well as our open source code, will spur further research into the fundamental and important problem of non-IID data in decentralized learning.}

%\appendix
%\clearpage

%\makeatletter
%\setlength{\@fptop}{0pt}
%\makeatother

%\twocolumn[
%\onecolumn

%\section*{Appendix}

\begin{subappendices}

\section{Details of Decentralized Learning Algorithms}
\label{appendix:decentral_algo}

\begin{algorithm}[h]
\caption{\gaia~\cite{DBLP:conf/nsdi/HsiehHVKGGM17} on node $k$ for vanilla momentum SGD}
\label{algo:gaia}
\begin{algorithmic}[1]
\Require initial weights $w_0 = \{w_0[0],..., w_0[M]\}$
\Require $K$ data partitions (or data centers); initial significance threshold $T_0$
\Require local minibatch size $B$; momentum $m$; learning rate $\eta$; local dataset $\mathcal{X}_k$
\State $u_0^k \gets 0$; $v_0^k \gets 0$
\State $w_0^k \gets w_0$
\For{$t = 0, 1, 2,...$}
\State $b \gets$ (sample $B$ data samples from $\mathcal{X}_k$)
\State $u_{t+1}^k \gets m \cdot u_t^k - \eta \cdot \bigtriangledown f(w_t^k, b)$
\State $w_{t+1}^k \gets w_{t}^k + u_{t+1}^k$
\State $v_{t+1}^k \gets v_{t}^k + u_{t+1}^k$ 
\Comment{Accumulate weight updates}
\For{$j = 0, 1,...M$}
\State $S \gets ||\frac{v_{t+1}^k}{w_{t+1}^k}|| > T_t$
\Comment{Check if accumulated updates are significant}
\State $\widetilde{v}_{t+1}^k[j] \gets v_{t+1}^k[j] \odot S$
\Comment{Share significant updates with other $P_k$}
\State $v_{t+1}^k[j] \gets v_{t+1}^k[j] \odot \neg S$
\Comment{Clear significant updates locally}
\EndFor
\For{$i = 0, 1,...K; i \neq k$}
\State $w_{t+1}^k \gets w_{t+1}^k + \widetilde{v}_{t+1}^i$
\Comment{Apply significant updates from other $P_k$}
\EndFor
\State $T_{t+1} \gets \texttt{update\_threshold}(T_t)$
\Comment{Decreases whenever the learning rate decreases}
\EndFor
\end{algorithmic}
\end{algorithm}

%]

\begin{algorithm}[h]
\caption{\fedavg~\cite{DBLP:conf/aistats/McMahanMRHA17} on node $k$ for vanilla momentum SGD}
\label{algo:fedavg}
\begin{algorithmic}[1]
\Require initial weights $w_0$; $K$ data partitions (or clients)
\Require local minibatch size $B$; local iteration number $Iter_{Local}$
\Require momentum $m$; learning rate $\eta$; local dataset $\mathcal{X}_k$
\State $u^k \gets 0$
\For{$t = 0, 1, 2,...$}
\State $w_t^k \gets w_t$
\Comment{Get the latest weight from the server}
\For{$i = 0,...Iter_{Local}$}
\State $b \gets$ (sample $B$ data samples from $\mathcal{X}_k$)
\State $u^k \gets m \cdot u^k - \eta \cdot \bigtriangledown f(w_t^k, b)$
\State $w_t^k \gets w_{t}^k + u^k$
\EndFor
\State $\texttt{all\_reduce:}\: w_{t+1} \gets \sum^{K}_{k=1} \frac{1}{K} w_t^k$
\Comment{Average weights among all partitions}
\EndFor
\end{algorithmic}
\end{algorithm}

\begin{algorithm}[h]
\caption{\dgc~\cite{DBLP:conf/ICLR/LinHMWD18} on node $k$ for vanilla momentum SGD}
\label{algo:dgc}
\begin{algorithmic}[1]
\Require initial weights $w_0 = \{w_0[0],..., w_0[M]\}$
\Require $K$ data partitions (or data centers); $s\%$ update sparsity
\Require local minibatch size $B$; momentum $m$; learning rate $\eta$; local dataset $\mathcal{X}_k$
\State $u_0^k \gets 0$; $v_0^k \gets 0$
\For{$t = 0, 1, 2,...$}
\State $b \gets$ (sample $B$ data samples from $\mathcal{X}_k$)
\State $g_{t+1}^k \gets - \eta \cdot \bigtriangledown f(w_t, b)$
\State $g_{t+1}^k \gets \texttt{gradient\_clipping}(g_{t+1}^k)$
\Comment{Gradient clipping}
\State $u_{t+1}^k \gets m \cdot u_t^k + g_{t+1}^k$
%\State $w_{t+1}^k \gets w_{t}^k + u_{t+1}^k$
\State $v_{t+1}^k \gets v_{t}^k + u_{t+1}^k$ 
\Comment{Accumulate weight updates}
\State $T \gets s\% \text{ of } ||v_{t+1}^k||$ 
\Comment{Determine the threshold for sparsified updates}
\For{$j = 0, 1,...M$}
\State $S \gets ||v_{t+1}^k|| > T$
\Comment{Check if accumulated updates are top $s\%$}
\State $\widetilde{v}_{t+1}^k[j] \gets v_{t+1}^k[j] \odot S$
\Comment{Share top updates with other $P_k$}
\State $v_{t+1}^k[j] \gets v_{t+1}^k[j] \odot \neg S$
\Comment{Clear top updates locally}
\State $u_{t+1}^k[j] \gets u_{t+1}^k[j] \odot \neg S$
\Comment{Clear the history of top updates (momentum correction)}
\EndFor
\State $w_{t+1} = w_t + \sum_{k=1}^K \widetilde{v}_{t+1}^k$
\Comment{Apply top updates from all $P_k$}
\EndFor
\end{algorithmic}
\end{algorithm}

\clearpage

\section{Training Parameters}
\label{appendix:training_parameter}

Tables \ref{tbl:training_parameter_cifar10}, \ref{tbl:training_parameter_imagenet}, \ref{tbl:training_parameter_face} list major training parameters for all the applications, models, and datasets in our study.

\begin{table}[h!]
  \centering
  \small
  \begin{tabular}{c c c c c c}
    \toprule
    \textbf{Model} & \scell{\textbf{Minibatch size} \\\textbf{per node} \\ \textbf{(5 nodes)}} & \textbf{Momentum} & \scell{\textbf{Weight}\\ \textbf{decay}} & \textbf{Learning rate} & \textbf{Total epochs} \\ \midrule
    AlexNet & 20 & 0.9 & 0.0005 & \scell{$\eta_0 = 0.0002$, divides by \\10 at epoch 64 and 96} & 128 \\ \midrule
    GoogLeNet & 20 & 0.9 & 0.0005 & \scell{$\eta_0 = 0.002$, divides by \\10 at epoch 64 and 96} & 128 \\ \midrule
    \scell{LeNet, BN-LeNet,\\GN-LeNet} & 20 & 0.9 & 0.0005 & \scell{$\eta_0 = 0.002$, divides by \\10 at epoch 64 and 96} & 128 \\ \midrule
    ResNet-20 & 20 & 0.9 & 0.0005 & \scell{$\eta_0 = 0.002$, divides by \\10 at epoch 64 and 96} & 128 \\ 
    \bottomrule \\
  \end{tabular}
  \caption{Major training parameters for \appimage over CIFAR-10}
  \label{tbl:training_parameter_cifar10}
\end{table}

\begin{table}[h!]
  \centering
  \small
  \begin{tabular}{c c c c c c}
    \toprule
    \textbf{Model} & \scell{\textbf{Minibatch size} \\\textbf{per node} \\ \textbf{(8 nodes)}} & \textbf{Momentum} & \scell{\textbf{Weight}\\ \textbf{decay}} & \textbf{Learning rate} & \textbf{Total epochs} \\ \midrule
    GoogLeNet & 32 & 0.9 & 0.0002 & \scell{$\eta_0 = 0.0025$, polynomial \\decay, power = 0.5} & 60 \\ \midrule
    ResNet-10 & 32 & 0.9 & 0.0001 & \scell{$\eta_0 = 0.00125$, polynomial \\decay, power = 1} & 64 \\ 
    \bottomrule \\
  \end{tabular}
  \caption[Major training parameters for \appimage over ImageNet]{Major training parameters for \appimage over ImageNet. Polynomial decay means $\eta = \eta_0 \cdot (1 - \frac{\text{iter}}{\text{max\_iter}})^{\text{power}}$.}
  \label{tbl:training_parameter_imagenet}
\end{table}

\begin{table}[h!]
  \centering
  \small
  \begin{tabular}{c c c c c c}
    \toprule
    \textbf{Model} & \scell{\textbf{Minibatch size} \\\textbf{per node} \\ \textbf{(4 nodes)}} & \textbf{Momentum} & \scell{\textbf{Weight}\\ \textbf{decay}} & \textbf{Learning rate} & \textbf{Total epochs} \\ \midrule
    center-loss & 64 & 0.9 & 0.0005 & \scell{$\eta_0 = 0.025$, divides by \\10 at epoch 4 and 6} & 7 \\ 
    \bottomrule \\
  \end{tabular}
  \caption{Major training parameters for \appface over CASIA-WebFace.}
  \label{tbl:training_parameter_face}
\end{table}

%\clearpage

\section{More Algorithm Hyper-Parameter Results}
\label{appendix:decentral_parameter}

In \xref{subsec:decentral_parameter} we presented results varying \gaia's $T_0$ hyper-parameter. 
In this section, we show results for \fedavg and \dgc, varying their respective hyper-parameters. 
We make the same observations as \xref{subsec:decentral_parameter} for these algorithms (Tables \ref{tbl:fedavg_hyper_parameter} and \ref{tbl:dgc_hyper_parameter}).

\begin{table}[h!]
  \centering
  \small
  \begin{tabular}{c cc cc cc cc}
    \toprule
    \multirow{2}{*}{Configuration} & \multicolumn{2}{c}{AlexNet} & \multicolumn{2}{c}{GoogLeNet} & \multicolumn{2}{c}{LeNet} & \multicolumn{2}{c}{ResNet20} \\
    \cmidrule(lr){2-3} \cmidrule(lr){4-5} \cmidrule(lr){6-7} \cmidrule(lr){8-9}
    & {IID} & {Non-IID} & {IID} & {Non-IID} & {IID} & {Non-IID} & {IID} & {Non-IID} \\ \midrule
    BSP &  74.9\% & 75.0\%	& 79.1\% & 78.9\% & 77.4\% & 76.6\% & 83.7\% & \bad{44.3\%} \\ \midrule
    $Iter_{Local} = 5$ & 73.7\% & \bad{62.8\%} & \bad{75.8\%} & \bad{68.9\%} & 79.7\% & \bad{67.3\%} & \bad{73.6\%} & \bad{31.3\%} \\ \midrule
    $Iter_{Local} = 10$ & 73.5\% & \bad{60.1\%} & \bad{76.4\%} & \bad{64.8\%} & 79.3\% & \bad{63.2\%} & \bad{73.4\%} & \bad{28.0\%} \\ \midrule
    $Iter_{Local} = 20$ & 73.4\% & \bad{59.4\%} & \bad{76.3\%} & \bad{64.0\%} & 79.1\% & \bad{10.1\%} & \bad{73.8\%} & \bad{28.1\%} \\ \midrule
    $Iter_{Local} = 50$ & 73.5\% & \bad{56.3\%} & \bad{75.9\%} & \bad{59.6\%} & 79.2\% & \bad{55.6\%} & \bad{74.0\%} & \bad{26.3\%} \\ \midrule
    $Iter_{Local} = 200$ & 73.7\% & \bad{53.2\%} & \bad{76.8\%} & \bad{52.9\%} & 79.4\% & \bad{54.2\%} & \bad{75.7\%} & \bad{27.3\%} \\ \midrule
    $Iter_{Local} = 500$ & 73.0\% & \bad{24.0\%} & \bad{76.8\%} & \bad{20.8\%} & 79.6\% & \bad{19.4\%} & \bad{74.1\%} & \bad{24.0\%} \\ \midrule
    $Iter_{Local} = 1000$ & 73.4\% & \bad{23.9\%} & \bad{76.1\%} & \bad{20.9\%} & 78.3\% & \bad{19.0\%} & \bad{74.3\%} & \bad{22.8\%}  \\     
    \bottomrule \\
  \end{tabular}
  \caption[CIFAR-10 Top-1 validation accuracy with various \fedavg hyper-parameters]{CIFAR-10 Top-1 validation accuracy with various \fedavg hyper-parameters. The configurations that lose more than 2\% accuracy are highlighted. Note that larger settings for $Iter_{Local}$ mean significantly greater communication savings.}
  \label{tbl:fedavg_hyper_parameter}
\end{table}

\begin{table}[h]
  \centering
  \small
  \begin{tabular}{c cc cc cc cc}
    \toprule
    \multirow{2}{*}{Configuration} & \multicolumn{2}{c}{AlexNet} & \multicolumn{2}{c}{GoogLeNet} & \multicolumn{2}{c}{LeNet} & \multicolumn{2}{c}{ResNet20} \\
    \cmidrule(lr){2-3} \cmidrule(lr){4-5} \cmidrule(lr){6-7} \cmidrule(lr){8-9}
    & {IID} & {Non-IID} & {IID} & {Non-IID} & {IID} & {Non-IID} & {IID} & {Non-IID} \\ \midrule
    BSP &  74.9\% & 75.0\%	& 79.1\% & 78.9\% & 77.4\% & 76.6\% & 83.7\% & \bad{44.3\%} \\ \midrule
    $E_{warm} = 8$ & 75.5\% & \bad{72.3\%} & 78.3\% & \bad{10.0\%} & 80.3\% & \bad{47.2\%} & \bad{10.0\%} & \bad{10.0\%} \\ \midrule
    $E_{warm} = 4$ & 75.5\% & 75.7\% & 79.4\% & \bad{61.6\%} & \bad{10.0\%} & \bad{47.3\%} & \bad{10.0\%} & \bad{10.0\%} \\ \midrule
    $E_{warm} = 3$ & 75.9\% & 74.9\% & 78.9\% & \bad{75.7\%} & \bad{64.9\%} & \bad{50.5\%} & \bad{10.0\%} & \bad{10.0\%} \\ \midrule
    $E_{warm} = 2$ & 75.7\% & 76.7\% & 79.0\% & \bad{58.7\%} & \bad{10.0\%} & \bad{47.5\%} & \bad{10.0\%} & \bad{10.0\%} \\ \midrule
    $E_{warm} = 1$ & 75.4\% & 77.9\% & 78.6\% & \bad{74.7\%} & \bad{10.0\%} & \bad{39.9\%} & \bad{10.0\%} & \bad{10.0\%}  \\ 
    \bottomrule \\
  \end{tabular}
  \caption[CIFAR-10 Top-1 validation accuracy with various \dgc hyper-parameters]{CIFAR-10 Top-1 validation accuracy with various \dgc hyper-parameters. The configurations that lose more than 2\% accuracy are highlighted. Note that smaller settings for $E_{warm}$ mean significantly greater communication savings.}
  \label{tbl:dgc_hyper_parameter}
\end{table}

\end{subappendices}

\chapter{Conclusion and Future Directions}
\label{ch:conclusion}

\section{Conclusion}

The goal of this thesis is to enable low-latency and low-cost ML
training and serving over real-world, large-scale data, which are
highly distributed and rapidly growing. These highly-distributed and
rapidly-growing data pose major computation, communication, and
statistical challenges to ML. In this thesis, we demonstrate that the
latency and cost of ML training and serving over such real-world data
can be improved by one to two orders of magnitude by designing ML
systems that exploit the characteristics of ML algorithms, ML model
structures, and ML training/serving data. We present three directions
to address the aforementioned challenges.

First, we present {\focus} (Chapter~\ref{ch:focus}), a system that
provides both low-cost and low-latency querying over large,
continuously-growing datasets such as videos. {\focus}' architecture
flexibly and effectively divides the query processing work between
ingest time and query time. At ingest time (on live videos), {\focus}
uses cheap techniques to construct an approximate index. At query
time, {\focus} leverages this approximate index to provide low
latency, but compensates for the lower accuracy of the cheap CNNs
through the judicious use of an expensive CNN. This architecture
enables orders-of-magnitude faster queries with only a small
investment at ingest time, and allows flexibly trading off ingest cost
and query latency. Our evaluations using \kff{real-world videos} from
traffic, surveillance, and news domains show that {\focus} reduces
ingest cost on average by $48\times$ (up to $92\times$) and makes
queries \kff{on average} $125\times$ (up to $607\times$) faster
compared to state-of-the-art baselines at two ends of the design
spectrum (ingest heavy or query heavy). The ideas and insights behind
{\focus} can be applied to designing efficient systems for many other
forms of querying on large and continuously-growing datasets in many
domains, \kff{such as audio, bioinformatics, and geoinformatics.}}

Second, we present {\gaia} (Chapter~\ref{ch:gaia}), a first general
geo-distributed ML system that (1) differentiates the communication
over a LAN from the communication over WANs to make efficient use of
the scarce and heterogeneous WAN bandwidth, and (2) is general and
flexible enough to deploy a wide range of ML algorithms while
requiring \emph{no} change to the ML algorithms themselves. We present
a new ML \synchronization model, \protocol (\protoabbrv), whose key
idea is to dynamically eliminate insignificant communication between
data centers while still guaranteeing the correctness of ML algorithms
by ensuring that all significant updates are synchronized in time. Our
experiments on our prototypes of {\gaia} running across 11 Amazon EC2
global regions and on a cluster that emulates EC2 WAN bandwidth show
that, compared to two two state-of-the-art distributed ML training
systems, {\gaia} (1) significantly improves performance, by
1.8--53.5$\times$, (2) \khii{has performance within 0.94--1.40$\times$
  of running the same ML algorithm on a local area network (LAN) in a
  single data center}, and (3) significantly reduces the monetary cost
of running the same ML algorithm on WANs, by 2.6--59.0$\times$.

Finally, we present a first detailed study and a system-level solution
(Chapter~\ref{ch:noniid}) on the problem of non-IID data partitions
for decentralized learning. Our study reveals three key
findings. First, to our knowledge, our study is the first to show that
the problem of non-IID data partitions is a fundamental and pervasive
challenge for decentralized learning, as it exists in all ML
applications, DNN models, training datasets, and decentralized
learning algorithms in our study. Second, we make a new observation
showing that the challenge of non-IID data partitions is particularly
problematic for DNNs with batch normalization, even under the most
conservative communication approach. Finally, we show that the
difficulty level of this problem varies with the degree of deviation
from IID. With these findings in mind, we present
\sscout, a system-level approach that adapts the communication
frequency of decentralized learning algorithms to the (skew-induced)
accuracy loss between data partitions.  We also show that group
normalization can recover much of the skew-induced accuracy loss of
batch normalization. We hope that our findings will facilitate
more solution developments for this important but heavily overlooked
challenge in decentralized learning.

\section{Future Research Directions}

This dissertation opens up several future research directions. In this
section, we discuss several future directions in which the idea and
approaches described in this thesis can be applied or extended to
tackle these problems for ML over highly-distributed and
continuously-growing data.

\subsection{ML Serving for Growing and Distributed Data}

Chapter~\ref{ch:focus} presents {\focus}, a system that provides
low-cost and low-latency ML serving over rapidly-growing
datasets. However, many rapidly-growing datasets are also distributed
in many places, such as traffic and enterprise cameras. This poses an
interesting dimension that is not considered by {\focus}:
communication between data generators (e.g., cameras) and ML serving
providers (e.g., cloud). As most modern cameras are equipped with some
processing capability, there is an opportunity to design an efficient
ML serving system for growing and distributed datasets. For example,
we can build approximate indexes using camera's processors, and only
transmit necessary information to the cloud. We can then intelligently
decide which part of the videos needs to be sent to the cloud based on
user queries. \kt{Similarly, we can design a system that runs part of
  the processing on edge clusters instead of data centers so that the
  system can support real-time object detection with much shorter
  turnaround latency. In general, building an end-to-end ML serving
  system for distributed and growing data is a promising
  research direction.}

%IGNORE
\ignore{
\subsection{Decentralized Learning Systems for Non-IID Data Partitions}

As Chapter~\ref{ch:noniid} discusses, ML training over non-IID data is
a fundamental problem for decentralized learning, as existing
decentralized learning algorithms (including {\gaia} in
Chapter~\ref{ch:gaia}) suffer from major model quality loss over
non-IID data partitions. On the contrary, while communication heavy
approaches such as BSP can retain model quality over non-IID data, its
communication overhead is unrealistic in most decentralized learning
settings. Hence, it is imperative to develop new decentralized
learning algorithms or systems that can retain model quality while
minimizing communication overheads. One possible approach is to
leverage our finding in Chapter~\ref{ch:noniid}, which shows the
difficulty level of this problem varies with the degree of deviation
from IID. Specifically, we can explore mechanisms to detect the degree
of non-IID, and then apply appropriate communication mechanisms. With
this approach, we can minimize communication overhead, especially if
the data partitions do not deviate too much from IID.
}

\subsection{ML Training Systems for Intermittent Networks}
\kt{ Chapter~\ref{ch:gaia} introduces \gaia to mitigate limited WAN
  bandwidth when training over geo-distributed data. However, there is
  another challenge with limited connectivity: intermittent network
  connection. While \gaia uses some mechanisms
  (\xref{subsec:protocol}) to ensure all data centers are always
  synchronized, this can lead to very high cost if connection to some
  data centers is lost for an extended period of time (as all other
  data centers are idle waiting). The other extreme approach is to
  disregard the disconnected participants and keep training, which may
  work well if there are thousands or millions of participants (e.g.,
  federated learning~\cite{DBLP:conf/aistats/McMahanMRHA17}) but may
  not be a good solution for few training participants (e.g.,
  geo-distributed learning). It is still unclear how to build a ML
  training system that can handle intermittent connection reliably and
  efficiently. Specifically, the system should be able to keep making
  progress while waiting for the disconnected participants, but allow
  the disconnected participants to catch up if their connections are
  recovered. Tackling this challenge will further enable practical ML
  training over highly-distributed data.}

\subsection{Training Local and Global Models for Non-IID Data Partitions}

An alternative solution to ML training over non-IID data partition is
to train local models that fit the data distribution in each data
partition, while leveraging data from other data partitions to improve
model quality (such as federated multi-task
learning~\cite{DBLP:conf/nips/SmithCST17} or \kt{semi-cyclic
  SGD~\cite{DBLP:conf/icml/EichnerKMST19}}). However, existing
approaches have major shortcomings as they are either not general for
all ML applications (such as DNNs), (2) not communication efficient,
or (3) unable to provide a global model, which is still important when
local models are ineffective. A better approach is to develop an ML
system that trains \emph{both} local and global models in a
communication efficient manner, \emph{at the same time}. One
possibility to design such a system is that we can leverage the idea
of multi-task learning to partition an ML model into local and global
part, and then we can apply decentralized learning algorithms (such as
{\gaia} in Chapter~\ref{ch:gaia}) and solutions like {\sscout}
(Chapter~\ref{ch:noniid}) to make training of the global part
communication efficient and resilient in the Non-IID settings. We can
then explore mechanisms to construct a complete global model based on
the global part and local part of the model.

\subsection{ML Training Systems for Non-IID Data over Space and Time}

Chapter~\ref{ch:noniid} only discusses one form of non-IID data, which
is non-IID data over \emph{space} with multiple partitions. More
generally, rapidly-growing data can also vary significantly over
\emph{time}, which adds another dimension to the non-IID data
problem. Prior work on non-IID data over time (i.e., continuous
learning~\cite{DBLP:journals/nn/ParisiKPKW19}) does not consider the
space dimension. As most timely and relevant ML data are continuously
generated at many different locations, it is important to explore ML
systems that can handle non-IID data in both space and time. The goals
of such a system are the capability to: (1) detect data change over
time in each location; (2) efficiently and incrementally update the
local and global models; and (3) tailor to the application requirements
on historical data. We believe the ideas presented in this thesis can
be extended to this interesting future direction.

\chapter*{Other Works of the Author}

{\ktb
During the course of my Ph.D., I have been interested in several topics beyond ML systems, such as novel hardware architectures and accelerators and their integration into the software stack. I had opportunities to work on these topics through collaboration with fellow graduate students and industrial collaborators. These projects not only broadened my horizon, they also helped me in learning research fundamentals. I would like to acknowledge these projects in this chapter.

In the early years of my Ph.D., I have worked on several projects on \emph{processing-in-memory (PIM)}, a promising paradigm that places computation close to data in memory. This provides a new opportunity to alleviate the main memory bottleneck in modern computers (also known as the ``memory wall''~\cite{DBLP:journals/sigarch/WulfM95}). In collaboration with Eiman Ebrahimi, Gwangsun Kim, and others, we designed mechanisms to enable GPU computation offloading to memory without burdening the programmer~\cite{DBLP:conf/isca/HsiehEKCOVMK16, DBLP:conf/sc/KimCOH17}. In collaboration with Amirali Boroumand, we proposed efficient cache coherence mechanisms for PIM architectures~\cite{DBLP:journals/cal/BoroumandGPHLHM17, DBLP:conf/isca/BoroumandGPHLAH19}. In collaboration with Vivek Seshadri, we proposed low-overhead mechanisms for bulk bitwise operations in DRAM~\cite{DBLP:journals/cal/SeshadriHBLKMGM15}. Finally, I also worked on architecting a pointer chasing accelerator in memory~\cite{DBLP:conf/iccd/HsiehKVCBGM16, DBLP:journals/corr/abs-1802-00320}. 

In collaboration with Nandita Vijaykumar, we have architected rich cross-layer abstractions to enhance programmability, performance portability, and performance in CPUs and GPUs. Three contributions are made in this line of work: (1) Expressive Memory~\cite{DBLP:conf/isca/VijaykumarJMHPE18}: a cross-layer abstraction to express and communicate higher-level program information from the application to the underlying OS/hardware to enhance memory optimizations; (2) The Locality Descriptor~\cite{DBLP:conf/isca/VijaykumarEHGM18}: a cross-layer abstraction to express data locality in GPUs; and (3) Zorua~\cite{DBLP:conf/micro/VijaykumarHPKSG16}: a framework to decouple the programming models from on-chip resource managements. We demonstrated significant performance benefits from enabling cross-layer optimizations.

In collaboration with Kevin Chang, we worked on mechanisms to reduce DRAM latency. We comprehensively characterized hundreds of DRAM chips and made several new observations about latency variation within DRAM. We further proposed a mechanism that exploits latency variation across DRAM cells within a DRAM chip to improve system performance~\cite{DBLP:conf/sigmetrics/ChangKHGHLLPKM16}.
}

\backmatter 

%\chapter{Conclusion}

%\appendix
%\include{appendix}

%\backmatter

%\renewcommand{\baselinestretch}{1.0}\normalsize

% By default \bibsection is \chapter*, but we really want this to show
% up in the table of contents and pdf bookmarks.
%\renewcommand{\bibsection}{\chapter{\bibname}}
%\renewcommand{\bibpreamble}{This text goes between the ``Bibliography''
%  header and the actual list of references}
%\bibliographystyle{plainnat}
%\bibliography{register} %your bib file

%\begin{small}
%\let\oldbibliography\thebibliography
%\renewcommand{\thebibliography}[1]{\oldbibliography{#1}
%\setlength{\itemsep}{0pt}} %Reducing spacing in the bibliography.
%\begin{small}
%\setstretch{0.85}
\bibliographystyle{IEEEtranS}
\bibliography{ref}
%\end{small}

\end{document}